\pdfoutput=1
\documentclass{article}

\usepackage[main, final]{neurips_2026}

\usepackage[utf8]{inputenc}
\usepackage[T1]{fontenc}
\usepackage{hyperref}
\usepackage{url}
\usepackage{graphicx}
\usepackage{booktabs}
\usepackage{arydshln}
\usepackage{amsfonts}
\usepackage{amssymb}
\usepackage{amsmath}
\usepackage{nicefrac}
\usepackage{microtype}
\usepackage{xcolor}
\usepackage{pifont}
\usepackage{multirow}
\usepackage{makecell}
\usepackage{stfloats}
\usepackage{wrapfig}
\usepackage{tabularx}
\usepackage{caption}
\usepackage{xspace}
\usepackage{float}
\usepackage{placeins}
\usepackage{needspace}
\usepackage{enumitem}
\usepackage{pgfplots}
\pgfplotsset{compat=1.18}
\definecolor{zoomyellow}{RGB}{255,215,0}

\newcommand{\eg}{e.g.\xspace}

\newcommand{\etal}{et~al.\xspace}

\newcommand{\modelname}{LOCO\xspace}
\newcommand{\benchname}{\modelname-bench\xspace}

\newcommand{\inputimg}[2][.192\linewidth]{%
	\begin{tikzpicture}[baseline=(img.south)]
		\node[inner sep=0] (img) {\includegraphics[width=#1]{#2}};
		\useasboundingbox (img.south west) rectangle (img.north east);
		\node[anchor=north, font=\scriptsize, text=black, fill=white, inner xsep=3pt, inner ysep=1pt, yshift=-2pt] at (img.north) {Input};
	\end{tikzpicture}%
}

\newlength{\rotimgh}
\newcommand{\rotlabel}[1]{\raisebox{\dimexpr 0.5\rotimgh - 0.5\height\relax}{\rotatebox{90}{#1}}}

\title{\modelname: Local Light-Aware Object Compositing with Spatially Varying Illumination-Augmented Data}

\author{%
  Jinseo Jeong\textsuperscript{1}, Hyunsoo Kim\textsuperscript{1}, Junseo Koo\textsuperscript{1},
  Junhyeog Yun\textsuperscript{1}, Gunhee Kim\textsuperscript{1}, Soo Ye Kim\textsuperscript{2}\thanks{Corresponding author.} \\
  \textsuperscript{1}Seoul National University \quad \textsuperscript{2}Adobe Research \\
  \noalign{\vskip 5pt}%
	Project page: \url{https://jinseo.kr/LOCO}\\
}

\begin{document}

\maketitle

\begin{figure}[h]
	\vspace{-20pt}
	\centering
	\includegraphics[width=\linewidth]{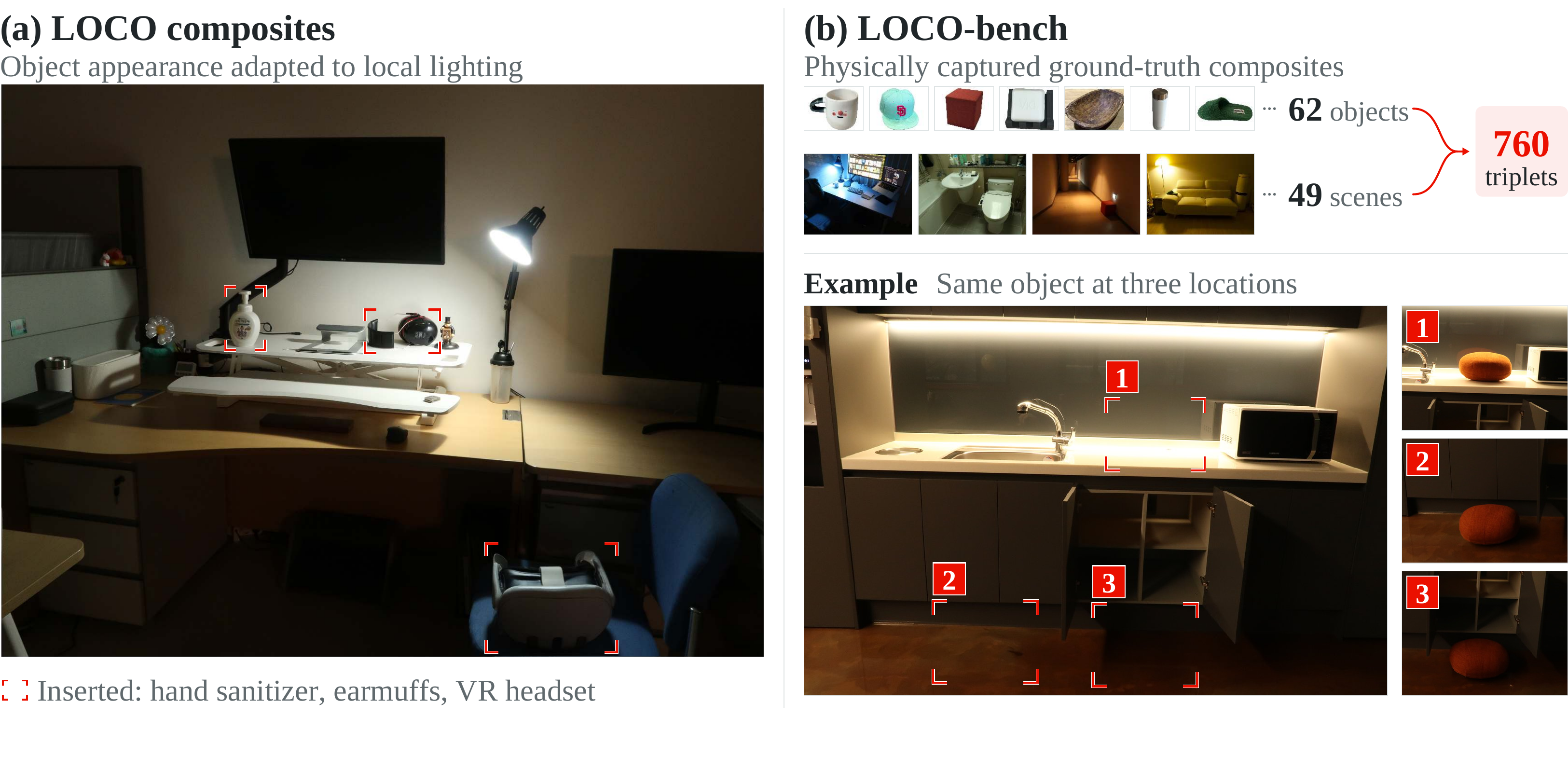}
	\caption{
		\textbf{Object compositing under spatially varying illumination.}
		\textit{(a)} Object compositing inserts a user-provided object into a background image at a chosen location.
		Within a single scene, surface brightness, color cast, and shadow direction vary with location—brighter near light sources and dimmer farther away.
		Our method, \modelname, adapts the inserted object's appearance to the local lighting at the inserted location.
		\textit{(b)} \benchname{} provides physically captured ground-truth composites of the same object at different locations within a shared scene.
		Numbered boxes mark the insertion locations; crops show the captured ground truth at each.
	}
	\label{fig:teaser}
\end{figure}

\begin{abstract}
	In indoor scenes, light sources and occlusions from scene geometry create \textit{spatially varying illumination}, including brightness gradients, cast shadows, and local color shifts that shape how every object appears.
	A less explored aspect of diffusion-based object compositing is whether an inserted object can adapt to this local illumination at the desired location.
	The core challenge is the scarcity of large-scale training data capturing how object appearance should adapt to location-dependent illumination.
	We present \modelname, a data augmentation framework for LOcal light-aware object COmpositing that leverages publicly available monocular videos as illumination sources.
	We treat each video frame as a virtual spotlight anchored at its camera pose, so that viewpoint variation across a video yields diverse light directions.
	With controllable beam shape, color, and intensity, this design enables large-scale generation of training data for placement-dependent illumination \textit{on the fly}.
	We also introduce \benchname, an indoor benchmark of physically captured composites under spatially varying illumination, where the same object is placed at multiple locations within each scene.
	Experiments on Laval Indoor SV HDR and \benchname show that \modelname outperforms prior methods in local illumination adaptation while preserving object identity.
\end{abstract}

\section{Introduction}
\label{sec:intro}
Within a single indoor scene, illumination varies from point to point.
A tabletop near a window may sit in bright daylight while a far corner falls into shadow~\citep{li2020inverse, Garon_2019_CVPR, srinivasan2020lighthousepredictinglightingvolumes}.

For an inserted object to appear naturally in the scene, its overall brightness, gradient lighting, and cast shadow must match the illumination at the chosen location.
Although this is an important aspect of object compositing, prior methods largely overlook this lighting effect.

Diffusion-based methods~\citep{ho2020denoisingdiffusionprobabilisticmodels,EsserKBEMSLLSBP24} have advanced object compositing in several directions, including geometric alignment~\citep{Song_2023_CVPR,chen2024anydoorzeroshotobjectlevelimage,yu2025objectmovergenerativeobjectmovement}, semantic coherence~\citep{tewel2025addit,lu2023tficondiffusionbasedtrainingfreecrossdomain,song2026insert,huang2025dreamfuseadaptiveimagefusion}, and color harmonization~\citep{wang2023semi,FreeCompose}.
Some recent approaches further improve global appearance alignment by leveraging diffusion-generated training data~\citep{huang2025dreamfuseadaptiveimagefusion, wang2025unicombine}.

\begin{figure}[b]
	\centering
	\scriptsize
	\renewcommand{\arraystretch}{1.1}
	\resizebox{\linewidth}{!}{%
		\begin{tabular}{@{} c @{\hspace{6pt}} c@{\hspace{2pt}}c @{}}
			\multirow{4}{*}{\makecell{Input                                                                                                                                                                                                                                    \\ \includegraphics[width=.320\linewidth]{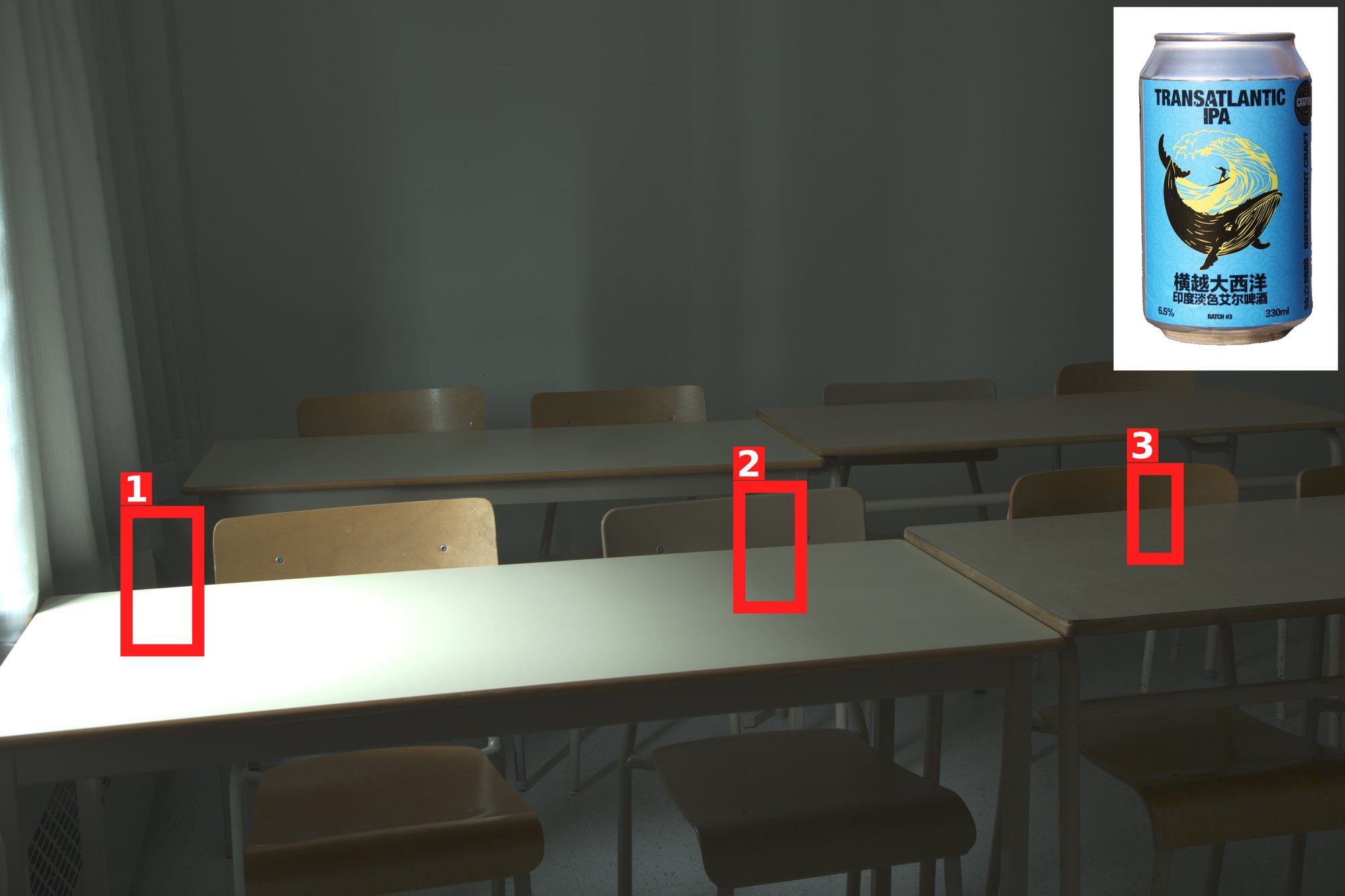}}}
			 & DreamFuse~\citep{huang2025dreamfuseadaptiveimagefusion}
			 & UniCombine~\citep{wang2025unicombine}                                                                                                                                                                                                                           \\
			 & \begin{tabular}[b]{@{}c@{\hspace{1pt}}c@{\hspace{1pt}}c@{}}
				   \includegraphics[width=.090\linewidth]{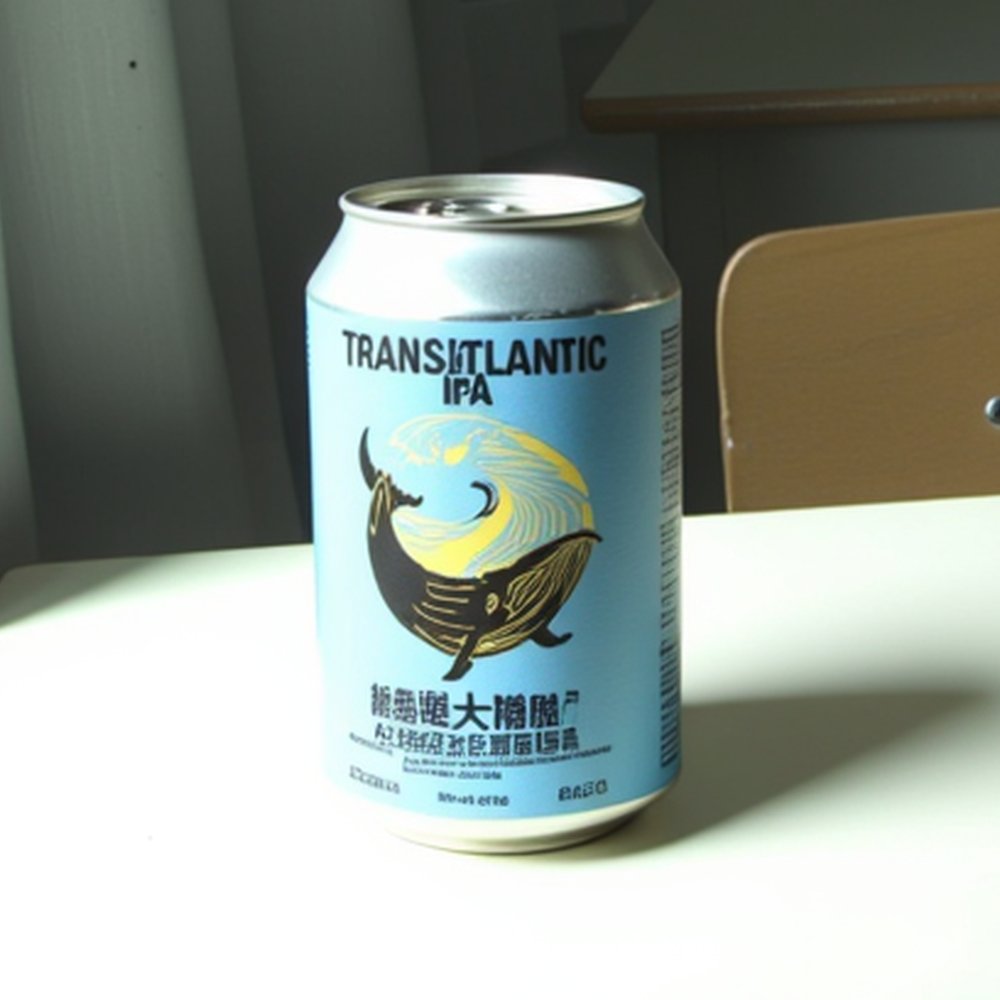} & \includegraphics[width=.090\linewidth]{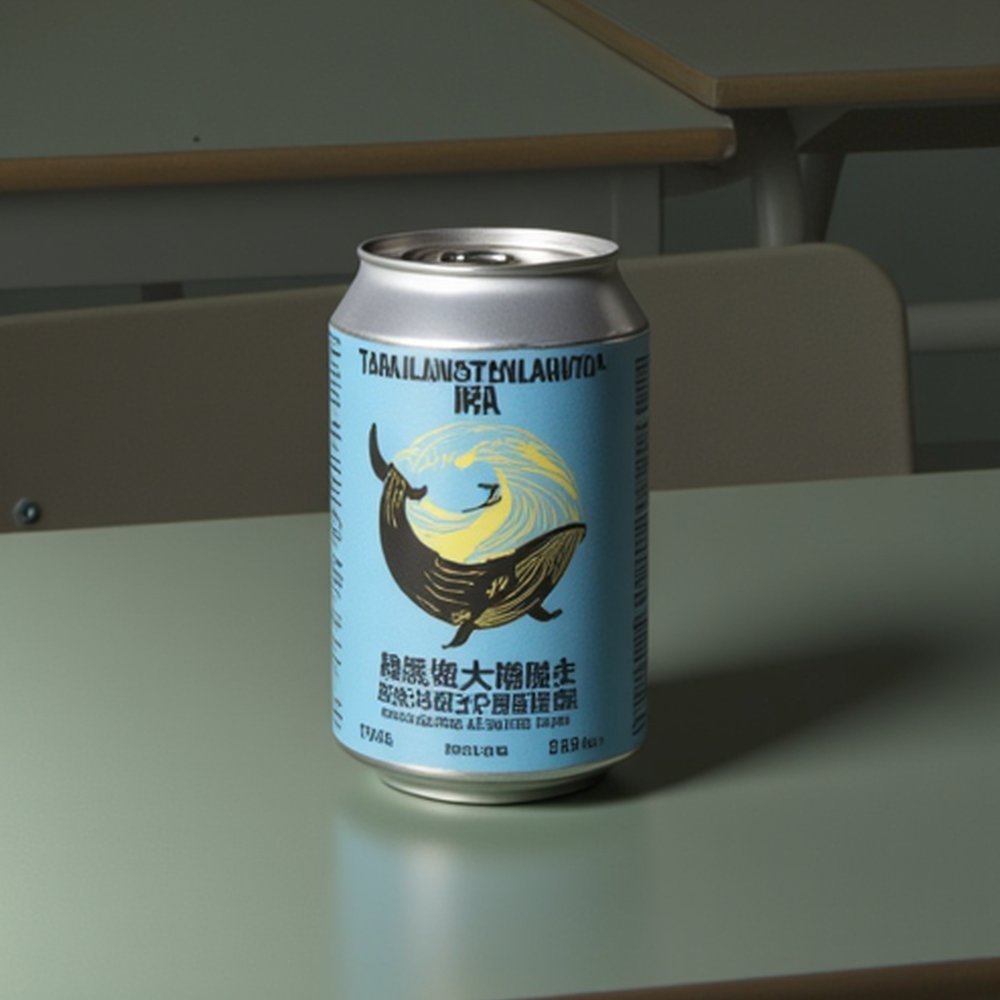} & \includegraphics[width=.090\linewidth]{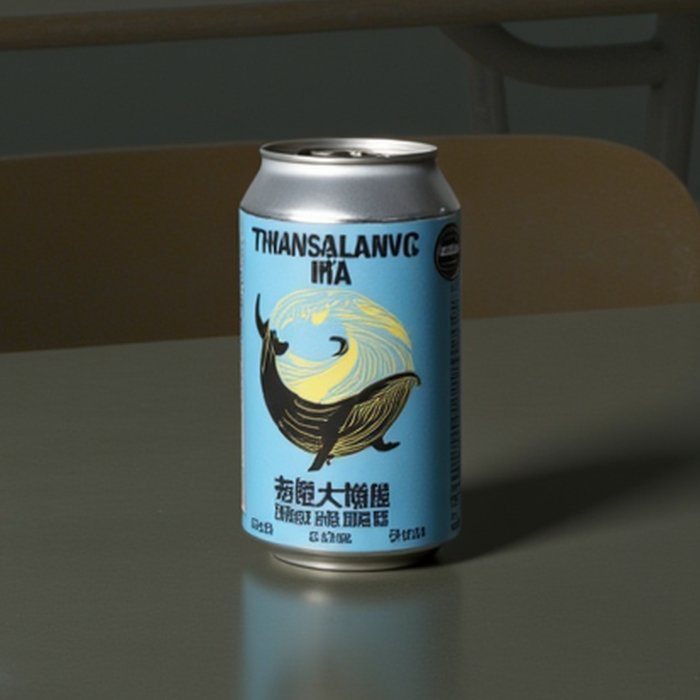} \\
			   \end{tabular}
			 & \begin{tabular}[b]{@{}c@{\hspace{1pt}}c@{\hspace{1pt}}c@{}}
				   \includegraphics[width=.090\linewidth]{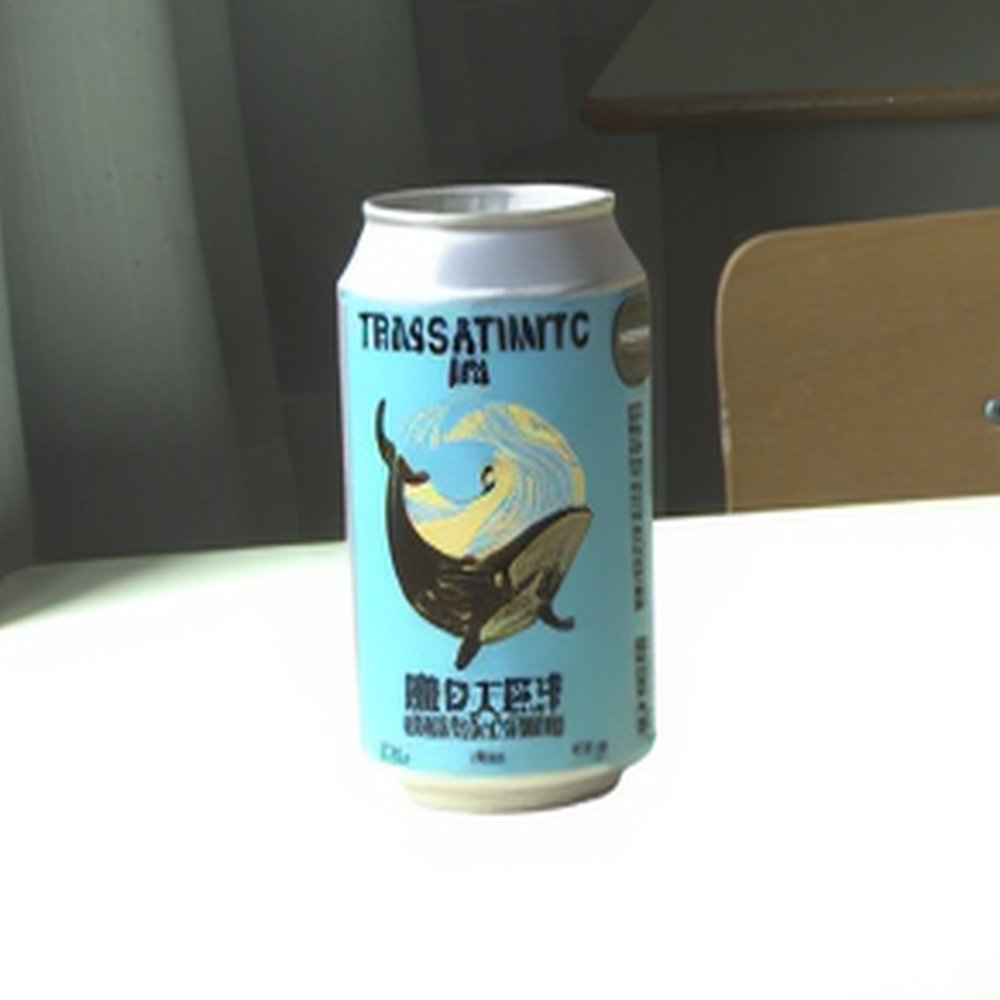} & \includegraphics[width=.090\linewidth]{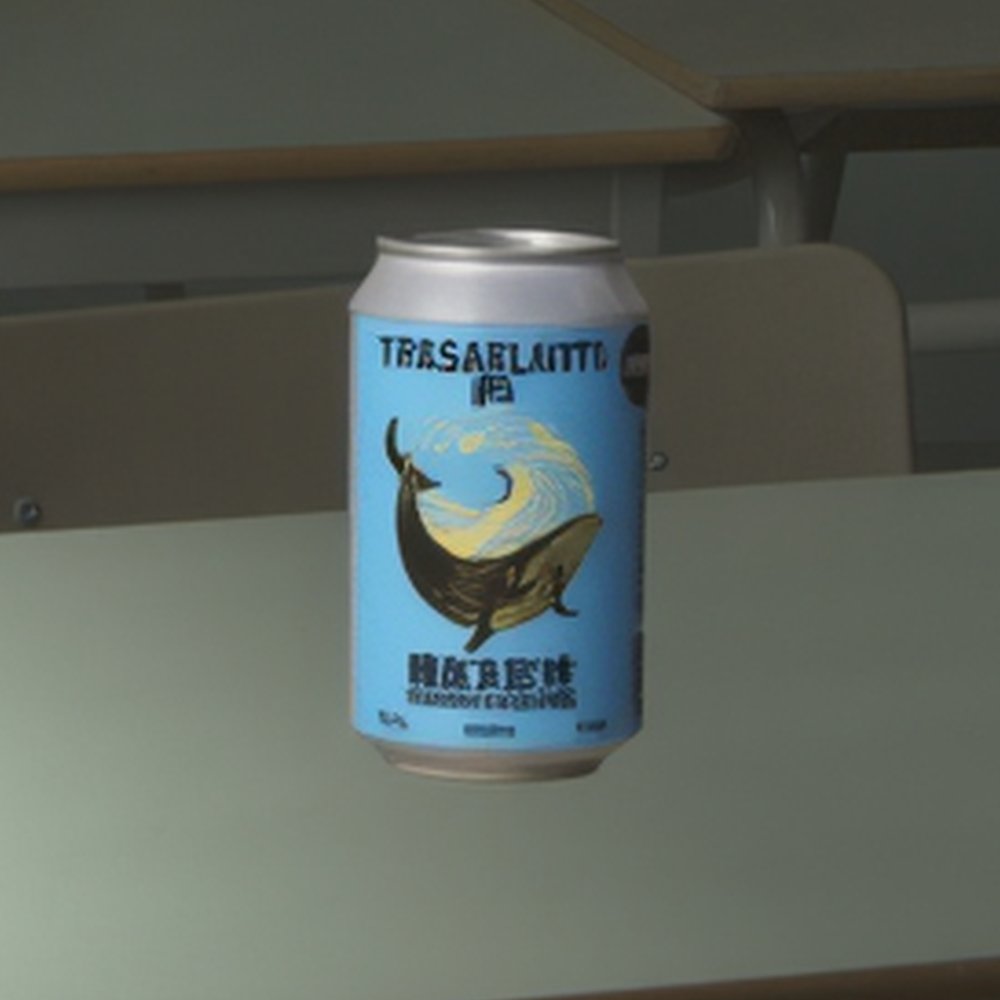} & \includegraphics[width=.090\linewidth]{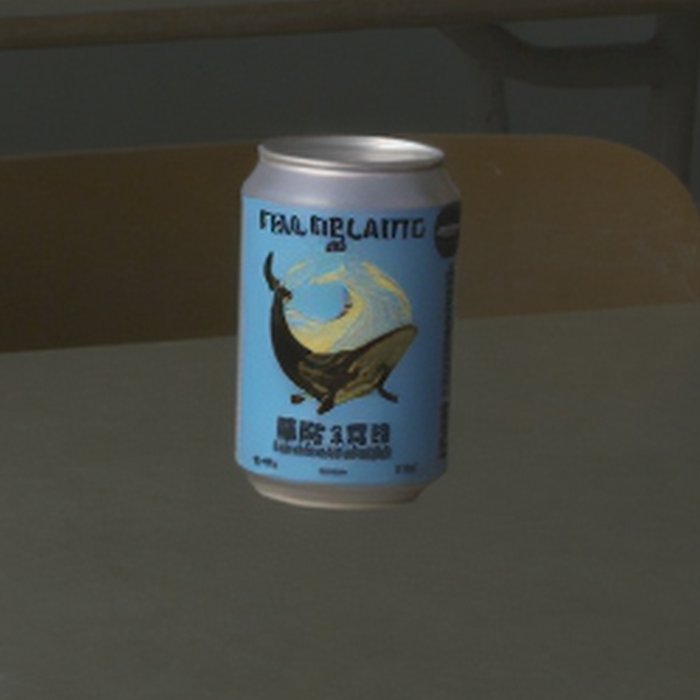} \\
			   \end{tabular}             \\
			 & OmniPaint~\citep{yu2025omnipaintmasteringobjectorientedediting}
			 & Insert Anything~\citep{song2026insert}                                                                                                                                                                                                                          \\
			 & \begin{tabular}[b]{@{}c@{\hspace{1pt}}c@{\hspace{1pt}}c@{}}
				   \includegraphics[width=.090\linewidth]{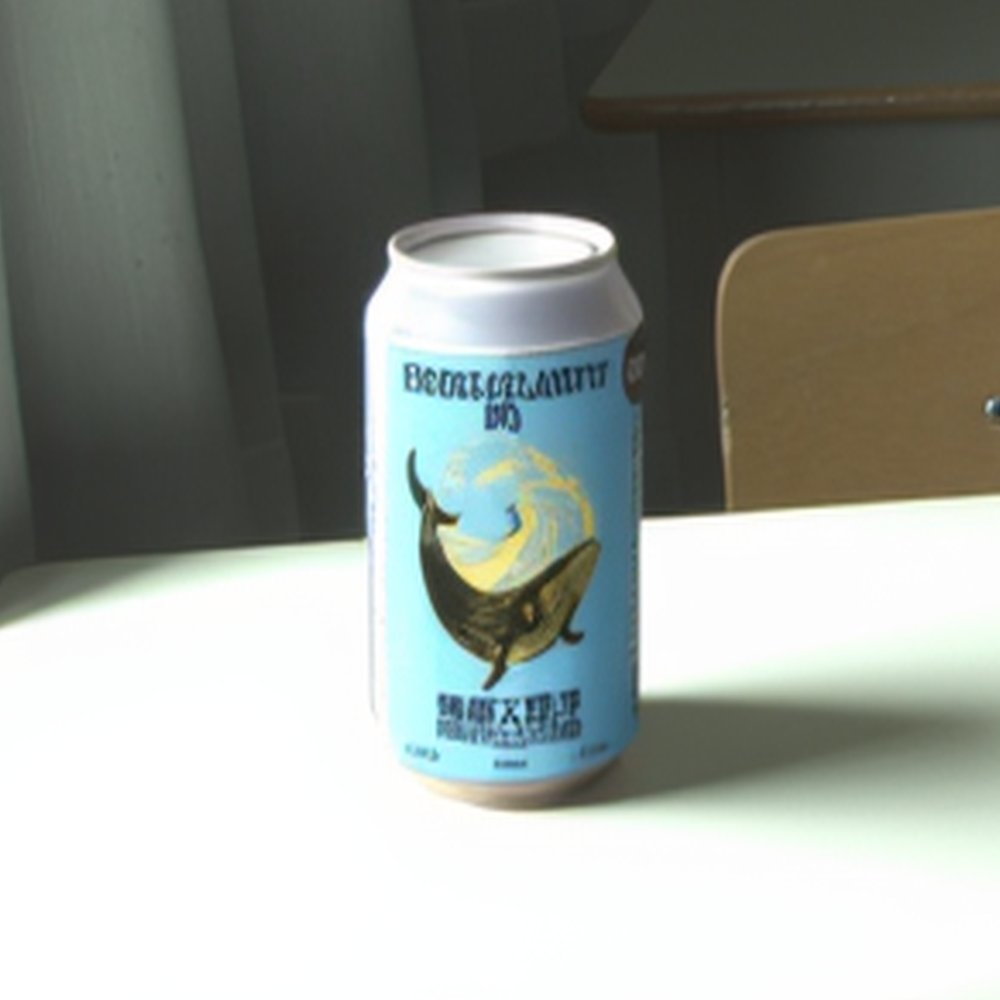} & \includegraphics[width=.090\linewidth]{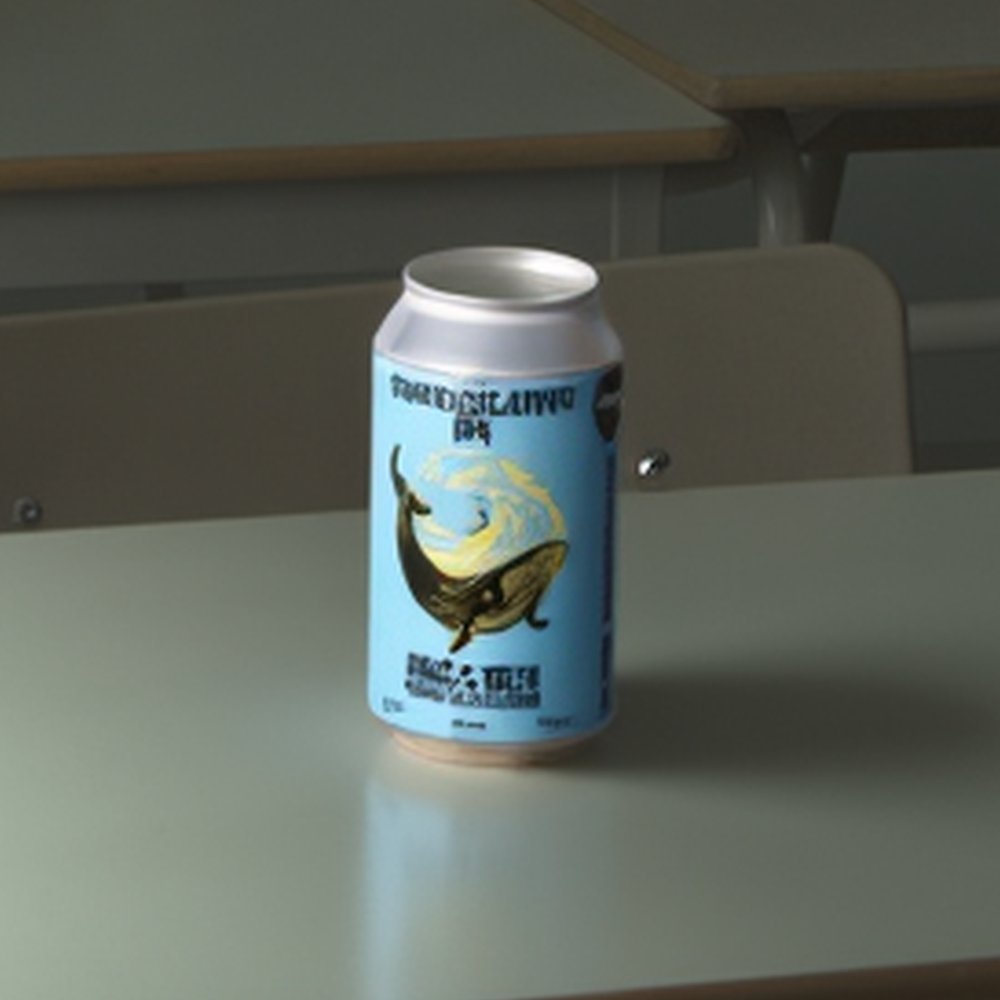} & \includegraphics[width=.090\linewidth]{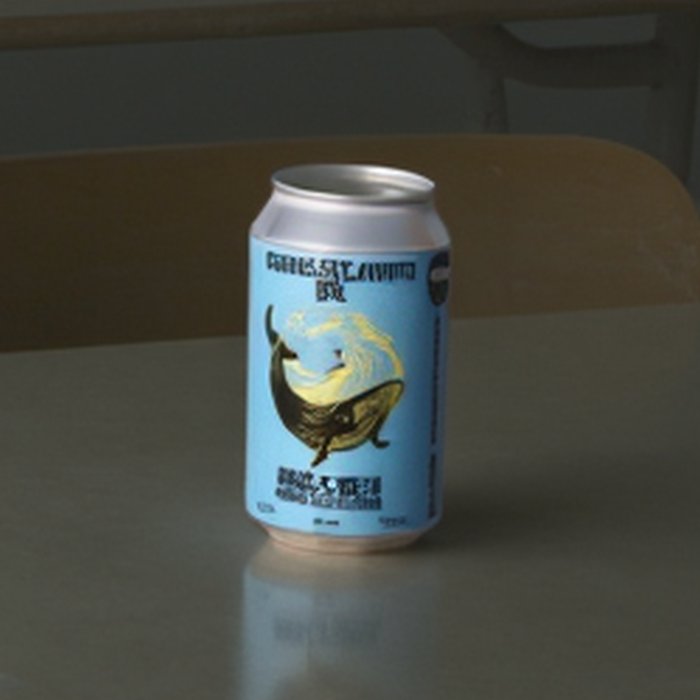} \\
			   \end{tabular}
			 & \begin{tabular}[b]{@{}c@{\hspace{1pt}}c@{\hspace{1pt}}c@{}}
				   \includegraphics[width=.090\linewidth]{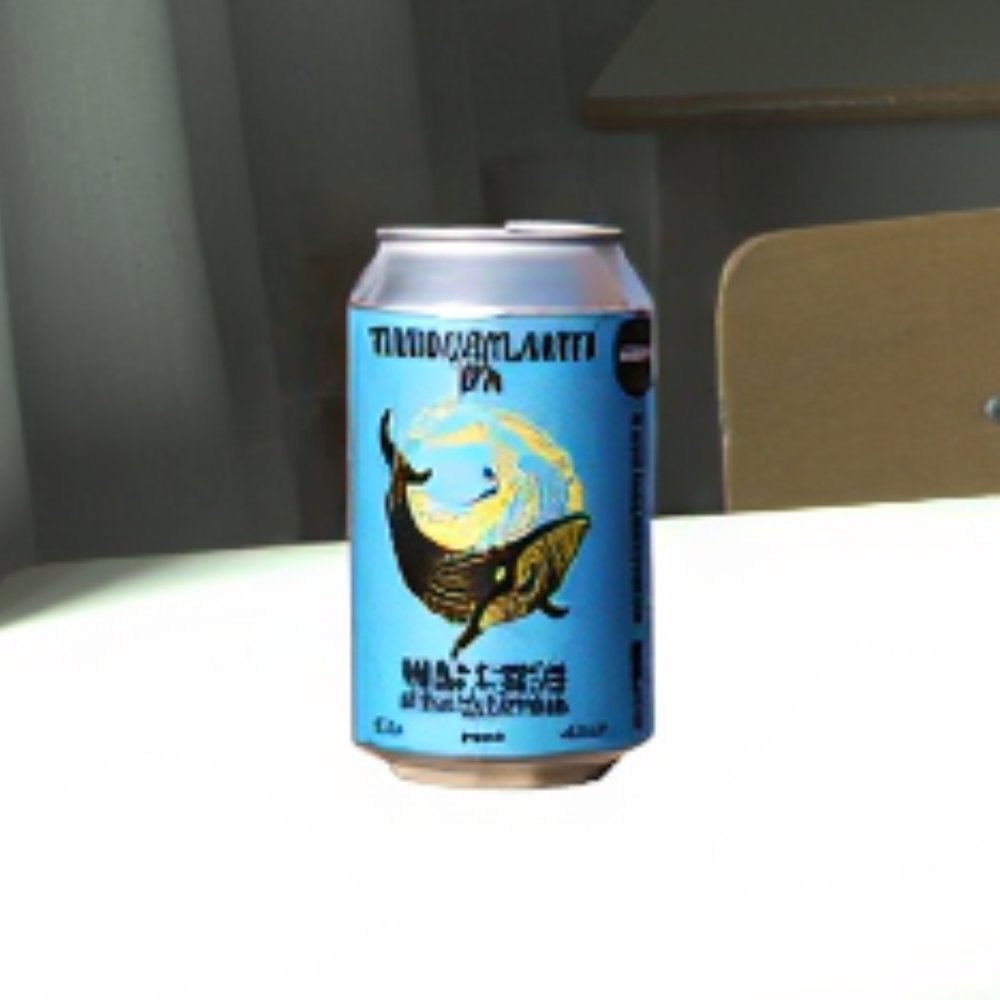} & \includegraphics[width=.090\linewidth]{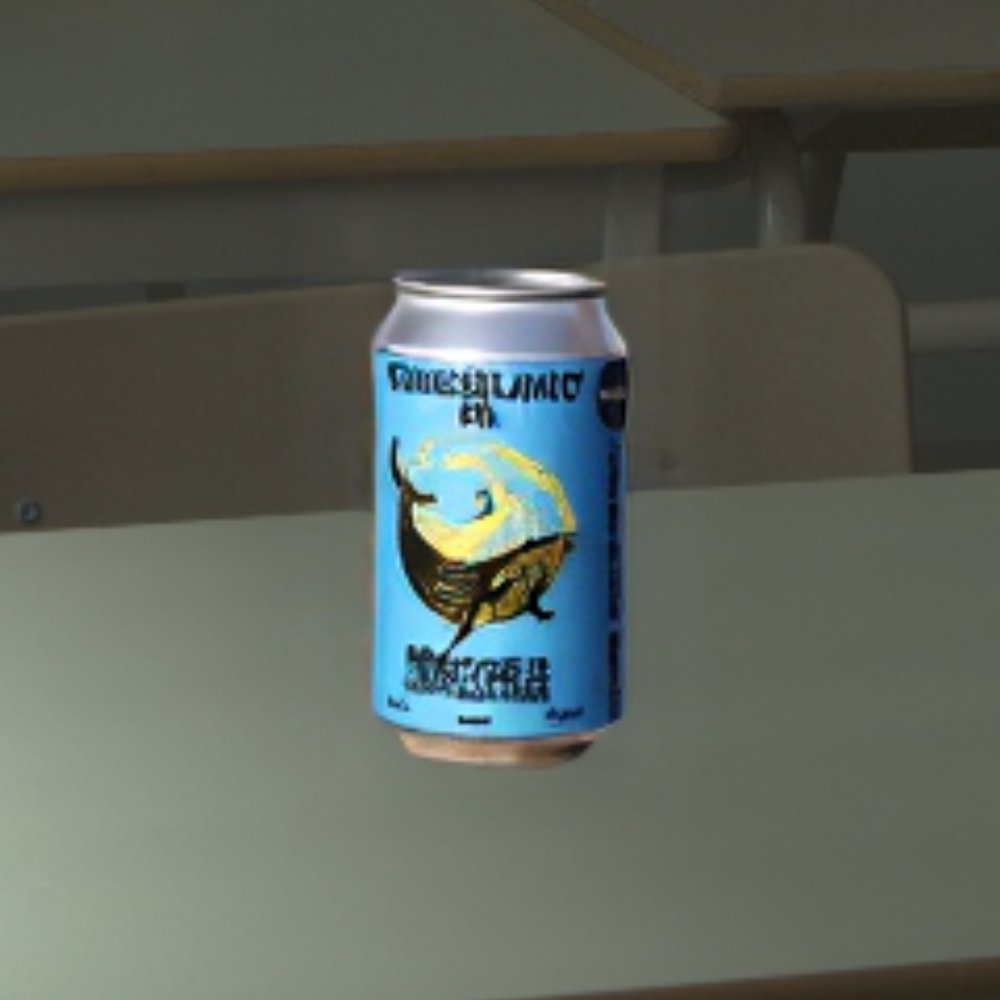} & \includegraphics[width=.090\linewidth]{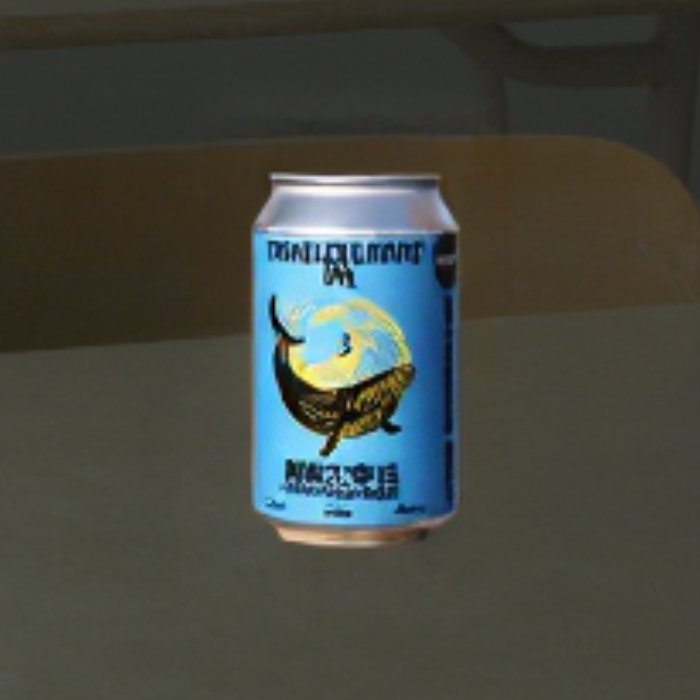} \\
			   \end{tabular} \\
		\end{tabular}%
	}
	\caption{
		\textbf{Same object inserted at three locations within one scene.}
		A DreamBooth~\citep{ruiz2023dreambooth} object is placed at three locations in a Laval Indoor SV HDR~\citep{Garon_2019_CVPR} scene, where local illumination varies across insertion location.
		Across all four existing methods, surface brightness and shadow direction show limited variation across the three locations, indicating that current models respond only weakly to insertion location.
	}

	\label{fig:per_site}
\end{figure}

However, lighting variation in their training data remains scene-wide rather than position-specific within a scene—each training image still carries a single global lighting state, providing no signal for how an inserted object's appearance should depend on its particular location.
Fig.~\ref{fig:per_site} illustrates this limitation: the inserted object's appearance does not adapt appropriately to different placements, even when local illumination differs substantially across locations within the same scene.

\begin{figure}[t]
	\centering
	\scriptsize
	\renewcommand{\arraystretch}{1}
	\begin{tabular}{@{}l@{\hspace{2pt}}*{5}{c@{\hspace{3pt}}}@{}}
		\toprule
		                         &
		\makecell{AnyDoor~\citep{chen2024anydoorzeroshotobjectlevelimage},                                                                                              \\ AnyInsertion~\citep{song2026insert}} &
		\makecell{ORIDa                                                                                                                                                 \\\citep{kim2025oridaobjectcentricrealworldimage}} &
		\makecell{ObjectMover                                                                                                                                           \\\citep{yu2025objectmovergenerativeobjectmovement}} &
		\makecell{DreamFuse                                                                                                                                             \\\citep{huang2025dreamfuseadaptiveimagefusion}} &
		\modelname                                                                                                                                                      \\

		\midrule
		Data acquisition         & \makecell{Segmentation} & \makecell{Photo capture} & \makecell{Game engine} & \makecell{Model generation} & \makecell{Public videos} \\
		Data domain              & Real                    & Real                     & Synthetic              & Synthetic                   & Real                     \\
		On-the-fly light control & \ding{55}               & \ding{55}                & \ding{55}              & \ding{55}                   & \ding{51}                \\
		Human-free labeling      & \ding{51}               & \ding{55}                & \ding{55}              & \ding{55}                   & \ding{51}                \\
		Spatial illumination     & \ding{55}               & \ding{55}                & $\triangle$            & \ding{55}                   & \ding{51}                \\
		\bottomrule
	\end{tabular}
	\vspace{5pt}

	\scriptsize
	\renewcommand{\arraystretch}{0}
	\begin{tabular}{@{}  c@{\hspace{1pt}}*{6}{c@{\hspace{3pt}}} @{}}
		\vspace{2pt}
		                                                                                                                                                          &
		\citep{chen2024anydoorzeroshotobjectlevelimage}                                                                                                           &
		\citep{song2026insert}                                                                                                                                    &
		\citep{kim2025oridaobjectcentricrealworldimage}                                                                                                           &
		\citep{yu2025objectmovergenerativeobjectmovement}                                                                                                         &
		\citep{huang2025dreamfuseadaptiveimagefusion}                                                                                                             &
		\modelname                                                                                                                                                  \\

		\rotatebox[origin=c]{90}{\makecell{B.G.}}                                                                                                                 &
		\raisebox{-0.5\height}{\includegraphics[width=.15\linewidth,height=.15\linewidth,clip,trim=0 0 0 0]{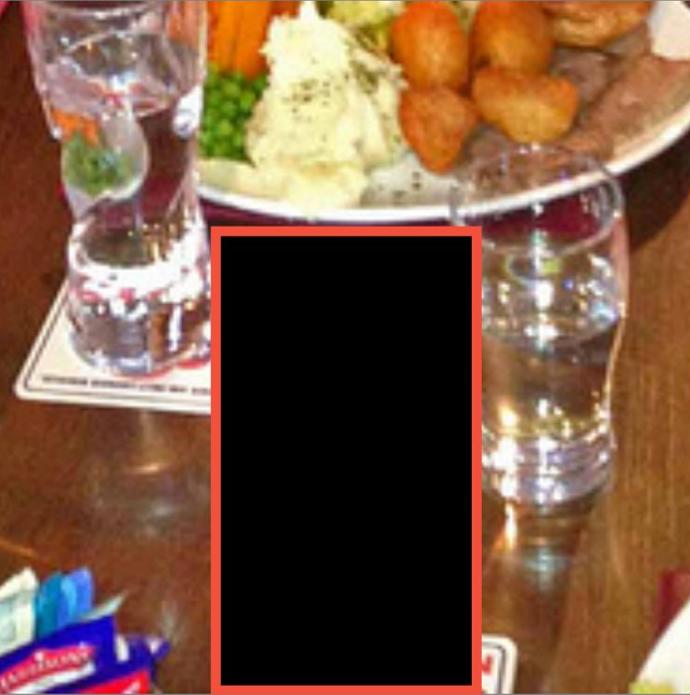}}          &
		\raisebox{-0.5\height}{\includegraphics[width=.15\linewidth,height=.15\linewidth,clip,trim=0 0 0 0]{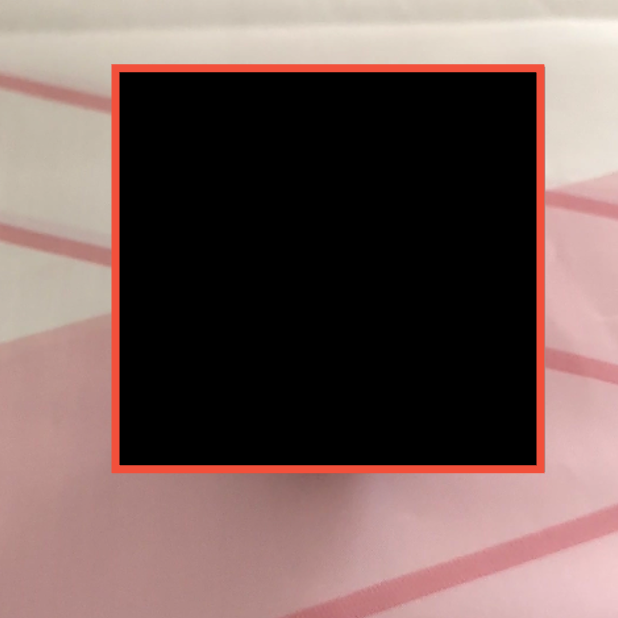}}     &
		\raisebox{-0.5\height}{\includegraphics[width=.15\linewidth,height=.15\linewidth,clip,trim=0 0 0 0]{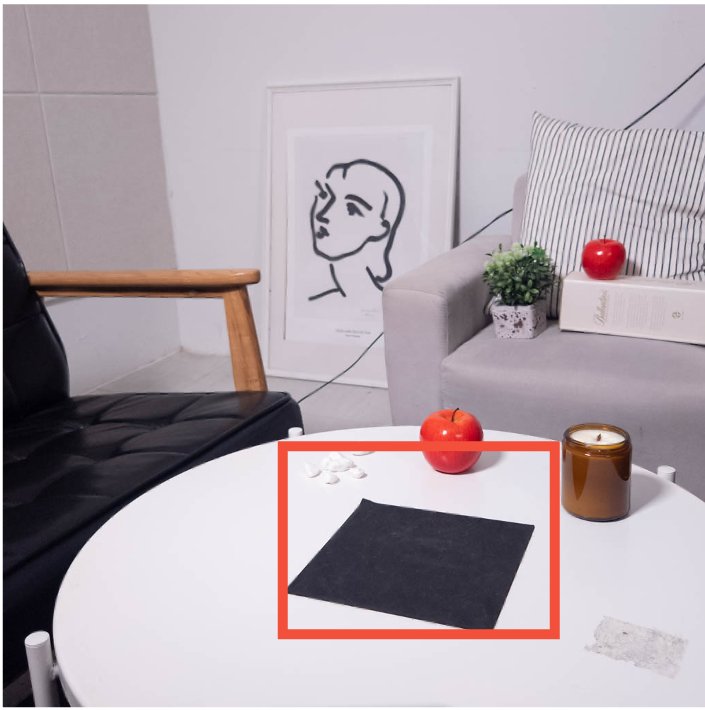}}                           &
		\raisebox{-0.5\height}{\includegraphics[width=.15\linewidth,height=.15\linewidth,clip,trim=0 0 0 0]{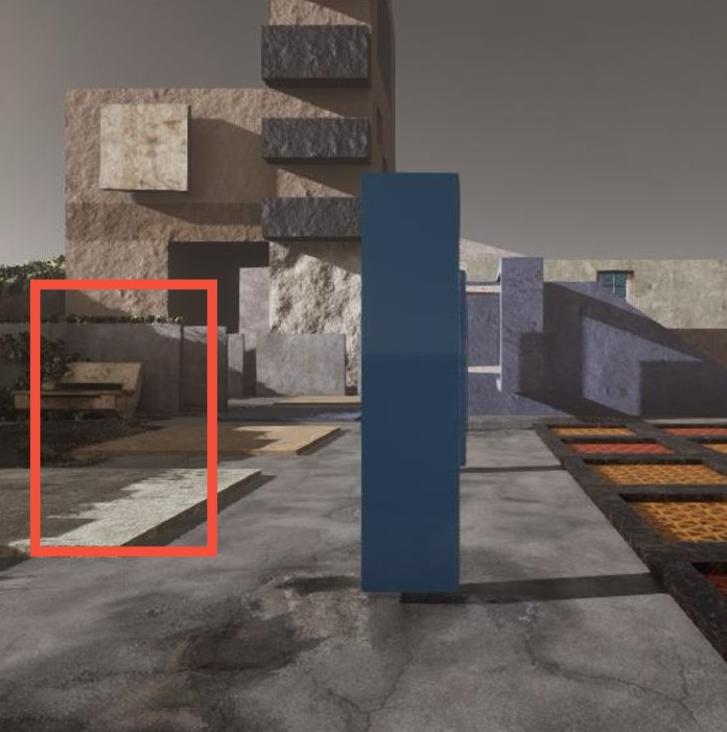}}        &
		\raisebox{-0.5\height}{\includegraphics[width=.15\linewidth,height=.15\linewidth,clip,trim=0 0 0 0]{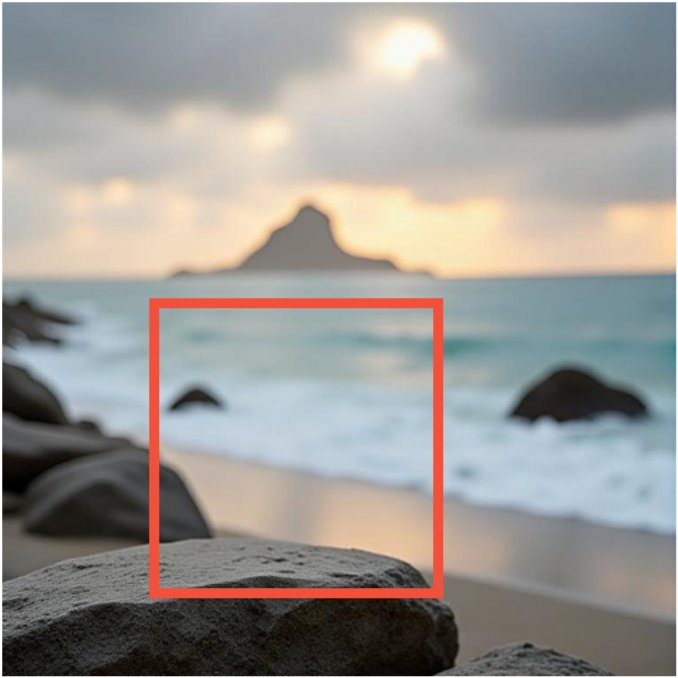}}                       &
		\raisebox{-0.5\height}{\includegraphics[width=.15\linewidth,height=.15\linewidth,clip,trim=0 0 0 0]{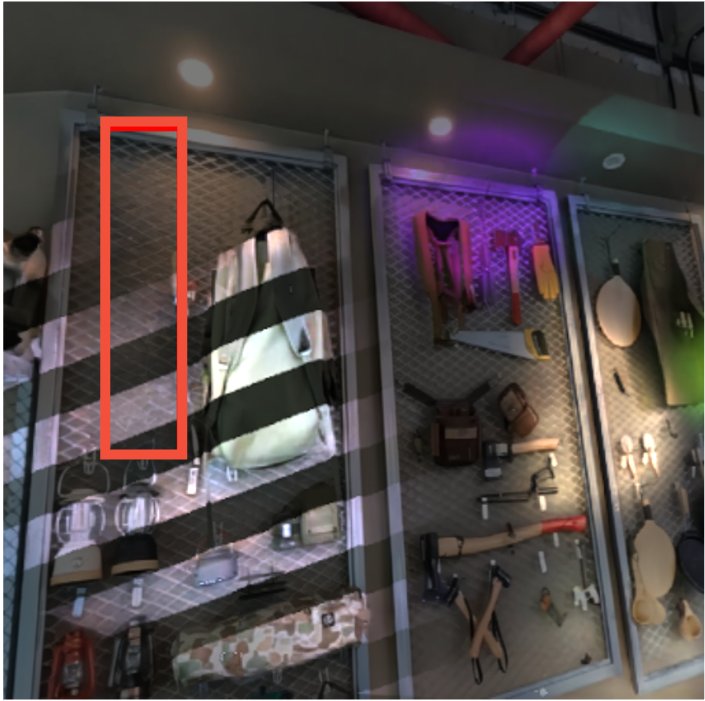}}
		\\

		\rotatebox[origin=c]{90}{\makecell{Object}}                                                                                                               &
		\raisebox{-0.5\height}{\includegraphics[width=.15\linewidth,height=.15\linewidth,clip,trim=0 0 0 0]{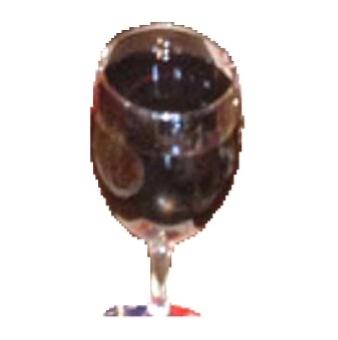}}         &
		\raisebox{-0.5\height}{\includegraphics[width=.15\linewidth,height=.15\linewidth,clip,trim=0 0 0 0]{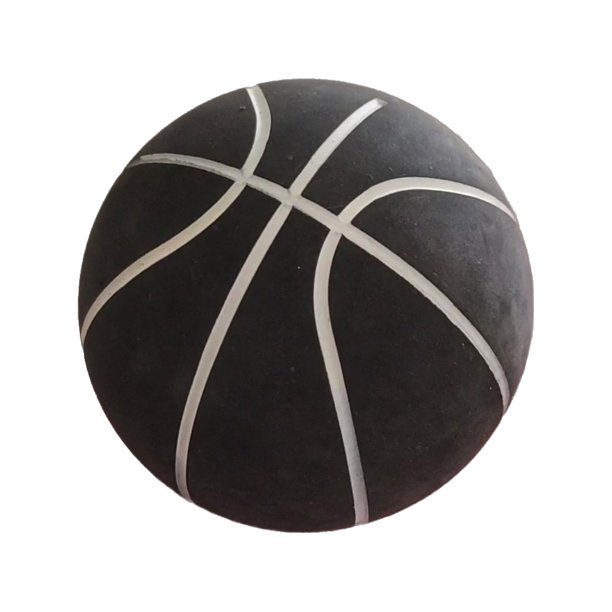}} &
		\raisebox{-0.5\height}{\includegraphics[width=.15\linewidth,height=.15\linewidth,clip,trim=0 0 0 0]{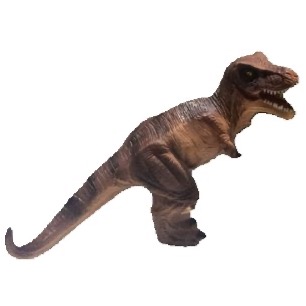}}                          &
		\raisebox{-0.5\height}{\includegraphics[width=.15\linewidth,height=.15\linewidth,clip,trim=0 0 0 0]{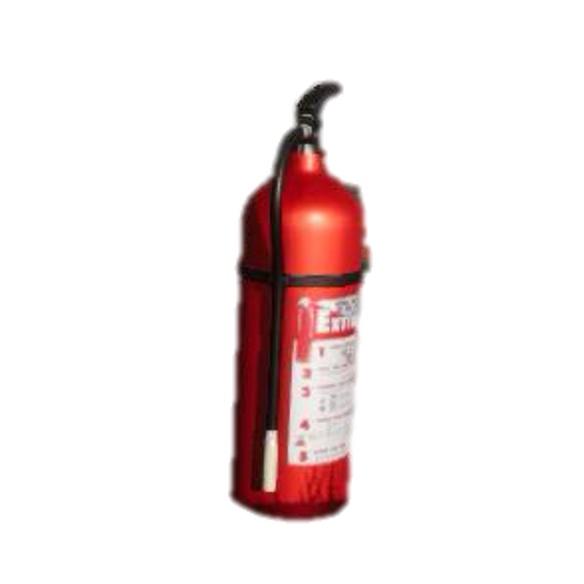}}       &
		\raisebox{-0.5\height}{\includegraphics[width=.15\linewidth,height=.15\linewidth,clip,trim=0 0 0 0]{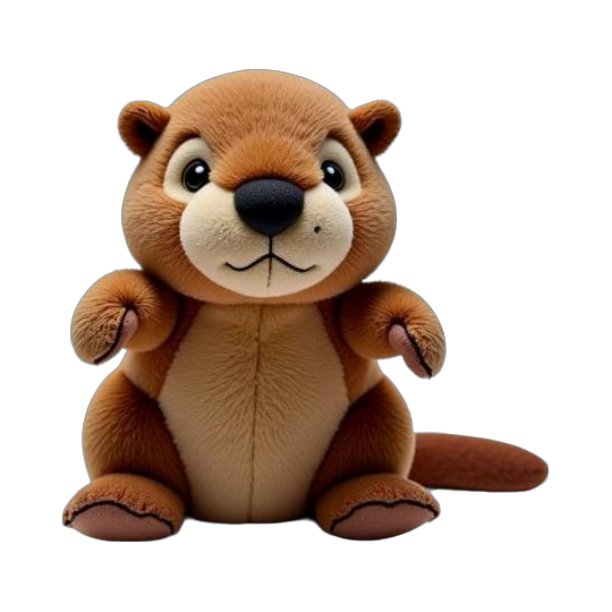}}       &
		\raisebox{-0.5\height}{\includegraphics[width=.15\linewidth,height=.15\linewidth,clip,trim=0 0 0 0]{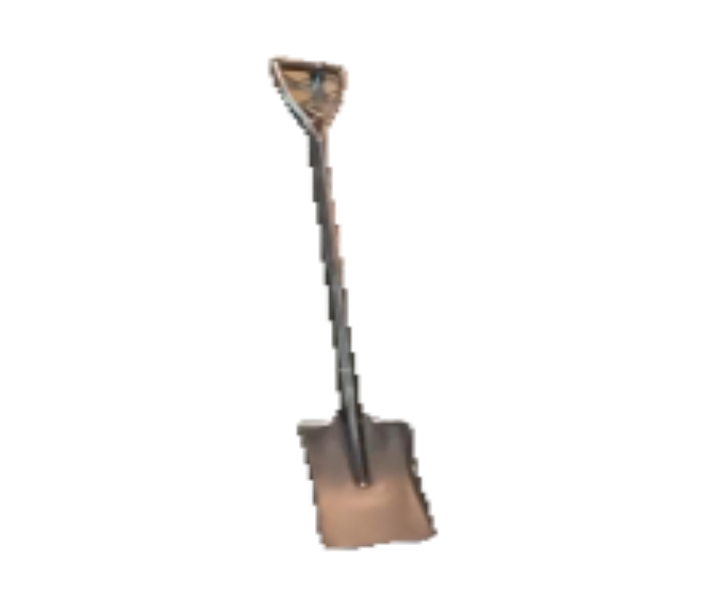}}
		\\

		\rotatebox[origin=c]{90}{\makecell{G.T.}}                                                                                                                 &
		\raisebox{-0.5\height}{\includegraphics[width=.15\linewidth,height=.15\linewidth,clip,trim=0 0 0 0]{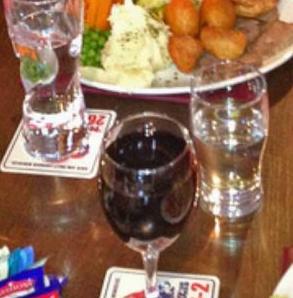}}          &
		\raisebox{-0.5\height}{\includegraphics[width=.15\linewidth,height=.15\linewidth,clip,trim=0 0 0 0]{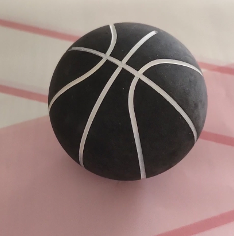}}     &
		\raisebox{-0.5\height}{\includegraphics[width=.15\linewidth,height=.15\linewidth,clip,trim=0 0 0 0]{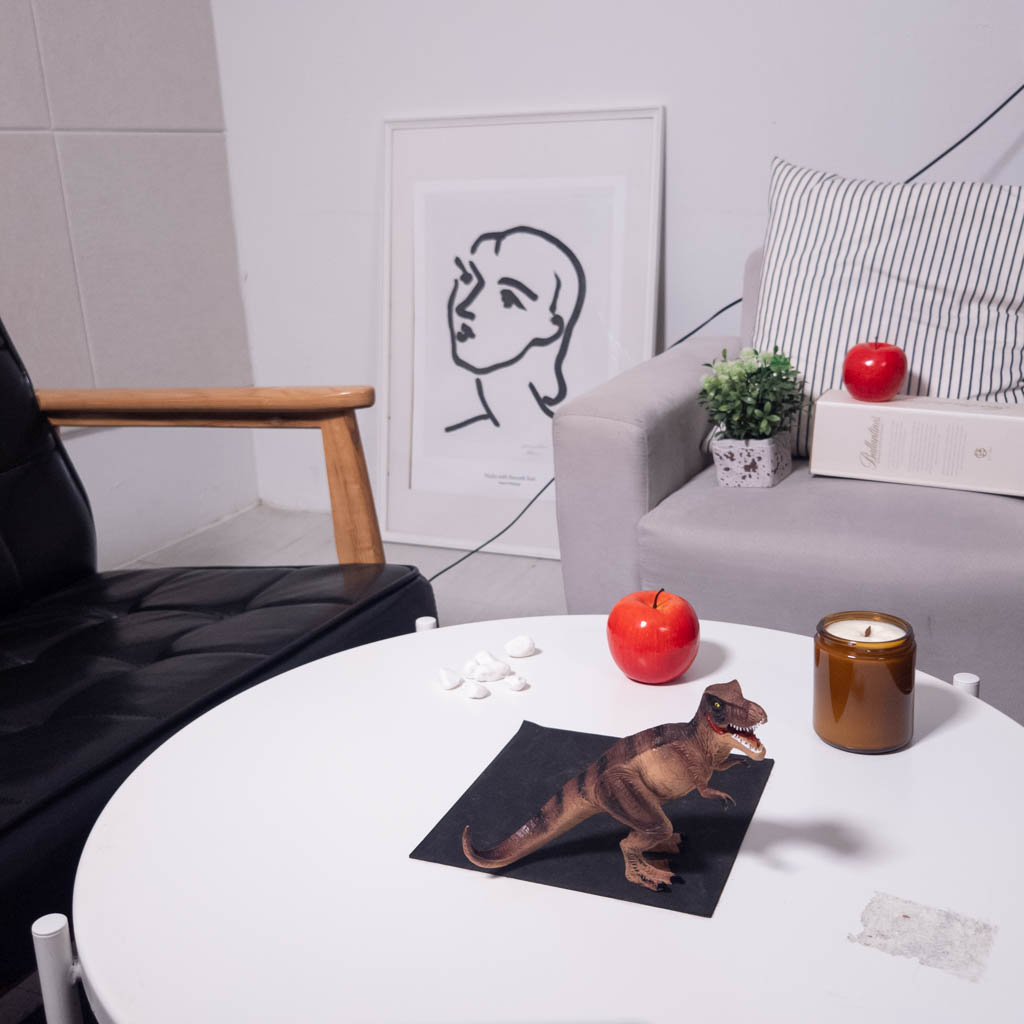}}                           &
		\raisebox{-0.5\height}{\includegraphics[width=.15\linewidth,height=.15\linewidth,clip,trim=0 0 0 0]{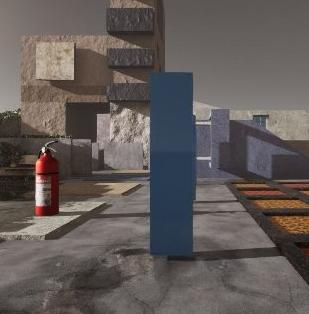}}        &
		\raisebox{-0.5\height}{\includegraphics[width=.15\linewidth,height=.15\linewidth,clip,trim=0 0 0 0]{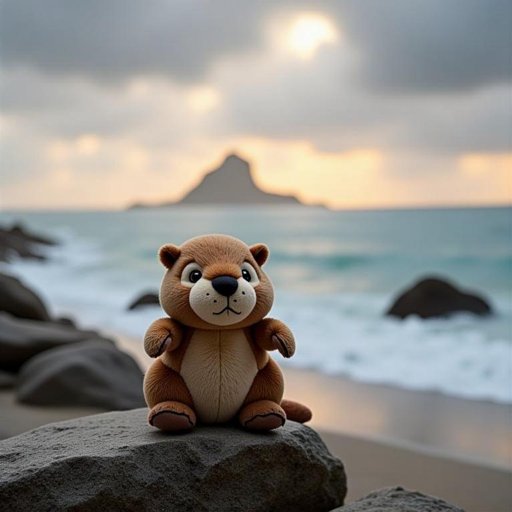}}                        &
		\raisebox{-0.5\height}{\includegraphics[width=.15\linewidth,height=.15\linewidth,clip,trim=0 0 0 0]{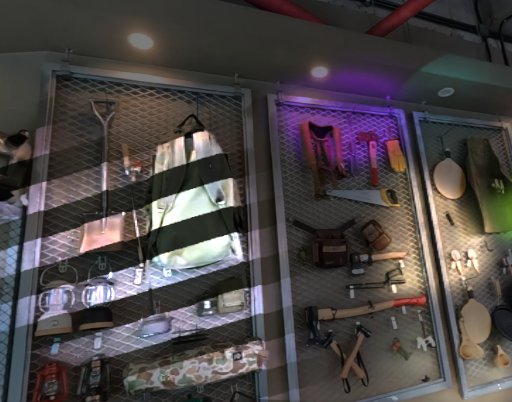}}

	\end{tabular}

	\caption{
		Comparison of data construction methods for object compositing.
		Prior approaches build training data from segmentation datasets or staged scenes, where spatially varying illumination is rarely captured; ObjectMover supports offline control of a global environment light, so its placement-dependent shading is driven primarily by geometric occlusion.
		\modelname\ instead augments real-world videos with online, multi-source spatially varying illumination.
	}
	\label{fig:data}
	\vspace{-10pt}
\end{figure}

The main obstacle for making compositing models lighting-aware is the lack of training data that link object appearance to location-dependent lighting at scale.
Two aspects of current dataset construction pipelines make such data difficult to obtain at scale (Fig.~\ref{fig:data}).

First, capturing structured, spatially varying illumination at training scale is costly. Real-world collection typically requires carefully staged scenes with repeatedly placing the same object across multiple locations, limiting the scale of data that can be collected.
Synthetic alternatives based on game engines~\citep{blender,unrealengine} offer controllable lighting, but their scale is similarly constrained by the cost of authoring complex scenes.
Diffusion-based data generation methods~\citep{huang2025dreamfuseadaptiveimagefusion} reduce manual scene setup, yet offer limited control over illumination and often require substantial human filtering.

Second, commonly used segmentation-derived datasets fail to preserve the local shading context during object insertion.
Datasets such as SA-Co~\citep{carion2026sam3segmentconcepts} and OpenImages~\citep{OpenImages} simulate insertion by masking foreground regions and filling the resulting holes.
This process removes local shading cues, including adjacent surfaces and ground-plane shading that reveal the lighting at the insertion location.

To overcome these limitations, we introduce \modelname, a physically grounded data augmentation framework for \textbf{LO}cal light-aware object \textbf{CO}mposition that leverages monocular videos as illumination sources.
Our idea is to reinterpret each video frame as a pose-anchored virtual spotlight, so that camera motion yields diverse light directions.
With controllable beam shape, color, and intensity, this design produces placement-dependent illumination via physically based rendering (PBR) over the recovered scene geometry.
Since illumination is generated on-the-fly rather than precomputed, the pipeline scales naturally as data augmentation during training.
The pipeline can be applied to any monocular video of a static scene, and we instantiate it on publicly available datasets (uCO3D~\citep{liu24uco3d}, DL3DV-10K~\citep{ling2024dl3dv}, Mip-NeRF 360~\citep{360mipnerf2022}, Tanks \& Temples~\citep{Knapitsch2017}) without staging scenes for spatially varying illumination.

We evaluate local light adaptation across multiple settings.
First, we use the Laval Indoor Spatially Varying HDR dataset~\citep{Garon_2019_CVPR}, widely used for spatially varying light estimation. Since Laval is not built for compositing, it provides no ground-truth composites; we therefore evaluate via a user study and a reference-free lighting-consistency measure.
Second, for objective evaluation with ground truth, we collect \benchname, an indoor benchmark of 760 physically captured composites in which each object is placed at multiple locations within a scene under spatially varying illumination.
Third, we evaluate object compositing on the ORIDa benchmark~\citep{kim2025oridaobjectcentricrealworldimage} to verify that improved illumination adaptation preserves object identity.
Across all settings, \modelname{} improves local illumination adaptation while preserving object identity.

Our main contributions are as follows:
\begin{enumerate}
	\item We study object compositing under spatially varying illumination, a setting less explored in prior generative compositing work.
	\item We present a physically grounded illumination-augmentation pipeline that turns real-world monocular videos into large-scale training data for local light-aware compositing.
	\item We collect \benchname, a benchmark of physically captured relit composites for objective evaluation under spatially varying illumination.
\end{enumerate}

\section{Related Work}
\label{sec:related}
\textbf{Generative Object Compositing}.
Recent diffusion-based approaches have advanced object compositing by improving spatial alignment~\citep{Song_2023_CVPR, chen2024anydoorzeroshotobjectlevelimage, yu2025objectmovergenerativeobjectmovement, tarres2025multitwinemultiobjectcompositingtext, winter2025objectmaterecurrencepriorobject}, color harmonization~\citep{wang2023semi, svindoorharmonization}, and plausibility cues such as shadows and reflections~\citep{ValencaZGHL23, winter2024objectdropbootstrappingcounterfactualsphotorealistic, hong2022shadowgenerationcompositeimage, liu2024shadowgenerationcompositeimage, tarres2024thinkingoutsidebboxunconstrained, wang2025metashadowobjectcenteredshadowdetection}.
However, most methods implicitly assume spatially uniform illumination, offering no explicit supervision for local lighting adaptation.
Several works introduce lighting variation in training data via synthetic 3D environments or game engines~\citep{michel2024object3ditlanguageguided3daware, wang2025metashadowobjectcenteredshadowdetection}, but these require costly scene authoring and operate with simplified lighting models. ObjectMover~\citep{yu2025objectmovergenerativeobjectmovement} uses an environment lighting setup with offline illumination control, limiting its ability to model diverse localized effects.
Other approaches address spatially varying illumination in specialized settings; SV Indoor Harmonization~\citep{svindoorharmonization} targets human subjects using 3D scans but does not generalize to arbitrary objects.
In contrast, our method targets general object compositing under spatially varying illumination through on-the-fly illumination augmentation from monocular videos, without staged capture or synthetic scene authoring.

\noindent
\textbf{Spatially-Varying Light Estimation}.
Another body of work explicitly estimates scene illumination. Early methods recover per-pixel illumination from single images~\citep{zhu2022irisformerdensevisiontransformers,physically2022,li2020inverse}, while recent approaches combine illumination estimation with differentiable rendering to jointly recover geometry, BRDFs, and lighting~\citep{zhu2022learning,wang2021learning,peng2025interactiveobjectinsertion,wang2022neurallightfieldestimation,bolduc2025GaSLight}.
These methods target physical scene reconstruction rather than object compositing, and typically rely on either multi-view inputs~\citep{li2023spatiotemporallyconsistenthdr,srinivasan2020lighthousepredictinglightingvolumes,fipt2023} or explicit 3D object assets~\citep{zhu2021spatiallyvaryingoutdoorlightingestimation,bai2023localglobalpanoramainpainting,liang2024photorealisticobjectinsertiondiffusionguided,wang2022neurallightfieldestimation}, limiting their use in practical editing scenarios.
Rather than estimating lighting explicitly, our method learns to adapt object appearance to local illumination directly in image space, enabling compositing from a single background image and an object mask.

\noindent
\textbf{Intrinsic-Based Editing}.
A third paradigm tackles compositing through disentangled intrinsic spaces such as albedo--shading or reflectance representations~\citep{zeng2024rgb, chen2024intrinsicanythinglearningdiffusionpriors, Zhang2024Latent, intrinsicdiffusion2024, kocsis2024intrinsicimagediffusionindoor}.
Compositing here follows a two-stage pipeline: inverse-rendering models estimate intrinsic representations, which are then rendered back to RGB via forward rendering~\citep{careagaCompositing, zhang2025zerocomp, DiffusionRenderer}.
However, single-image inverse and forward rendering are each ill-posed, and chaining them accumulates errors~\citep{Lyu_2025, DiffusionRenderer}.
These methods also assume geometric compatibility between object and scene, limiting applicability to constrained settings such as planar placement or object cloning.
\modelname\ instead operates directly in RGB space without intrinsic decomposition, enabling end-to-end compositing under spatially varying illumination for arbitrary object poses and geometry.

\section{Proposed Method: \modelname}
\label{sec:method}
As illustrated in Fig. \ref{fig:data}, the goal of object compositing is, given a background image $I_{\text{bg}}$, an insertion mask $M$, and an object image $I_{\text{obj}}$,
to synthesize a composite image $I_{\text{comp}}$ whose object appearance matches the local illumination at the insertion region.
We formulate this as a conditional diffusion problem with the rectified-flow objective~\citep{liu2023flowstraightfastlearning,lipman2023flowmatchinggenerativemodeling}.
Our \modelname framework synthesizes illumination-augmented training data from monocular videos via physically based rendering (PBR), diversifying local lighting scenarios at training time.

\begin{figure}[t]
	\centering
	\scriptsize
	\includegraphics[width=\linewidth]{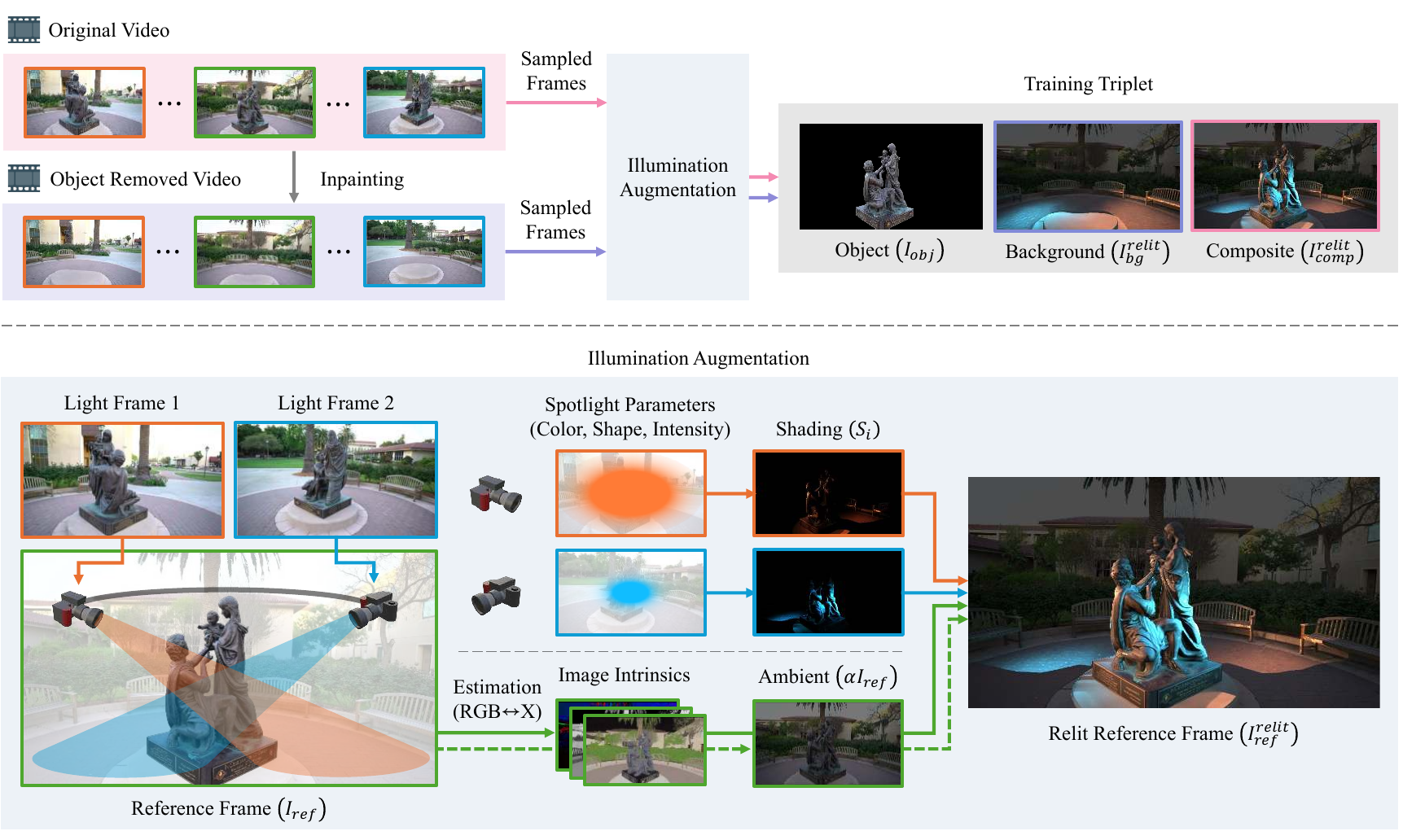}
	\caption{
		The \modelname pipeline.
		(Top) From a monocular video, we generate training triplets via illumination augmentation.
		(Bottom)
		Frames from the same video are treated as light views and instantiated as virtual spotlights with randomized color, shape, and intensity. Using estimated scene geometry and intrinsics, each light view projects spatially varying illumination onto a reference view via physically based relighting with complex lighting effects. This process enables efficient on-the-fly illumination augmentation.
	}
	\label{fig:pipeline}
	\vspace{-10pt}
\end{figure}

\subsection{The Illumination Augmentation Pipeline}
\label{sec:augmentation}

Fig.~\ref{fig:pipeline} illustrates the proposed illumination augmentation pipeline, which performs \emph{scene preparation} offline and applies \emph{relighting} on the fly during augmentation.

\noindent
\textbf{Scene Preparation.}
Our pipeline applies to any monocular video of a static scene with sufficient viewpoint coverage.%
We use publicly available large-scale video datasets—uCO3D~\citep{liu24uco3d}, DL3DV-10K~\citep{ling2024dl3dv}, Mip-NeRF 360~\citep{360mipnerf2022}, and Tanks \& Temples~\citep{Knapitsch2017}—as primary data sources, supplemented by casually captured videos of everyday scenes.
Foreground objects are segmented using Detic~\citep{zhou2022detecting} and SAM2~\citep{ravi2025sam2segmentimages}, then removed with RoRem~\citep{li2024RORem} to obtain clean backgrounds.
For relighting, we estimate and cache approximate scene geometry and reflectance parameters, including camera poses~\citep{colmap}, depth~\citep{bochkovskii2025depthprosharpmonocular,mast3r_eccv24}, BRDF~\citep{zeng2024rgb}, and original scene illumination~\citep{li2020inverse,Phongthawee2024DiffusionLight}.
These quantities are computed once offline and reused throughout the augmentation.
The filtering pipeline for these estimates is detailed in Appendix~\ref{app:data_filtering}.

Each processed video yields training triplets
$(I_{\text{bg}}, I_{\text{obj}}, I_{\text{comp}})$,
where $I_{\text{comp}}$ is a video frame,
$I_{\text{obj}}$ denotes the foreground segmented from $I_{\text{comp}}$,
and $I_{\text{bg}}$ is the object-removed background of the same frame.
We uniformly sample up to 100 frames per scene, filter low-quality objects, and obtain ${\sim}$24K relightable objects, each observed from ${\sim}$28 viewpoints, yielding ${\sim}$671K unaugmented training triplets.
At training time, we apply on-the-fly relighting and the permutation process described below.

\noindent
\textbf{Relighting.}
We treat illumination as a dynamically resampled augmentation variable applied to both $I_{\text{bg}}$ and $I_{\text{comp}}$, ensuring that the input background and the supervision target share the same lighting configuration.
To induce structured variation, we reinterpret other frames from the same video as pose-anchored \emph{virtual spotlights}.
Given reference frame $I_{\text{ref}}$, we sample $N$ frames whose camera poses define candidate light positions, so that camera motion naturally yields diverse lighting directions.
Since these poses lie in free space, face the scene, and are registered to the geometry, they enable occlusion-aware shading and shadows without manual light placement.
We randomly perturb spotlight parameters such as beam shape, color, and intensity to produce diverse placement-dependent illumination patterns.
Examples of relit realizations from a single reference frame are shown in Appendix Fig.~\ref{fig:relight_diversity}.
Parameter ranges and sampling details are provided in Appendix~\ref{app:implementation}.

To render illumination consistently with scene geometry, we employ rasterization-based shadow mapping.
Given the reference depth map $D_{\text{ref}}$, each pixel $x$ is back-projected to 3D and reprojected into the $i$-th light frame:
\begin{equation}
	p(x) = o + D_{\text{ref}}(x)\,\omega_o(x), \quad
	\tilde{x}_i = K_i [R_i \mid t_i]\,[p(x), 1]^\top ,
\end{equation}
where $o$ is the reference camera center, $\omega_o$ the viewing direction, and $(K_i, R_i, t_i)$ the light-frame camera parameters.
Visibility is determined via depth comparison,
$V_i(x) = \mathbf{1}(D_i(\tilde{x}_i) \ge \|p(x) - o_i\|_2)$, enabling occlusion-aware shadow casting.

For each sampled light, we compute the incident radiance and resulting shading:
\begin{align}
	\label{eq:light_shading}
	L_i(x,\omega_i) = \lambda_i\,V_i(x)\,M_i(\tilde{x}_i)\,\phi(x), \quad
	S_i(x) = L_i(x,\omega_i)\,R(x,\omega_o,\omega_i),
\end{align}
where $\lambda_i$ controls intensity and color, $M_i$ denotes the beam shape mask, $\phi(x)$ models distance falloff, and $R$ is the simplified Disney BRDF~\citep{DisneyBRDF}.

Following prior work~\citep{DiffusionRenderer, liang2024photorealisticobjectinsertiondiffusionguided, wang2022neurallightfieldestimation},
we adopt a ratio-based illumination transfer, where each light contributes a multiplicative shading ratio to the reference image.
The final relit image is obtained by summing these per-light contributions, with a small ambient term to prevent excessive darkening:
\begin{equation}
	\label{eq:illumination_transfer}
	I^{\text{relit}}_{\text{ref}}(x) =
	\sum_{i=1}^{N} E_i(x)\, I_{\text{ref}}(x) + \alpha I_{\text{ref}}(x),
	\quad
	\text{where }
	E_i(x) = \frac{S_i(x)}{S_{\text{orig}}(x)}.
\end{equation}
Here, $S_{\text{orig}}(x)$ denotes the shading under the original scene illumination, and $\alpha$ is an ambient strength.
$S_{\text{orig}}$ is independent of the sampled light configurations and therefore precomputed offline.
Relighting requires only rasterization (depth reprojection and per-pixel shading), adding just 1.5$\%$ to the training step time (Appendix~\ref{app:data_analysis}).

\begin{figure}[t]
	\centering
	\scriptsize
	\includegraphics[width=\linewidth]{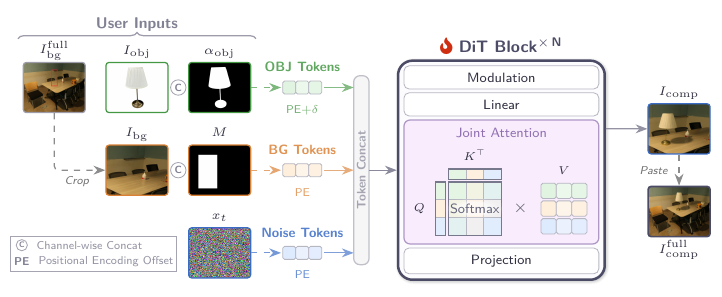}
	\vspace{-10pt}
	\caption{
		Conditioning layout.
		The object and the background crop, each channel-concatenated with its mask, are tokenized and processed jointly with the noisy latent $x_t$.
		$\alpha_{\text{obj}}$: object alpha matte; $M$: insertion mask; PE: positional encoding, offset by $\delta$ on object tokens.
	}
	\vspace{-10pt}
	\label{fig:model_arc}
\end{figure}

\begin{wrapfigure}{r}{0.5\linewidth}
	\vspace{-10pt}
	\centering
	\scriptsize
	\begin{tabular}{@{} c@{\hspace{1pt}} *{3}{c@{\hspace{2pt}}} @{}}
		                                                                                                     & B.G. \& BBox & Object & G.T. \\
		\rotatebox[origin=c]{90}{\makecell{Raw Triplet                                                                                      \\ (Before Aug.)}}                                                                                      &
		\raisebox{-0.5\height}{\includegraphics[width=.27\linewidth]{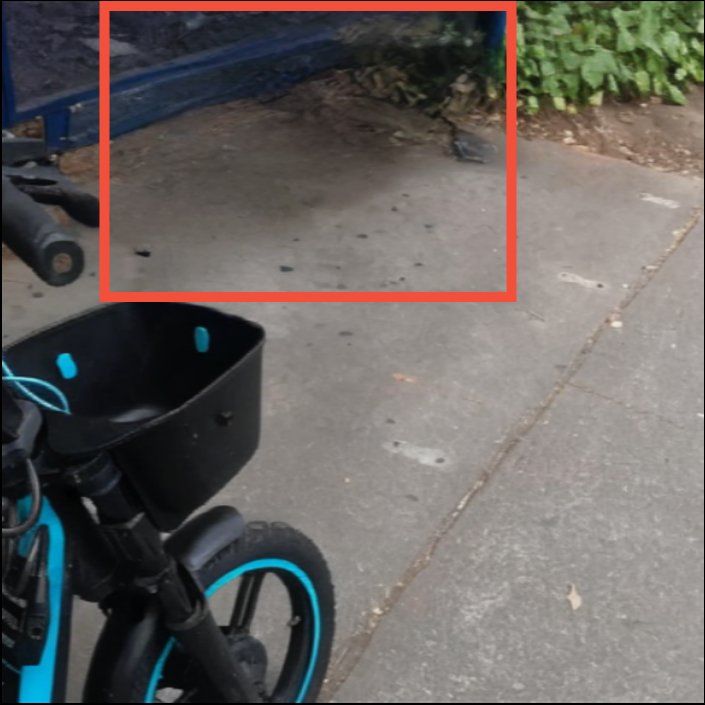}}  &
		\raisebox{-0.5\height}{\includegraphics[width=.27\linewidth]{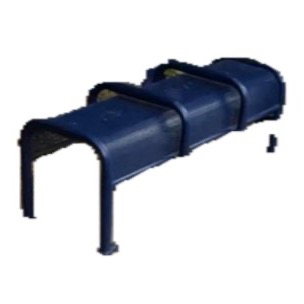}} &
		\raisebox{-0.5\height}{\includegraphics[width=.27\linewidth]{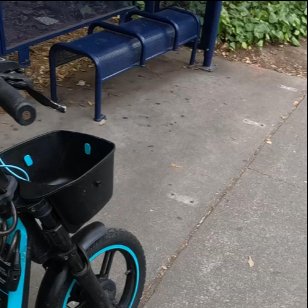}}                                 \\
		\rotatebox[origin=c]{90}{Illum. Aug.}                                                                &
		\raisebox{-0.5\height}{\includegraphics[width=.27\linewidth]{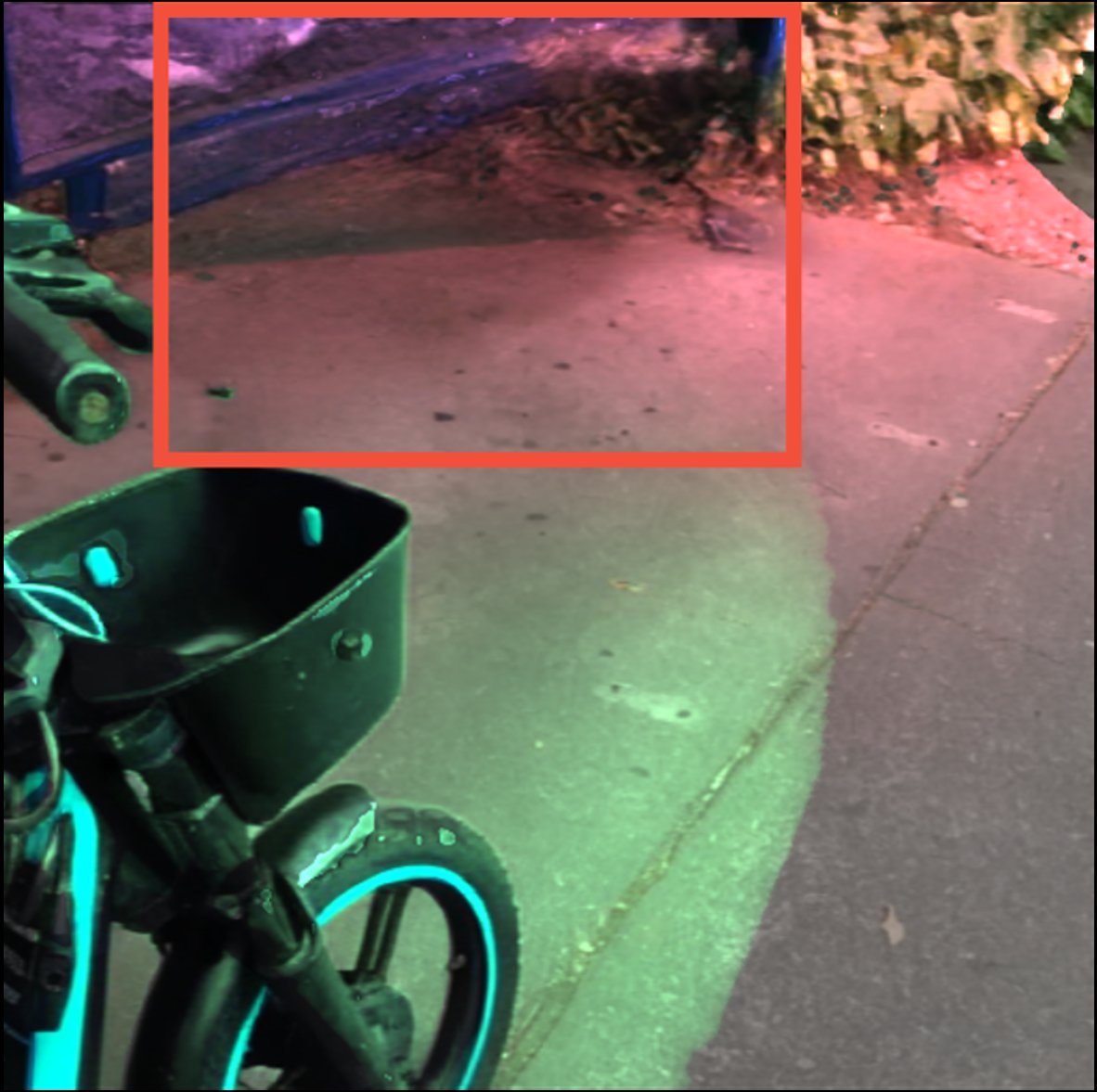}}     &
		\raisebox{-0.5\height}{\includegraphics[width=.27\linewidth]{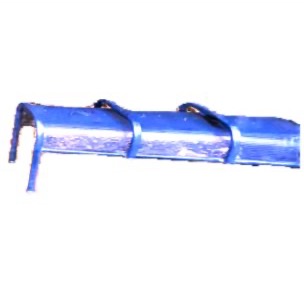}}    &
		\raisebox{-0.5\height}{\includegraphics[width=.27\linewidth]{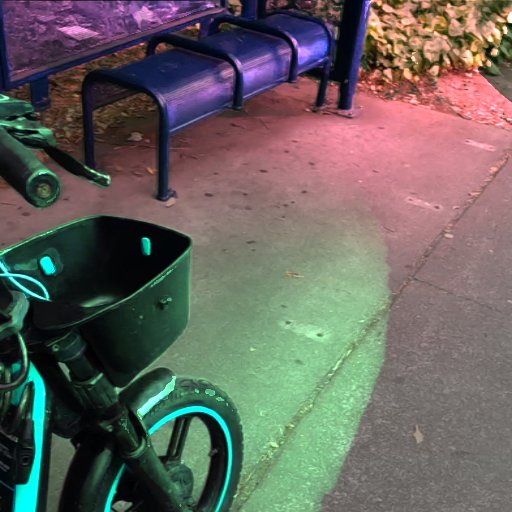}}                                    \\
	\end{tabular}
	\caption{
		Training triplet example.
		The raw triplet shown (top) lacks spatially varying illumination and provides little supervision for adapting the object's appearance to the local lighting at the insertion location.
		Illumination augmentation (bottom) relights the scene under a sampled spotlight and pairs the relit background with an object from a different view and lighting.
	}
	\label{fig:aug}
	\vspace{-8pt}
\end{wrapfigure}
\noindent
\textbf{Light and view batch permutation.}
Within each mini-batch, we recombine objects and backgrounds across light and view dimensions.
For each scene, we sample $V$ reference frames, each producing an original and a relit $(I_{\text{bg}}, I_{\text{comp}})$ pair, yielding $2V$ in total.
Training triplets are formed by pairing one of these pairs with a foreground $I_{\text{obj}}$ segmented from any of the $2V$ composites, allowing the objects and backgrounds to come from different views or lighting conditions.
This expands the per-scene triplet count from $V$ to $(2V)^2$; Fig.~\ref{fig:aug} shows a resulting triplet, where the object comes from a different view and lighting than the background.
Such mismatched lighting discourages the model from simply copying the object's input shading. Instead, the background provides local lighting cues, while the corresponding relit composite serves as supervision for adapting the object's appearance to those conditions.

\noindent
\textbf{Conditioning and training objective.}
\label{sec:training_strategies}
We condition the diffusion model $f_\theta$ on $c = \{I_{\text{obj}}, I_{\text{bg}}, M\}$: the object image, a background crop around the insertion region, and their associated masks.
Following prior compositing work~\citep{chen2024anydoorzeroshotobjectlevelimage,song2026insert}, we define $I_{\text{bg}}$ as a square crop enclosing the insertion bbox, with side length proportional to the bbox size.
Each image is encoded by a VAE~\citep{kingma2014auto}; the corresponding mask is bilinearly resized to the latent grid, concatenated channel-wise with the latent, and projected to DiT token dimensions.
All tokens are concatenated into a single sequence and processed by joint attention in the DiT backbone.
To preserve spatial alignment between $x_t$ and the background while keeping the object distinguishable, we share positional embeddings between $x_t$ and the background and offset the object by a fixed $\delta$, following OmniControl~\citep{tan2025ominicontrol}.
The model is supervised by the rectified-flow velocity loss
\begin{equation}
	\mathcal{L}_{\text{diff}}
	= \mathbb{E}_{t,\epsilon}\!\left[
		\|f_\theta(x_t, c, t) - v\|_2^2
		\right],
\end{equation}
where $x_t = (1{-}t)\, x_0 + t\, \epsilon$, $v = \epsilon - x_0$ is the velocity target, and $x_0$, $\epsilon$, and $t$ denote the ground-truth latent, Gaussian noise, and the time variable, respectively.

\section{Experiments}
\label{sec:experiments}

\subsection{Experimental Setup}
\noindent
\textbf{Benchmarks.}
We evaluate \modelname across multiple settings, spanning a conventional compositing benchmark, a light-estimation benchmark, and our newly collected relit-composite benchmark.
(1) \textit{Conventional object compositing}:
The ORIDa evaluation set~\citep{kim2025oridaobjectcentricrealworldimage} provides real-world object--background pairs with ground-truth composites.
As these scenes lack spatially varying illumination, they serve to evaluate general compositing capabilities such as identity preservation.
(2) \textit{Scenes with spatially varying illumination}:
We insert 30 DreamBooth objects~\citep{ruiz2023dreambooth} into 20 scenes from Laval Indoor SV HDR dataset~\citep{Garon_2019_CVPR} at 3--5 positions per scene corresponding to the dataset's mirror-ball capture points, yielding 2{,}280 (object, background) pairs.
Since ground-truth composites are unavailable, we evaluate via a user study and measure local lighting agreement before and after compositing.
(3) \textit{\benchname}:
We introduce \benchname, a benchmark of 760 triplets (object image, background, ground-truth composite) under controlled spatially varying illumination, collected from 62 objects across 49 scenes.
Backgrounds and ground-truth composites are physically
captured from fixed viewpoints, while object references
are photographed under different lighting conditions.
Each object is placed at multiple locations per scene
(up to 8, median 3), enabling direct evaluation of
location-dependent appearance adaptation;
construction details are in Appendix~\ref{app:dataset}.

\noindent
\textbf{Metrics.}
For identity preservation on ORIDa, we report CLIP~\citep{clip} and DINO~\citep{simeoni2025dinov3} similarity to the ground-truth composite, following \citet{chen2024anydoorzeroshotobjectlevelimage,wang2025unicombine,yu2025omnipaintmasteringobjectorientedediting}.
On \benchname{}, we use DreamSim distance~\citep{fu2023dreamsim} to the ground truth as the primary perceptual metric, capturing both identity and appearance (unlike DINO; see Appendix~\ref{app:dreamsim}).
For Laval, which lacks ground-truth composites, we design a reference-free illumination consistency (IC) metric: the PSNR between estimated local lighting before and after compositing using a pretrained per-pixel lighting estimator~\citep{li2020inverse}.
Computation details and validation of the metric are provided in Appendix~\ref{app:icmetric}.
On \benchname{}, to compare more directly against the captured composite, we additionally measure, inside the insertion box, the light direction implied by the estimator's environment maps (L2 and angular error of the intensity-weighted mean direction), and the brightness, hue, and saturation of the bbox-mean color in HSV. %
Note that IC is reference-free, whereas these measurements compare against the ground truth.
We also conduct a pairwise user study comparing \modelname against each baseline.
Participants rate three aspects: \emph{harmony} (whether the inserted object's shading looks natural in the scene), \emph{identity} (whether the object's distinctive features are preserved), and \emph{quality} (overall compositing quality).
On Laval, judgments are made without a reference; on \benchname, the ground-truth composite is shown alongside.
Each participant evaluates 10 trials per dataset, with 38 valid participants on \benchname{} and 33 on Laval. Further study details are provided in Appendix~\ref{app:user_study} (Fig.~\ref{fig:user_study_ui}).

\noindent
\textbf{Baselines.}
We compare against open-sourced state-of-the-art \textit{generative compositing} methods: AnyDoor~\citep{chen2024anydoorzeroshotobjectlevelimage}, Personalize Anything~\citep{feng2025personalize}, Insert-Anything~\citep{song2026insert}, Unicombine~\citep{wang2025unicombine}, DreamFuse~\citep{huang2025dreamfuseadaptiveimagefusion}, OmniPaint~\citep{yu2025omnipaintmasteringobjectorientedediting}, and FreeCompose~\citep{FreeCompose}.
For a thorough comparison, we evaluate each baseline with four background crop sizes ($1.5\times$, $3.0\times$, $4.5\times$, and $6.0\times$ the insertion mask) and report its best-performing configuration; per-scale results are in Appendix~\ref{app:baseline_analysis}.
We provide additional comparisons with harmonization~\citep{wang2023semi} and intrinsic-based editing methods~\citep{zeng2024rgb,zhang2025zerocomp,DiffusionRenderer} in Appendix~\ref{app:baseline_analysis}.

\begin{figure}[t]
	\centering
	\scriptsize
	\renewcommand{\arraystretch}{1}

	\begin{minipage}{\linewidth}
		\centering
		\begin{tabular}{@{} *{5}{c@{\hspace{3pt}}} @{}}
			                                                                                                                 &
			\makecell{Anydoor}                                                                                               &
			\makecell{Personalize                                                                                              \\Anything} &
			\makecell{Insert                                                                                                   \\Anything} &
			\modelname                                                                                                         \\

			\inputimg{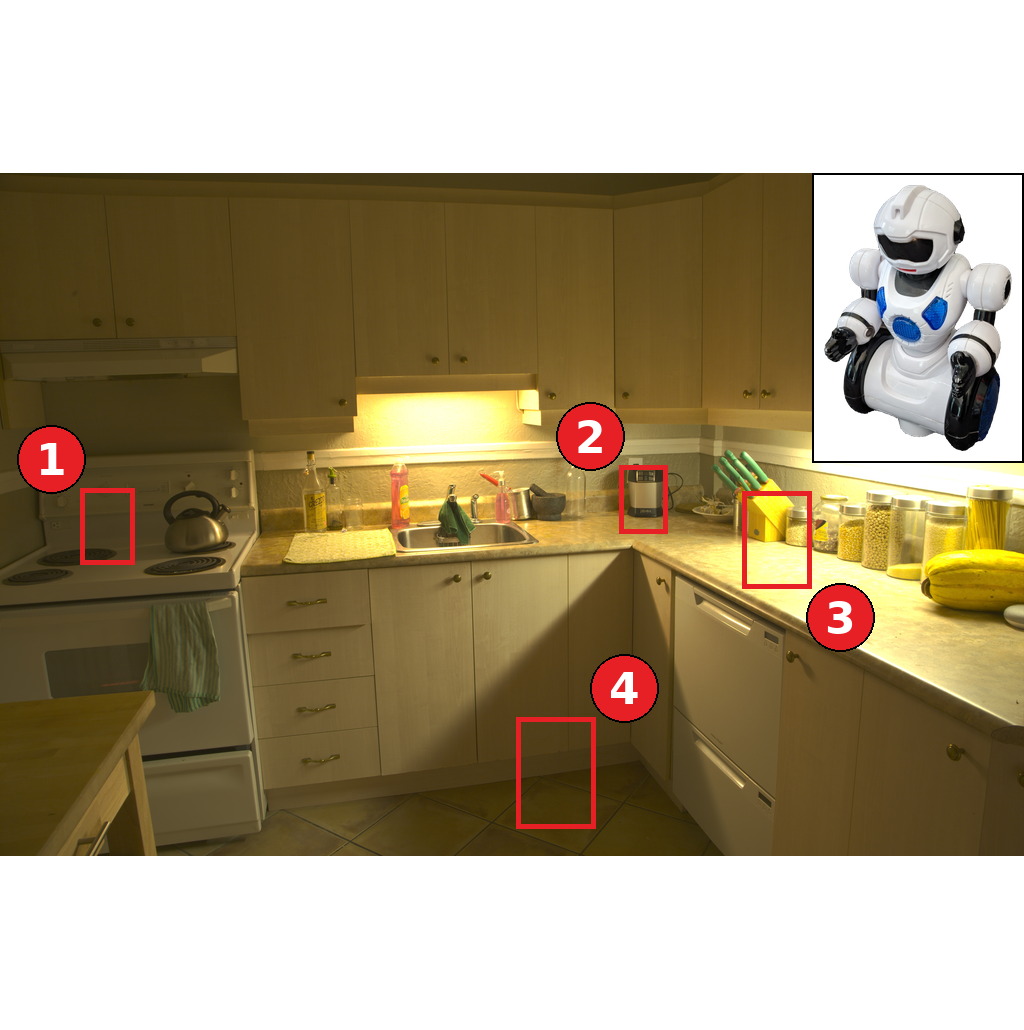}                                            &
			\includegraphics[width=.192\linewidth]{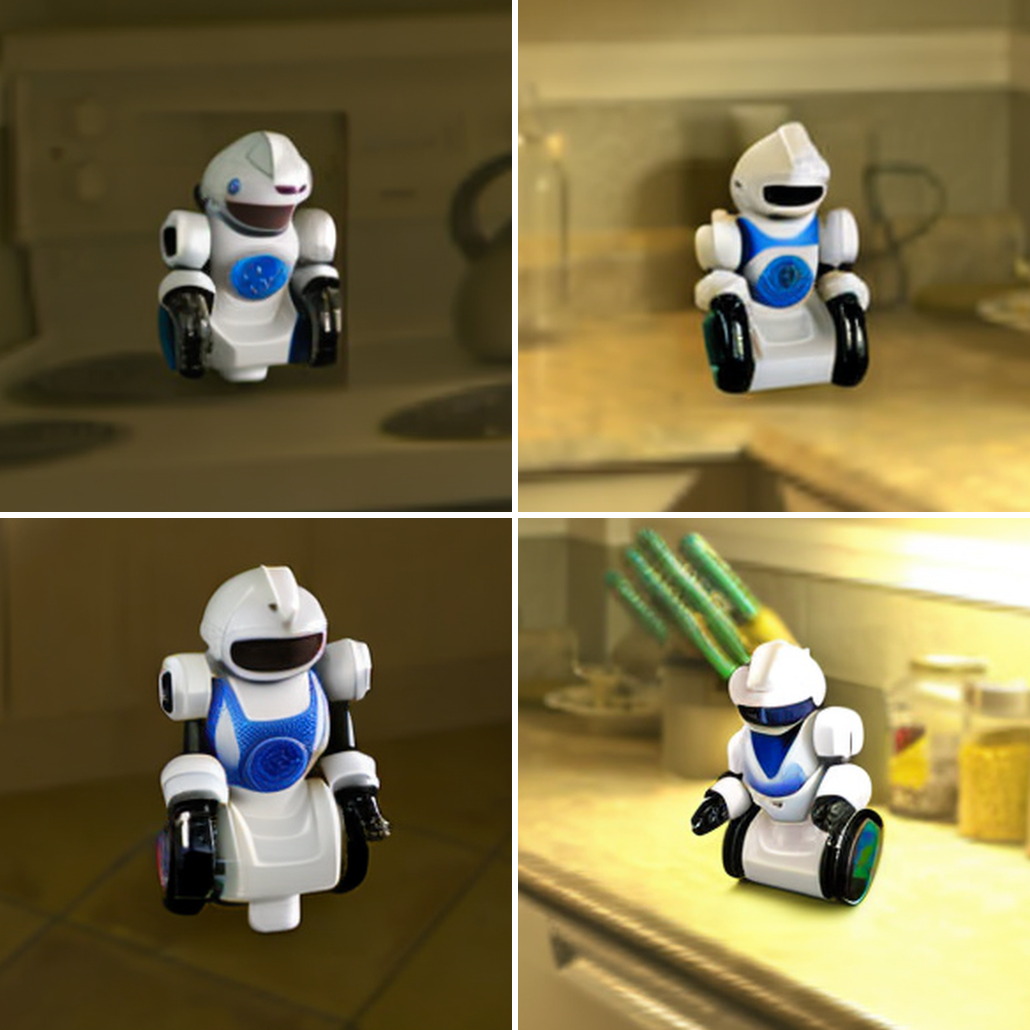}             &
			\includegraphics[width=.192\linewidth]{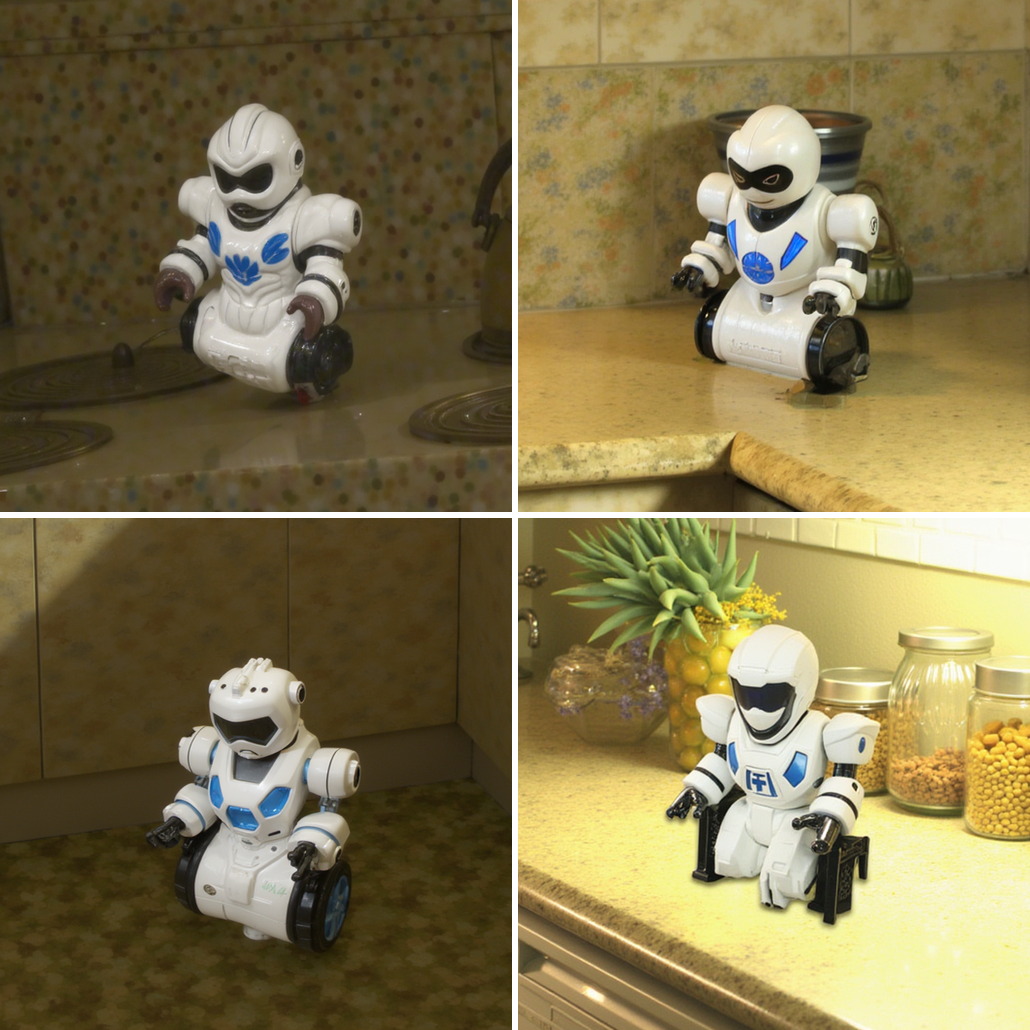} &
			\includegraphics[width=.192\linewidth]{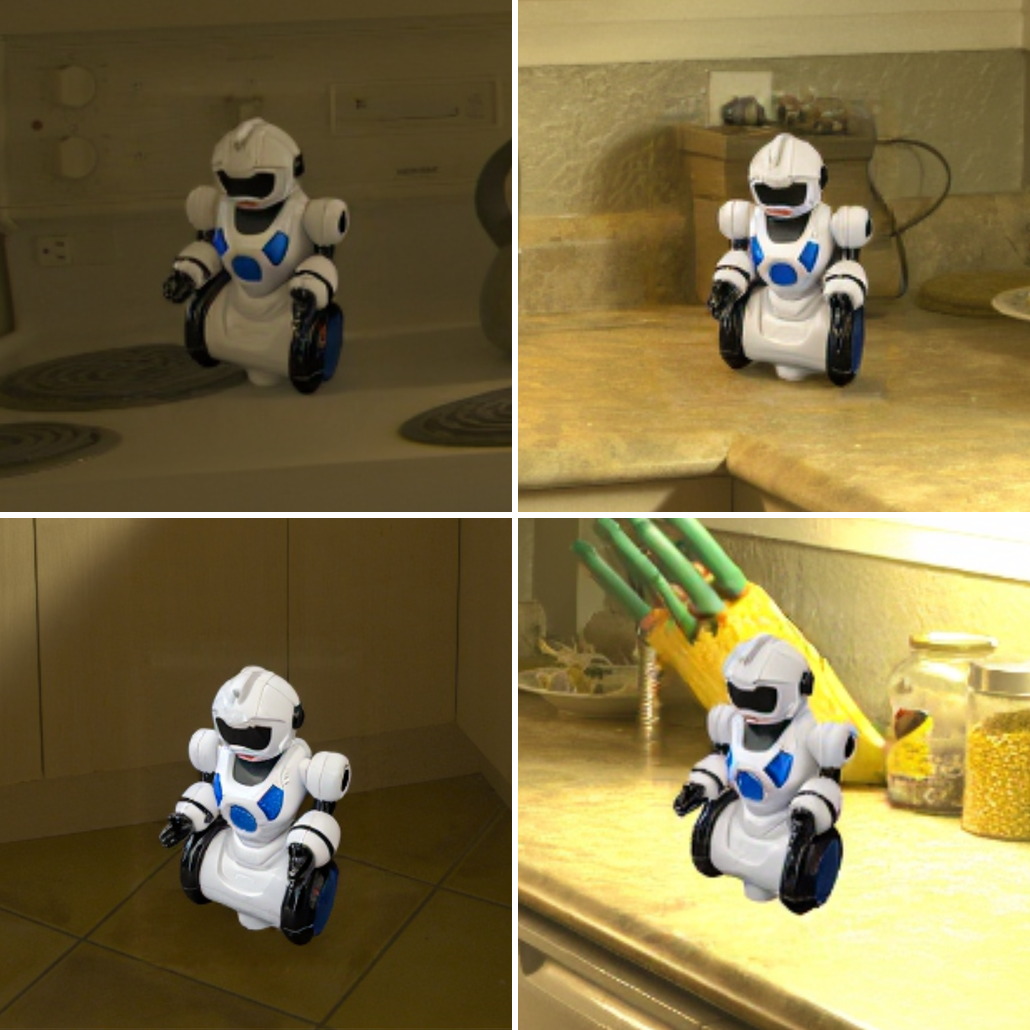}      &
			\includegraphics[width=.192\linewidth]{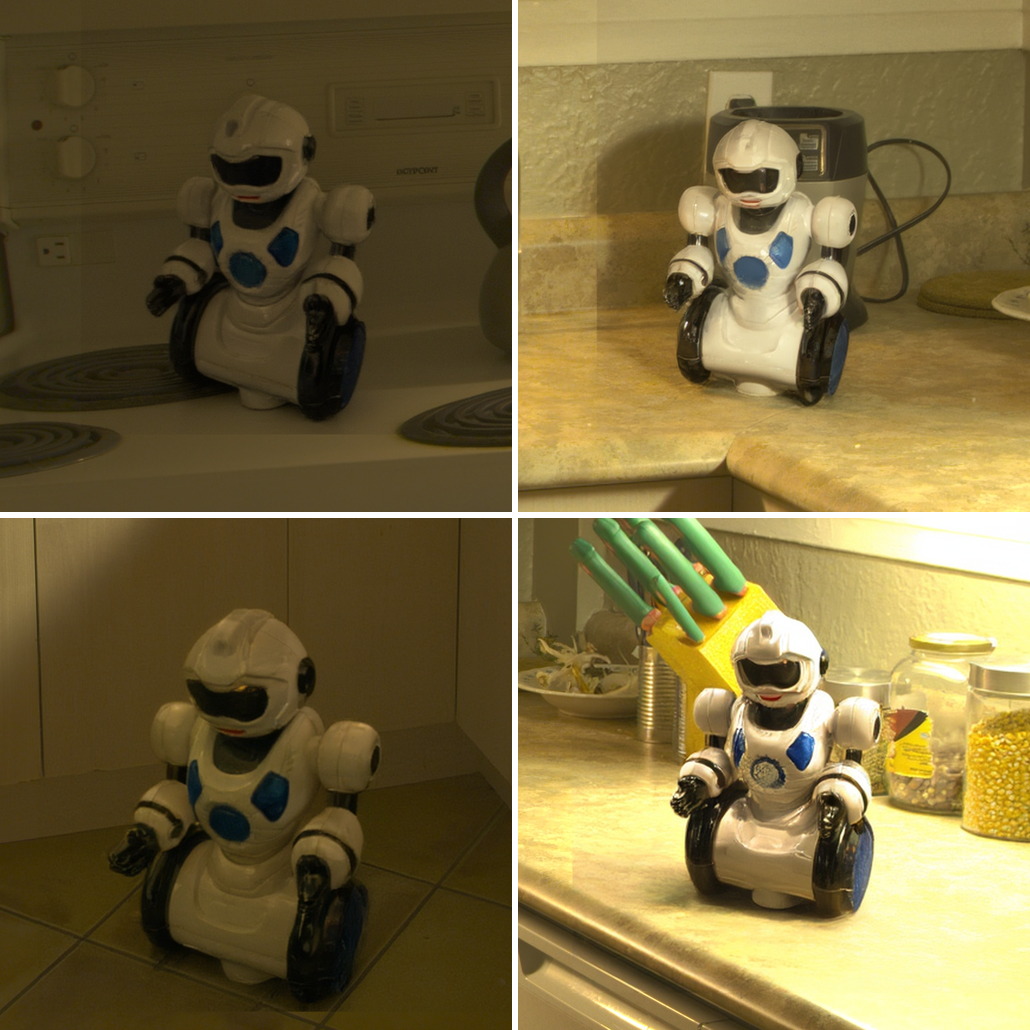}                  \\

			                                                                                                                 &
			\makecell{UniCombine}                                                                                            &
			\makecell{DreamFuse}                                                                                             &
			\makecell{OmniPaint}                                                                                             &
			\modelname                                                                                                         \\

			\inputimg{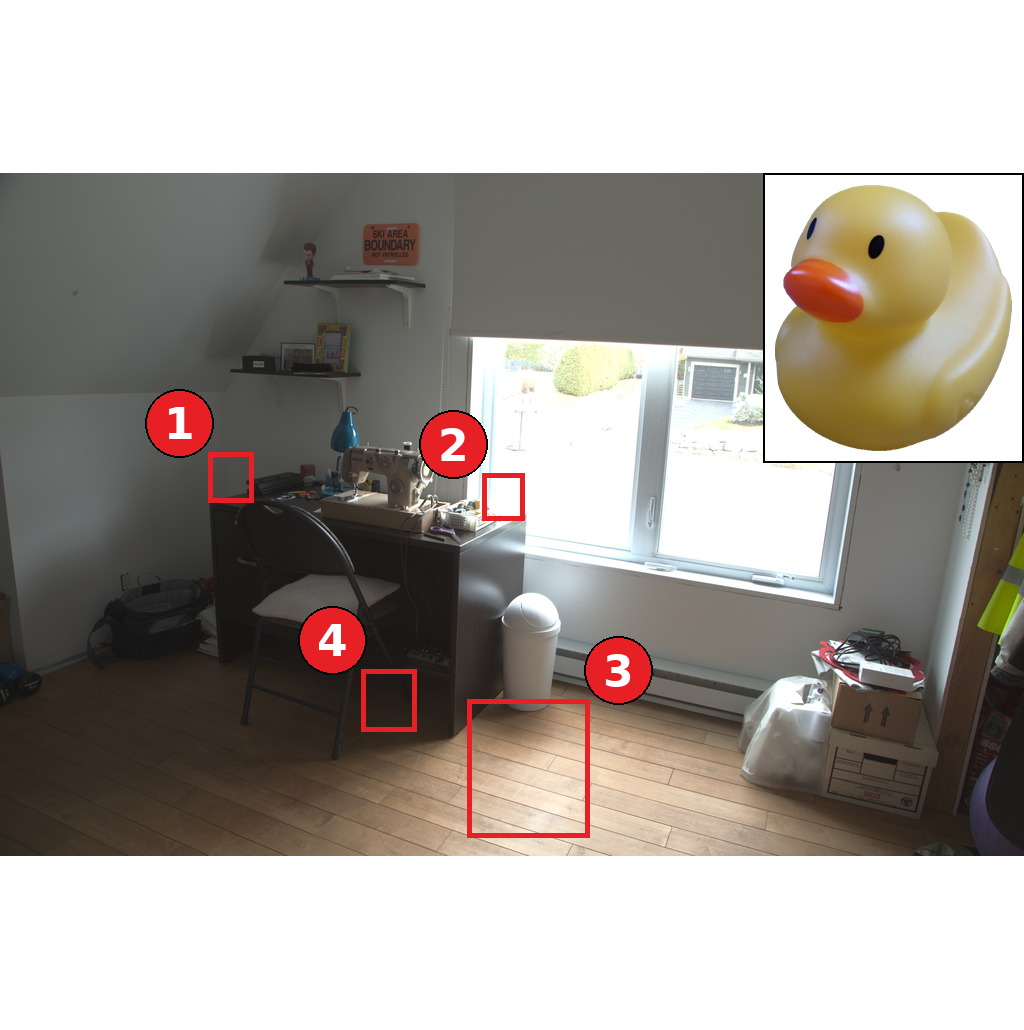}                                             &
			\includegraphics[width=.192\linewidth]{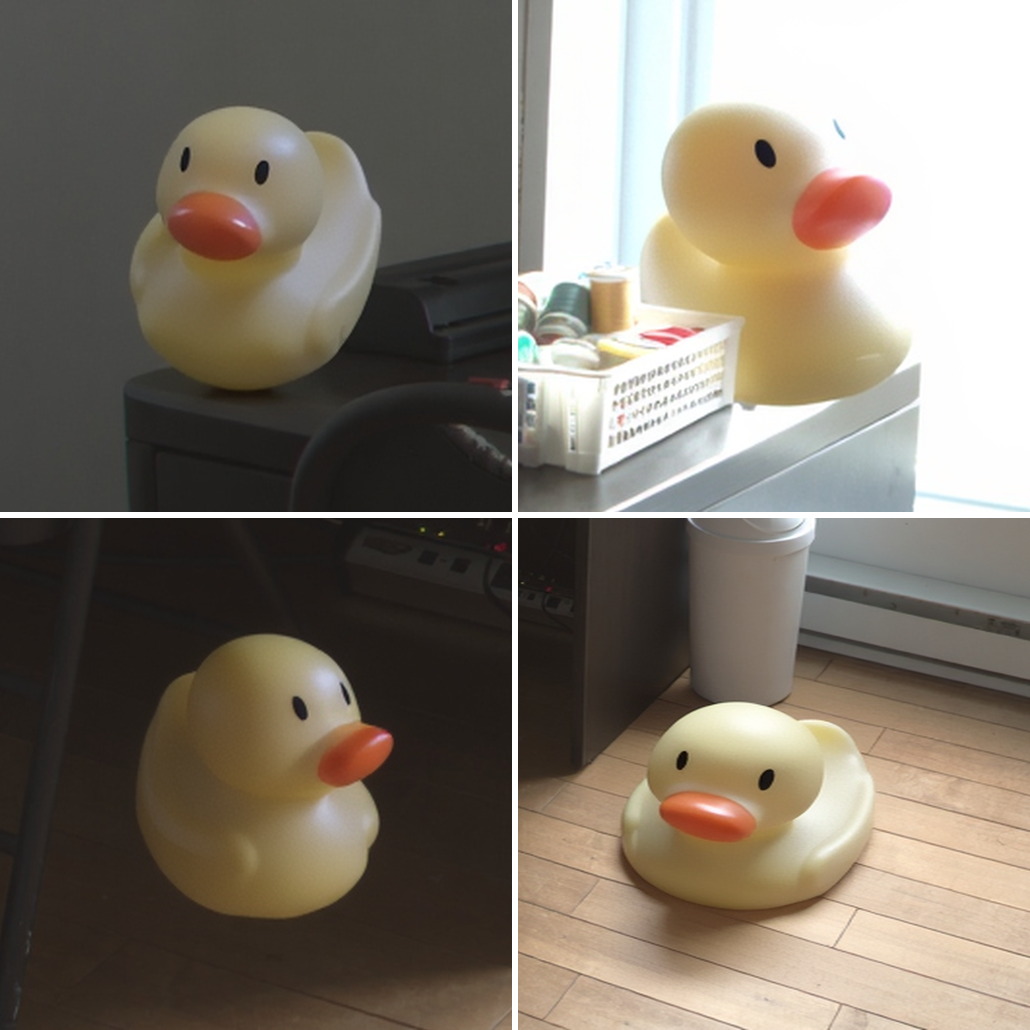}           &
			\includegraphics[width=.192\linewidth]{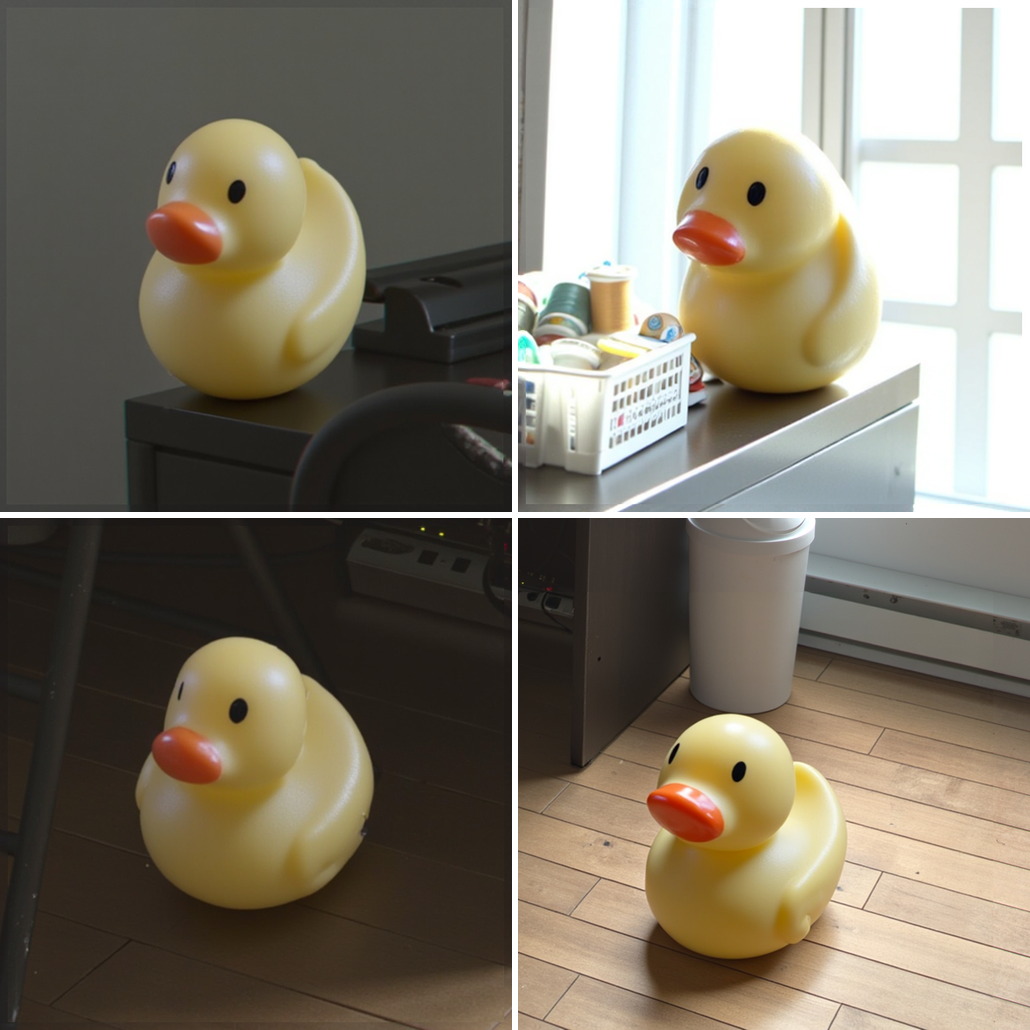}            &
			\includegraphics[width=.192\linewidth]{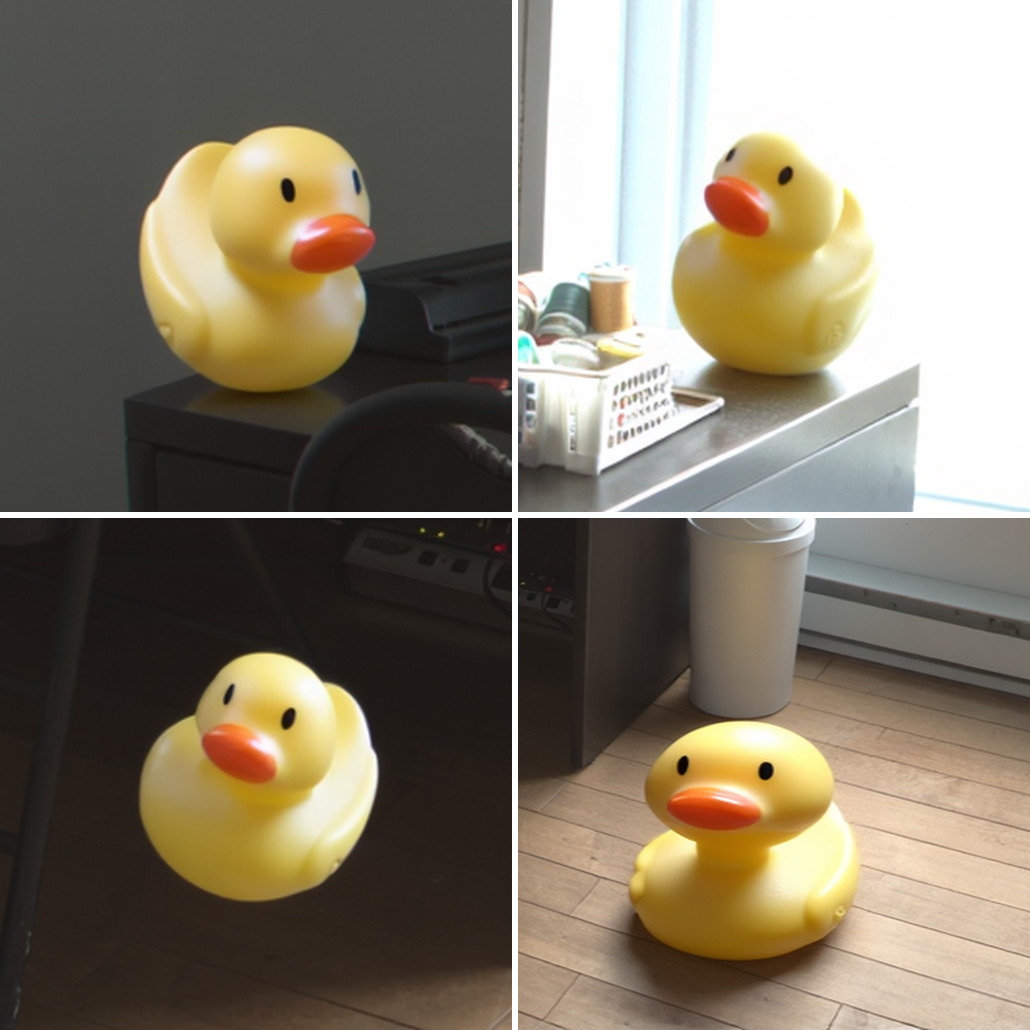}            &
			\includegraphics[width=.192\linewidth]{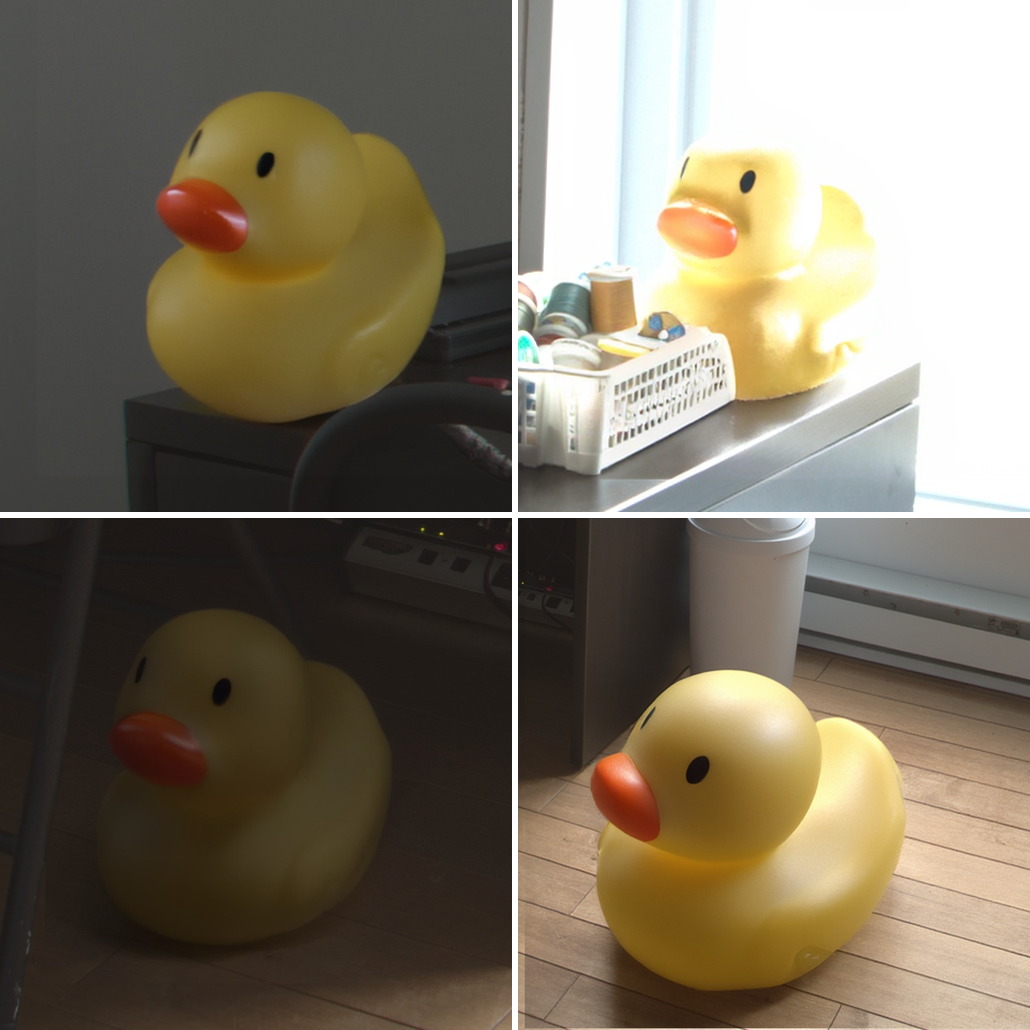}                   \\
		\end{tabular}
	\end{minipage}

	\par\vspace{2pt}{\footnotesize (a) Laval Indoor SV HDR}

	\vspace{7pt}

	\begin{minipage}{\linewidth}
		\centering
		\begin{tabular}{@{} *{6}{c@{\hspace{3pt}}} @{}}
			                                                                                                            &
			\makecell{G.T.}                                                                                             &
			\makecell{Anydoor}                                                                                          &
			\makecell{Personalize                                                                                         \\Anything} &
			\makecell{Insert                                                                                              \\Anything} &
			\modelname                                                                                                    \\

			\begin{tikzpicture}[baseline=(img.south)]
				\node[inner sep=0] (img) {\includegraphics[width=.162\linewidth]{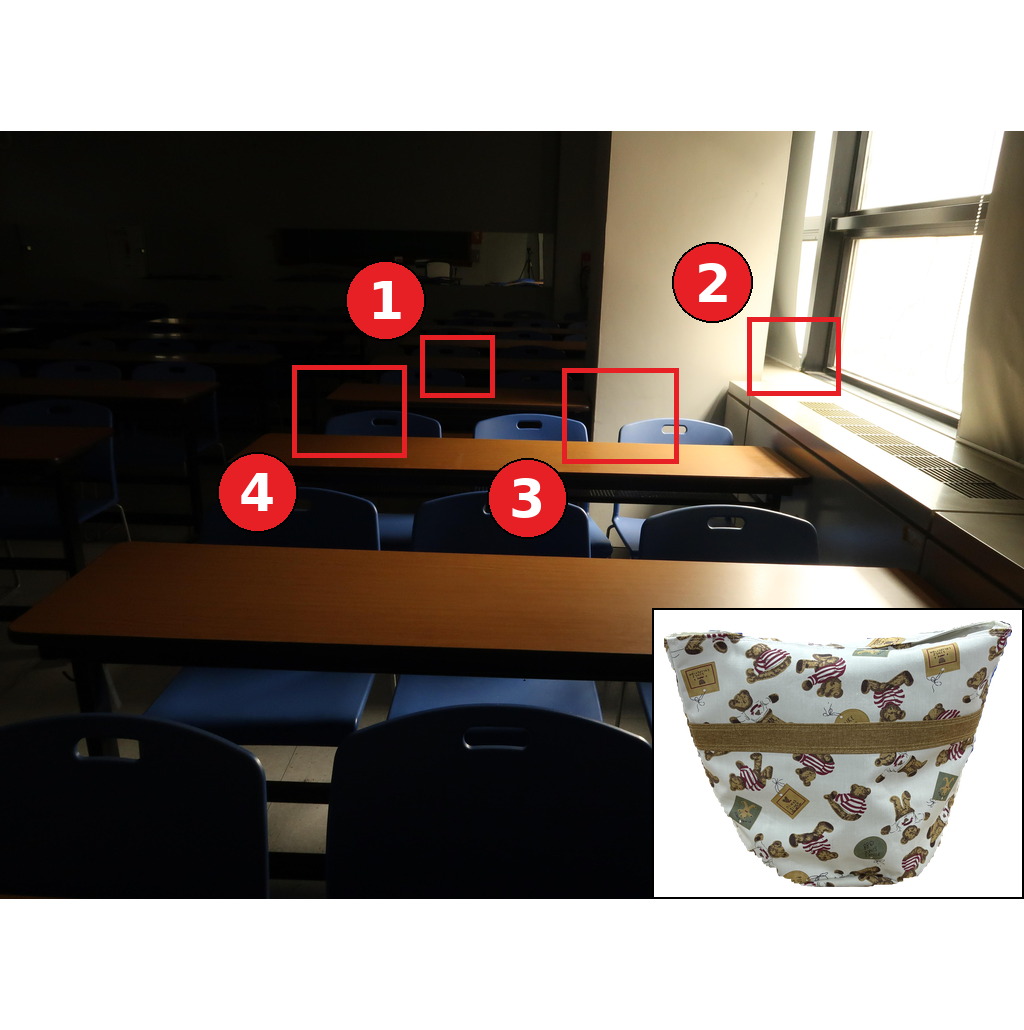}};
				\useasboundingbox (img.south west) rectangle (img.north east);
				\node[anchor=north, font=\scriptsize, text=black, fill=white, inner xsep=3pt, inner ysep=1pt, yshift=2pt] at (img.north) {Input};
			\end{tikzpicture}                            &
			\includegraphics[width=.162\linewidth]{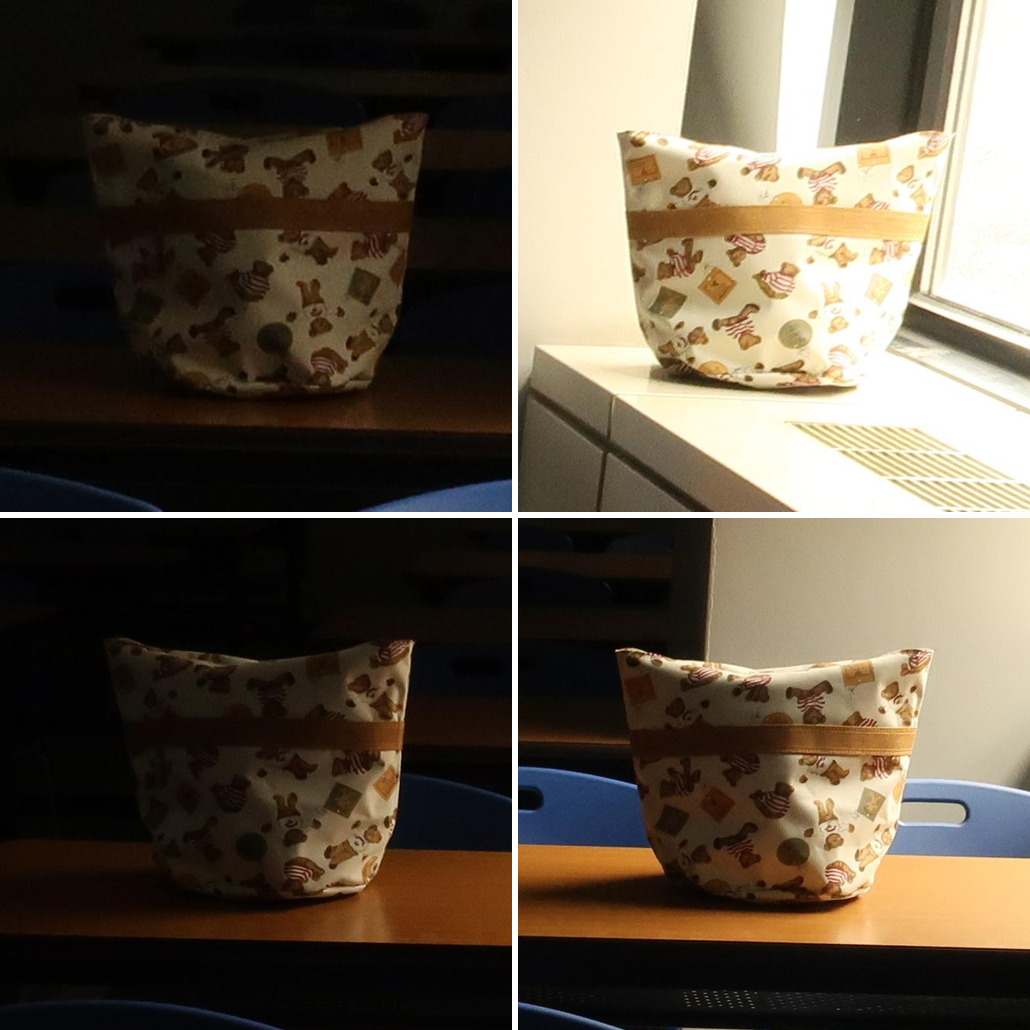}                  &
			\includegraphics[width=.162\linewidth]{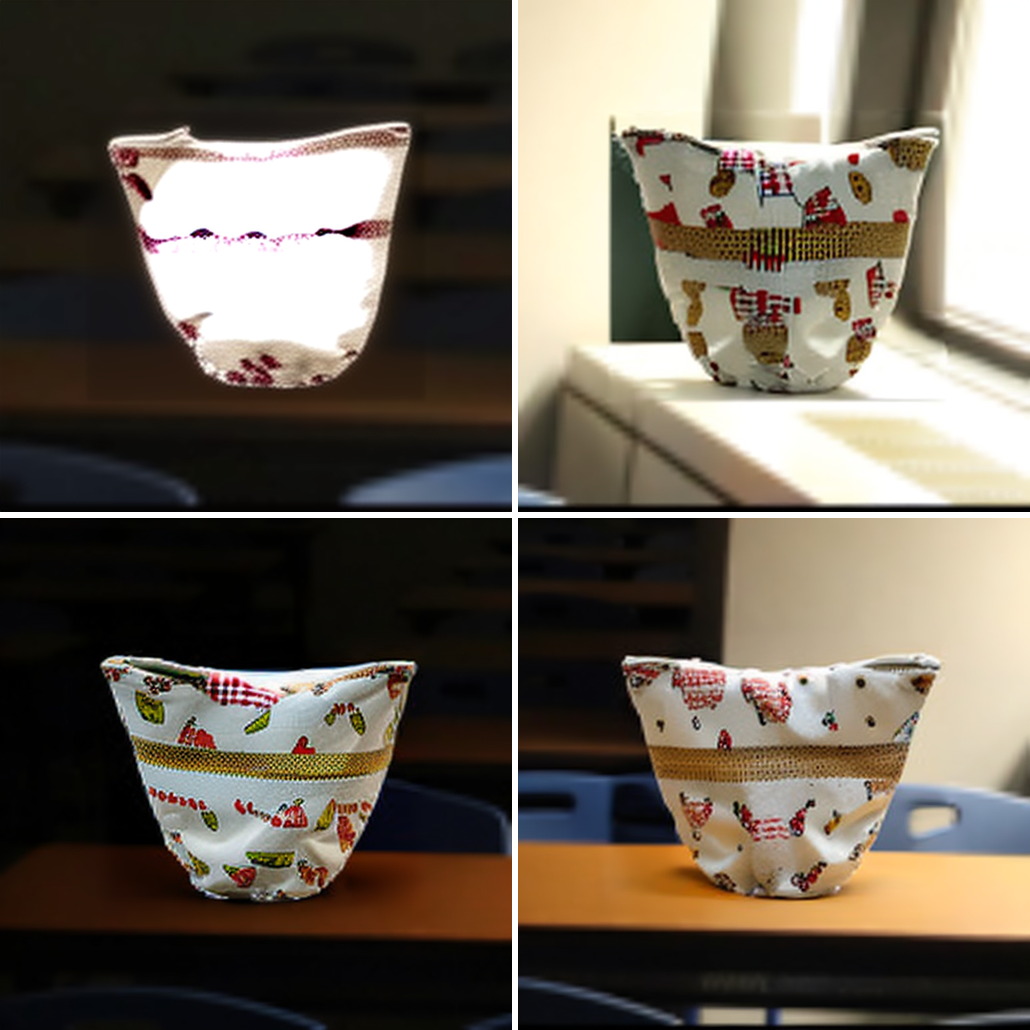}             &
			\includegraphics[width=.162\linewidth]{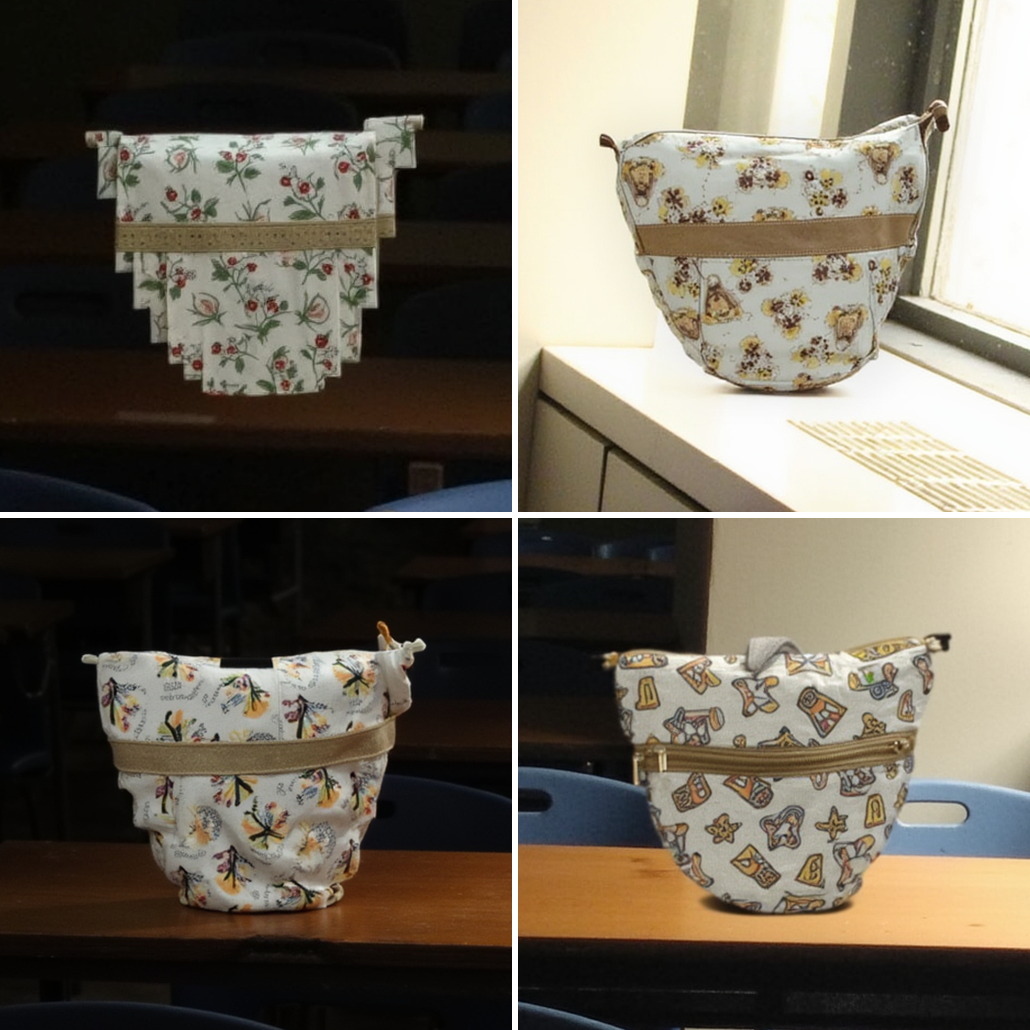} &
			\includegraphics[width=.162\linewidth]{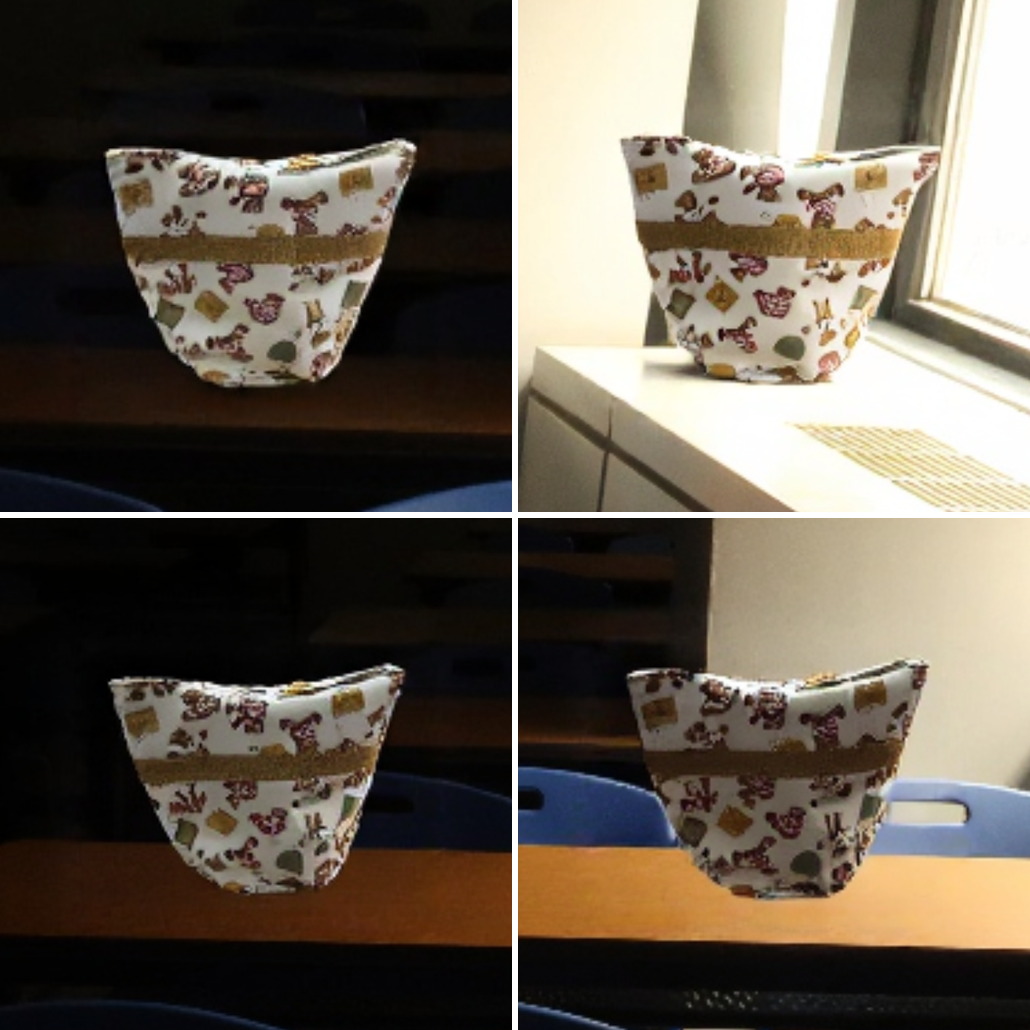}      &
			\includegraphics[width=.162\linewidth]{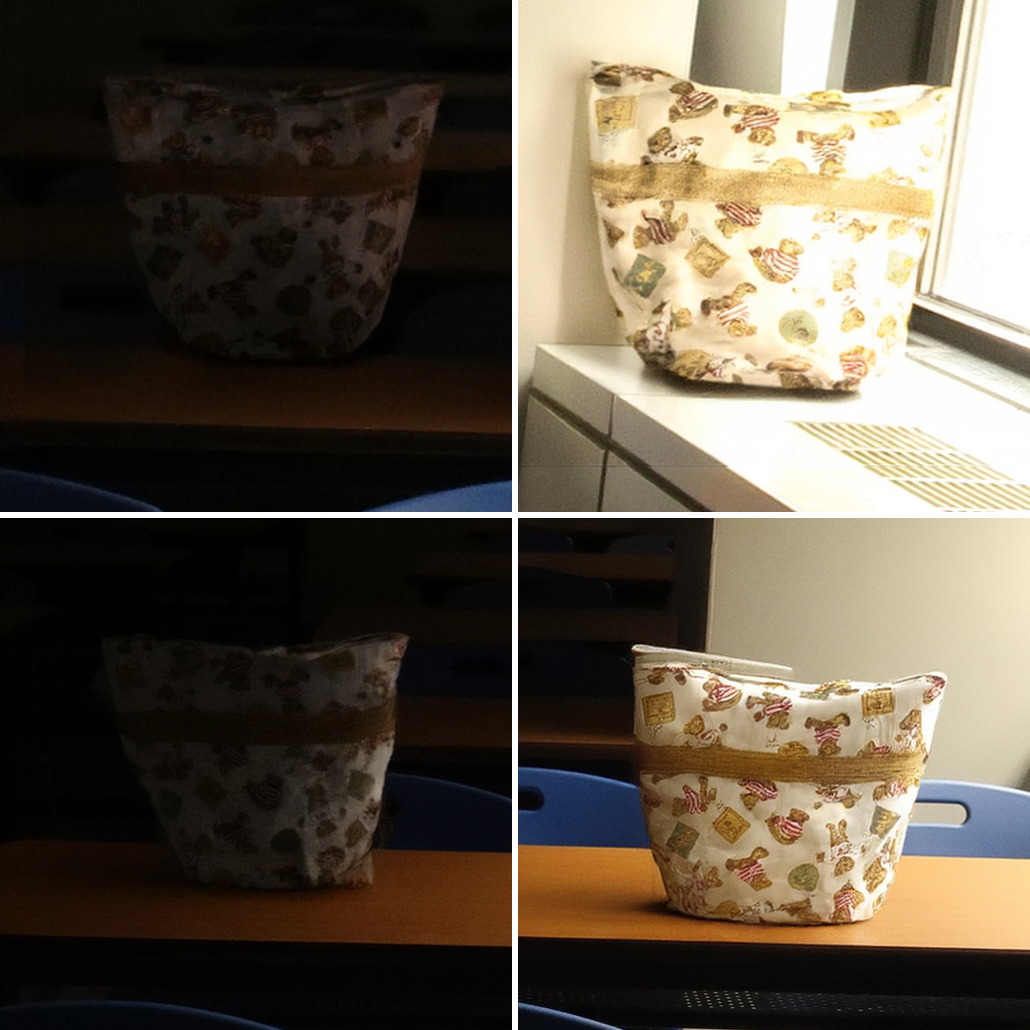}                  \\

			                                                                                                            &
			\makecell{G.T.}                                                                                             &
			\makecell{UniCombine}                                                                                       &
			\makecell{DreamFuse}                                                                                        &
			\makecell{OmniPaint}                                                                                        &
			\modelname                                                                                                    \\

			\begin{tikzpicture}[baseline=(img.south)]
				\node[inner sep=0] (img) {\includegraphics[width=.162\linewidth]{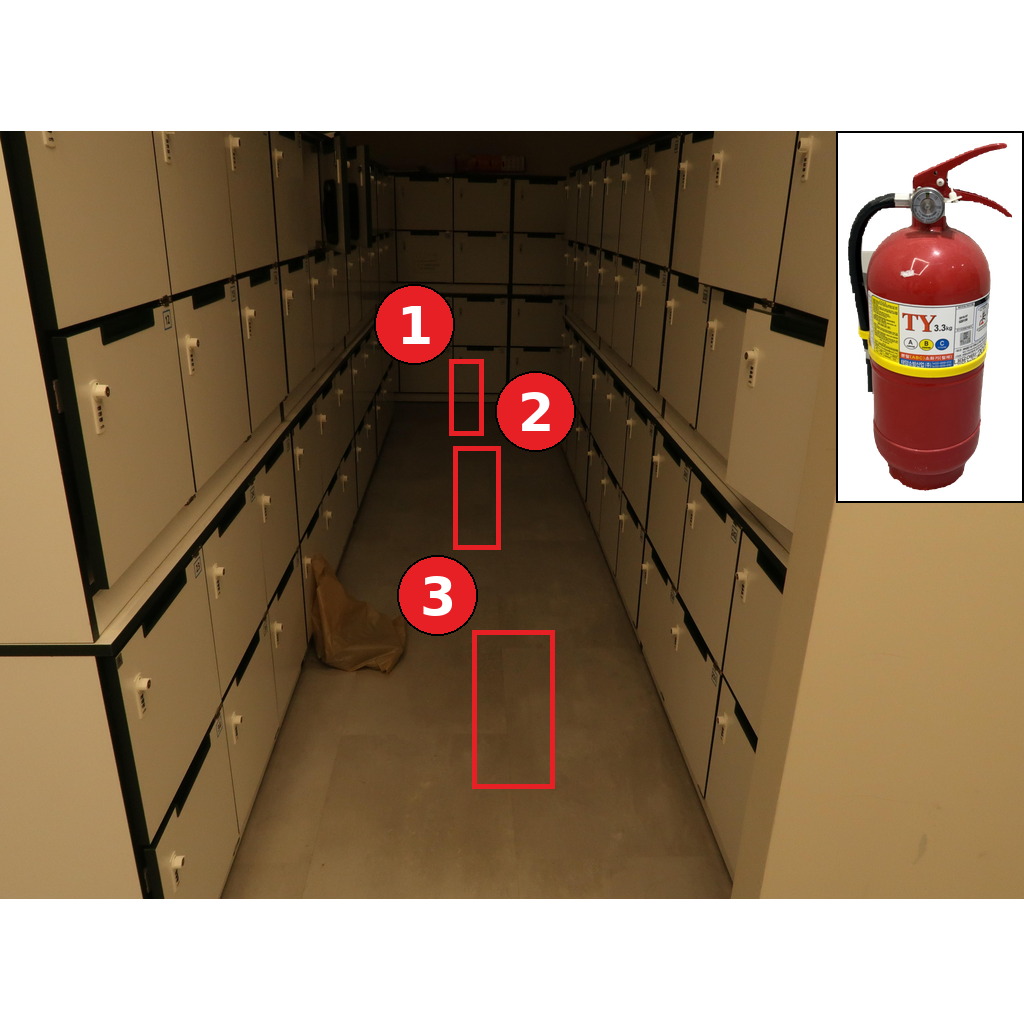}};
				\useasboundingbox (img.south west) rectangle (img.north east);
				\node[anchor=north, font=\scriptsize, text=black, fill=white, inner xsep=3pt, inner ysep=1pt, yshift=2pt] at (img.north) {Input};
			\end{tikzpicture}                            &
			\includegraphics[width=.162\linewidth]{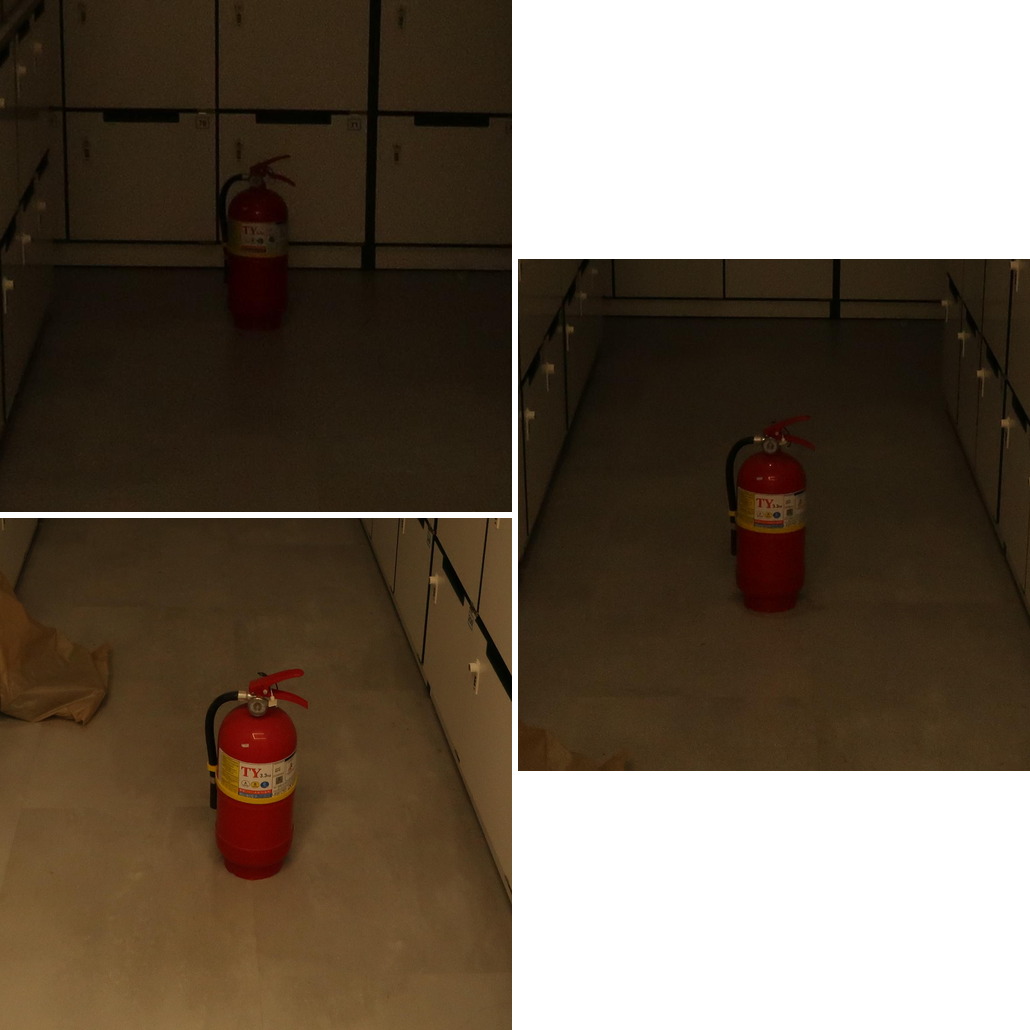}                  &
			\includegraphics[width=.162\linewidth]{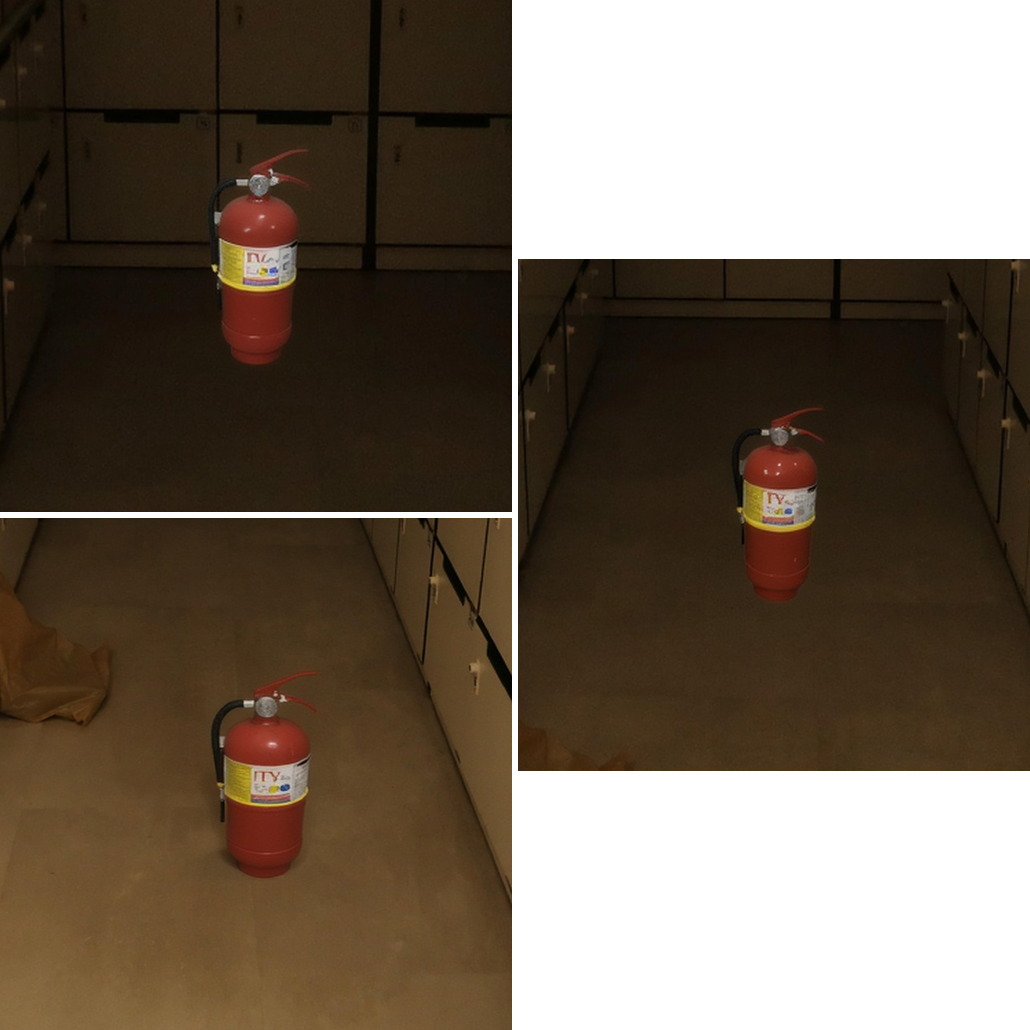}          &
			\includegraphics[width=.162\linewidth]{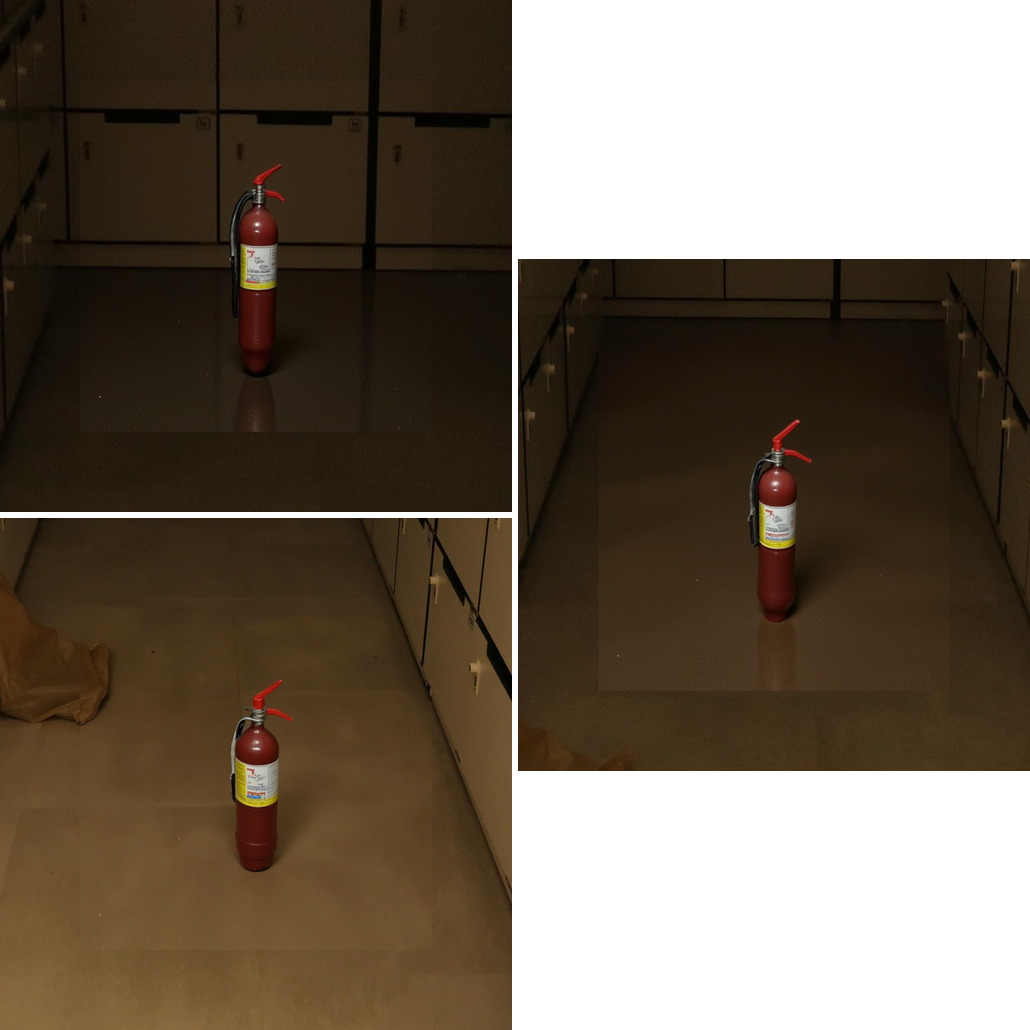}           &
			\includegraphics[width=.162\linewidth]{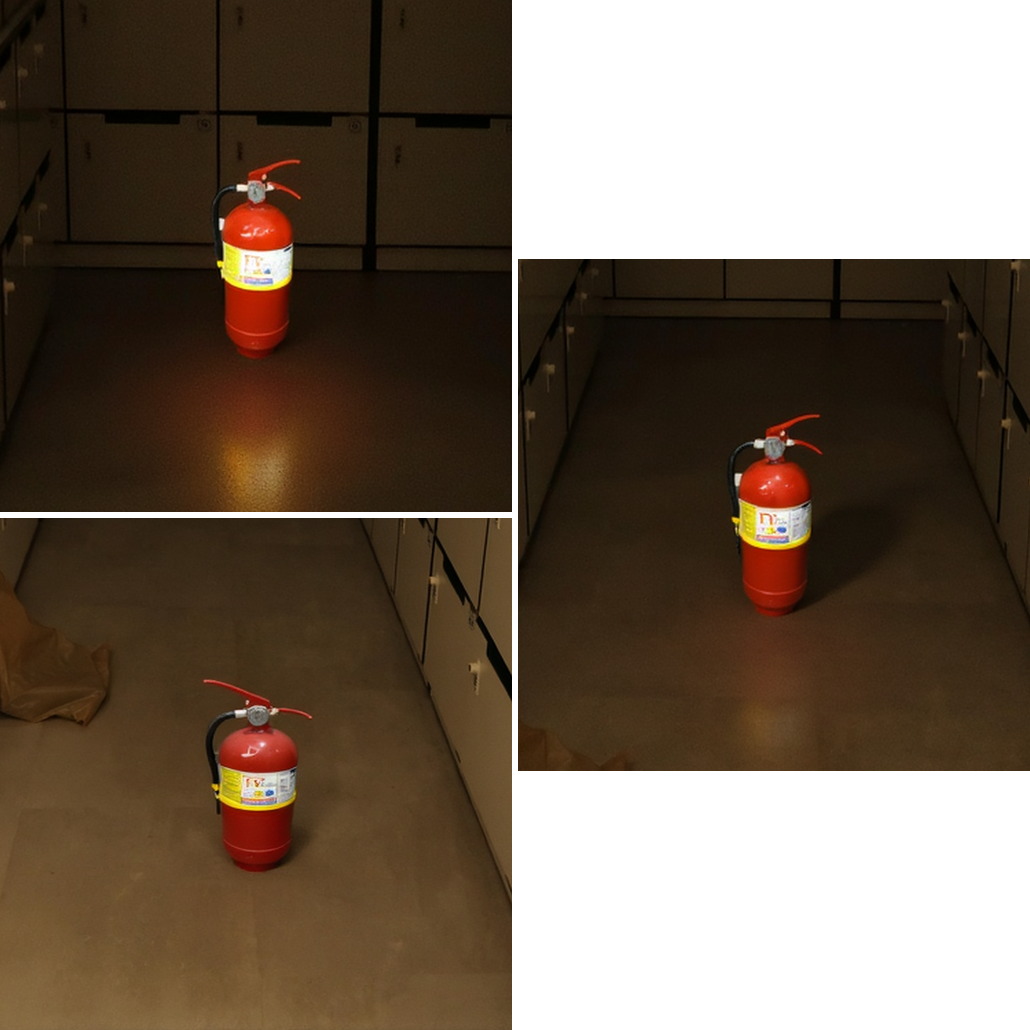}           &
			\includegraphics[width=.162\linewidth]{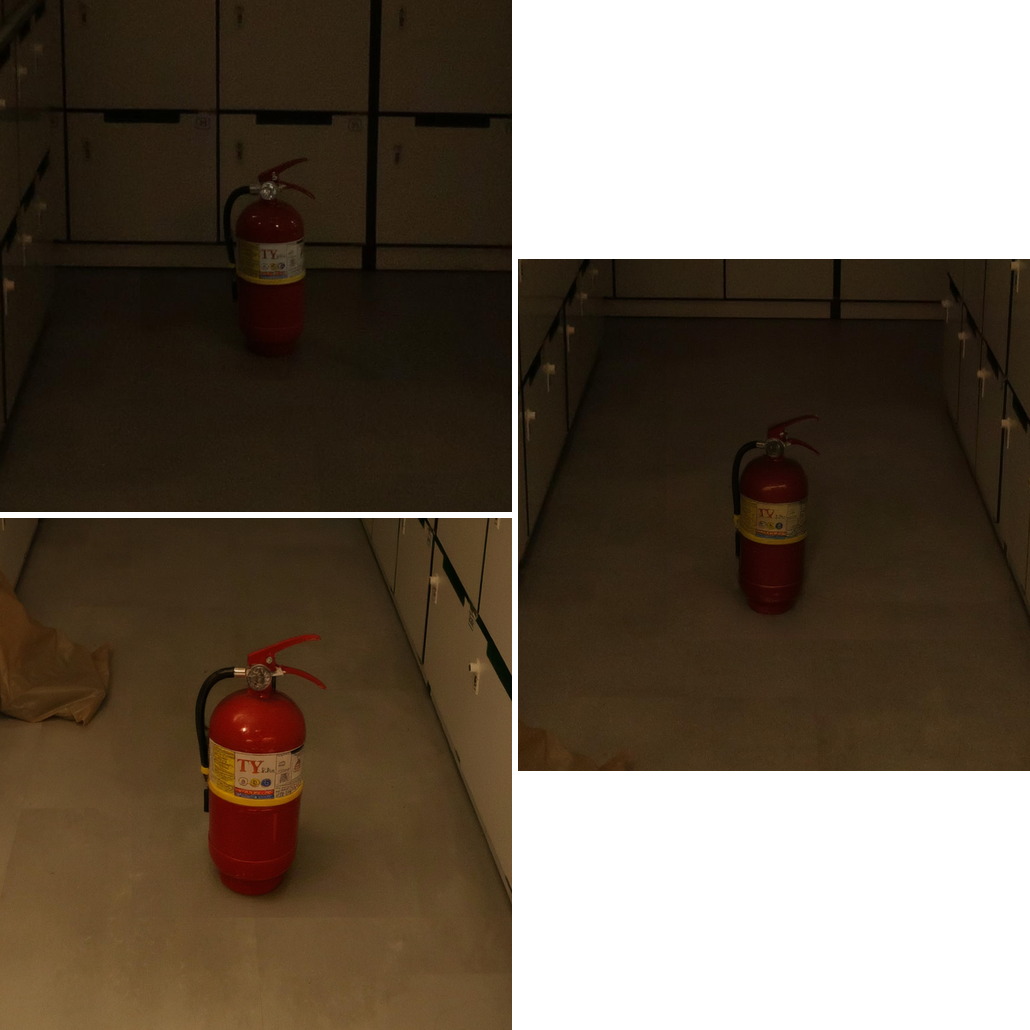}                  \\
		\end{tabular}
	\end{minipage}

	\par\vspace{2pt}{\footnotesize (b) \benchname{}}

	\caption{
		\modelname{} adapts object appearance to local illumination at each insertion location.
		Surfaces darken at shadowed locations (\eg, Laval kitchen probe~4), while shading and color follow the local lighting (\eg, probes~2--3).
		In contrast, baselines show limited variation across locations; for example, Insert Anything mainly rescales overall brightness, with limited adaptation of shading to the local illumination.
		Cropped results follow the input numbering in clockwise order from the top-left.
		Ground-truth composites are available for \benchname but not for Laval.
	}
	\label{fig:composite}
	\vspace{-10pt}
\end{figure}

\noindent
\textbf{Implementation Details.}
\label{sec:impl_details}
For fair comparison with recent compositing baselines~\citep{feng2025personalize,song2026insert,wang2025unicombine,huang2025dreamfuseadaptiveimagefusion,yu2025omnipaintmasteringobjectorientedediting}, we adopt the Flux1-Dev backbone~\citep{flux2024}.
Training uses rank-16 LoRA~\citep{hu2022lora} in bfloat16 with AdamW~\citep{loshchilov2018decoupled}.
Starting from an AnyDoor-pretrained checkpoint as a generic generative-compositing prior, we fine-tune on a 4:1 mixture of our illumination-augmented data (\S \ref{sec:augmentation}) and the training sets of AnyDoor, AnyInsertion, ORIDa, and DreamFuse, which provide higher-resolution object crops complementary to our video-extracted data.
During training, the background crop ranges from the bounding box to the full image, and is fixed to $1.5\times$ at inference.
For each scene, we sample two reference frames and form 16 training triplets via batch permutation described in \S \ref{sec:augmentation}.
Additional implementation details are provided in Appendix~\ref{app:implementation}.

\subsection{Comparisons with Existing Methods}
\label{sec:results}
\noindent
\textbf{Appearance Adaptation.}
Fig.~\ref{fig:composite} shows \modelname{} adapting object appearance to local illumination across insertion locations, with consistent changes in brightness, shading, and color.
Additional qualitative results and estimated irradiance maps~\citep{zeng2024rgb} are provided in Appendix (Fig.~\ref{fig:composite_additional}, \ref{fig:irradiance}).
Quantitatively, \modelname{} achieves the lowest DreamSim distance on \benchname{} ($0.265$ vs.\ Insert Anything $0.312$, the strongest baseline; Tab.~\ref{tab:quantitative_comparison}).
In user studies shown on the right side of Tab.~\ref{tab:quantitative_comparison}, our model is preferred over all baselines in harmony and overall quality.
The reference-free IC metric is consistent: \modelname{} improves over the strongest baseline by $1.77$ on Laval (vs.\ DreamFuse, $13.27$) and $2.19$ on \benchname{} (vs.\ UniCombine, $11.80$).
Direct comparison with ground-truth composites (Tab.~\ref{tab:lighting_measurements}) further validates \modelname{}'s adaptation to local illumination, reducing errors by 12\% in light direction, 38\% in brightness, and 22\% in hue relative to the best-performing baselines.
Additional comparisons show that \modelname{} also outperforms harmonization and intrinsic-based editing methods across IC, DreamSim, and ground-truth-based illumination and appearance metrics (Appendix Tabs.~\ref{tab:additional_baselines}, \ref{tab:lighting_measurements_full} and Fig.~\ref{fig:additional_baselines}).

\begin{table}[t]
	\caption{
		(Left) Quantitative comparison on scenes with spatially varying illumination.
		DreamSim~\citep{fu2023dreamsim} measures perceptual distance to ground-truth composites in \benchname{}; IC is a reference-free metric (PSNR between estimated local lighting before and after compositing) used additionally on Laval~\citep{Garon_2019_CVPR}, which lacks ground-truth composites.
		(Right) Pairwise user study, \modelname{}-win\,/\,tie\,/\,baseline-win (\%); larger of W/L in bold.
		No-Ref denotes no-reference realism ranking on Laval, and Ref denotes reference-based comparison on \benchname with ground-truth composites.
	}
	\label{tab:quantitative_comparison}
	\vspace{4pt}
	\scriptsize
	\centering
	\setlength{\tabcolsep}{4.5pt}
	\begin{tabular}{c c c @{\hspace{3pt}} c || *{6}c}
		\toprule
		                                   & \makecell{Laval}                       & \multicolumn{2}{c}{\makecell{\benchname}} &
		\multicolumn{3}{c}{Laval (No-Ref)} & \multicolumn{3}{c}{\benchname{} (Ref)}                                                                                                                                                                                                                                                      \\
		\cmidrule(lr){2-2}\cmidrule(lr){3-4}\cmidrule(lr){5-7} \cmidrule(lr){8-10}
		                                   & IC$\uparrow$                           & IC$\uparrow$                              & DreamSim$\downarrow$
		                                   & Harmony                                & Identity                                  & Quality              & Harmony                    & Identity                  & Quality                                                                                                                \\
		\midrule
		AnyDoor                            & 10.25                                  & 7.88                                      & 0.461                & \textbf{96}\,/\,02\,/\,02  & \textbf{76}\,/\,15\,/\,09 & \textbf{91}\,/\,02\,/\,07 & \textbf{88}\,/\,06\,/\,06 & \textbf{88}\,/\,06\,/\,06          & \textbf{92}\,/\,02\,/\,06 \\
		Personalize Anything               & 11.28                                  & 9.09                                      & 0.419                & \textbf{100}\,/\,00\,/\,00 & \textbf{61}\,/\,23\,/\,16 & \textbf{96}\,/\,04\,/\,00 & \textbf{85}\,/\,09\,/\,06 & \textbf{74}\,/\,11\,/\,15          & \textbf{91}\,/\,04\,/\,05 \\
		Insert Anything                    & 12.12                                  & 11.10                                     & \underline{0.312}    & \textbf{94}\,/\,06\,/\,00  & \textbf{37}\,/\,33\,/\,30 & \textbf{98}\,/\,00\,/\,02 & \textbf{65}\,/\,14\,/\,21 & \textbf{43}\,/\,14\,/\,\textbf{43} & \textbf{67}\,/\,14\,/\,19 \\
		UniCombine                         & 13.17                                  & \underline{11.80}                         & 0.342                & \textbf{90}\,/\,08\,/\,02  & \textbf{63}\,/\,25\,/\,12 & \textbf{79}\,/\,13\,/\,08 & \textbf{75}\,/\,06\,/\,19 & \textbf{58}\,/\,09\,/\,33          & \textbf{67}\,/\,08\,/\,25 \\
		DreamFuse                          & \underline{13.27}                      & 11.51                                     & 0.335                & \textbf{82}\,/\,13\,/\,05  & \textbf{64}\,/\,16\,/\,20 & \textbf{71}\,/\,07\,/\,22 & \textbf{70}\,/\,09\,/\,21 & \textbf{68}\,/\,21\,/\,11          & \textbf{79}\,/\,12\,/\,09 \\
		OmniPaint                          & 11.92                                  & 9.54                                      & 0.349                & \textbf{94}\,/\,02\,/\,04  & \textbf{57}\,/\,29\,/\,14 & \textbf{90}\,/\,02\,/\,08 & \textbf{72}\,/\,11\,/\,17 & \textbf{60}\,/\,13\,/\,27          & \textbf{72}\,/\,13\,/\,15 \\
		FreeCompose                        & 11.37                                  & 9.15                                      & 0.393                & \textbf{100}\,/\,00\,/\,00 & \textbf{84}\,/\,07\,/\,09 & \textbf{95}\,/\,05\,/\,00 & \textbf{91}\,/\,09\,/\,00 & \textbf{74}\,/\,13\,/\,13          & \textbf{93}\,/\,05\,/\,02 \\
		\midrule
		\textbf{\modelname}                & \textbf{15.04}                         & \textbf{13.99}                            & \textbf{0.265}       & --                         & --                        & --                        & --                        & --                                 & --                        \\
		\bottomrule
	\end{tabular}
	\vspace{-10pt}
\end{table}

\noindent
\textbf{Identity Preservation.}
On ORIDa, \modelname{} nearly matches Insert Anything in CLIP but trails it in DINO (Tab.~\ref{tab:identity_preservation}).
This aligns with the benchmark and metric: ORIDa lacks spatially varying lighting, and DINO emphasizes shape and texture similarity, favoring predictions that preserve the input appearance (Sec.~\ref{app:dreamsim}).
As shown in Fig.~\ref{fig:composite}, Insert Anything often retains the input object's original shading, mainly adjusting overall brightness rather than re-lighting the object (e.g., Laval and \benchname{} probe~3).
This yields strong identity scores on ORIDa but lower illumination consistency on Laval (IC 12.12 vs.\ 15.04; Tab.~\ref{tab:quantitative_comparison}).
In the user study, identity preferences between \modelname{} and Insert Anything are comparable (Laval: $37$/$33$/$30$; \benchname{}: $43$/$14$/$43$; Tab.~\ref{tab:quantitative_comparison}).
Overall, \modelname{} maintains competitive identity preservation while better adapting object appearance to local lighting.

\begin{table}[t]
	\scriptsize
	\centering
	\begin{minipage}[t]{0.56\linewidth}
		\centering
		\captionof{table}{
			Illumination and appearance accuracy on \benchname{}. %
		}
		\label{tab:lighting_measurements}
		\vspace{4pt}
		\setlength{\tabcolsep}{3.5pt}
		\begin{tabular}{l ccccc}
			\toprule
			                     & \multicolumn{2}{c}{Light direction} & \multicolumn{3}{c}{Color (HSV)}                                                                       \\
			\cmidrule(lr){2-3}\cmidrule(lr){4-6}
			                     & L2$\downarrow$                      & Angle ($^\circ$)$\downarrow$    & Hue ($^\circ$)$\downarrow$ & Bright.$\downarrow$ & Sat.$\downarrow$ \\
			\midrule
			AnyDoor              & 0.191                               & 34.29                           & 36.96                      & 0.177               & 0.253            \\
			Personalize Anything & 0.188                               & 32.82                           & 22.88                      & 0.172               & 0.248            \\
			Insert Anything      & 0.147                               & 26.36                           & 14.33                      & 0.107               & 0.175            \\
			UniCombine           & 0.150                               & 26.55                           & 13.06                      & 0.103               & 0.178            \\
			DreamFuse            & 0.149                               & 26.23                           & 10.17                      & 0.099               & 0.148            \\
			OmniPaint            & 0.171                               & 30.30                           & 14.48                      & 0.150               & 0.194            \\
			FreeCompose          & 0.202                               & 33.93                           & 22.58                      & 0.158               & 0.214            \\
			\midrule
			\textbf{\modelname}  & \textbf{0.129}                      & \textbf{23.00}                  & \textbf{7.96}              & \textbf{0.061}      & \textbf{0.070}   \\
			\bottomrule
		\end{tabular}
	\end{minipage}
	\hfill
	\begin{minipage}[t]{0.40\linewidth}
		\centering
		\setlength{\tabcolsep}{6pt}
		\captionof{table}{
			Compositing on the ORIDa~\citep{kim2025oridaobjectcentricrealworldimage} evaluation set.
		}
		\label{tab:identity_preservation}
		\vspace{4pt}
		\begin{tabular}{lcc}
			\toprule
			                     & CLIP $\uparrow$  & DINO $\uparrow$  \\
			\midrule
			Anydoor              & 86.6             & 79.4             \\
			Personalize Anything & 74.4             & 52.2             \\
			Insert Anything      & \textbf{90.9}    & \textbf{92.1}    \\
			Unicombine           & 87.9             & 86.3             \\
			DreamFuse            & 83.1             & 67.7             \\
			OmniPaint            & 89.4             & 87.2             \\
			FreeCompose          & 81.0             & 78.2             \\
			\midrule
			\modelname           & \underline{90.8} & \underline{89.3} \\
			\bottomrule
		\end{tabular}
	\end{minipage}
	\vspace{-8pt}
\end{table}

\subsection{Ablation Study}
\label{sec:ablation}
\begin{figure}[t]
	\centering
	\scriptsize
	\renewcommand{\arraystretch}{1}

	\begin{minipage}{0.49\linewidth}
		\centering
		\begin{minipage}[c]{0.44\linewidth}
			\centering
			\begin{tikzpicture}[baseline=(img.south)]
				\node[inner sep=0] (img) {\includegraphics[width=\linewidth]{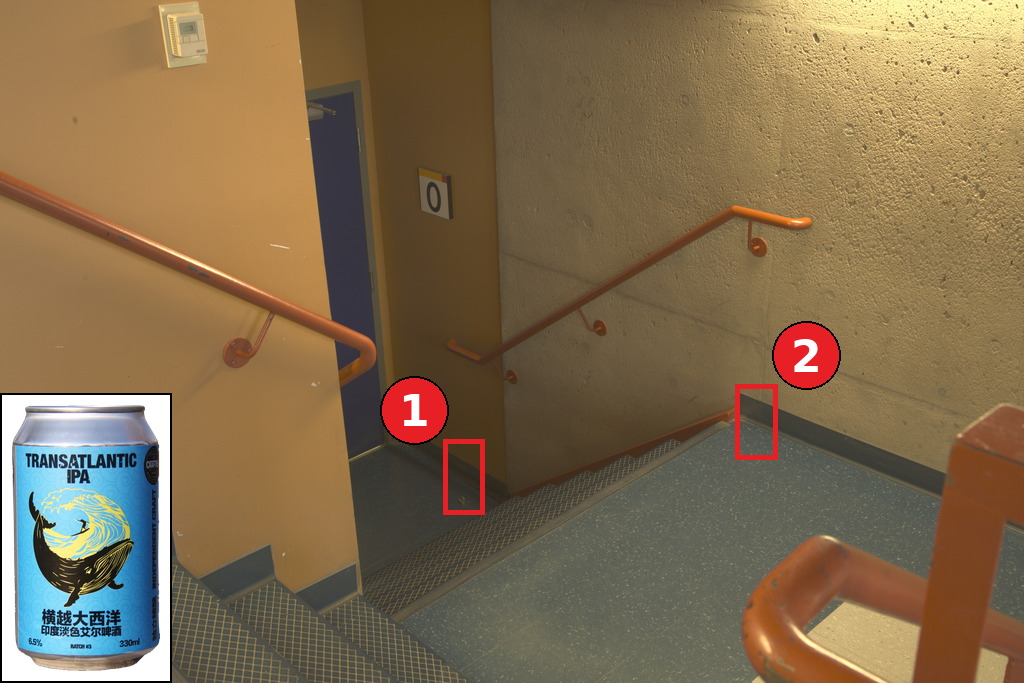}};
				\useasboundingbox (img.south west) rectangle (img.north east);
				\node[anchor=south, font=\scriptsize, text=black, fill=white, inner xsep=3pt, inner ysep=1pt, yshift=2pt] at (img.north) {Input};
			\end{tikzpicture}
		\end{minipage}%
		\hfill
		\begin{minipage}[c]{0.54\linewidth}
			\centering
			\begin{tabular}{@{} c@{\hspace{2pt}}c @{}}
				\makecell{w/o Illum. Aug.}                                                     &
				\modelname                                                                       \\
				\includegraphics[width=.49\linewidth]{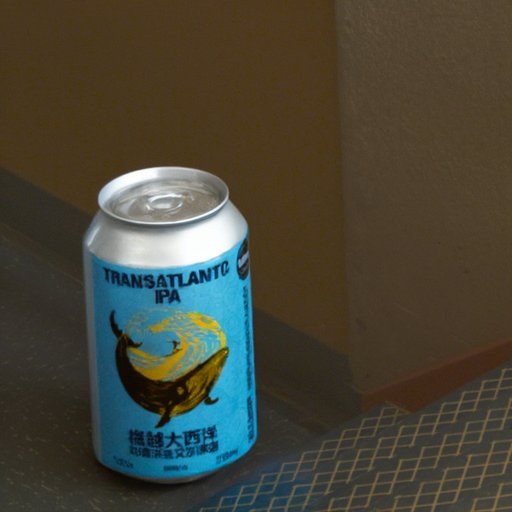} &
				\includegraphics[width=.49\linewidth]{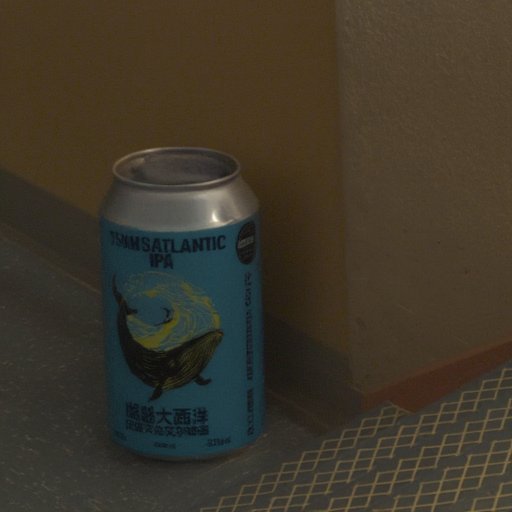}    \\
				\includegraphics[width=.49\linewidth]{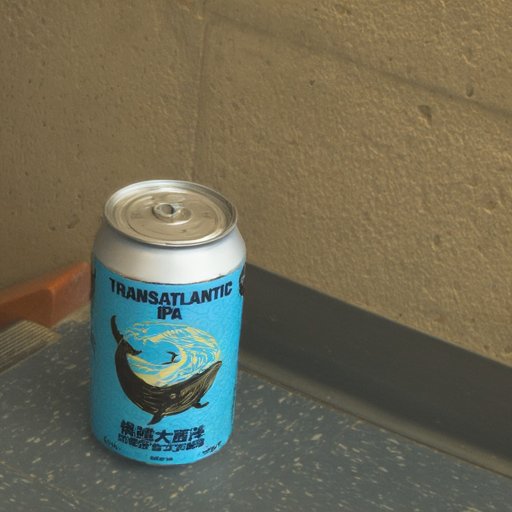} &
				\includegraphics[width=.49\linewidth]{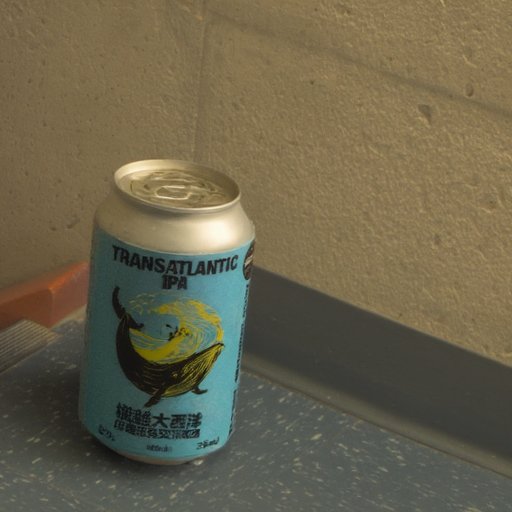}    \\
			\end{tabular}
		\end{minipage}
	\end{minipage}
	\hfill
	\begin{minipage}{0.49\linewidth}
		\centering
		\begin{minipage}[c]{0.44\linewidth}
			\centering
			\smash{\begin{tikzpicture}[baseline=(bboxbot)]
					\node[inner sep=0, anchor=north] (img) at (0.5\linewidth, 0) {\includegraphics[width=\linewidth]{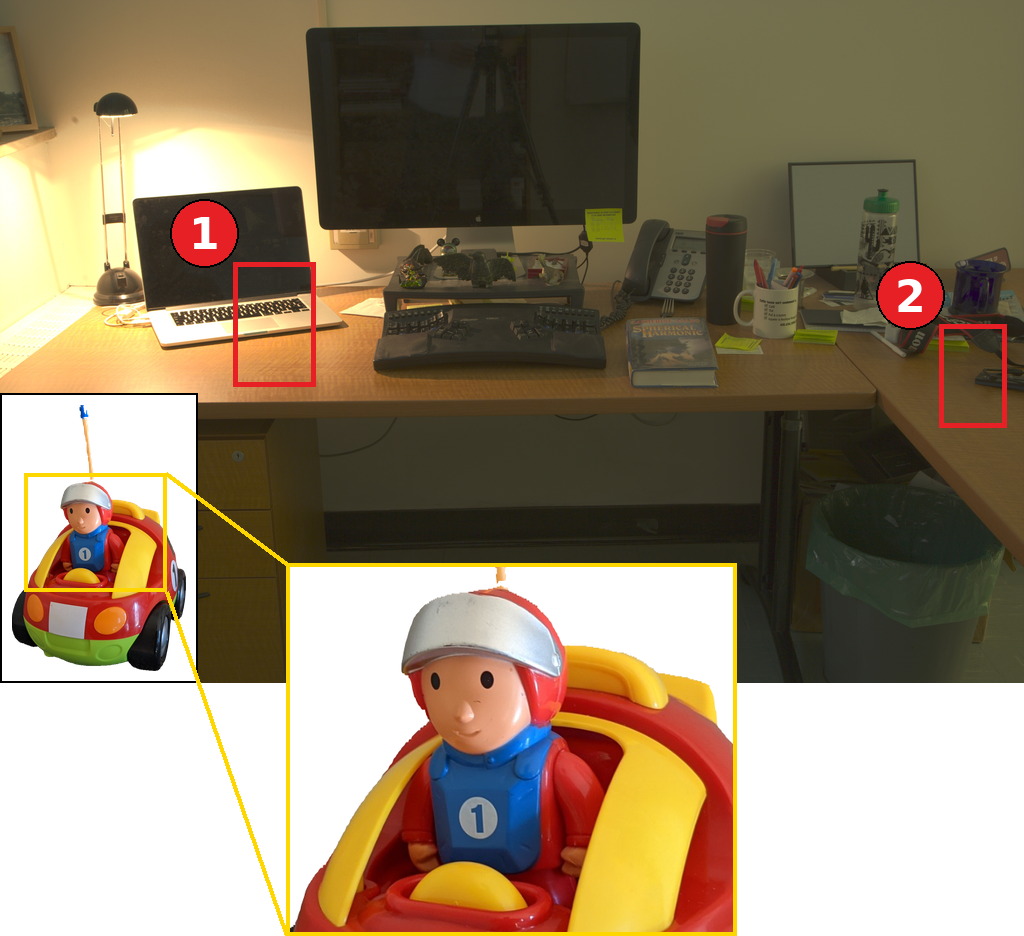}};
					\useasboundingbox (0, 0) rectangle (\linewidth, -0.667\linewidth);
					\coordinate (bboxbot) at (0.5\linewidth, -0.667\linewidth);
					\node[anchor=south, font=\scriptsize, text=black, fill=white, inner xsep=3pt, inner ysep=1pt, yshift=2pt] at (img.north) {Input};
				\end{tikzpicture}}%
			\rule{0pt}{0.667\linewidth}%
		\end{minipage}%
		\hfill
		\begin{minipage}[c]{0.54\linewidth}
			\centering
			\begin{tabular}{@{} c@{\hspace{2pt}}c @{}}
				\makecell{w/o Prior                                                                   \\ Comp. Data} &
				\modelname                                                                            \\
				\includegraphics[width=.49\linewidth]{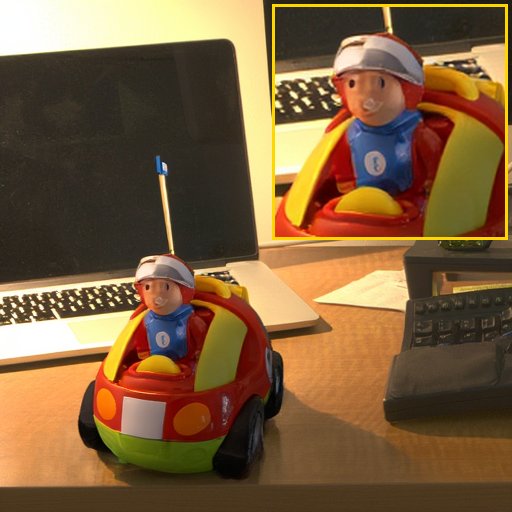} &
				\includegraphics[width=.49\linewidth]{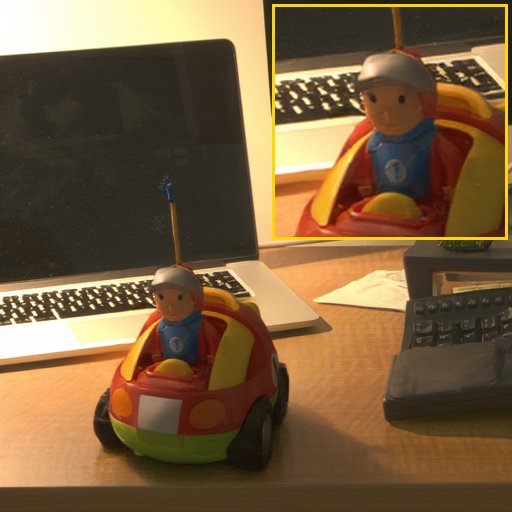}    \\
				\includegraphics[width=.49\linewidth]{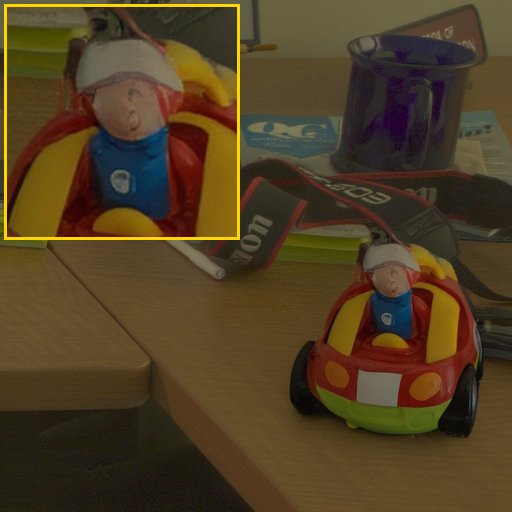} &
				\includegraphics[width=.49\linewidth]{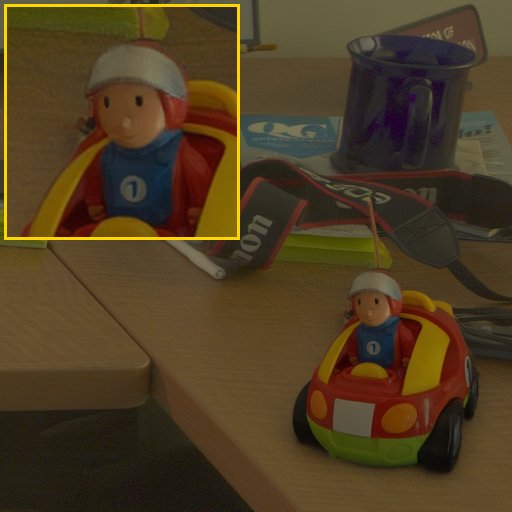}    \\
			\end{tabular}
		\end{minipage}
	\end{minipage}

	\caption{
		Qualitative ablation of illumination augmentation and prior compositing data.
		(Left) Illumination augmentation enables placement-dependent shading: without it, the inserted object remains brighter than the dim lighting at the insertion location and is inconsistent with the surrounding illumination gradients.
		(Right) Mixing prior compositing data better preserves details of the reference object (\eg, facial features and surface textures) that are less preserved with augmented triplets alone.
	}

	\label{fig:ablation_qual}
	\vspace{-10pt}
\end{figure}

We isolate the effect of illumination augmentation by varying the training data along two axes (Tab.~\ref{tab:data_source}):
(i) fine-tuning on our raw triplets, with and without illumination augmentation;
(ii) fine-tuning on prior compositing data, with and without our illumination-augmented triplets.
On raw triplets, augmentation lowers \benchname{} DreamSim from $0.372$ to $0.308$.
On prior compositing data, DreamSim decreases from $0.309$ to $0.265$ when augmented triplets are added.
This improvement is consistent with our data exhibiting a larger input--target appearance gap than prior datasets (Tab.~\ref{tab:appearance_gap}: $0.504$ vs.\ $0.17$--$0.30$), providing stronger supervision for lighting adaptation.
The reference-free Laval IC moves in the same direction across both axes (raw triplets $10.47\to14.51$ with augmentation; prior data $13.26\to15.04$).
Taken together, these results suggest complementary roles for the two data sources: illumination augmentation primarily improves adaptation to local lighting, while prior compositing data help preserve object identity and fine details.
Fig.~\ref{fig:ablation_qual} provides qualitative evidence
for this distinction. Without illumination augmentation,
the can retains similar shading across insertion locations,
appearing overly bright in the dim region despite the
surrounding illumination gradient. In contrast, training
without prior compositing data still adapts object appearance
to local lighting but less faithfully preserves fine details
of the toy figure, including its eyes, nose, helmet,
and the number on its shirt.

\begin{table}[tb]
	\centering

	\begin{minipage}[t]{0.56\linewidth}
		\centering
		\scriptsize

		\captionof{table}{
			Effect of illumination augmentation, applied either to our raw triplets (top) or mixed with prior compositing data (bottom).
			Each row accumulatively adds one component to the previous configuration.
		}
		\label{tab:data_source}
		\vspace{10pt}
		\begin{tabular}{lccc}
			\toprule
			                                                                                                              & Laval        & \multicolumn{2}{c}{\benchname}         \\
			\cmidrule(lr){2-2} \cmidrule(lr){3-4}
			                                                                                                              & IC$\uparrow$ &
			IC$\uparrow$                                                                                                  &
			DreamSim$\downarrow$                                                                                                                                                  \\
			\midrule
			Raw Triplets (No Aug.)                                                                                        & 10.47        & 9.24                           & 0.372 \\
			\quad + Illum. Aug.                                                                                           & 14.51        & 13.52                          & 0.308 \\
			\quad \makebox[0pt][l]{+ Prior Comp. Data}\hphantom{+ Our Illum. Aug. Data} $\rightarrow$ \textbf{\modelname} & 15.04        & 13.99                          & 0.265 \\
			\midrule
			Prior Comp. Data                                                                                              & 13.26        & 11.83                          & 0.309 \\
			\quad + Our Illum. Aug. Data $\rightarrow$ \textbf{\modelname}                                                & 15.04        & 13.99                          & 0.265 \\
			\bottomrule
		\end{tabular}
	\end{minipage}
	\hfill
	\begin{minipage}[t]{0.42\linewidth}
		\centering
		\scriptsize
		\setlength{\tabcolsep}{4pt}
		\captionof{table}{
			Input--target appearance gap measured by DreamSim distance between the input object and the ground-truth composite. Larger values indicate stronger supervision for appearance adaptation.
		}
		\label{tab:appearance_gap}
		\vspace{10pt}
		\begin{tabular}{lc}
			\toprule
			Dataset      & Appearance gap $\uparrow$ \\
			\midrule
			ORIDa        & 0.219                     \\
			AnyInsertion & 0.173                     \\
			DreamFuse    & 0.299                     \\
			\modelname{} & \textbf{0.504}            \\
			\bottomrule
		\end{tabular}
	\end{minipage}
\end{table}

\section{Conclusion}
\label{sec:conclusion}
We presented \modelname, a framework for local light-aware object compositing.
Treating monocular video frames as pose-anchored virtual spotlights can turn publicly available videos into illumination-augmented training data, equipping the compositing model with location-dependent shading.
By varying illumination during training, our approach provides supervision for adapting object appearance to local lighting without requiring physically staged captures for each lighting configuration.
Experiments on \benchname with ground-truth relit composites, Laval Indoor SV HDR, and ORIDa show that \modelname outperforms prior methods in illumination adaptation while maintaining object identity.
We will release the augmentation pipeline and \benchname{} to support further research on local light-aware compositing.

\textbf{Limitations:}
Our relighting pipeline relies on estimated scene geometry
and reflectance, which can introduce shading artifacts
for thin structures and highly specular surfaces.
The Disney BRDF models surface reflectance but not
transmission, leaving transparent and translucent
materials outside the scope of the augmentation.
Outdoor scenes pose a separate challenge, where a
distant sun produces near-parallel illumination rather
than the localized falloff modeled by our virtual spotlights.
These limitations are left for future work.

\begin{ack}
	This work was partly supported by Institute of Information \& communications Technology Planning \& Evaluation (IITP) grants funded by the Korea government (MSIT) (No.~RS-2022-II220156, Fundamental research on continual meta-learning for quality enhancement of casual videos and their 3D metaverse transformation; No.~RS-2026-25524173, Ultra-Long-Term Hierarchical Memory and Reasoning Architecture for Next-Generation Omnimodal Agents; No.~RS-2025-25442338, AI Star Fellowship Support Program (Seoul National Univ.); No.~RS-2025-02263841, Development of a Real-time Multimodal Framework for Comprehensive Deepfake Detection Incorporating Common Sense Error Analysis),
	by National IT Industry Promotion Agency (NIPA) grant funded by the Korea government (MSIT) (No.~RS-2026-25621677, Physical Intelligence Action Model and Integrated Framework Technology Development),
	and by the ``Advanced GPU Utilization Support Program'' funded by the Government of the Republic of Korea (Ministry of Science and ICT).
\end{ack}

\bibliographystyle{plainnat}
\bibliography{main}

\clearpage
\appendix
\begingroup
\hypersetup{hidelinks}
\renewcommand{\topfraction}{0.9}
\renewcommand{\bottomfraction}{0.8}
\renewcommand{\textfraction}{0.08}
\renewcommand{\floatpagefraction}{0.65}
\setlength{\textfloatsep}{12pt plus 2pt minus 2pt}
\setlength{\floatsep}{10pt plus 2pt minus 2pt}
\setlength{\intextsep}{10pt plus 2pt minus 2pt}

\section{Supplementary Material}
\label{app:suppl}

This supplementary material is organized into four groups.

\noindent
\begin{minipage}[t]{0.48\linewidth}
	\textbf{Method}
	\begin{itemize}[leftmargin=1.2em,itemsep=1pt,topsep=2pt]
		\item \S\ref{app:implementation} Implementation details.
		\item \S\ref{app:data_analysis} Illumination augmentation analysis.
	\end{itemize}
	\vspace{4pt}
	\textbf{Experiments}
	\begin{itemize}[leftmargin=1.2em,itemsep=1pt,topsep=2pt]
		\item \S\ref{app:baseline_analysis} Further baseline analysis.
		\item \S\ref{app:loco_analysis} Further analysis of \modelname.
		\item \S\ref{app:eval_protocol} Evaluation protocol details.
	\end{itemize}
\end{minipage}\hfill
\begin{minipage}[t]{0.48\linewidth}
	\textbf{Data \& benchmark}
	\begin{itemize}[leftmargin=1.2em,itemsep=1pt,topsep=2pt]
		\item \S\ref{app:dataset} \benchname{} construction.
		\item \S\ref{app:scene_prep} Scene preparation details.
		\item \S\ref{app:data_stats} Training data statistics.
		\item \S\ref{app:data_filtering} Training data filtering.
	\end{itemize}
	\vspace{4pt}
	\textbf{Discussion}
	\begin{itemize}[leftmargin=1.2em,itemsep=1pt,topsep=2pt]
		\item \S\ref{app:limitations} Limitations.
		\item \S\ref{app:broader_impact} Broader impact.
		\item \S\ref{app:image_attribution} Image attribution.
	\end{itemize}
\end{minipage}

\FloatBarrier
\subsection{Implementation Details}
\label{app:implementation}

This subsection collects the design and training-loop details needed to reproduce \modelname: the diffusion backbone, the training schedule, inference settings, and the parameters that drive the illumination-augmentation pipeline.

\vspace{4pt}
\noindent
\textbf{Architecture.}
Our base model is Flux1-Dev~\citep{flux2024}, fine-tuned with rank-16 LoRA~\citep{hu2022lora} in bfloat16.
The conditioning encoding and rectified-flow objective are described in Sec.~\ref{sec:training_strategies}.

\vspace{4pt}
\noindent
\textbf{Training.}
We first pre-train on the AnyDoor~\citep{chen2024anydoorzeroshotobjectlevelimage} training set for 10K iterations to obtain a generic generative-compositing prior.
Starting from this checkpoint, we fine-tune for 20K iterations on a 4:1 mixture of our illumination-augmented data (Sec.~\ref{sec:augmentation}) and the training sets of AnyDoor, AnyInsertion~\citep{song2026insert}, ORIDa~\citep{kim2025oridaobjectcentricrealworldimage}, and DreamFuse~\citep{huang2025dreamfuseadaptiveimagefusion}.
Both phases use a per-GPU batch size of 32 across 4$\times$B200 with gradient accumulation of 2, and AdamW~\citep{loshchilov2018decoupled} with learning rate $1{\times}10^{-4}$, $\beta_1{=}0.9$, $\beta_2{=}0.999$, and weight decay $1{\times}10^{-4}$.
The full schedule takes approximately 4 days.
Images are resized to $512\times512$.
The background crop is a square enclosing the bbox with side length $s \cdot \max(w_{\text{bbox}}, h_{\text{bbox}})$; during training, $s \sim \mathcal{U}[1.1, 10]$, set to span the scene-to-bbox scales observed across the training data.
During fine-tuning, we sample two reference frames per scene; with original and relit variants, this yields four (view, lighting) pairs, from which objects are cross-permuted to produce up to $4\times 4=16$ training triplets per scene.

\vspace{4pt}
\noindent
\textbf{Inference.}
We run 28 Euler sampling steps with guidance scale $3.5$.
The latent is initialized from Gaussian noise and denoised over these steps.
The background crop is taken at a fixed scale of $s = 1.5$.
To form the final composite, we decode the $512\times512$ crop prediction and blend its object region into the original background using the insertion mask, dilated and Gaussian-blurred at the boundary to smooth the transition.
Pixels outside the blending region retain the original background.
Higher-resolution refinement is discussed in Appendix~\ref{app:limitations}.

\Needspace{12\baselineskip}
\vspace{4pt}
\noindent
\textbf{BRDF formulation.}
The reflectance $R$ in Eq.~1 follows a simplified Disney BRDF~\citep{DisneyBRDF} with diffuse and specular terms,
\begin{equation}
	R(x,\omega_o,\omega_i) = \bigl(f_d(x) + f_s(x,\omega_o,\omega_i)\bigr)\,(\omega_i \cdot \mathbf{n}),
\end{equation}
where $\mathbf{n}$ is the surface normal at $x$.

The diffuse term $f_d$ is a Fresnel-modulated Lambertian lobe with surface albedo $\mathbf{a}$; the specular term $f_s$ follows the Cook--Torrance microfacet model with half vector $\mathbf{h} = \mathrm{normalize}(\omega_i + \omega_o)$ and roughness $r$,
\begin{equation}
	f_d(x) = \frac{(1 - F)\,\mathbf{a}(x)}{\pi},
	\qquad
	f_s(x,\omega_o,\omega_i) = \frac{D(\mathbf{h},\mathbf{n},r)\;F(\mathbf{h},\omega_o)\;G(\omega_i,\omega_o,\mathbf{n},r)}{4\,(\mathbf{n}\cdot\omega_i)\,(\mathbf{n}\cdot\omega_o)}.
\end{equation}
\Needspace{8\baselineskip}
The microfacet sub-terms use a spherical-Gaussian normal distribution $D$, the Schlick Fresnel approximation~\citep{schlick1994} $F$ with dielectric base reflectance $F_0 = 0.04$, and the Smith--Schlick--GGX geometry term $G$ with $k = r^2/2$,
\begin{align}
	D(\mathbf{h},\mathbf{n},r)        & = \frac{1}{\pi\,r^4}\,\exp\!\Bigl(\frac{2}{r^4}\,\bigl((\mathbf{h}\cdot\mathbf{n}) - 1\bigr)\Bigr), \\
	F(\mathbf{h},\omega_o)            & = F_0 + (1 - F_0)\,(1 - \mathbf{h}\cdot\omega_o)^5,                                                 \\
	G(\omega_i,\omega_o,\mathbf{n},r) & = \prod_{d\,\in\,\{i,o\}} \frac{\mathbf{n}\cdot\omega_d}{(\mathbf{n}\cdot\omega_d)(1-k)+k}.
\end{align}
All dot products are clamped to $[0,1]$ for numerical stability.

\vspace{4pt}
\noindent
\textbf{Lighting parameters.}
Beam shapes are drawn from triangles, rectangles, and circles, then perturbed by random translation, rotation, and scaling, and combined with horizontal, vertical, or diagonal stripe masks of randomized thickness for varied patterns.
Beam color is sampled in HSV: hue uniform in $[0,1]$, saturation in $[0, 0.35]$, value $1.0$, biasing the default sample toward near-white tones close to typical natural light; with probability $0.3$ we instead draw a vivid color with saturation in $[0.5, 1.0]$ to cover artificial colored lighting (\eg, smart bulbs, neon signage).
Since estimated reflectance and depth scale vary per scene, we set $\lambda_i$ adaptively: per light view, we scale the beam intensity so the $90$th-percentile relit pixel matches a fixed target, keeping apparent brightness comparable across scenes.
For distance attenuation, we use $\phi(x) = 1/\bigl(1 + d^{\gamma}\bigr)$, where $d$ is the source-to-surface distance and $\gamma > 0$ controls the rate of attenuation, ensuring a nonsingular falloff.
The ambient term $\alpha$ is sampled in $[10^{-3},\, 0.2]$.
From frames in which the object is visible, we sample up to 10 candidate light frames and apply them sequentially, accepting a candidate only if its lit region does not overlap with any previously accepted ones.
We also reject lighting configurations that brighten the foreground object but barely illuminate the surrounding background; the rejection criteria and relighting quality checks are detailed in Appendix~\ref{app:data_filtering}.

\FloatBarrier
\subsection{Analysis of Illumination Augmentation}
\label{app:data_analysis}

We analyze three properties of the augmented training data (spatially varying illumination, the input--target appearance gap, and lighting-direction coverage) and report the pipeline's computational footprint.
Together these properties characterize the supervision signal: the augmentation needs to
(a) produce spatially varying shading within each frame, so the model can infer local lighting cues at the insertion location,
(b) widen the input--target appearance gap so the model cannot copy-and-paste, and (c) cover diverse lighting directions so the cues do not collapse to a single direction.

\vspace{4pt}
\noindent
\textbf{Spatially varying illumination.}
We check two properties: (i) the augmentation actually changes the image, and (ii) the change is spatially varying rather than a uniform brightness shift.
For (i), pixel variance across multiple relit realizations of the same reference frame $I_{\text{ref}}$ increases from $0.00$ to $33.07$ with augmentation (Tab.~\ref{tab:variance}, top); Fig.~\ref{fig:relight_diversity} shows the same effect qualitatively: a single reference frame yields visibly different composites under different sampled lighting realizations.
For (ii), spatially varying illumination produces bright and dark regions within the same frame, fattening the tails of the intensity histogram.
We capture this with histogram contrast, the fraction of pixels beyond $\mu \pm 3\sigma$ computed per image.
With augmentation, contrast rises from $0.0047$ to $0.0151$, whereas the global brightness or color jitter common in prior data pipelines~\citep{Song_2023_CVPR,chen2024anydoorzeroshotobjectlevelimage,kim2025oridaobjectcentricrealworldimage} would yield $0$ contrast change.

\vspace{4pt}
\noindent
\textbf{Input--target appearance gap.}
Training data teaches lighting adaptation only when the input object's appearance differs from that in the target composite; otherwise, the supervision provides little signal for adaptation.
We measure this gap as DreamSim distance between the input object and the ground-truth composite, where larger values indicate stronger supervision.
Our triplets reach $0.504$, well above ORIDa~\citep{kim2025oridaobjectcentricrealworldimage}, AnyInsertion~\citep{song2026insert}, and DreamFuse~\citep{huang2025dreamfuseadaptiveimagefusion} (Tab.~\ref{tab:variance}, bottom).

\vspace{4pt}
\noindent
\textbf{Lighting-direction coverage.}
Fig.~\ref{fig:variance} plots the angular distribution of our sampled light-view directions at the insertion region, derived from the camera poses of the reference and light frames.
Frontal directions dominate, with additional coverage from side and above directions, while directions from below are rare, reflecting typical camera viewpoints in monocular videos.
This diversity provides varied lighting cues and prevents collapse to a single direction.

\begin{figure}[htbp]
	\centering
	\footnotesize

	\begin{minipage}[c]{0.42\linewidth}
		\centering

		\begin{minipage}{\linewidth}
			\centering
			\begin{tabular}{lcc}
				\toprule
				           & Pixel Var. $\uparrow$ & Contrast$ \uparrow$ \\
				\midrule
				No Aug.    & 0.00                  & 0.0047              \\
				\modelname & 33.074                & 0.0151              \\
				\bottomrule
			\end{tabular}
		\end{minipage}

		\vspace{6pt}

		\begin{minipage}{\linewidth}
			\centering
			\begin{tabular}{lc}
				\toprule
				             & Appearance gap $\uparrow$ \\
				\midrule
				ORIDa        & 0.219                     \\
				AnyInsertion & 0.173                     \\
				DreamFuse    & 0.299                     \\
				\modelname   & 0.504                     \\
				\bottomrule
			\end{tabular}
		\end{minipage}
		\captionof{table}{
			(Top) Pixel variance and contrast across relit realizations.
			(Bottom) Input--target appearance gap (DreamSim); reproduced from Tab.~\ref{tab:appearance_gap}. Higher values indicate stronger variation and a wider gap, respectively.
		}
		\label{tab:variance}
	\end{minipage}
	\hfill
	\begin{minipage}[c]{0.54\linewidth}
		\centering
		\includegraphics[width=\linewidth]{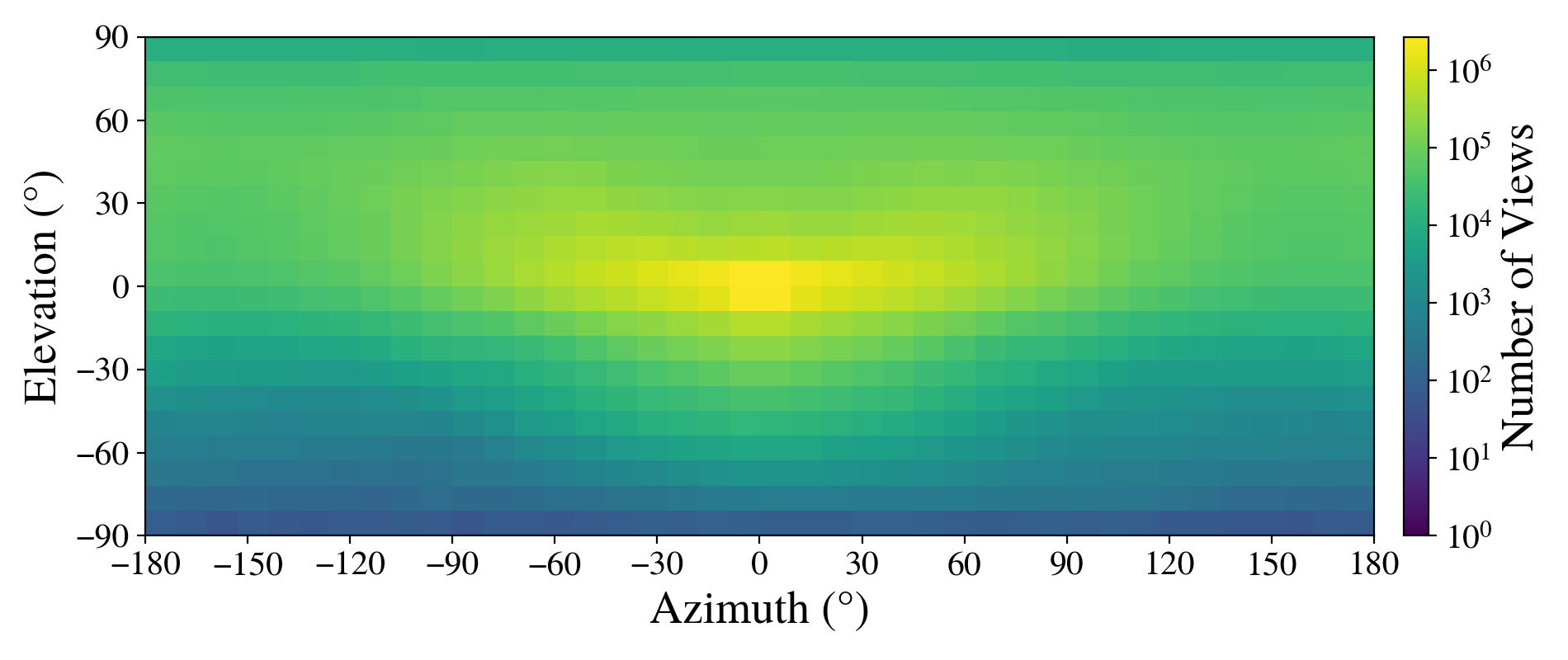}
		\caption{
			Sampled light-view directions, with frontal views dominant and additional coverage from lateral and oblique directions.
		}
		\label{fig:variance}
	\end{minipage}
\end{figure}

\begin{figure}[htbp]
	\centering
	\scriptsize
	\begin{tabular}{@{} c@{\hspace{2pt}}c @{\hspace{10pt}} c@{\hspace{2pt}}c @{}}
		\makecell{B.G. \& BBox}                                                                         & G.T.                                                                                              & \makecell{B.G. \& BBox}                                                                         & G.T.                                                                                              \\
		\includegraphics[width=.22\linewidth]{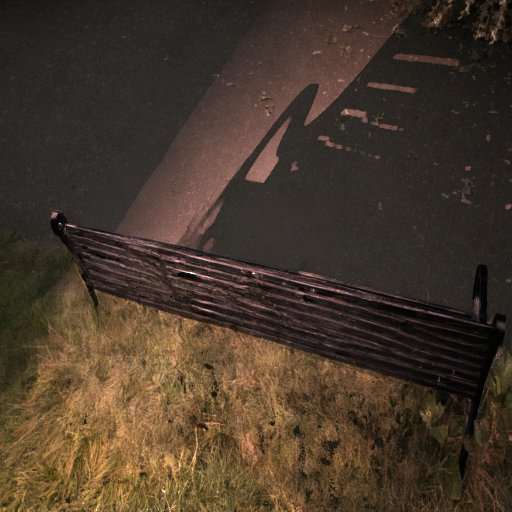} & \includegraphics[width=.22\linewidth]{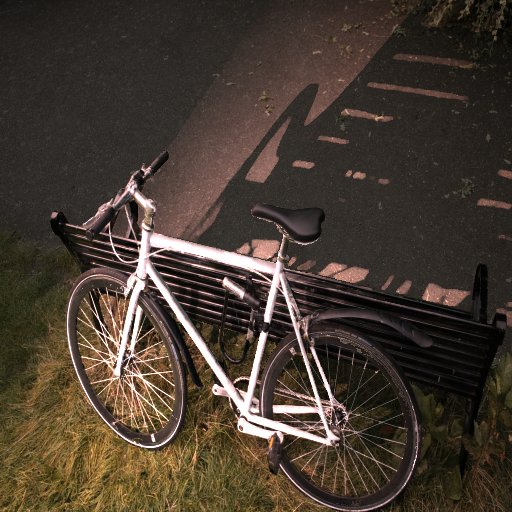} & \includegraphics[width=.22\linewidth]{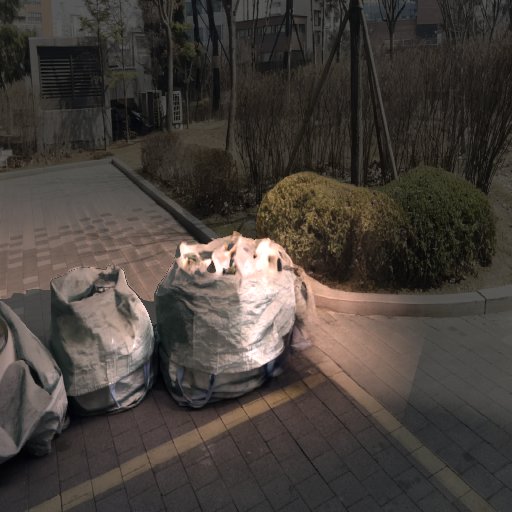} & \includegraphics[width=.22\linewidth]{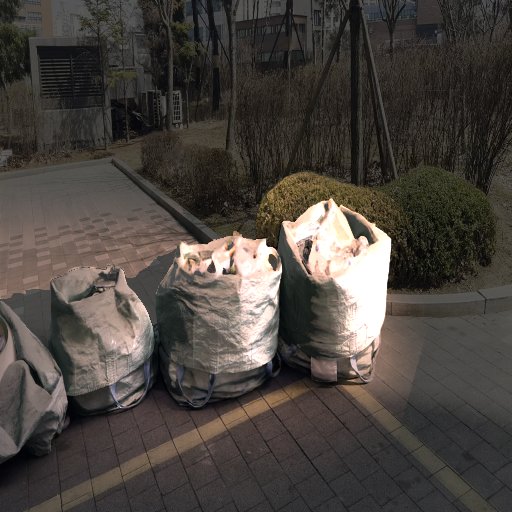} \\
		\includegraphics[width=.22\linewidth]{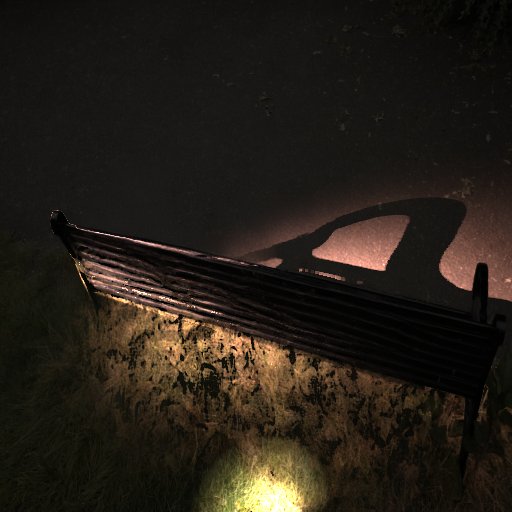} & \includegraphics[width=.22\linewidth]{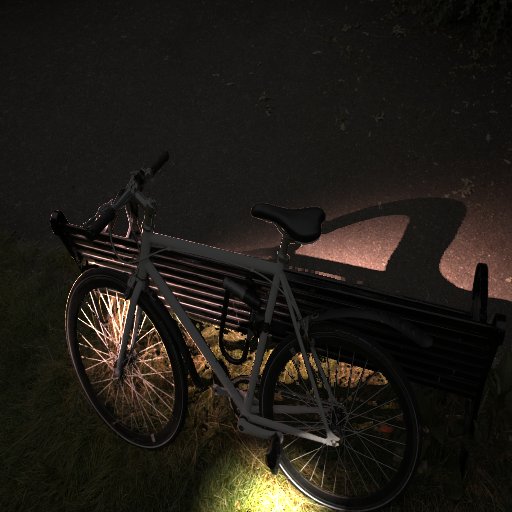} & \includegraphics[width=.22\linewidth]{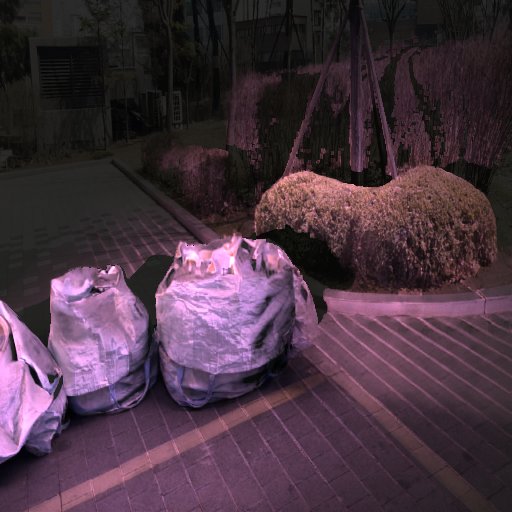} & \includegraphics[width=.22\linewidth]{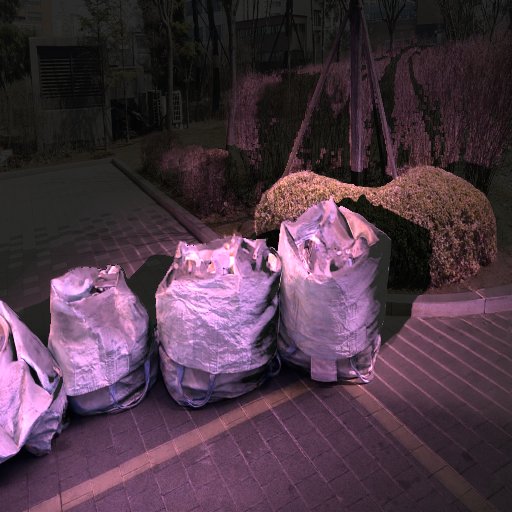} \\
		\includegraphics[width=.22\linewidth]{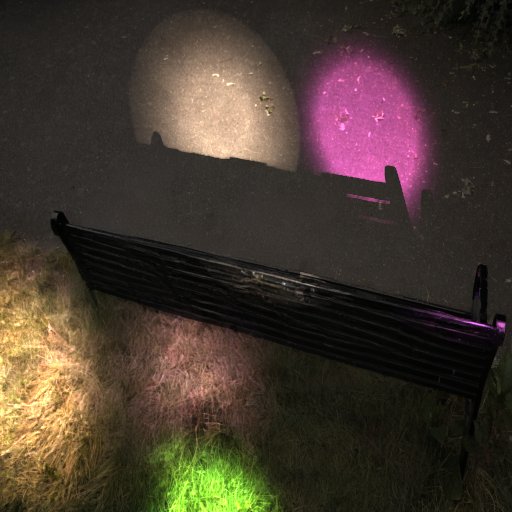} & \includegraphics[width=.22\linewidth]{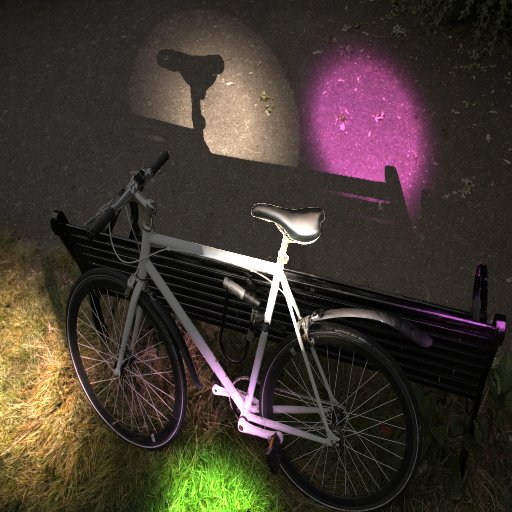} & \includegraphics[width=.22\linewidth]{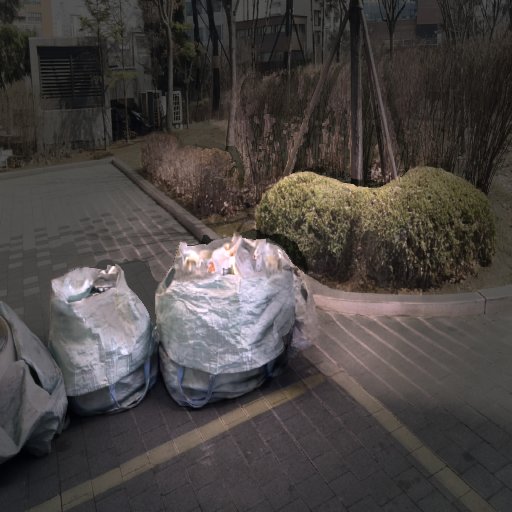} & \includegraphics[width=.22\linewidth]{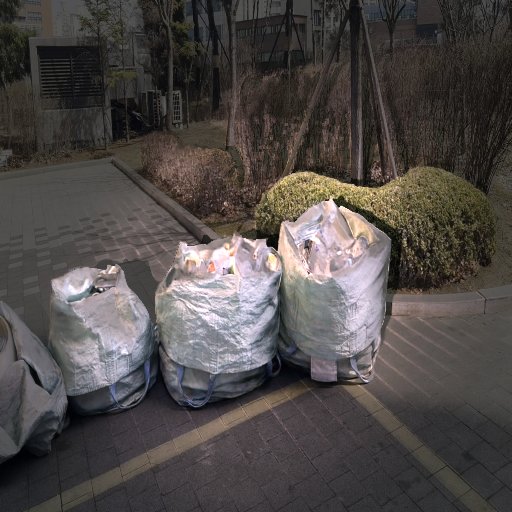} \\
		\includegraphics[width=.22\linewidth]{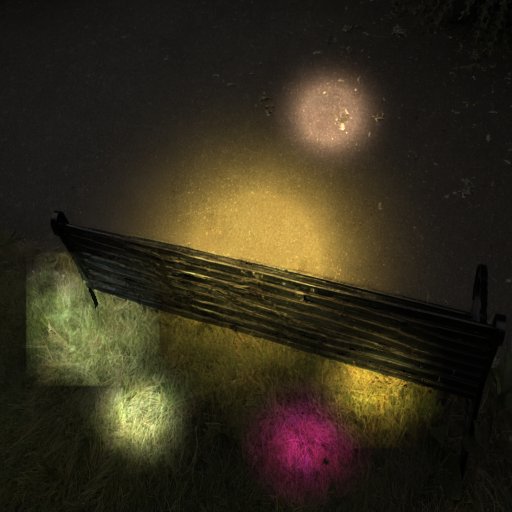} & \includegraphics[width=.22\linewidth]{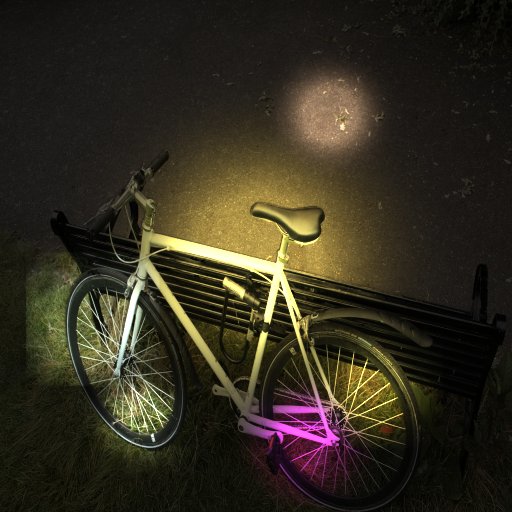} & \includegraphics[width=.22\linewidth]{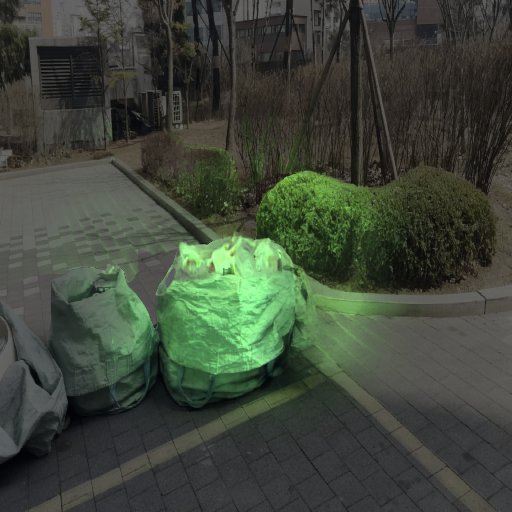} & \includegraphics[width=.22\linewidth]{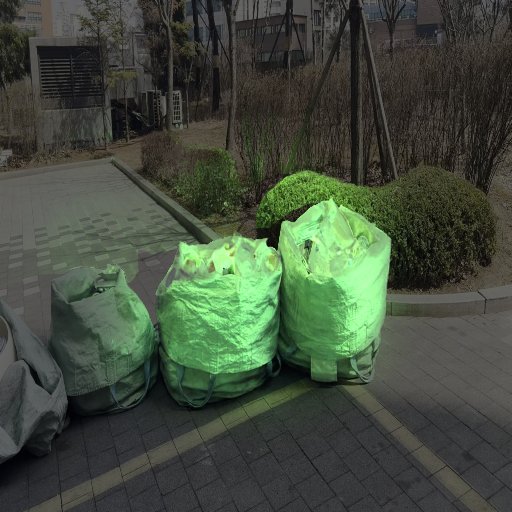} \\
	\end{tabular}
	\caption{
		Appearance diversity from a single reference frame under sampled lighting.
		Each panel shows one reference frame; the four rows are independent realizations of $(I_{\text{bg}}, I_{\text{comp}})$ produced by the augmentation pipeline (Sec.~\ref{sec:augmentation}), differing only in the sampled spotlight.
	}
	\label{fig:relight_diversity}
\end{figure}

\vspace{4pt}
\noindent
\textbf{Computational cost.}
The augmentation pipeline runs on the fly within the training loop rather than as a separate offline stage.
We measure wall-clock cost on a single NVIDIA B200: each iteration loads two reference frames with 5 sampled light views, yielding 16 training triplets (Flux1-Dev with rank-16 LoRA in bf16; 45 measured steps after 5 warmup).
Per iteration, augmentation takes a median of $83$\,ms while the full step (augmentation, forward, backward, parameter update) takes $5.40$\,s; augmentation occupies approximately $1.5\%$ of step time.
Peak GPU memory growth during augmentation is $1.08$\,GB (median; $1.11$\,GB max), small relative to the model's training footprint.
Under the full training batch (two scenes per GPU, $32$ variants), costs scale near-linearly: median augmentation $168$\,ms and full step $10.74$\,s, with a similar ${\sim}1.5\%$ ratio.
This low overhead allows augmentation to run in the loop, exposing the model to freshly relit triplets at each step.

\FloatBarrier
\subsection{Further Baseline Analysis}
\label{app:baseline_analysis}
This subsection extends the baseline coverage of the main paper by evaluating generative compositing baselines at different background-context crop scales and comparing against harmonization, intrinsic-based editing, and general-purpose multi-reference editing.

\vspace{4pt}
\noindent
\textbf{Effect of background context scale.}
Background crop scale affects methods that operate on local context.
We evaluate the generative compositing baselines at four scales ($1.5\times$, $3.0\times$, $4.5\times$, and $6.0\times$ the bbox side) on the evaluation set of ORIDa (identity) and \benchname{} (illumination).
Identity is measured using CLIP ViT-L/14~\citep{clip} and DINOv3 ViT-H/16~\citep{simeoni2025dinov3}.
Results are reported in Tab.~\ref{tab:bg_scale_orida} and Tab.~\ref{tab:bg_scale_bench}; main-paper comparisons use each method's best-performing scale.

\begin{table}[htbp]
	\centering
	\footnotesize
	\setlength{\tabcolsep}{3pt}
	\caption{
		Effect of background context scale on identity preservation (ORIDa).
		Each baseline is evaluated across four crop scales; the best-performing scale is used in the main paper.
	}
	\label{tab:bg_scale_orida}
	\vspace{2pt}
	\begin{tabular}{l*{4}{cc}}
		\toprule
		                     & \multicolumn{2}{c}{$1.5\times$} & \multicolumn{2}{c}{$3.0\times$} & \multicolumn{2}{c}{$4.5\times$} & \multicolumn{2}{c}{$6.0\times$}                                                                     \\
		\cmidrule(lr){2-3}\cmidrule(lr){4-5}\cmidrule(lr){6-7}\cmidrule(lr){8-9}
		                     & CLIP$\uparrow$                  & DINO$\uparrow$                  & CLIP$\uparrow$                  & DINO$\uparrow$                  & CLIP$\uparrow$ & DINO$\uparrow$ & CLIP$\uparrow$ & DINO$\uparrow$ \\
		\midrule
		AnyDoor              & 86.6                            & 79.4                            & 86.0                            & 74.6                            & 86.1           & 74.5           & 86.0           & 74.5           \\
		Personalize Anything & 74.4                            & 52.2                            & 71.7                            & 34.5                            & 69.9           & 28.7           & 69.4           & 27.6           \\
		Insert Anything      & 90.9                            & 92.1                            & 91.1                            & 89.4                            & 90.1           & 88.9           & 90.0           & 90.4           \\
		UniCombine           & 86.8                            & 85.0                            & 87.9                            & 86.3                            & 86.2           & 81.4           & 83.9           & 75.1           \\
		DreamFuse            & 83.1                            & 67.7                            & 79.2                            & 57.4                            & 76.3           & 46.7           & 76.6           & 46.6           \\
		OmniPaint            & 89.3                            & 87.2                            & 89.4                            & 87.2                            & 87.8           & 85.1           & 87.0           & 84.0           \\
		FreeCompose          & 81.0                            & 78.2                            & 79.7                            & 71.3                            & 80.1           & 71.8           & 80.0           & 72.2           \\
		\bottomrule
	\end{tabular}
\end{table}

\begin{table}[htbp]
	\centering
	\footnotesize
	\setlength{\tabcolsep}{4pt}
	\caption{
		Effect of background context scale on appearance adaptation (\benchname).
		Each baseline is evaluated across four crop scales; the best-performing scale is used in the main paper.
	}
	\label{tab:bg_scale_bench}
	\vspace{2pt}
	\begin{tabular}{l*{4}{c}}
		\toprule
		                     & $1.5\times$          & $3.0\times$          & $4.5\times$          & $6.0\times$          \\
		\cmidrule(lr){2-5}
		                     & DreamSim$\downarrow$ & DreamSim$\downarrow$ & DreamSim$\downarrow$ & DreamSim$\downarrow$ \\
		\midrule
		AnyDoor              & 0.477                & 0.461                & 0.476                & 0.501                \\
		Personalize Anything & 0.419                & 0.461                & 0.489                & 0.513                \\
		Insert Anything      & 0.359                & 0.323                & 0.313                & 0.312                \\
		UniCombine           & 0.350                & 0.342                & 0.399                & 0.467                \\
		DreamFuse            & 0.335                & 0.351                & 0.377                & 0.394                \\
		OmniPaint            & 0.357                & 0.349                & 0.365                & 0.393                \\
		FreeCompose          & 0.393                & 0.410                & 0.430                & 0.448                \\
		\bottomrule
	\end{tabular}
\end{table}

\FloatBarrier
\vspace{4pt}
\noindent
\textbf{Harmonization and intrinsic-based editing.}
To extend baseline coverage beyond the generative compositing methods in the main paper, we add two complementary paradigms:
(1) \textit{harmonization}: PIH~\citep{wang2023semi}, which transfers color and tone to match the surrounding scene;
(2) \textit{intrinsic-based editing}: RGBX~\citep{zeng2024rgb}, ZeroComp~\citep{zhang2025zerocomp}, and DiffusionRenderer~\citep{DiffusionRenderer}, which decompose the image into intrinsics (\eg, albedo, shading) and re-render under target lighting.
Both paradigms run on full images so they retain global lighting context.
We use official weights and default hyperparameters for every baseline.

\Needspace{4\baselineskip}
Tab.~\ref{tab:additional_baselines} reports IC and DreamSim~\citep{fu2023dreamsim} on \benchname{} under the same protocol as the main paper, and Fig.~\ref{fig:additional_baselines} aligns the same insertion under each method.
On IC, \modelname{} scores higher than all four baselines, exceeding the next-best ZeroComp by $0.29$\,PSNR; on DreamSim, \modelname{} ($0.265$) is lower than every baseline ($0.351$--$0.554$).

\begin{table}[htbp]
	\centering
	\footnotesize
	\caption{Quantitative comparison with harmonization and intrinsic-based editing methods on \benchname.}
	\label{tab:additional_baselines}
	\vspace{2pt}
	\begin{tabular}{lcc}
		\toprule
		                    & IC$\uparrow$ & DreamSim$\downarrow$ \\
		\midrule
		PIH                 & 11.79        & 0.351                \\
		RGBX                & 13.13        & 0.554                \\
		ZeroComp            & 13.70        & 0.447                \\
		DiffusionRenderer   & 12.38        & 0.389                \\
		\midrule
		\textbf{\modelname} & 13.99        & 0.265                \\
		\bottomrule
	\end{tabular}
\end{table}

\begin{figure}[htbp]
	\centering
	\scriptsize
	\begin{tabular}{@{} *{6}{c@{\hspace{2pt}}} @{}}
		\makecell{Input}                                                                                         &
		\makecell{PIH}                                                                                           &
		\makecell{RGBX}                                                                                          &
		\makecell{ZeroComp}                                                                                      &
		\makecell{Diffusion                                                                                        \\Renderer} &
		\makecell{\modelname}                                                                                      \\

		\includegraphics[width=.16\linewidth]{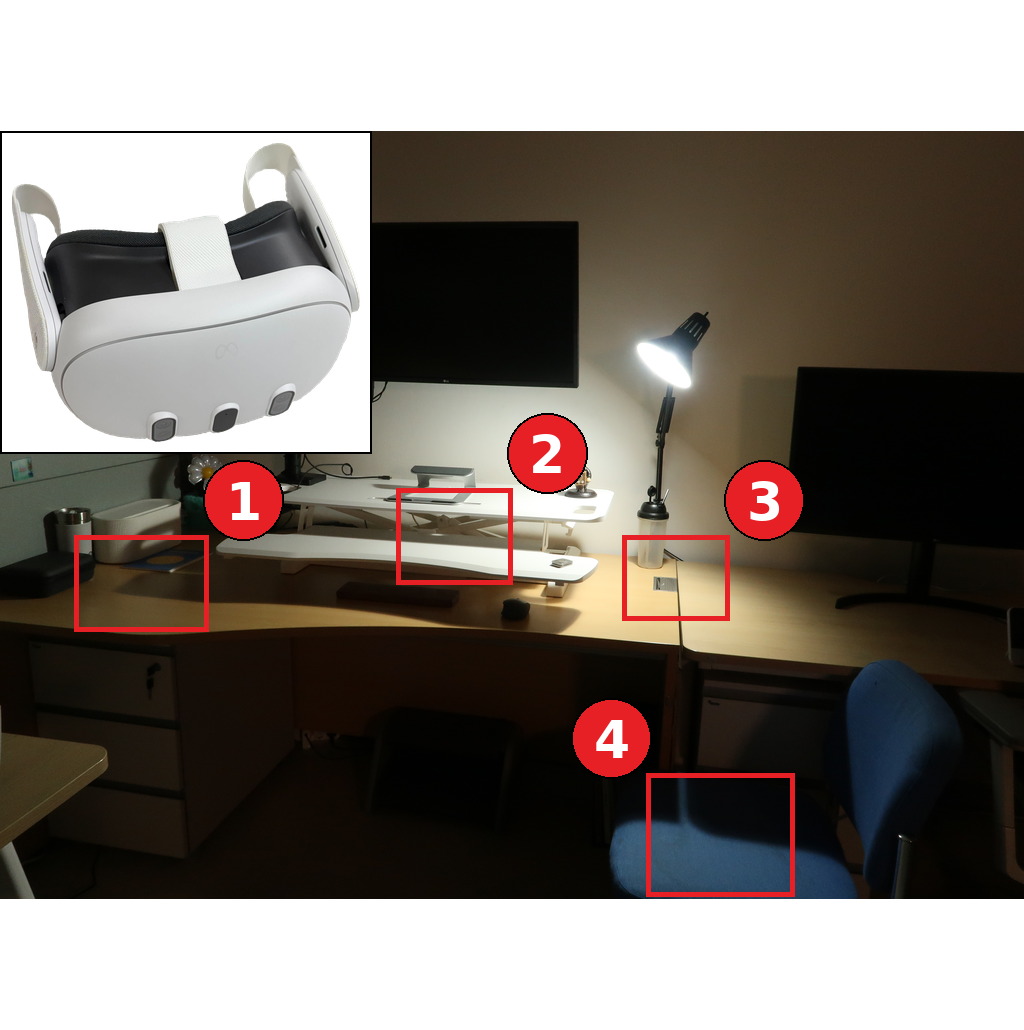}             &
		\includegraphics[width=.16\linewidth]{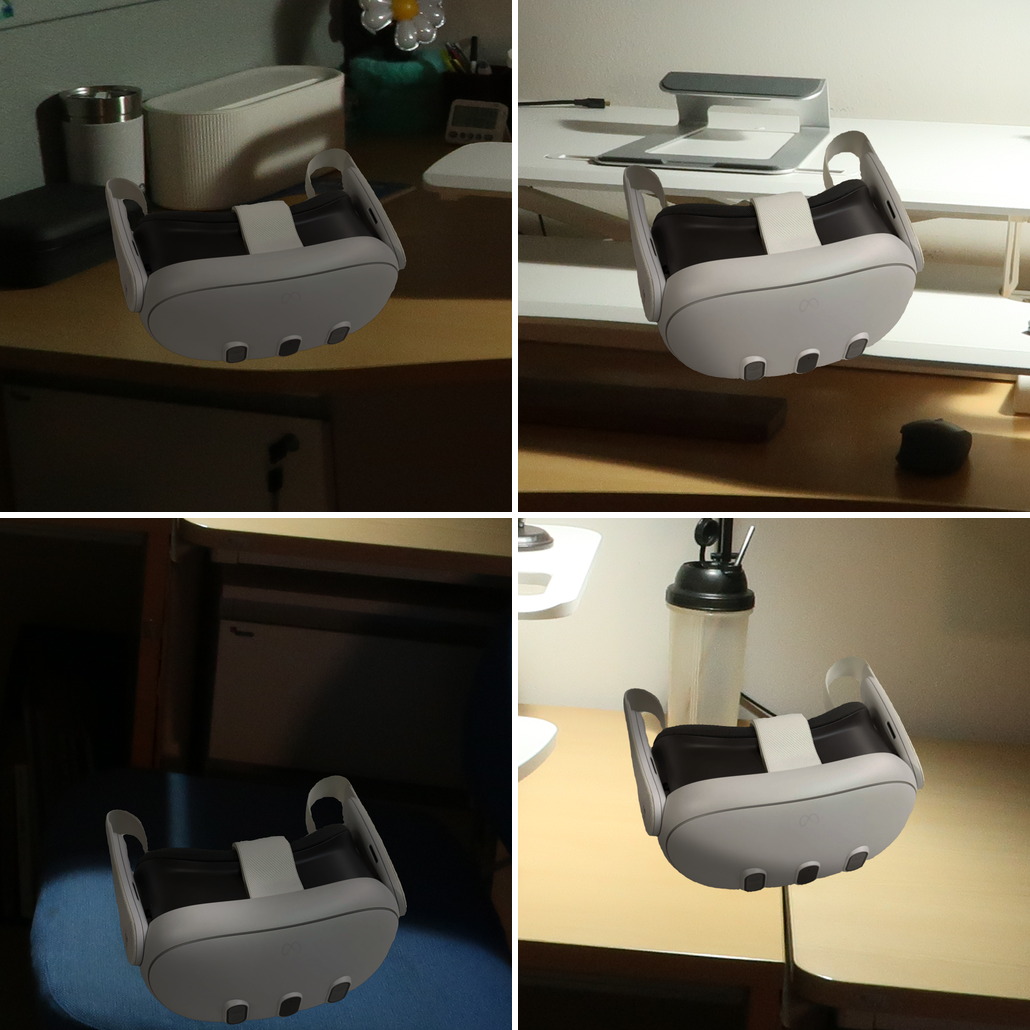}               &
		\includegraphics[width=.16\linewidth]{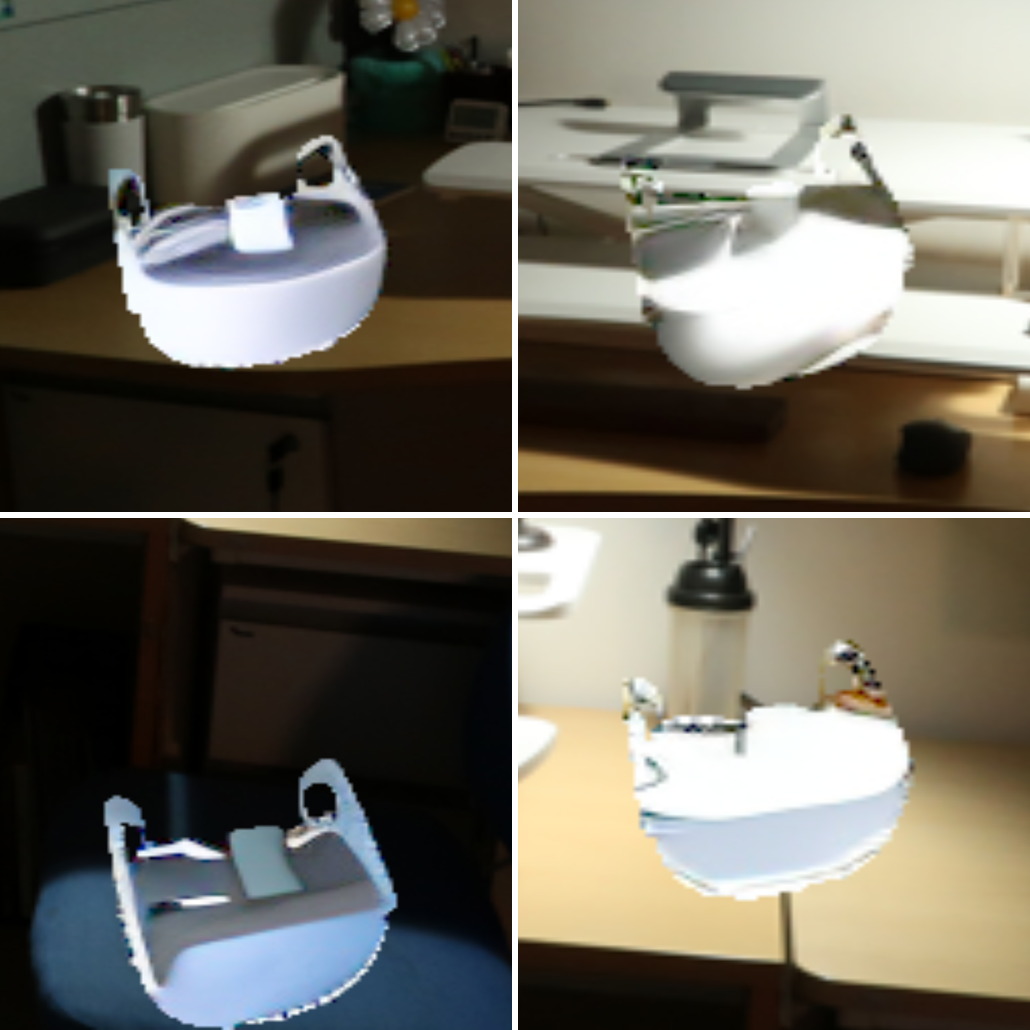}              &
		\includegraphics[width=.16\linewidth]{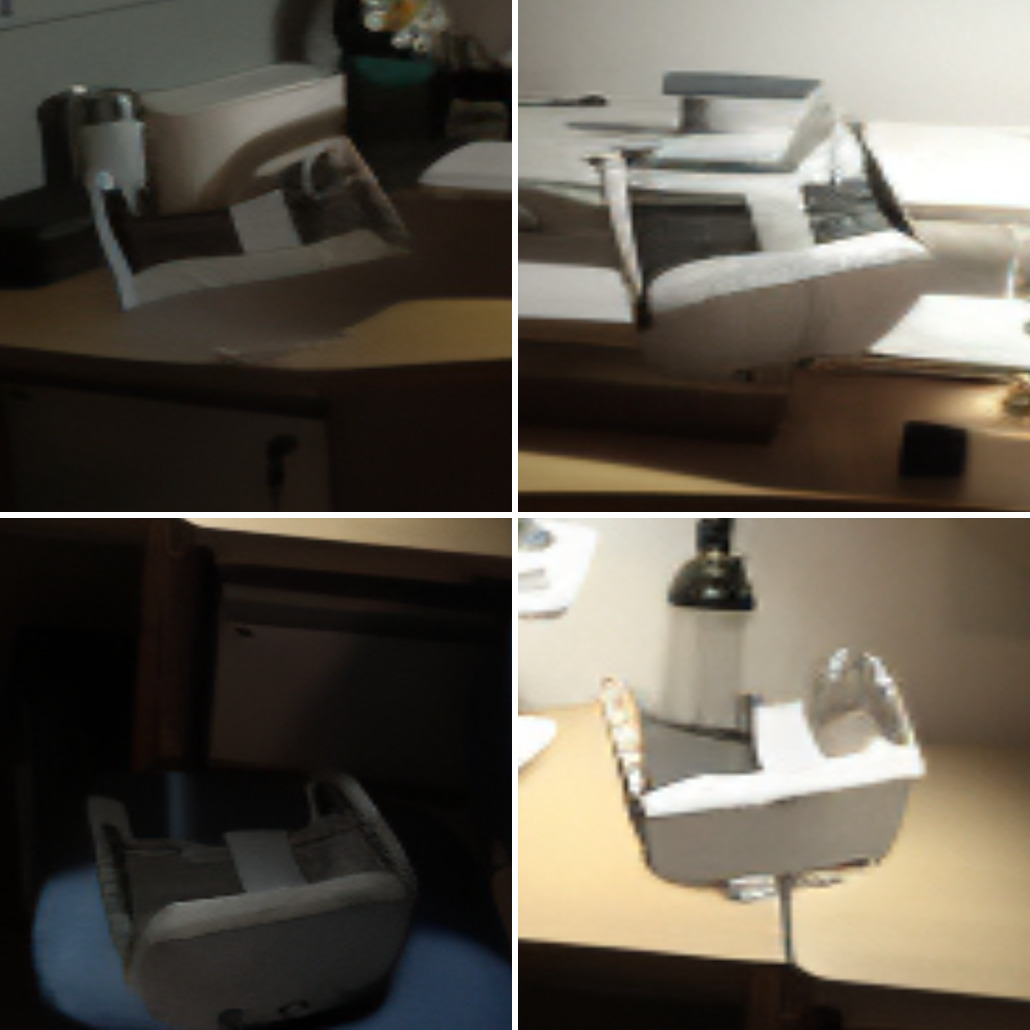}          &
		\includegraphics[width=.16\linewidth]{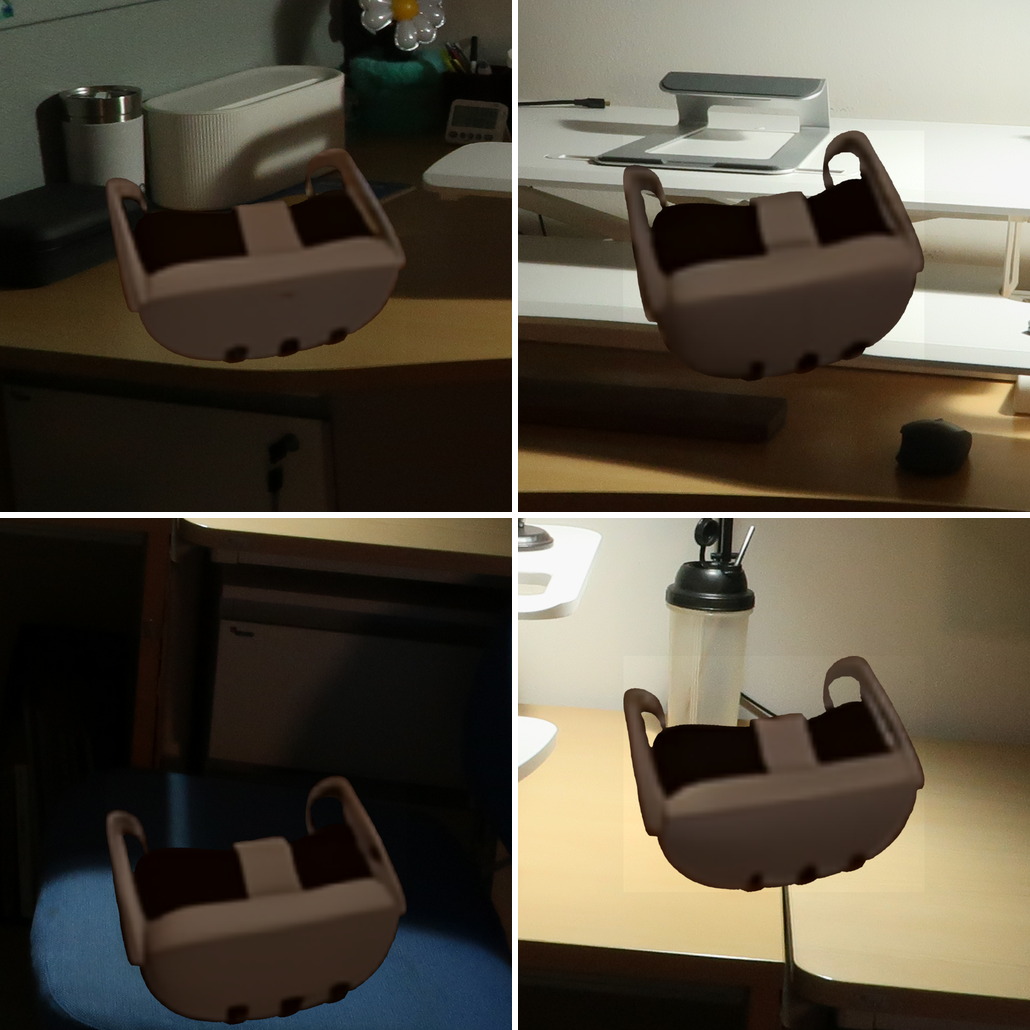} &
		\includegraphics[width=.16\linewidth]{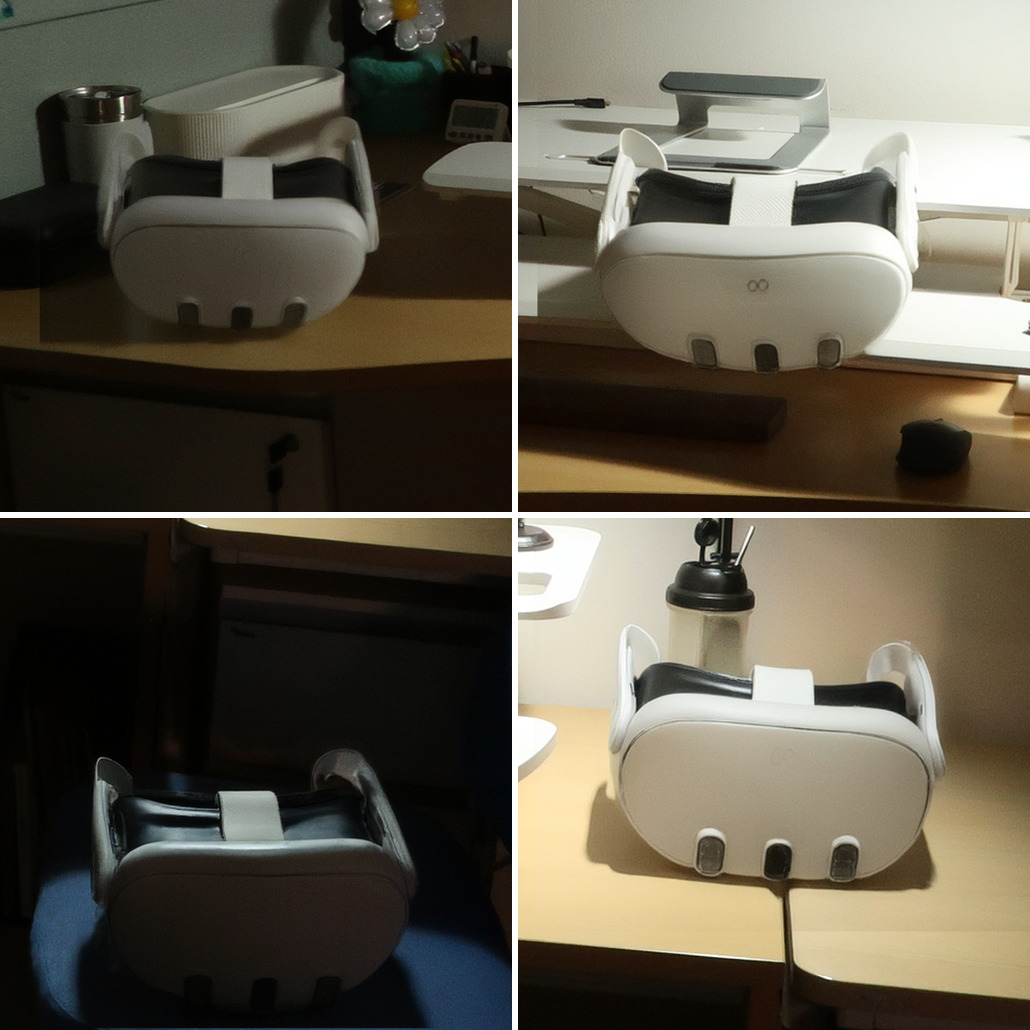}                \\
	\end{tabular}
	\caption{
		PIH adjusts the overall color and tone of the inserted object but leaves shading and material cues unchanged.
		RGBX, ZeroComp, and DiffusionRenderer adjust brightness and shading through inverse-then-forward rendering, but do not explicitly model geometric placement on the supporting surface, so the inserted object often appears visually detached and lacks contact cues such as cast shadows.
	}
	\label{fig:additional_baselines}
\end{figure}

\FloatBarrier
\vspace{4pt}
\noindent
\textbf{General-purpose multi-reference editing.}
We evaluate FLUX-2-Klein-9B~\citep{flux2klein2026} using two reference images: a background crop with the insertion bounding box marked in red, and a masked object reference.
The model is instructed to insert the reference object at the marked location, adapt its brightness, shading, color, and cast shadow to the local lighting, remove the red marker, and preserve the rest of the scene.
The full prompt is:
\begin{quote}
	Insert the object from the second image into the first image, exactly inside the region marked by the red box. Place it naturally on the supporting surface. Adapt the object's brightness, shading, color, and cast shadow to match the local lighting at that spot in the scene. Remove the red box marker. Do not change anything else in the scene.
\end{quote}
FLUX-2-Klein-9B achieves a lower DreamSim distance than all specialized compositing baselines in Tab.~\ref{tab:quantitative_comparison} ($0.293$ vs.\ the best baseline's $0.312$).
\modelname{} achieves a lower DreamSim distance ($0.265$) and higher IC on both benchmarks (Tab.~\ref{tab:general_editor}).

\begin{table}[htbp]
	\centering
	\footnotesize
	\caption{Comparison with a general-purpose multi-reference editor.}
	\label{tab:general_editor}
	\begin{tabular}{lccc}
		\toprule
		Method              & Laval IC$\uparrow$ & \benchname{} IC$\uparrow$ & \benchname{} DreamSim$\downarrow$ \\
		\midrule
		FLUX-2-Klein-9B     & 12.48              & 11.05                     & 0.293                             \\
		\textbf{\modelname} & \textbf{15.04}     & \textbf{13.99}            & \textbf{0.265}                    \\
		\bottomrule
	\end{tabular}
\end{table}

\clearpage
\subsection{Further Analysis of \modelname}
\label{app:loco_analysis}

This subsection examines the effect of the video training corpus, followed by qualitative comparisons against the main-paper baselines, an estimated-irradiance analysis of shading consistency, and comparisons with captured ground truth on challenging \benchname{} samples.

\vspace{4pt}
\noindent
\textbf{Contribution of Public Video Data.}
\phantomsection
\label{app:video_source_analysis}
The augmentation pipeline can use ordinary monocular videos, ranging from a small set of casual captures to a large public corpus.
Here we examine whether a small collection of casual captures provides effective training supervision and what is gained by expanding to a larger public video corpus, complementing the training-data ablation in Sec.~\ref{sec:ablation}.

Tab.~\ref{tab:video_source_ablation} compares three fine-tuning configurations: prior compositing data alone; those data combined with augmented triplets from our 51 casual captures; and the full training mixture, which additionally uses the public video datasets in Tab.~\ref{tab:data_stats}.
The casual-capture configuration uses no public videos for augmentation; the full configuration adds DL3DV, uCO3D, Mip-NeRF 360, and Tanks\&Temples together.
All configurations use the same initial checkpoint trained on the AnyDoor dataset and retain the same prior compositing datasets.
Both video configurations use the same augmentation and light/view permutation pipeline.
The casual captures are separate from \benchname{} and contribute 12{,}433 of the 670{,}899 unaugmented training triplets (1.9\%).

Training with prior compositing data alone gives results close to the best generative baseline scores listed in the table.
With augmented triplets from only our 51 casual captures, the model surpasses these references on all three metrics: Laval IC increases from $13.26$ to $14.65$, \benchname{} IC from $11.83$ to $13.50$, and \benchname{} DreamSim improves from $0.309$ to $0.288$.
Adding the public videos further improves these metrics to $15.04$, $13.99$, and $0.265$, respectively.
Most of the IC improvement is already achieved with the augmented casual captures, while adding public videos yields a further DreamSim reduction comparable to the initial improvement ($0.309\to0.288$ and $0.288\to0.265$, respectively).
These results show that our pipeline provides effective illumination supervision from a small collection of captures, while a larger and more diverse video corpus further improves perceptual fidelity.

\begin{table}[htbp]
	\centering
	\caption{
		Effect of the video collection used for augmented training triplets.
		All three training configurations start from the same checkpoint trained on the AnyDoor dataset and retain the prior compositing data from Tab.~\ref{tab:data_source}.
		The two video configurations use the same illumination augmentation and light/view permutation pipeline.
		The reference baseline row combines the per-column best generative compositing results from Tab.~\ref{tab:quantitative_comparison}: DreamFuse for Laval IC, UniCombine for \benchname{} IC, and Insert Anything for DreamSim.
	}
	\label{tab:video_source_ablation}
	\footnotesize
	\vspace{2pt}
	\begin{tabularx}{\linewidth}{@{}Xccc@{}}
		\toprule
		                                                            & Laval          & \multicolumn{2}{c}{\benchname}                        \\
		\cmidrule(lr){2-2}\cmidrule(lr){3-4}
		Video collection used for augmentation                      & IC$\uparrow$   & IC$\uparrow$                   & DreamSim$\downarrow$ \\
		\midrule
		Best generative baseline (per column)                       & 13.27          & 11.80                          & 0.312                \\
		\midrule
		None (prior compositing data only)                          & 13.26          & 11.83                          & 0.309                \\
		Our 51 casual captures                                      & 14.65          & 13.50                          & 0.288                \\
		Full corpus: casual captures and public videos (\modelname) & \textbf{15.04} & \textbf{13.99}                 & \textbf{0.265}       \\
		\bottomrule
	\end{tabularx}
\end{table}

\vspace{4pt}
\noindent
\textbf{Additional composite samples.}
Fig.~\ref{fig:composite_additional} extends the main-paper comparison (Fig.~\ref{fig:composite}) with additional Laval and \benchname{} scenes, where \modelname{} adapts object appearance to local illumination across insertion locations with consistent changes in brightness and cast shadow.

\begin{figure}[htbp]
	\centering
	\scriptsize
	\renewcommand{\arraystretch}{1}

	\begin{minipage}{\linewidth}
		\centering
		\begin{tabular}{@{} *{5}{c@{\hspace{3pt}}} @{}}
			                                                                                                                  &
			\makecell{AnyDoor}                                                                                                &
			\makecell{Personalize                                                                                               \\Anything} &
			\makecell{Insert                                                                                                    \\Anything} &
			\modelname                                                                                                          \\

			\inputimg{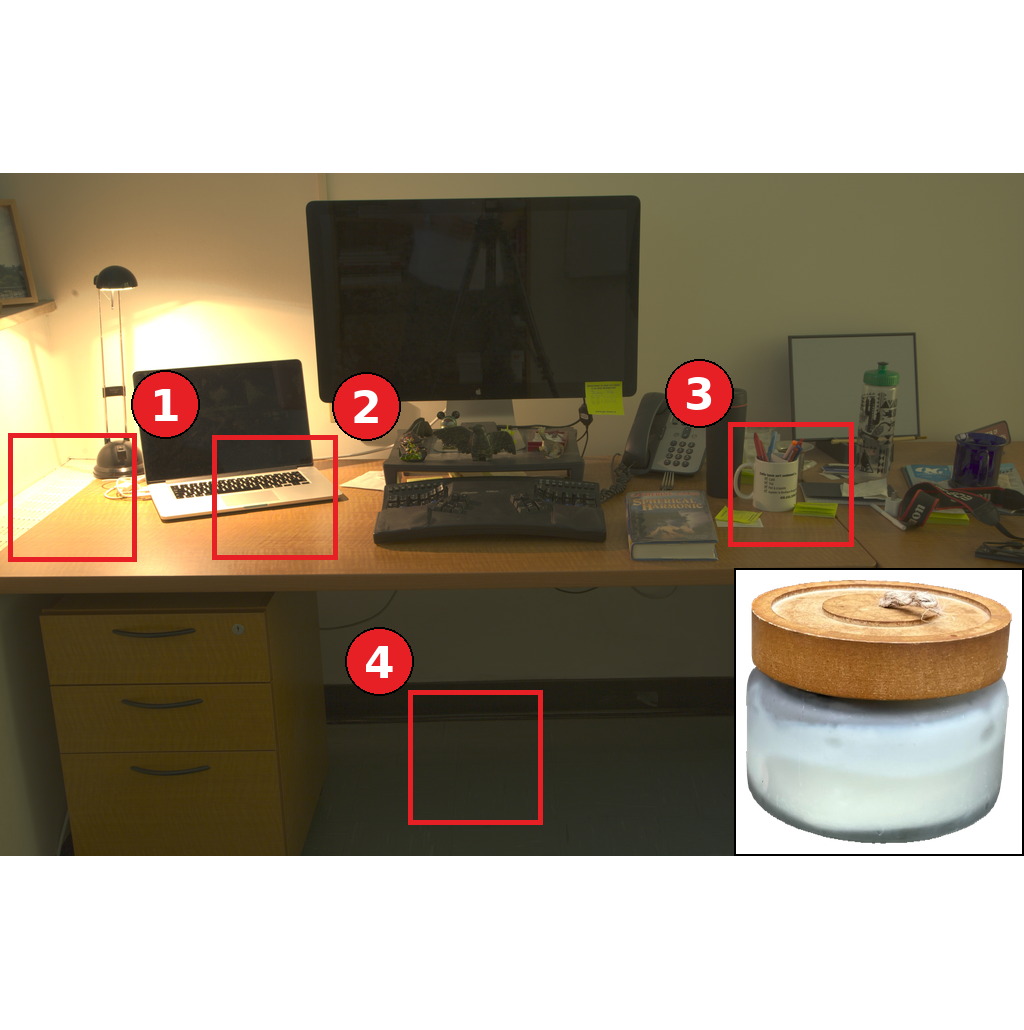}                                            &
			\includegraphics[width=.192\linewidth]{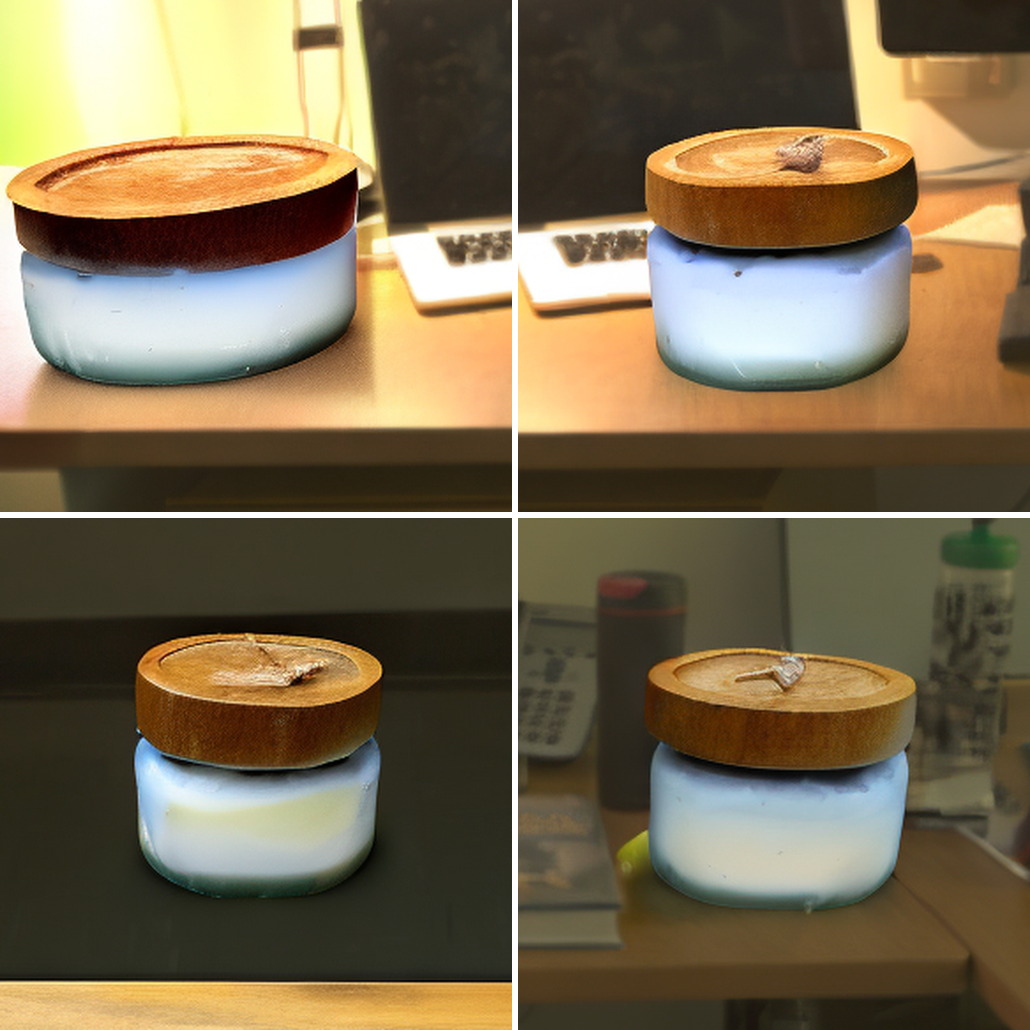}             &
			\includegraphics[width=.192\linewidth]{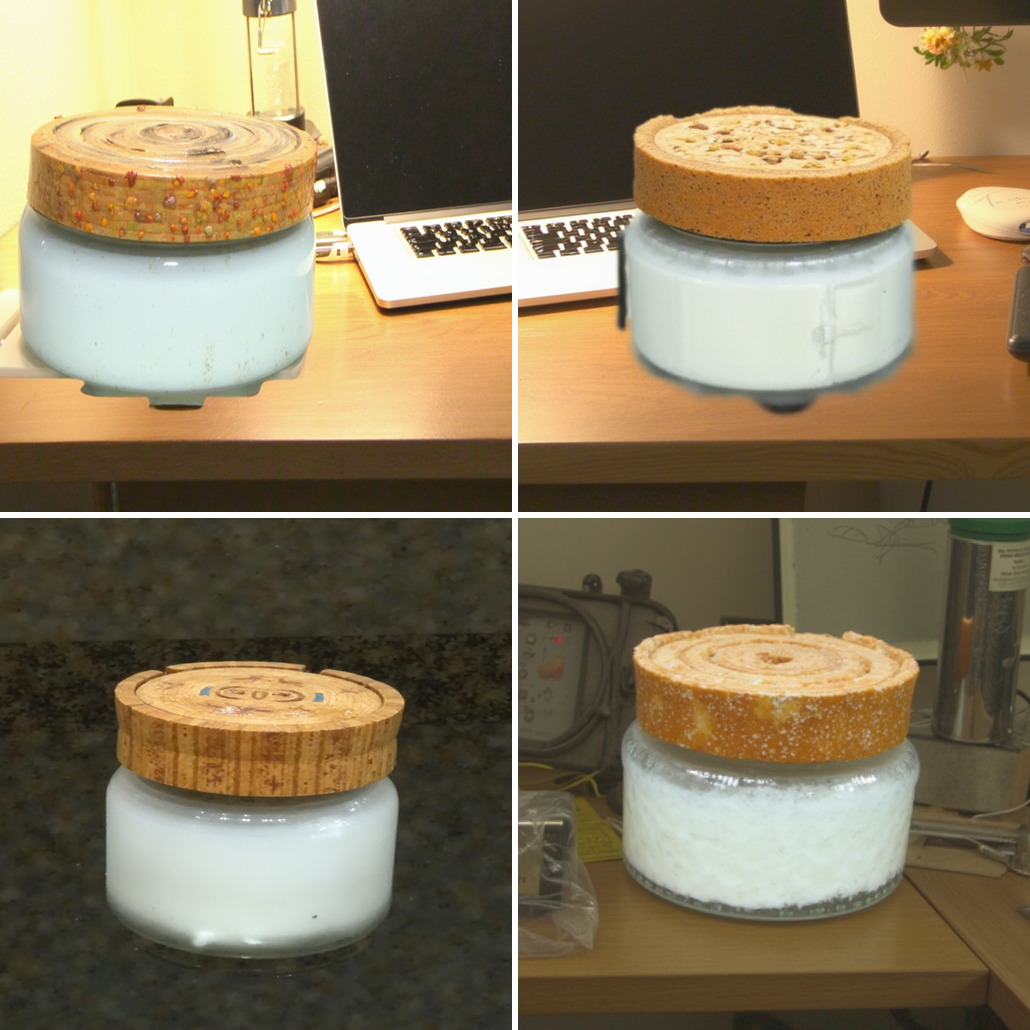} &
			\includegraphics[width=.192\linewidth]{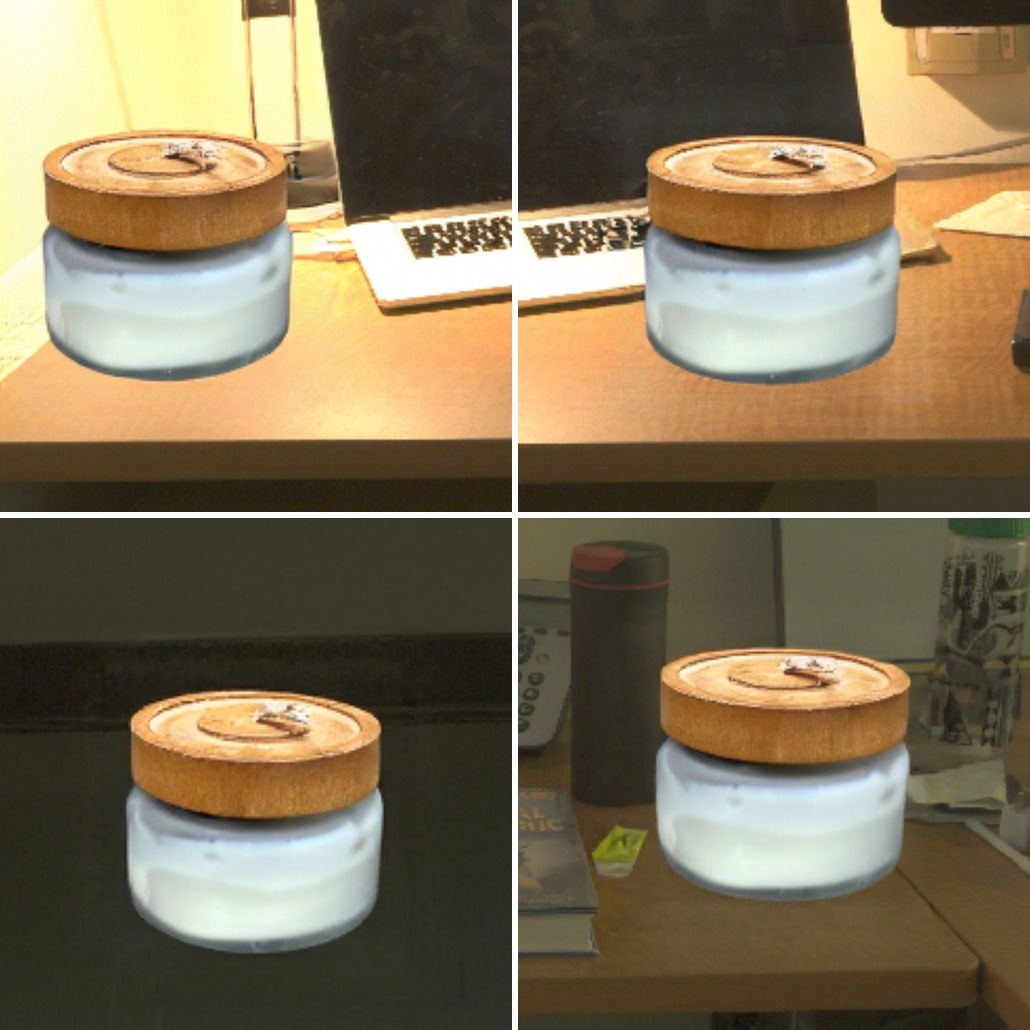}      &
			\includegraphics[width=.192\linewidth]{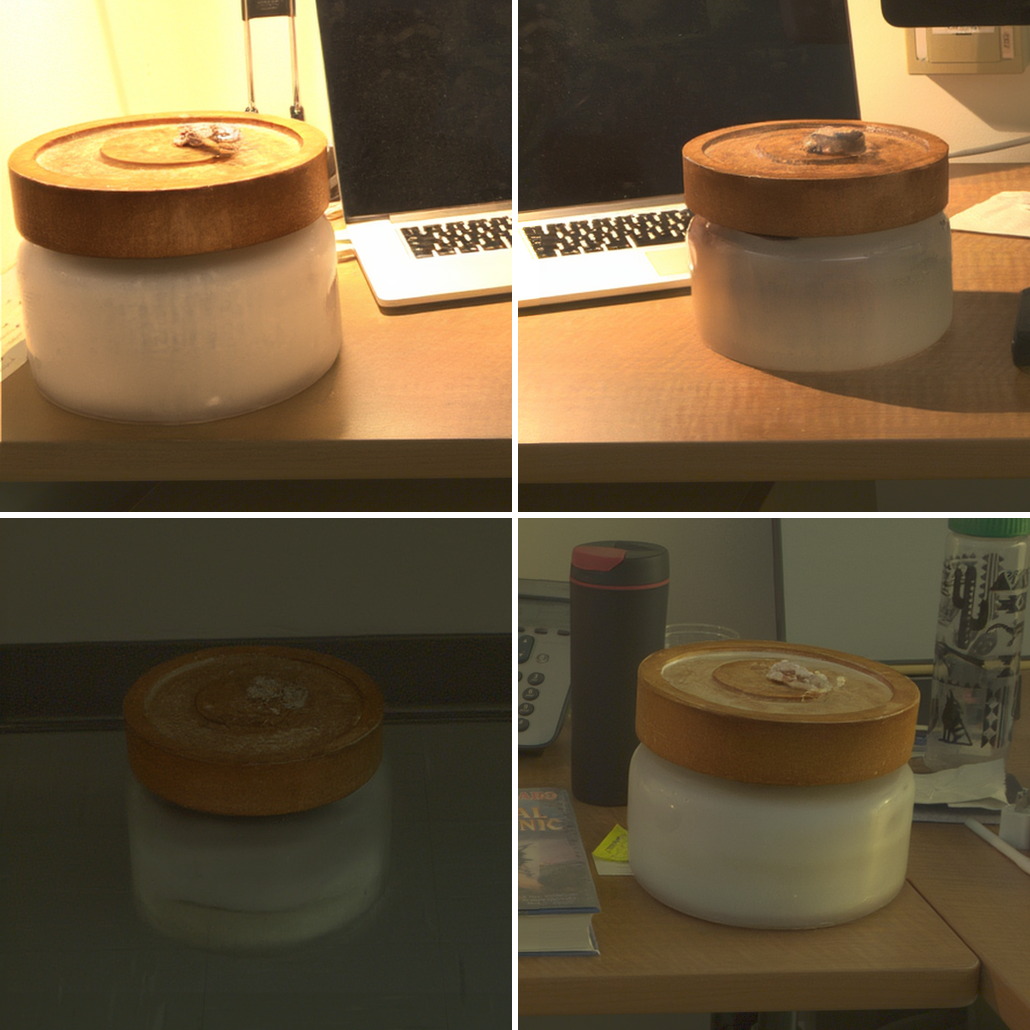}                  \\

			                                                                                                                  &
			\makecell{UniCombine}                                                                                             &
			\makecell{DreamFuse}                                                                                              &
			\makecell{OmniPaint}                                                                                              &
			\modelname                                                                                                          \\

			\inputimg{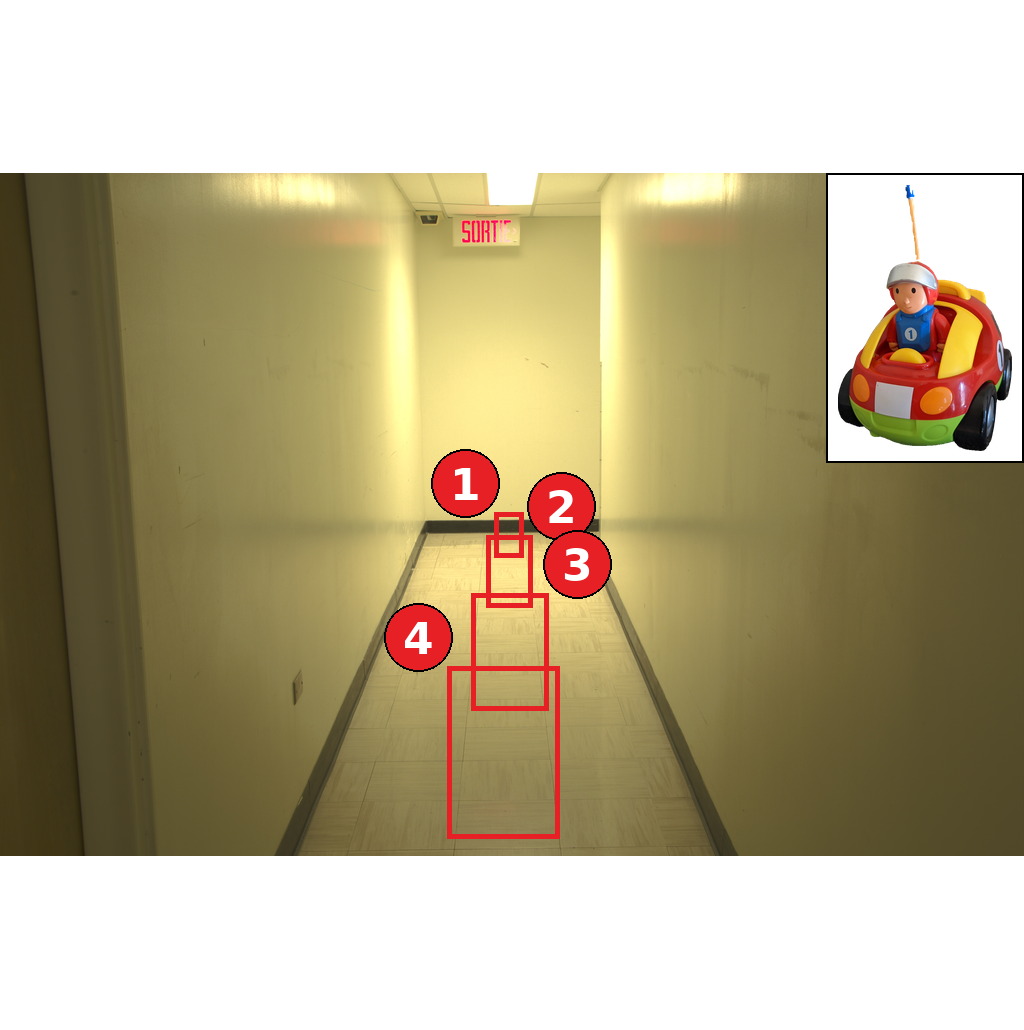}                                             &
			\includegraphics[width=.192\linewidth]{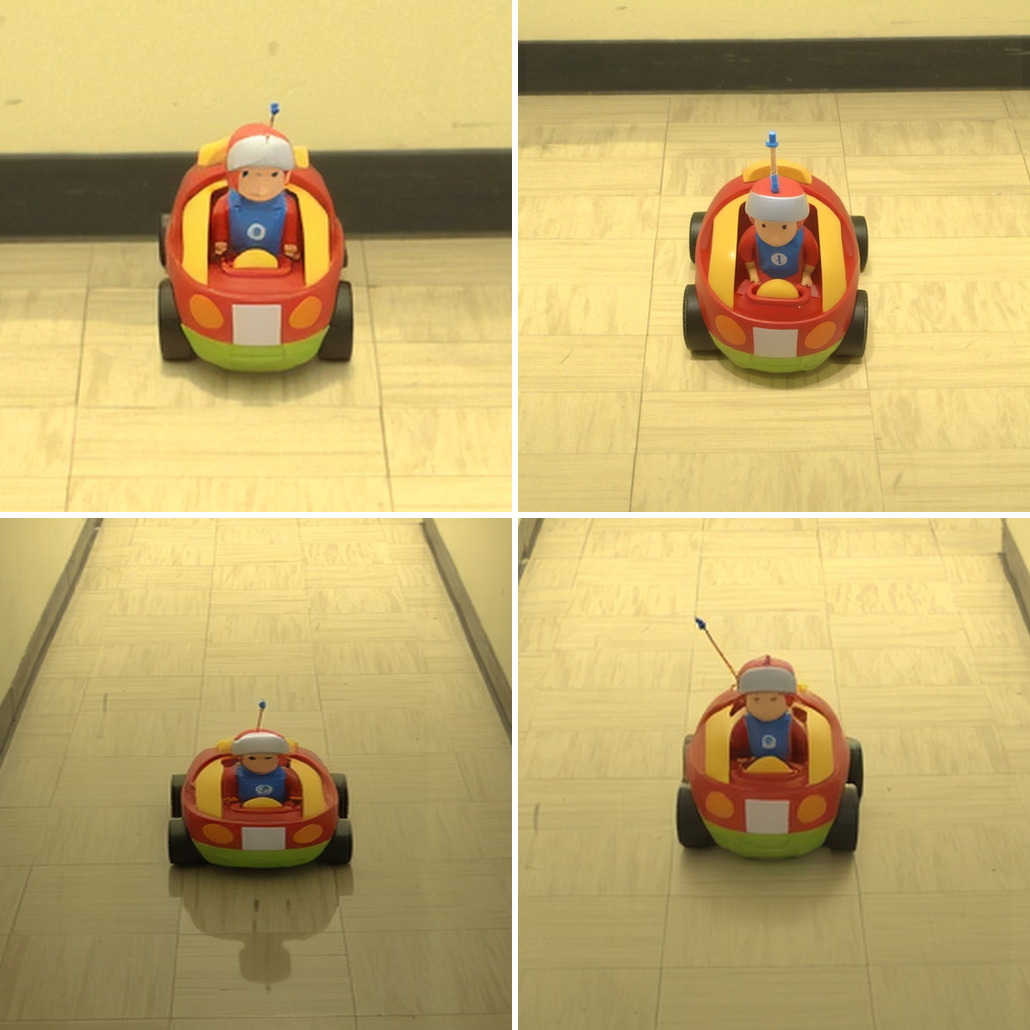}           &
			\includegraphics[width=.192\linewidth]{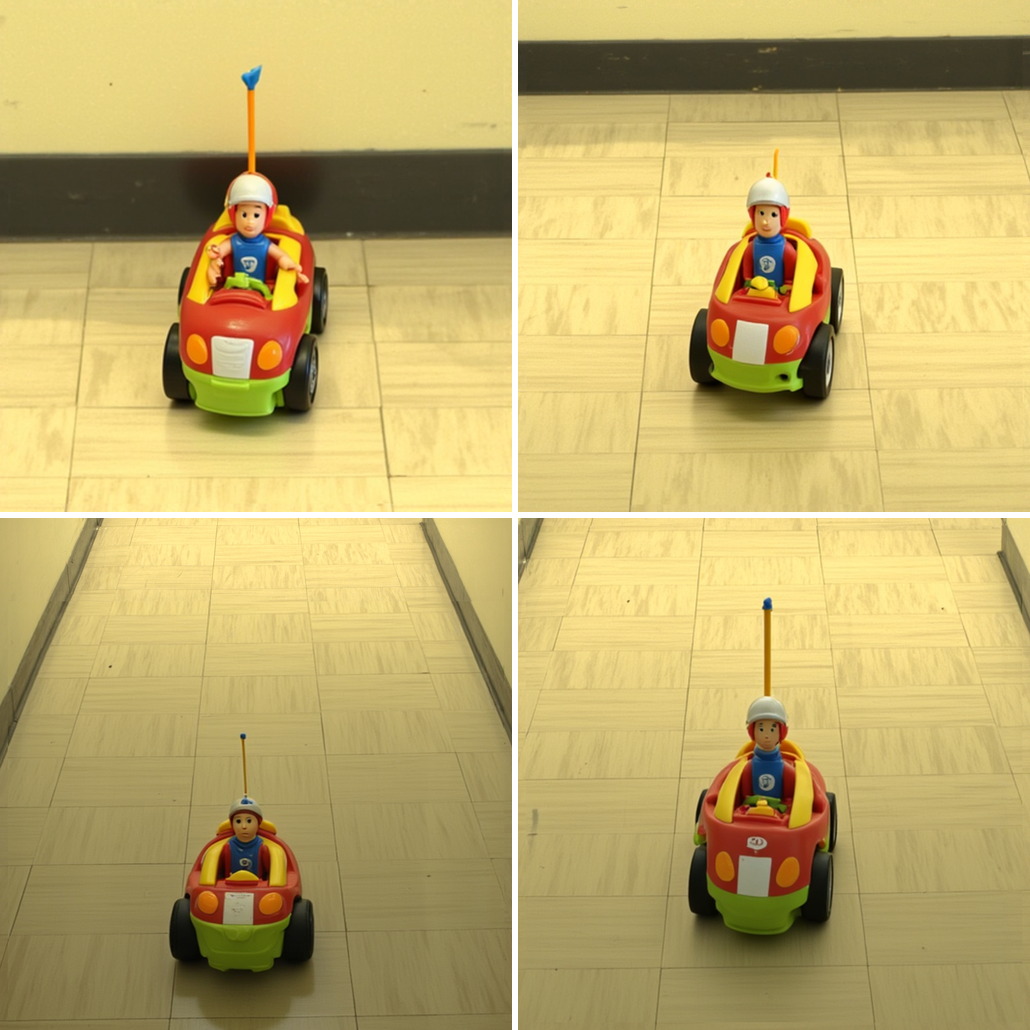}            &
			\includegraphics[width=.192\linewidth]{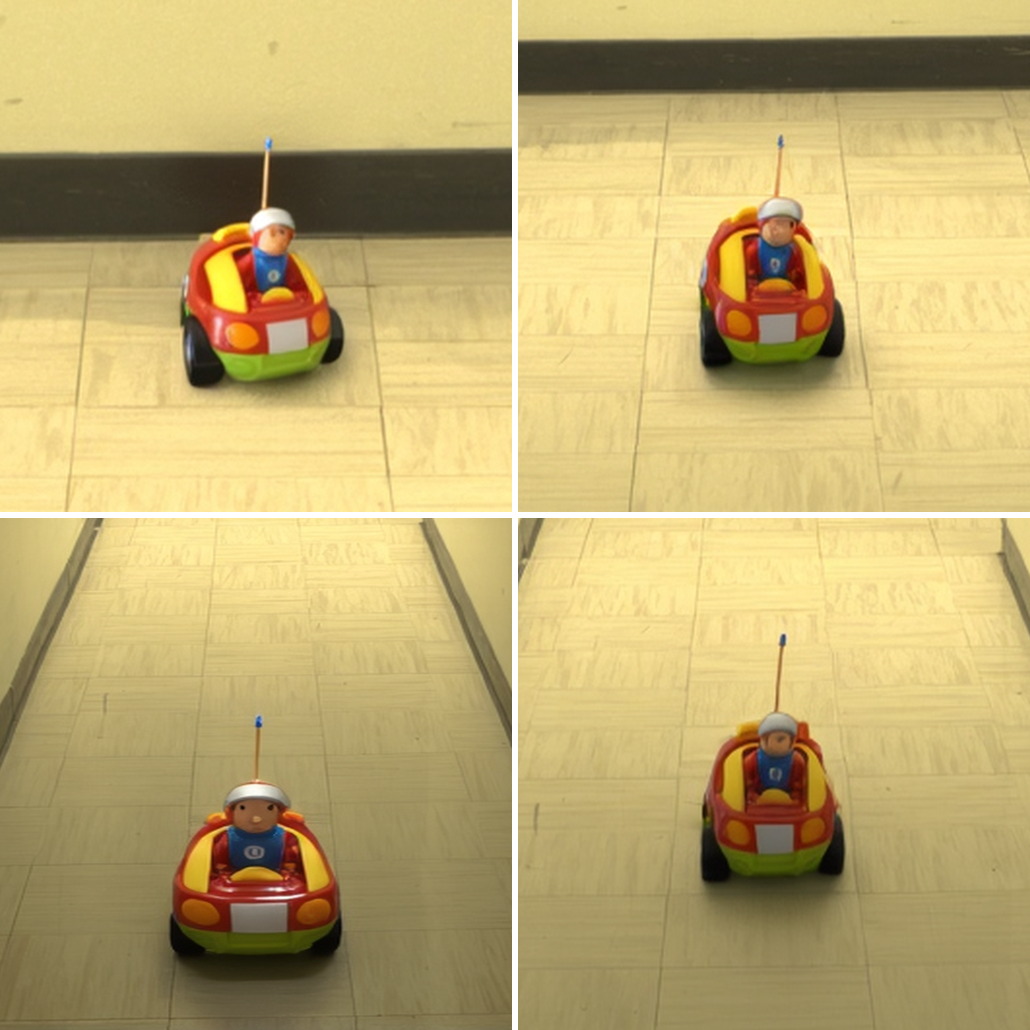}            &
			\includegraphics[width=.192\linewidth]{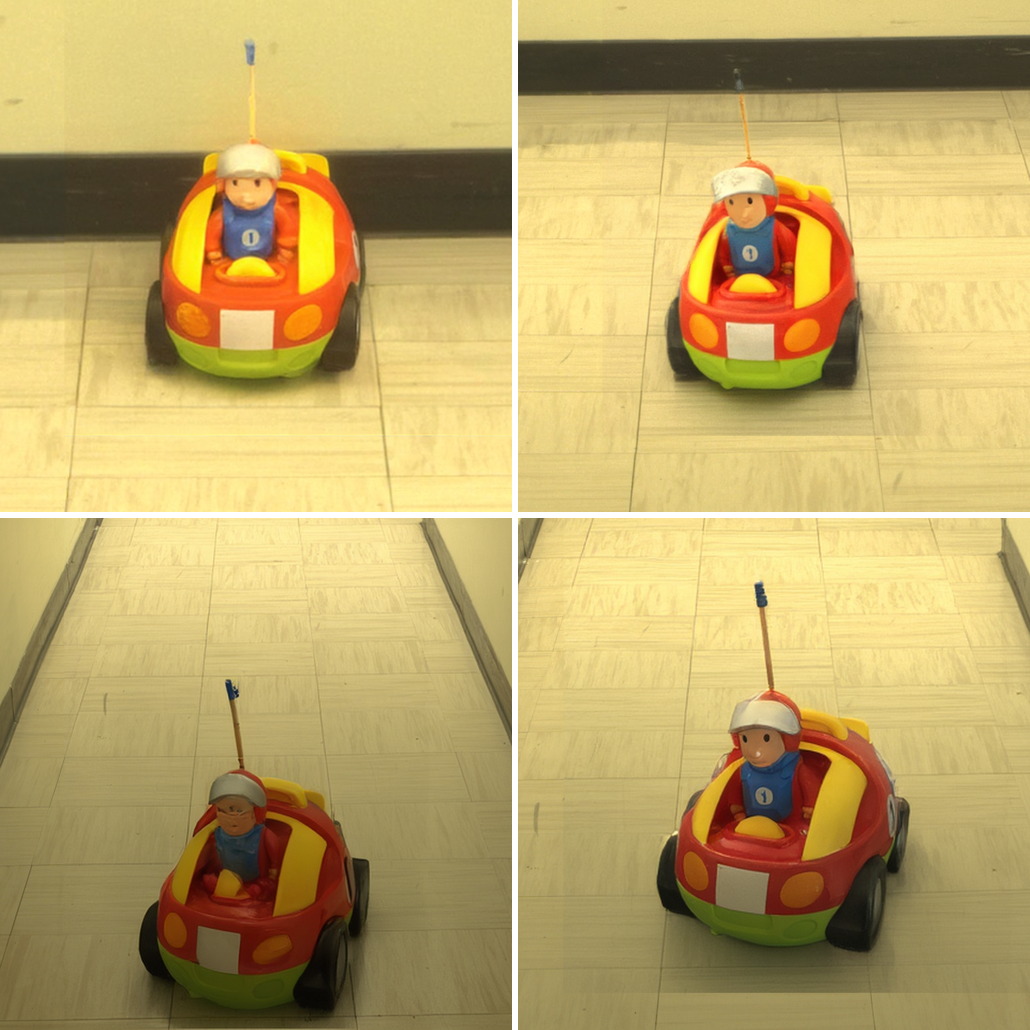}                   \\
		\end{tabular}
	\end{minipage}

	\par\vspace{2pt}{\footnotesize (a) Laval Indoor SV HDR}

	\vspace{7pt}

	\begin{minipage}{\linewidth}
		\centering
		\begin{tabular}{@{} *{6}{c@{\hspace{3pt}}} @{}}
			                                                                                                            &
			\makecell{G.T.}                                                                                             &
			\makecell{AnyDoor}                                                                                          &
			\makecell{Personalize                                                                                         \\Anything} &
			\makecell{Insert                                                                                              \\Anything} &
			\modelname                                                                                                    \\

			\begin{tikzpicture}[baseline=(img.south)]
				\node[inner sep=0] (img) {\includegraphics[width=.158\linewidth]{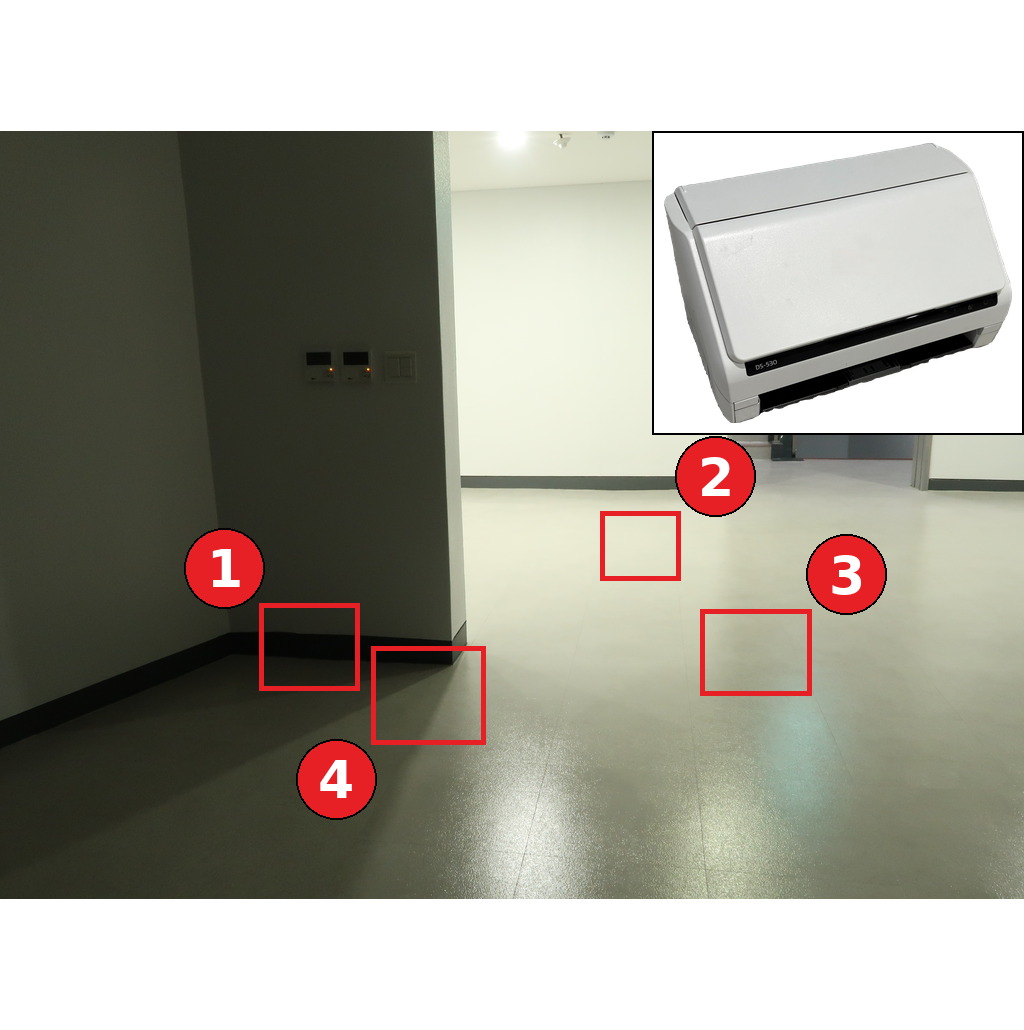}};
				\useasboundingbox (img.south west) rectangle (img.north east);
				\node[anchor=north, font=\scriptsize, text=black, fill=white, inner xsep=3pt, inner ysep=1pt, yshift=2pt] at (img.north) {Input};
			\end{tikzpicture}                            &
			\includegraphics[width=.158\linewidth]{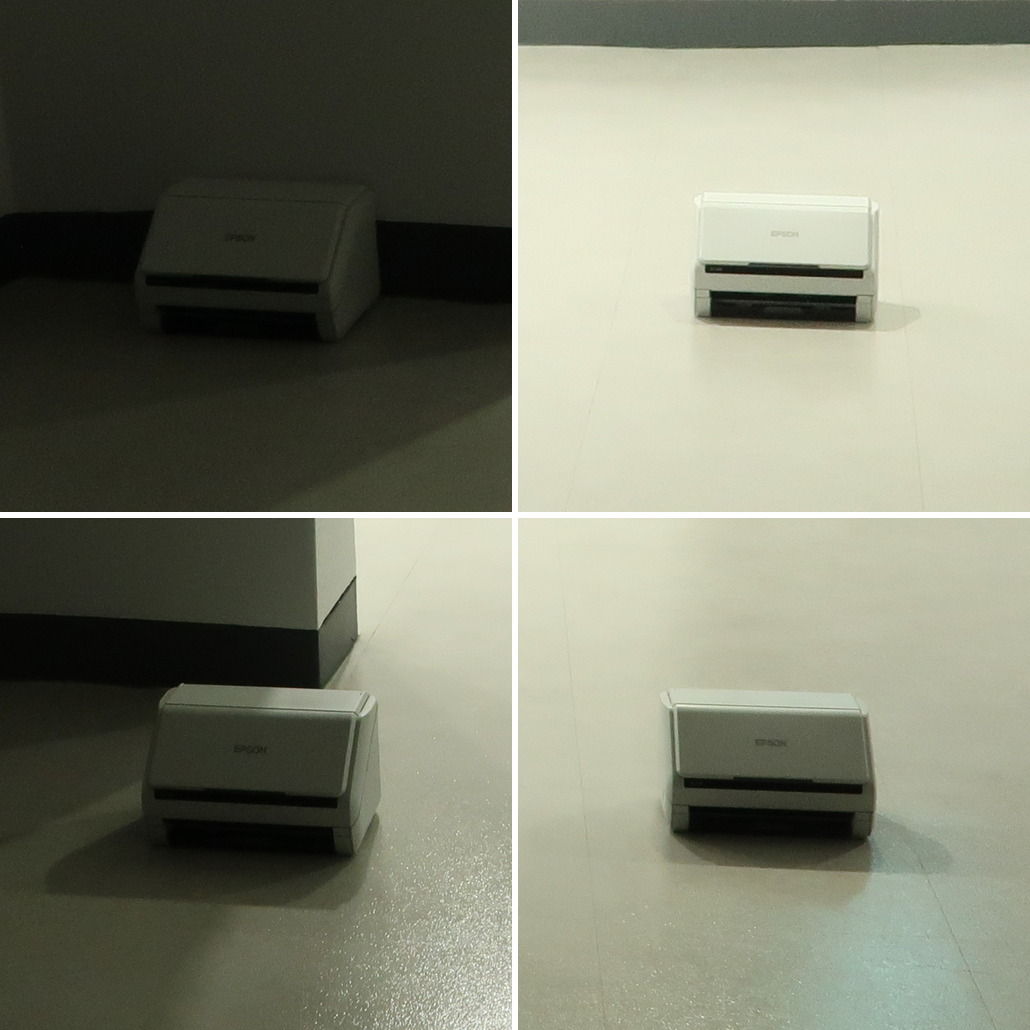}                  &
			\includegraphics[width=.158\linewidth]{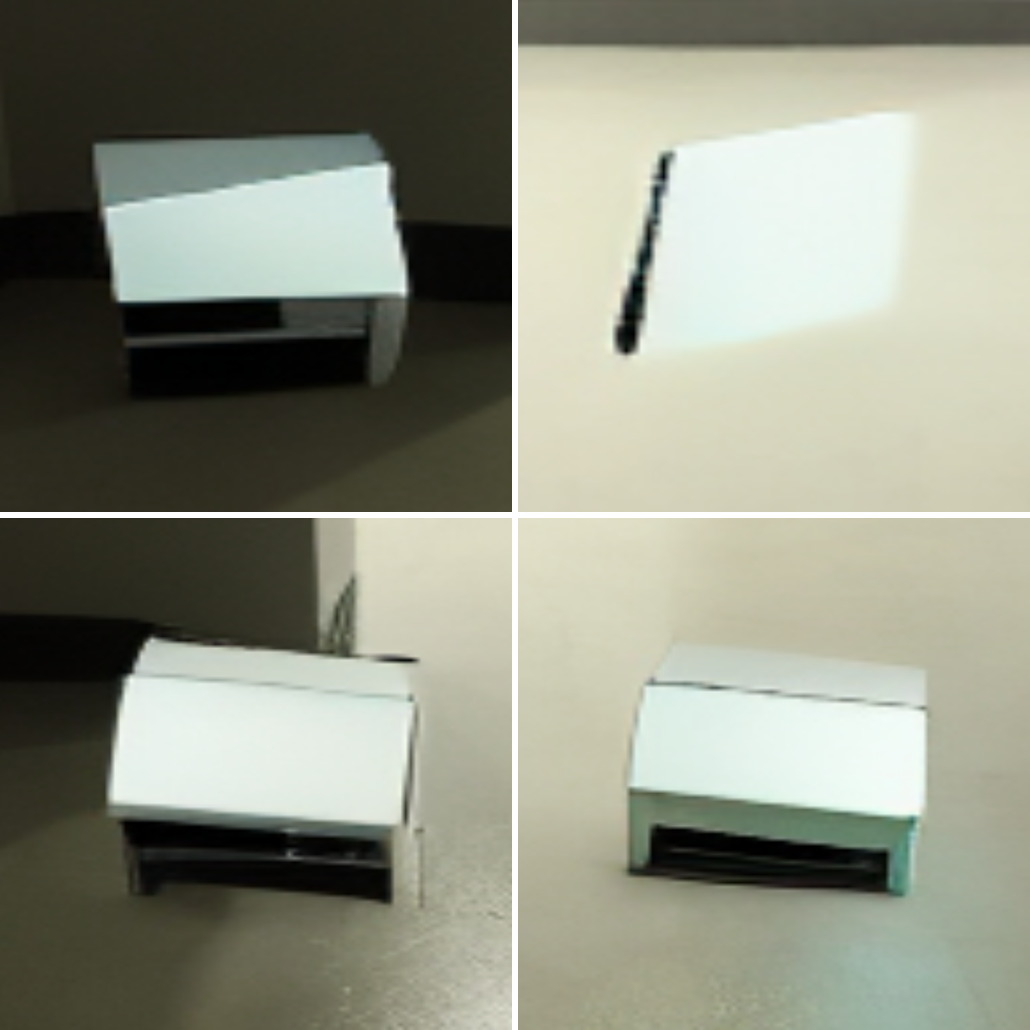}             &
			\includegraphics[width=.158\linewidth]{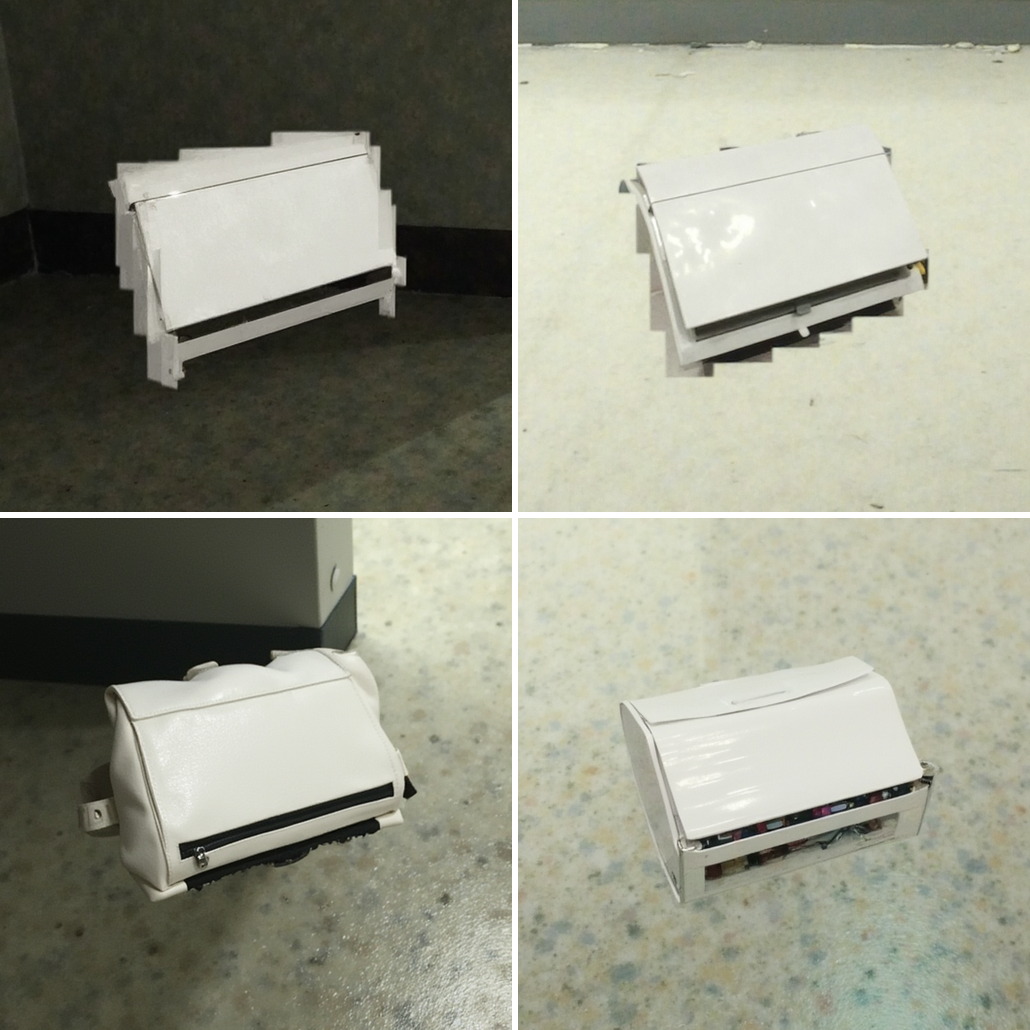} &
			\includegraphics[width=.158\linewidth]{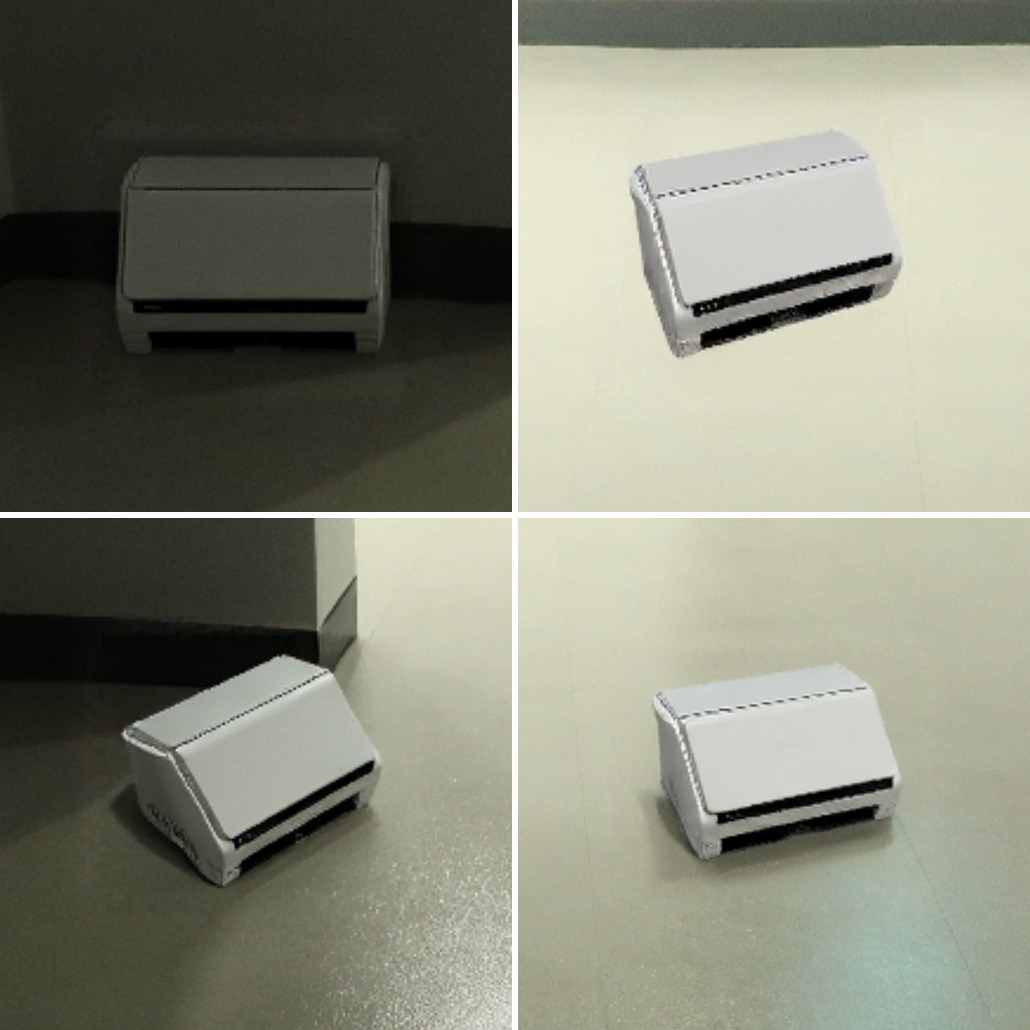}      &
			\includegraphics[width=.158\linewidth]{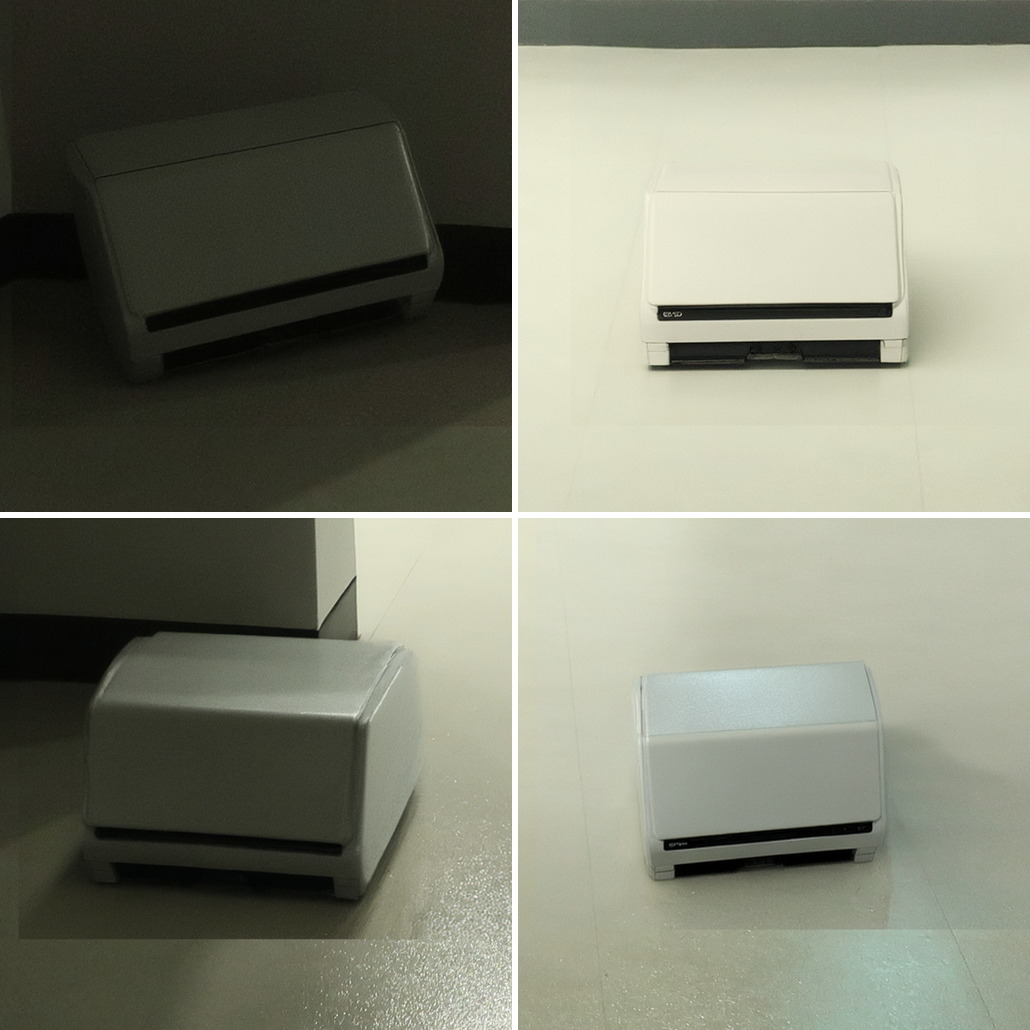}                  \\

			                                                                                                            &
			\makecell{G.T.}                                                                                             &
			\makecell{UniCombine}                                                                                       &
			\makecell{DreamFuse}                                                                                        &
			\makecell{OmniPaint}                                                                                        &
			\modelname                                                                                                    \\

			\begin{tikzpicture}[baseline=(img.south)]
				\node[inner sep=0] (img) {\includegraphics[width=.158\linewidth]{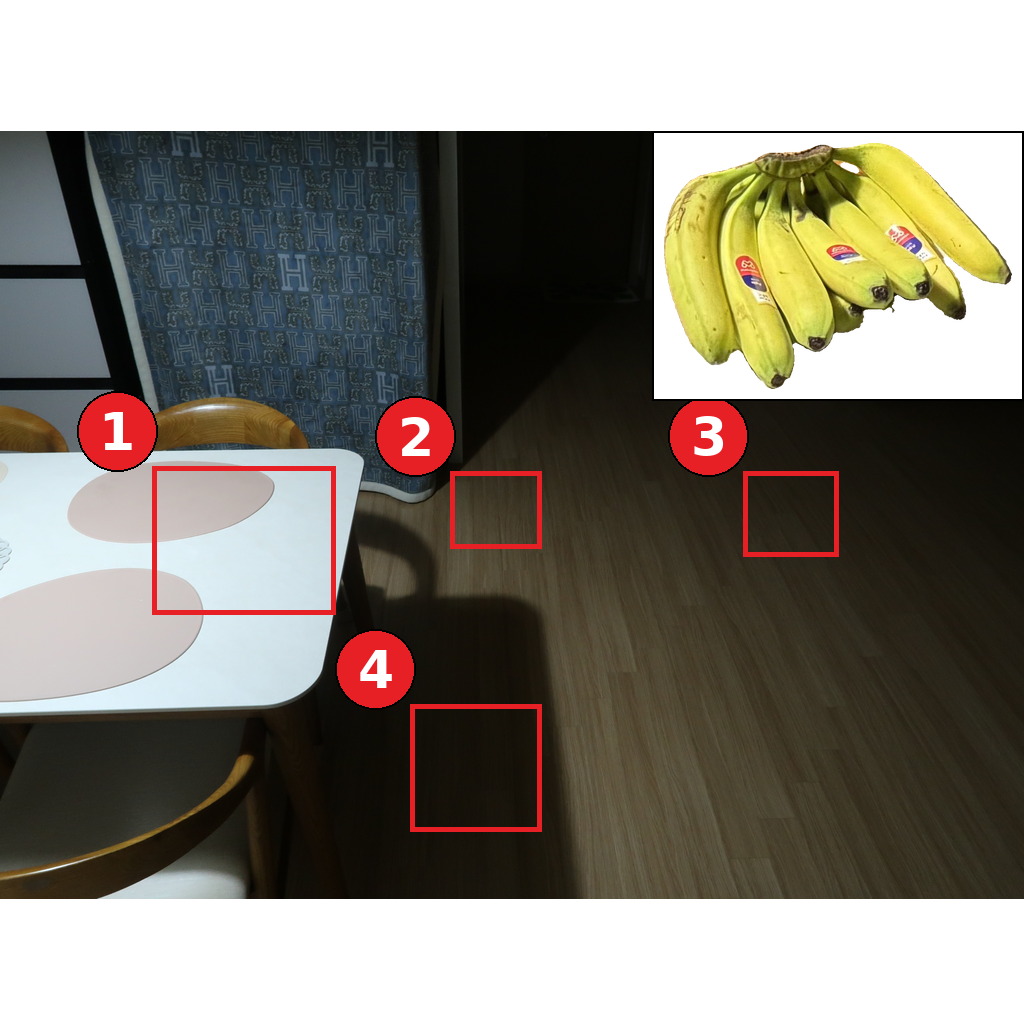}};
				\useasboundingbox (img.south west) rectangle (img.north east);
				\node[anchor=north, font=\scriptsize, text=black, fill=white, inner xsep=3pt, inner ysep=1pt, yshift=2pt] at (img.north) {Input};
			\end{tikzpicture}                            &
			\includegraphics[width=.158\linewidth]{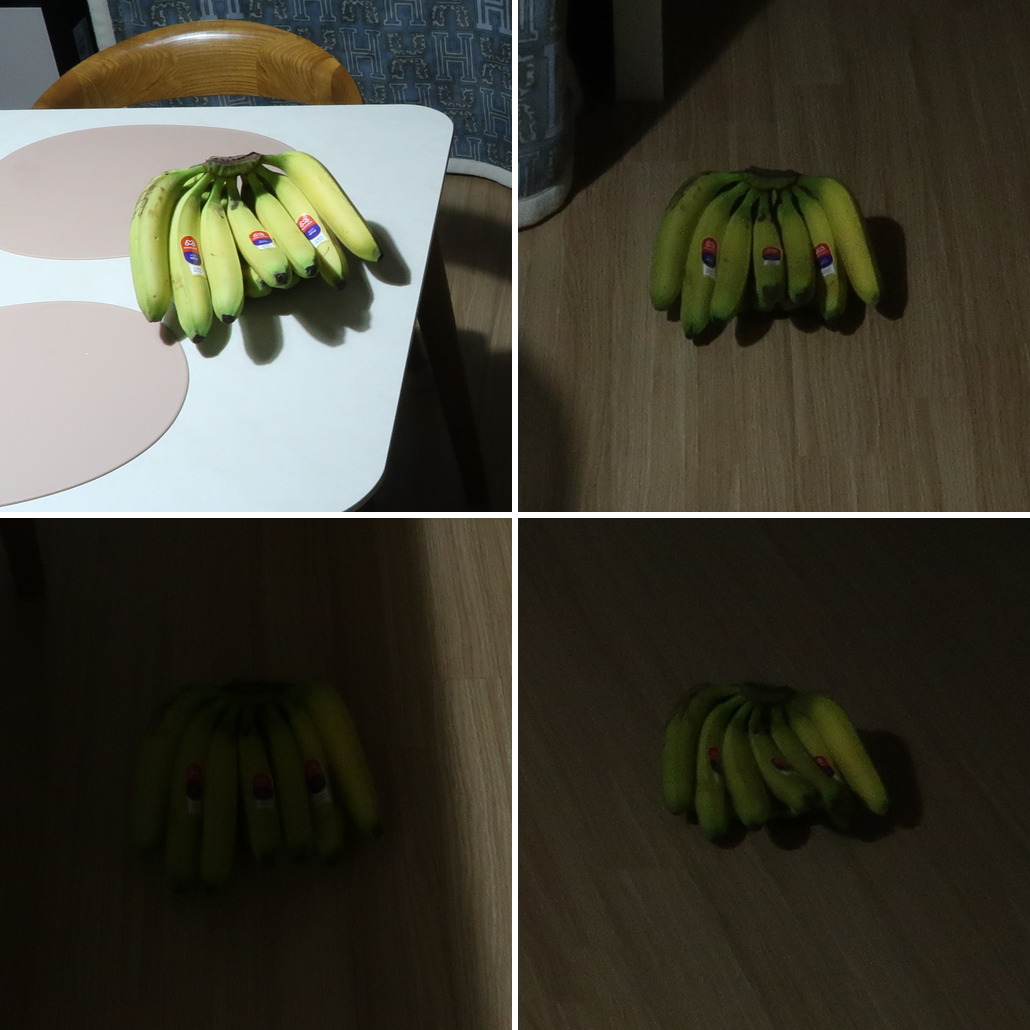}                   &
			\includegraphics[width=.158\linewidth]{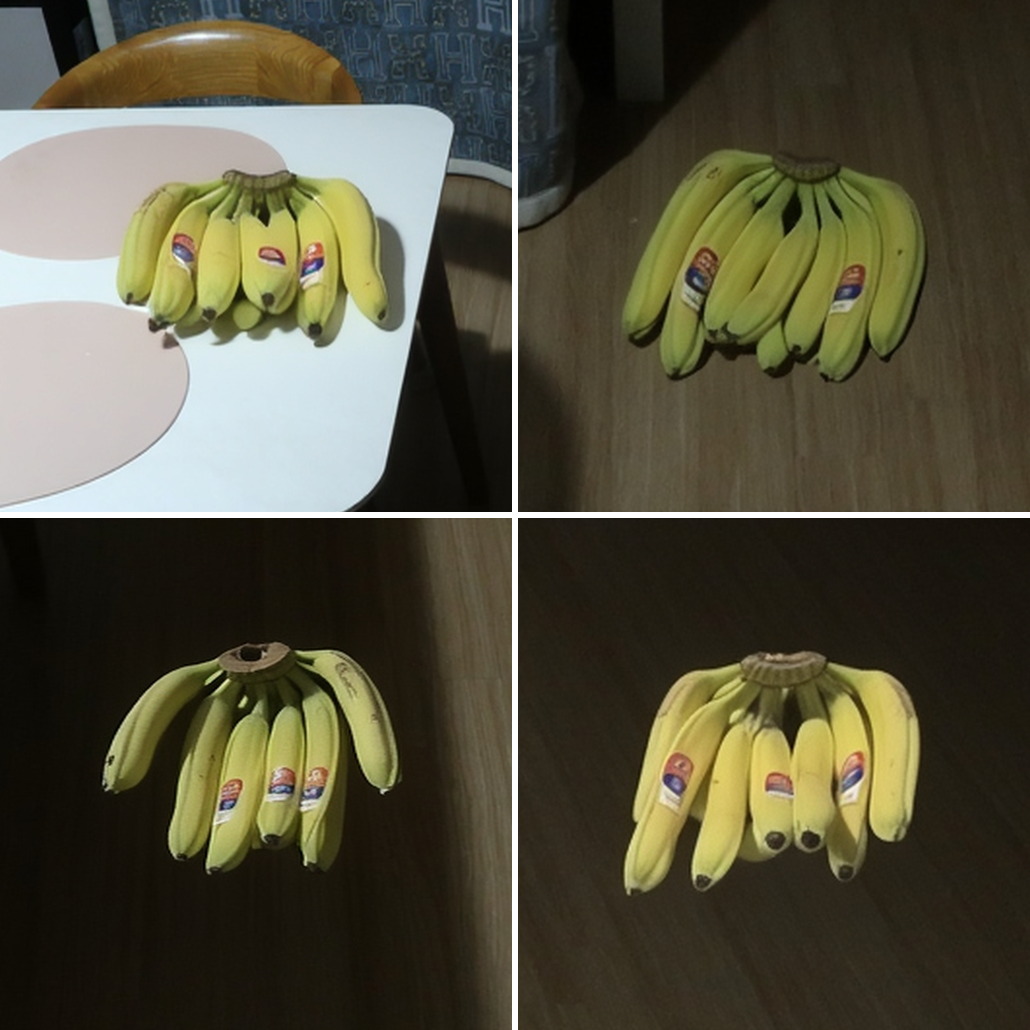}           &
			\includegraphics[width=.158\linewidth]{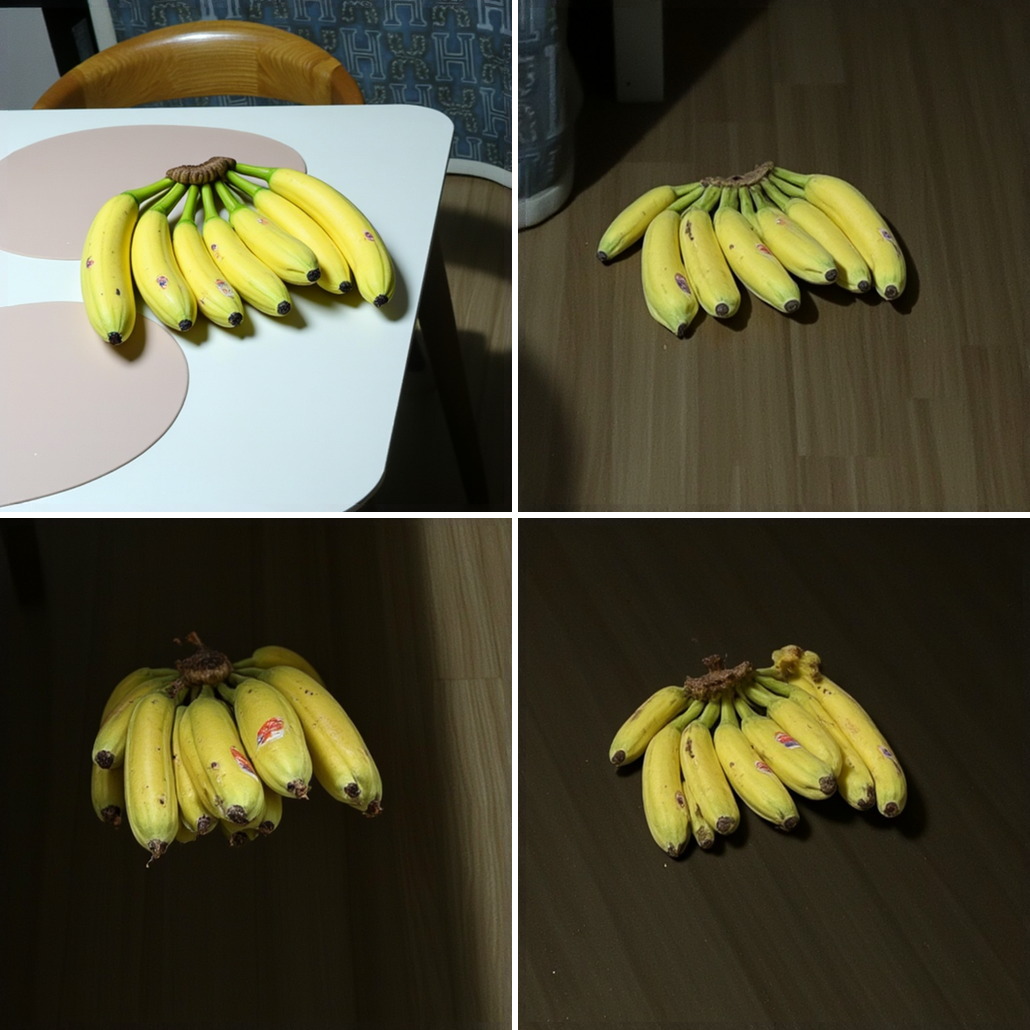}            &
			\includegraphics[width=.158\linewidth]{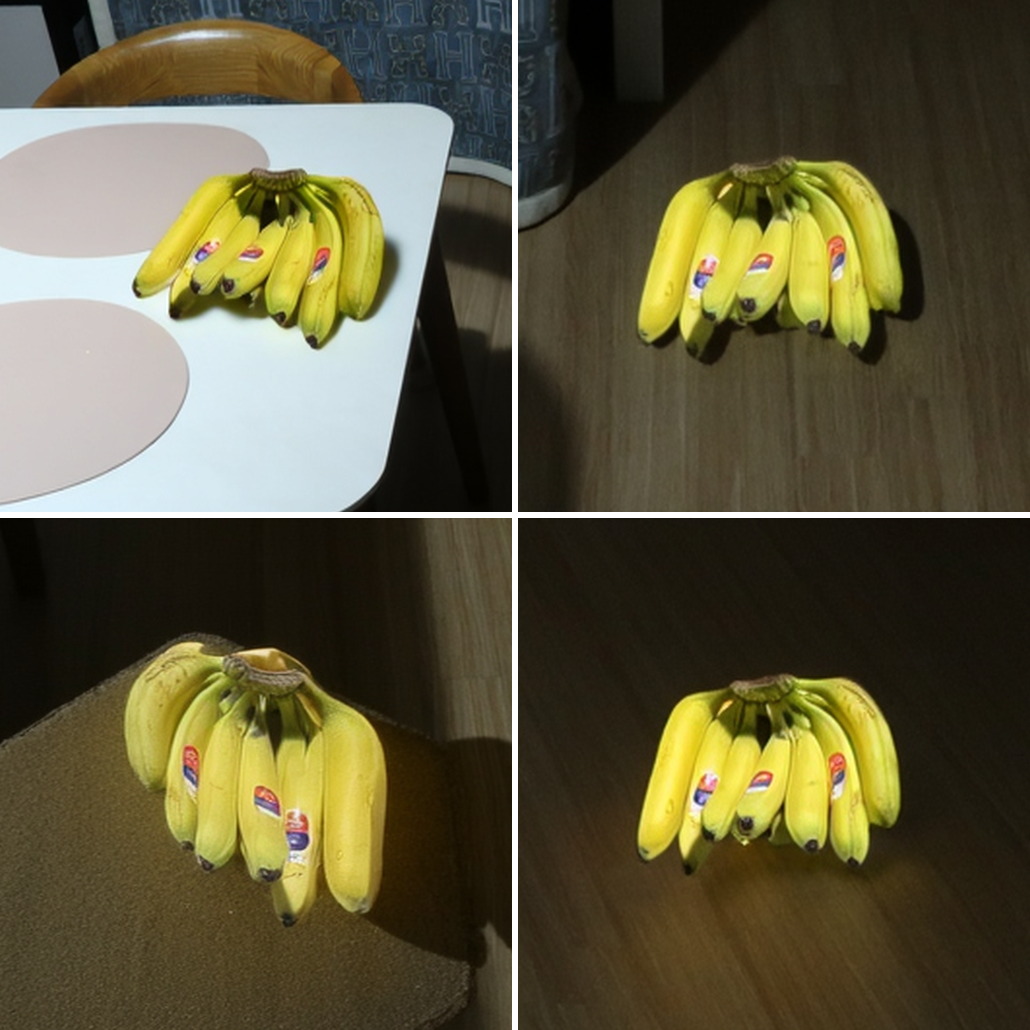}            &
			\includegraphics[width=.158\linewidth]{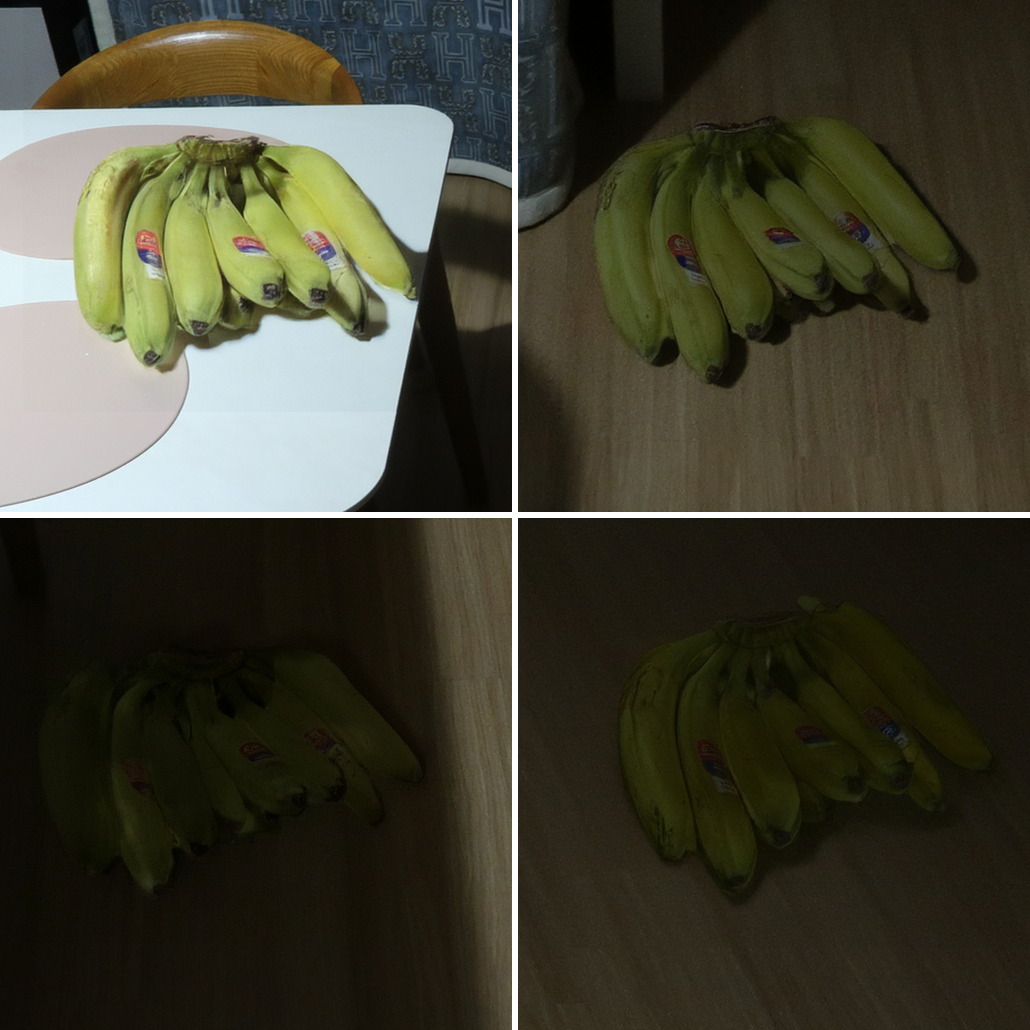}                   \\
		\end{tabular}
	\end{minipage}

	\par\vspace{2pt}{\footnotesize (b) \benchname{}}

	\caption{
		\modelname{} adapts object appearance to local illumination at each insertion location.
		Surface brightness and cast shadow vary with placement.
		In contrast, baselines respond only weakly to the local light context: they fail to match the local brightness (Laval top, \benchname{} bottom) or to produce a plausible cast shadow (Laval top, \benchname{} top).
		Cropped results follow the input numbering in clockwise order from the top-left.
		Ground-truth composites are available for \benchname{} but not for Laval.
	}
	\label{fig:composite_additional}
\end{figure}

\FloatBarrier
\Needspace{14\baselineskip}
\vspace{4pt}
\noindent
\textbf{Estimated irradiance.}
As an additional qualitative comparison on local lighting consistency, we estimate an irradiance map from each composite using RGBX~\citep{zeng2024rgb} and visualize it alongside the composite (Fig.~\ref{fig:irradiance}; top: composite, bottom: estimated irradiance).
For baselines, the irradiance on the inserted object often differs from its surroundings, with the boundary discontinuity becoming more pronounced as the composite's shading deviates further from the local lighting.
In contrast, \modelname{} yields irradiance that aligns with the surroundings and transitions smoothly across the boundary, reflecting consistent adaptation to insertion location.

The estimator can be unreliable in regions with clipped highlights, where the recovered irradiance may reflect estimation artifacts rather than true illumination.
For example, in the clock panel (second row), the desk surface around probe~1 contains strong highlights, and the estimates are noisy across all methods.

\begin{figure}[htbp]
	\centering
	\scriptsize
	\settoheight{\rotimgh}{\includegraphics[width=.17\linewidth]{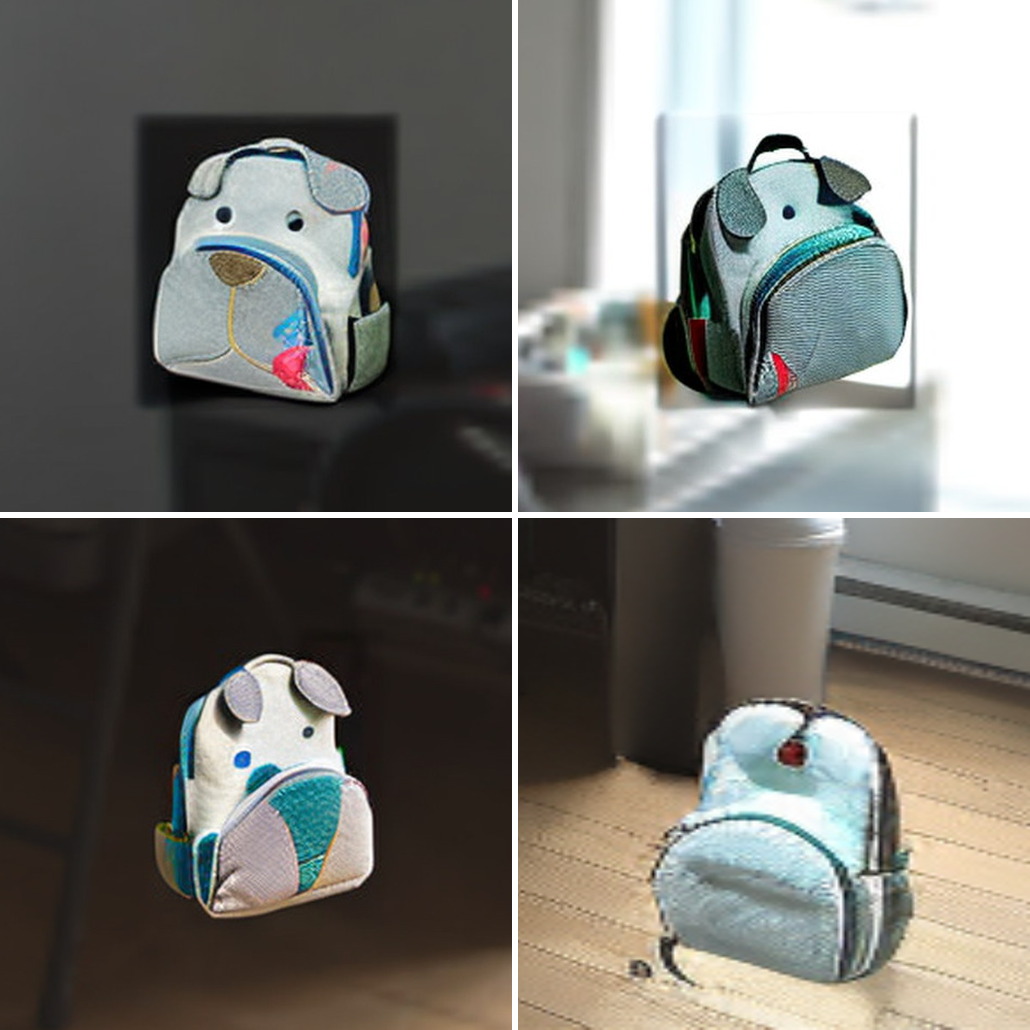}}
	\makebox[.25\linewidth][c]{}\hspace{3pt}%
	\makebox[.17\linewidth][c]{\makecell{AnyDoor}}\hspace{3pt}%
	\makebox[.17\linewidth][c]{\makecell{Personalize\\Anything}}\hspace{3pt}%
	\makebox[.17\linewidth][c]{\makecell{Insert\\Anything}}\hspace{3pt}%
	\makebox[.17\linewidth][c]{\modelname}\hspace{3pt}%
	\hphantom{\rotatebox{90}{Composite}}\\[2pt]
	\begin{minipage}[c]{.25\linewidth}
		\centering
		\begin{tikzpicture}[baseline=(img.south)]
			\node[inner sep=0] (img) {\includegraphics[width=\linewidth]{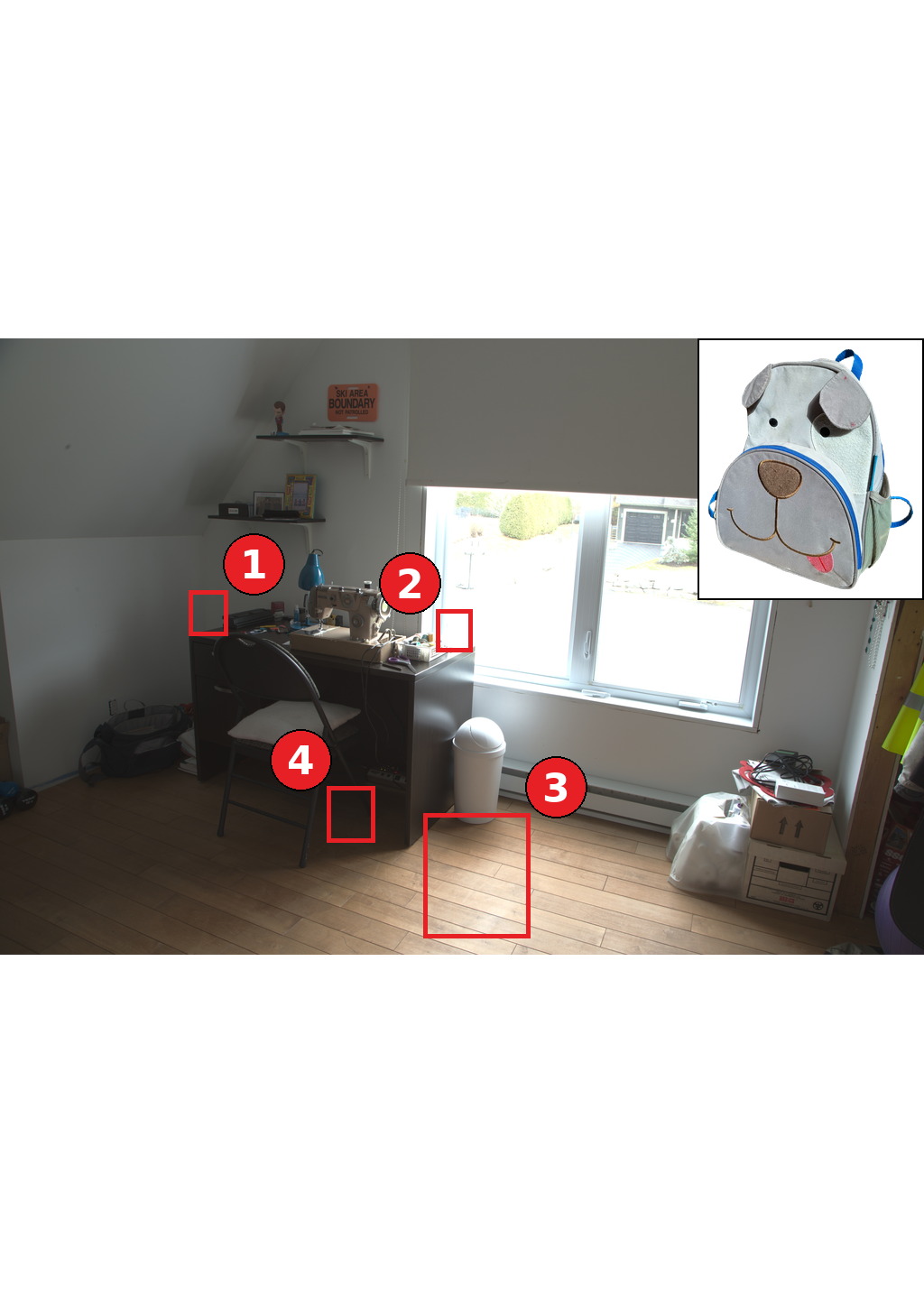}};
			\useasboundingbox (img.south west) rectangle (img.north east);
			\node[anchor=north, font=\scriptsize, text=black, fill=white, inner xsep=3pt, inner ysep=1pt, yshift=-20pt] at (img.north) {Input};
		\end{tikzpicture}%
	\end{minipage}\hspace{3pt}%
	\begin{tabular}[c]{@{}*{4}{c@{\hspace{3pt}}}c@{}}
		\includegraphics[width=.17\linewidth]{figures/experiment/composite/laval_backpackdog_scene17_anydoor.jpg}                        &
		\includegraphics[width=.17\linewidth]{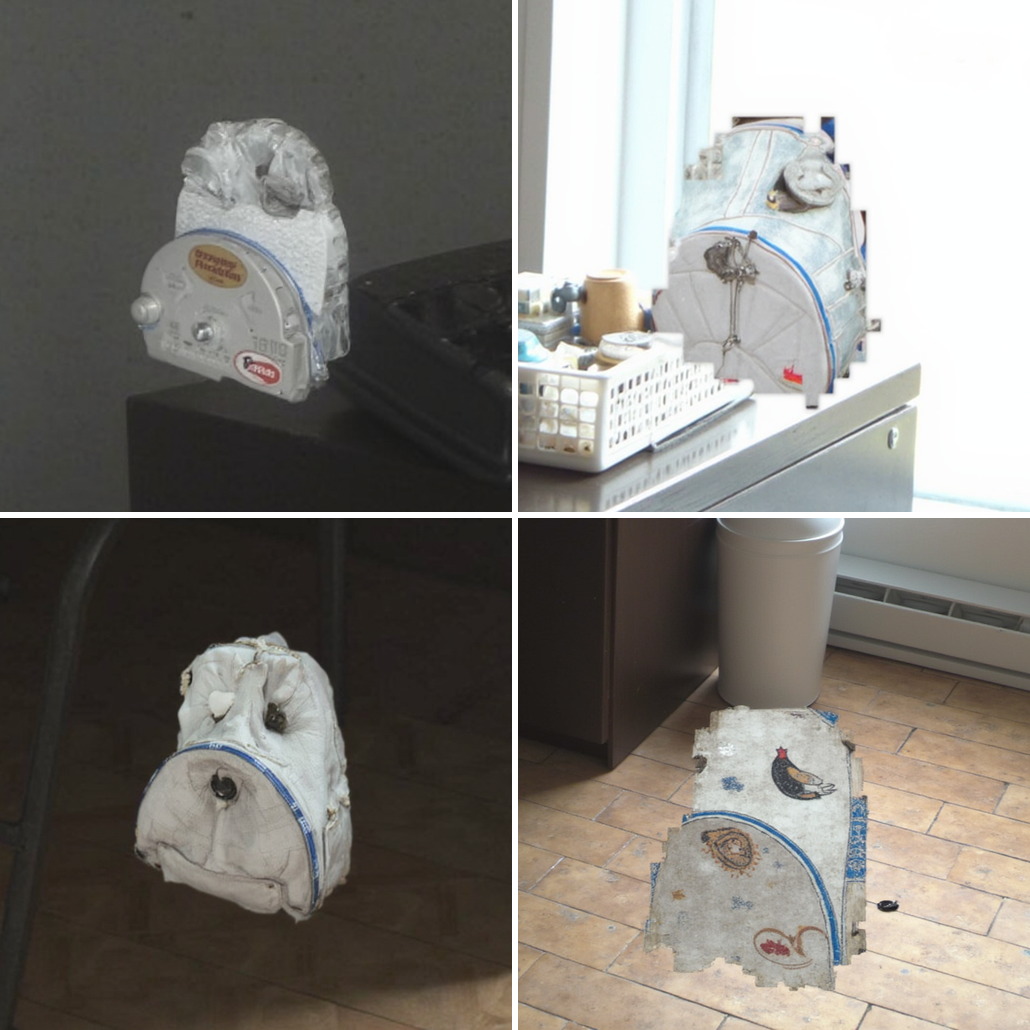}            &
		\includegraphics[width=.17\linewidth]{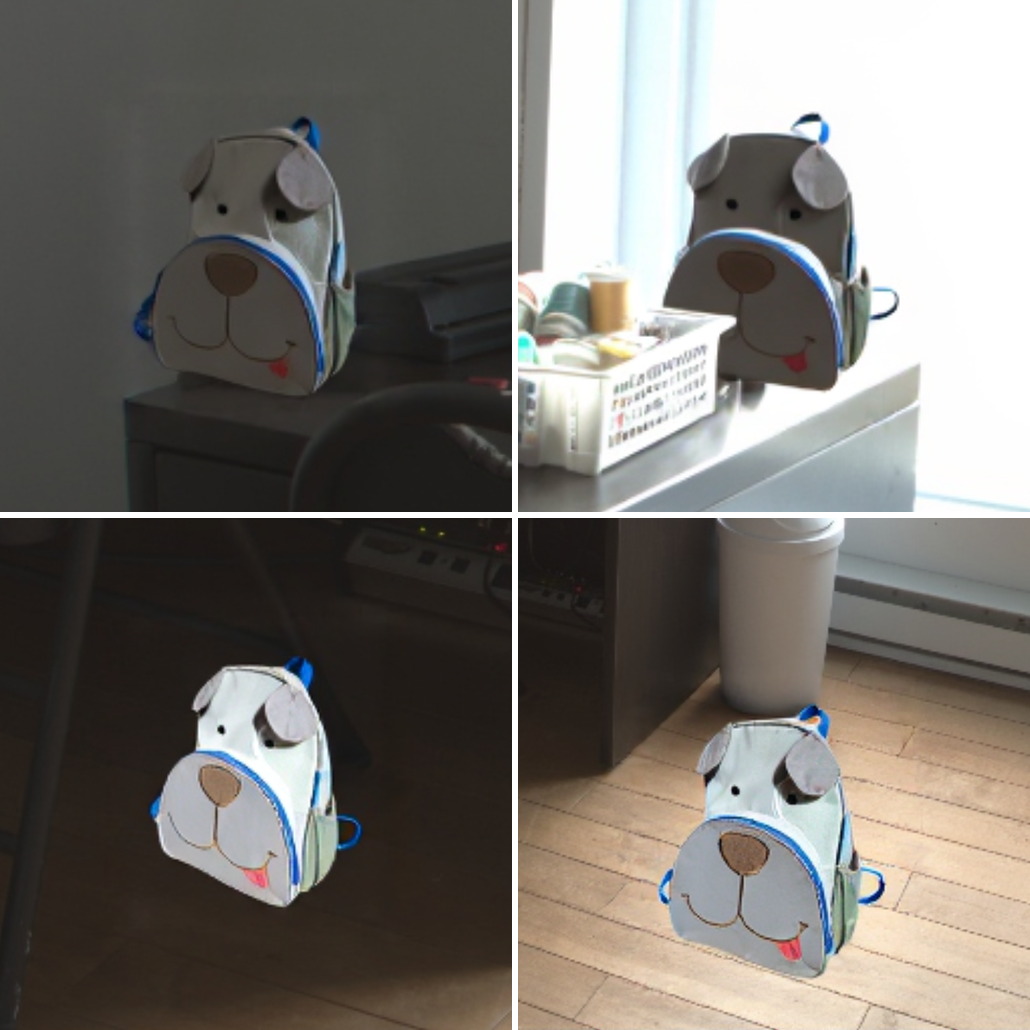}                 &
		\includegraphics[width=.17\linewidth]{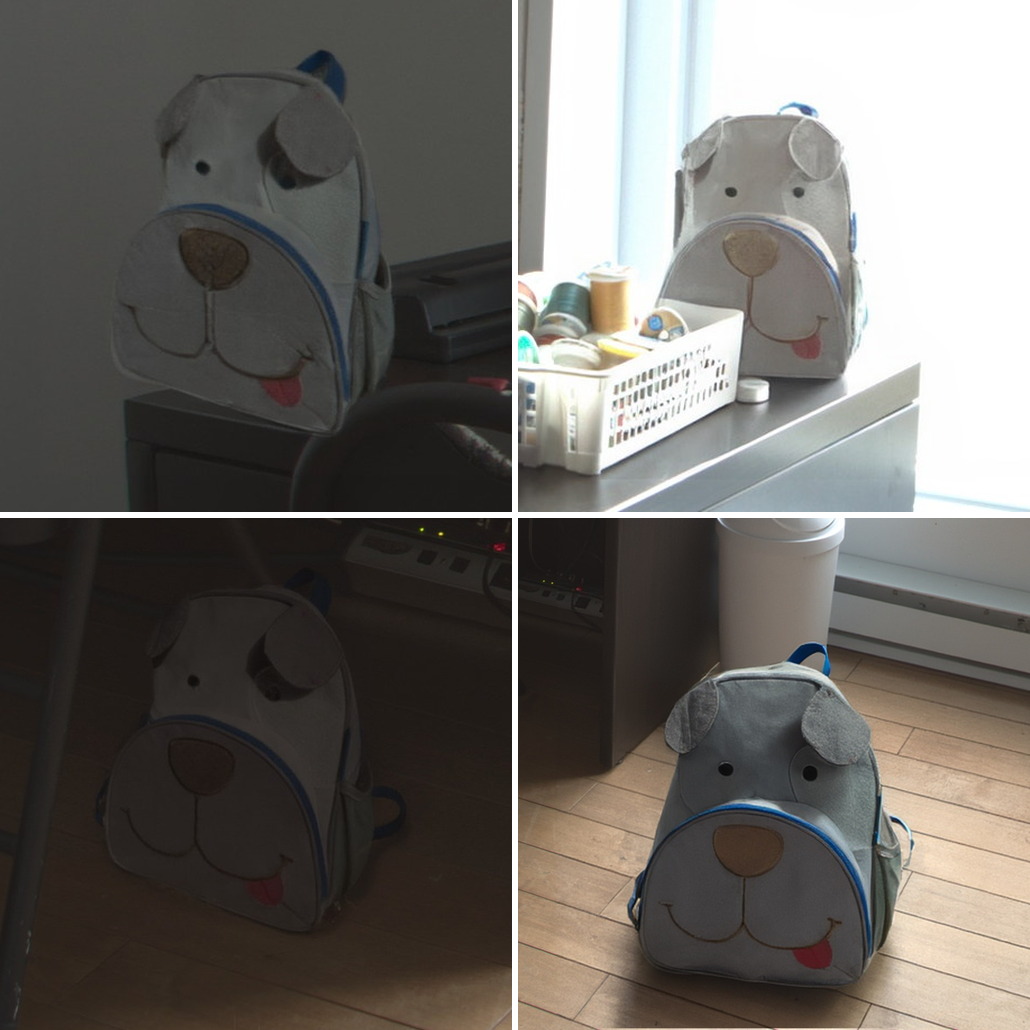}                           &
		\rotlabel{Composite}                                                                                                               \\
		\includegraphics[width=.17\linewidth]{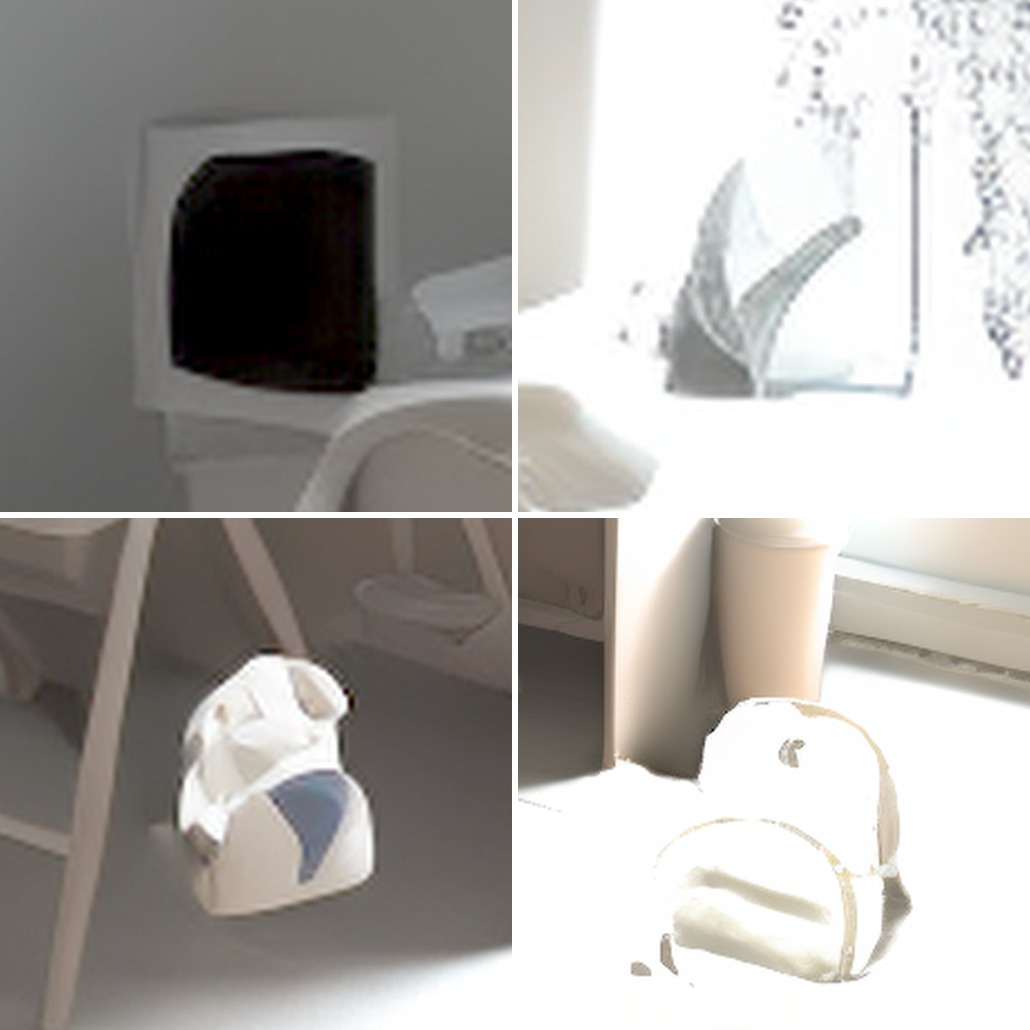}             &
		\includegraphics[width=.17\linewidth]{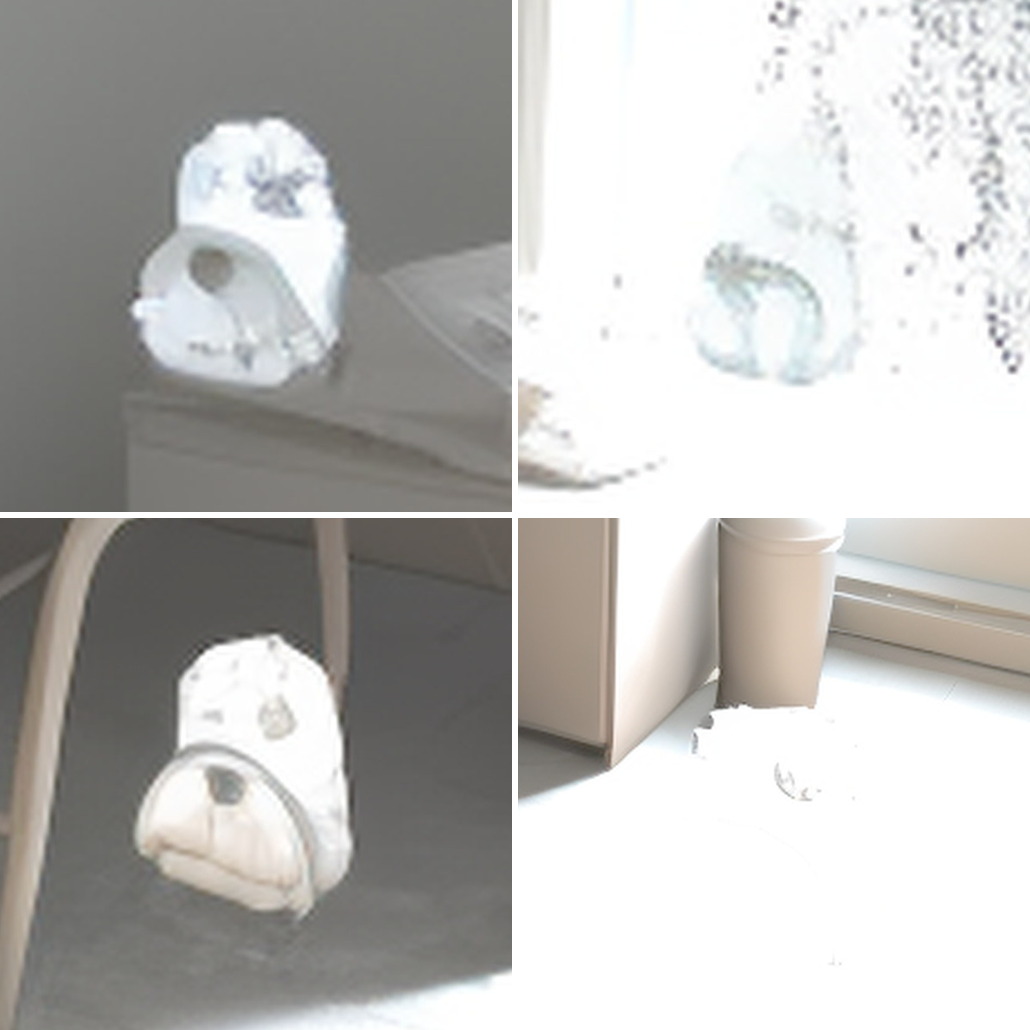} &
		\includegraphics[width=.17\linewidth]{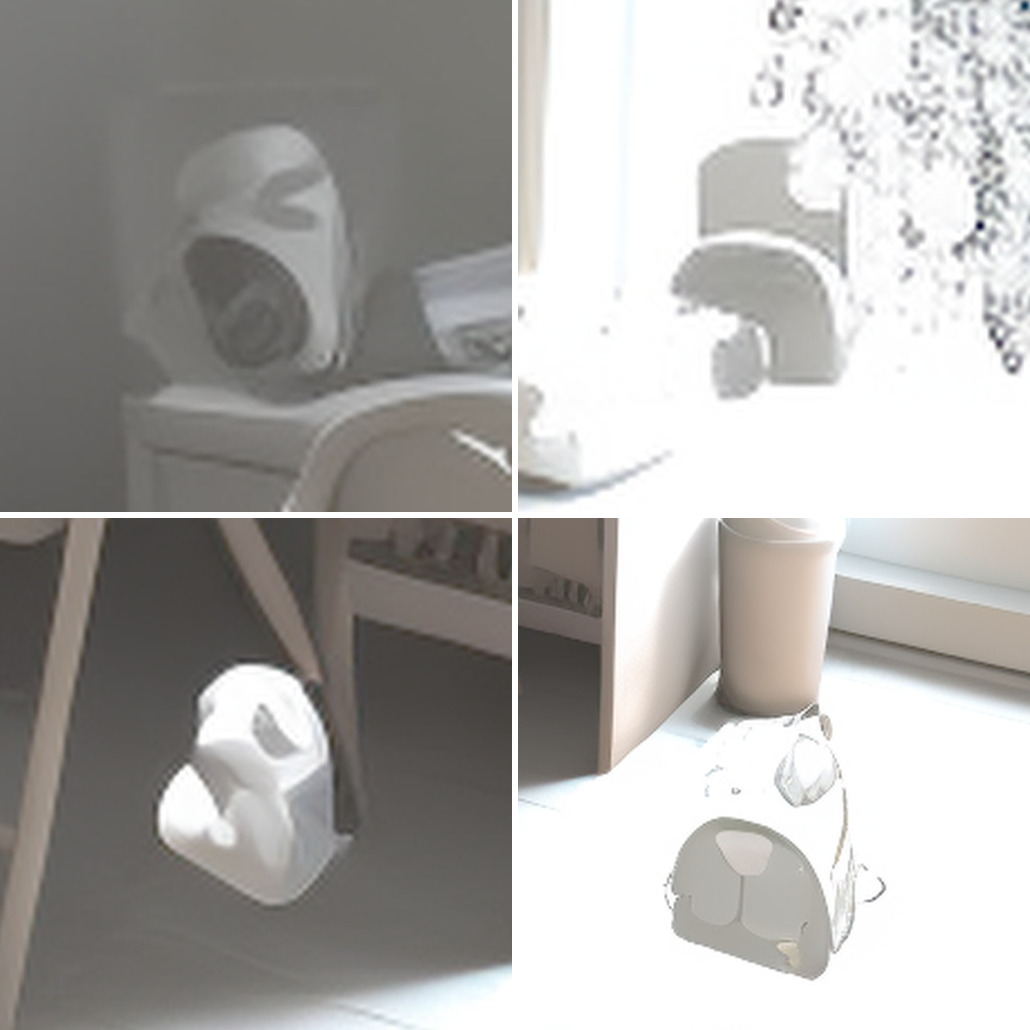}      &
		\includegraphics[width=.17\linewidth]{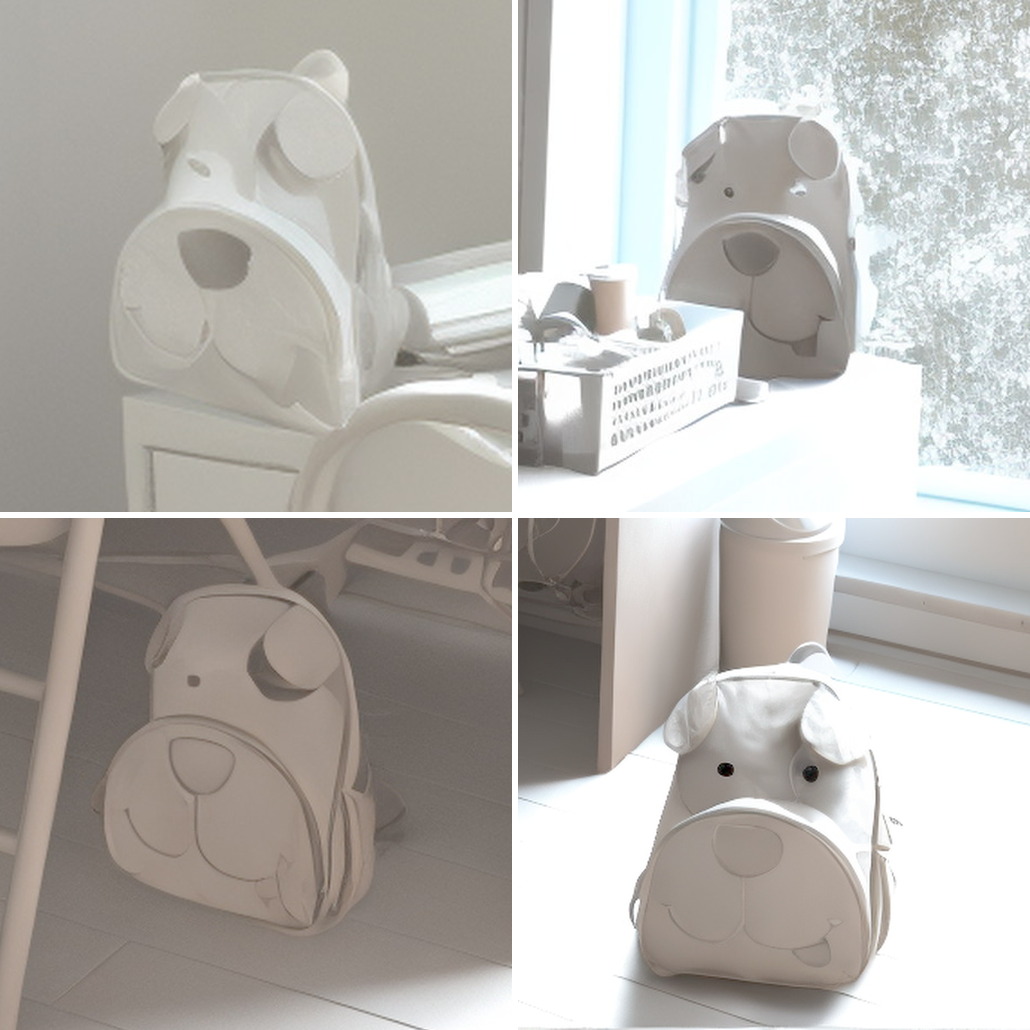}                &
		\rotlabel{Irradiance}                                                                                                              \\
	\end{tabular}\\[6pt]
	\makebox[.25\linewidth][c]{}\hspace{3pt}%
	\makebox[.17\linewidth][c]{\makecell{UniCombine}}\hspace{3pt}%
	\makebox[.17\linewidth][c]{\makecell{DreamFuse}}\hspace{3pt}%
	\makebox[.17\linewidth][c]{\makecell{OmniPaint}}\hspace{3pt}%
	\makebox[.17\linewidth][c]{\modelname}\hspace{3pt}%
	\hphantom{\rotatebox{90}{Composite}}\\[2pt]
	\begin{minipage}[c]{.25\linewidth}
		\centering
		\begin{tikzpicture}[baseline=(img.south)]
			\node[inner sep=0] (img) {\includegraphics[width=\linewidth]{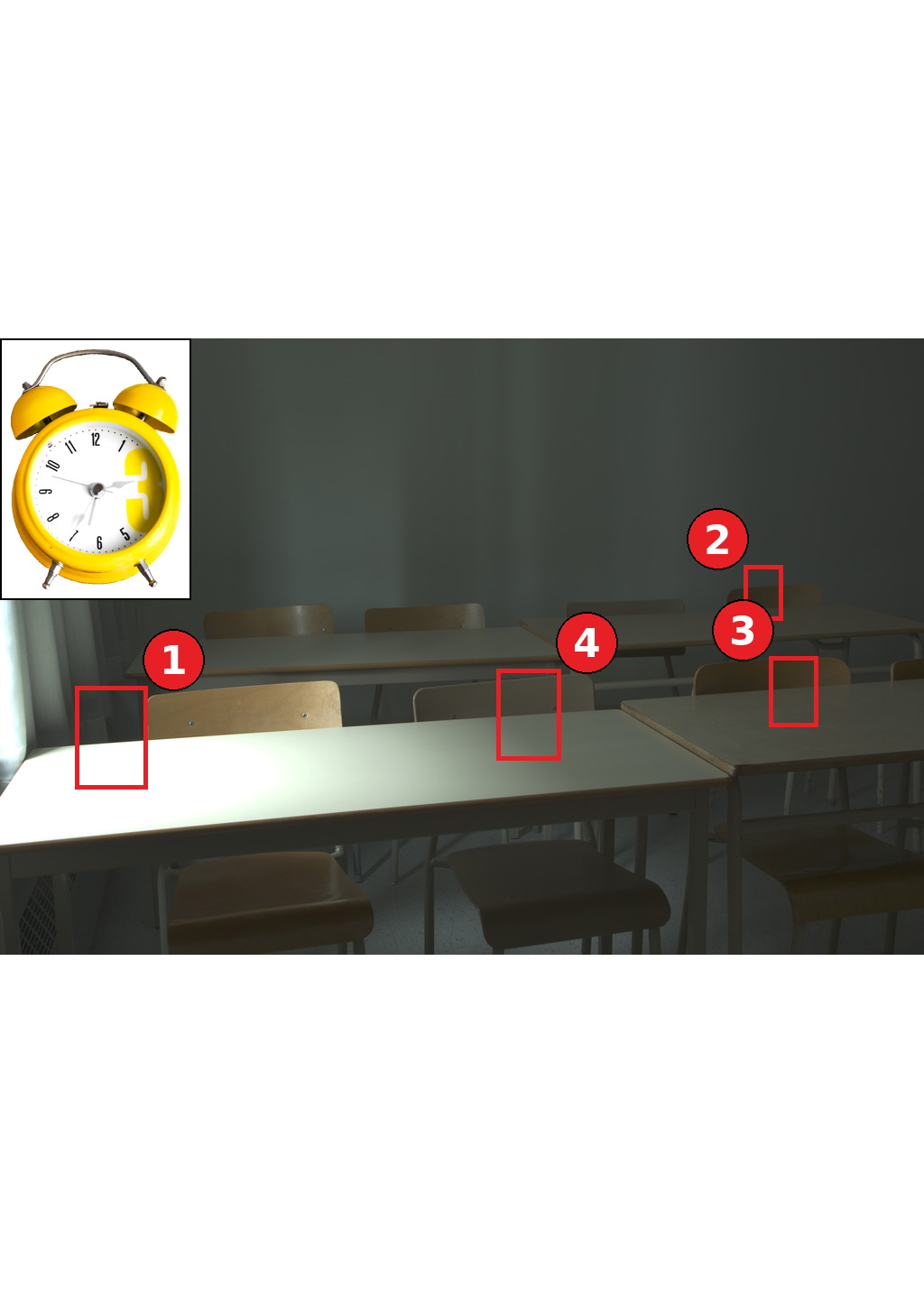}};
			\useasboundingbox (img.south west) rectangle (img.north east);
			\node[anchor=north, font=\scriptsize, text=black, fill=white, inner xsep=3pt, inner ysep=1pt, yshift=-20pt] at (img.north) {Input};
		\end{tikzpicture}%
	\end{minipage}\hspace{3pt}%
	\begin{tabular}[c]{@{}*{4}{c@{\hspace{3pt}}}c@{}}
		\includegraphics[width=.17\linewidth]{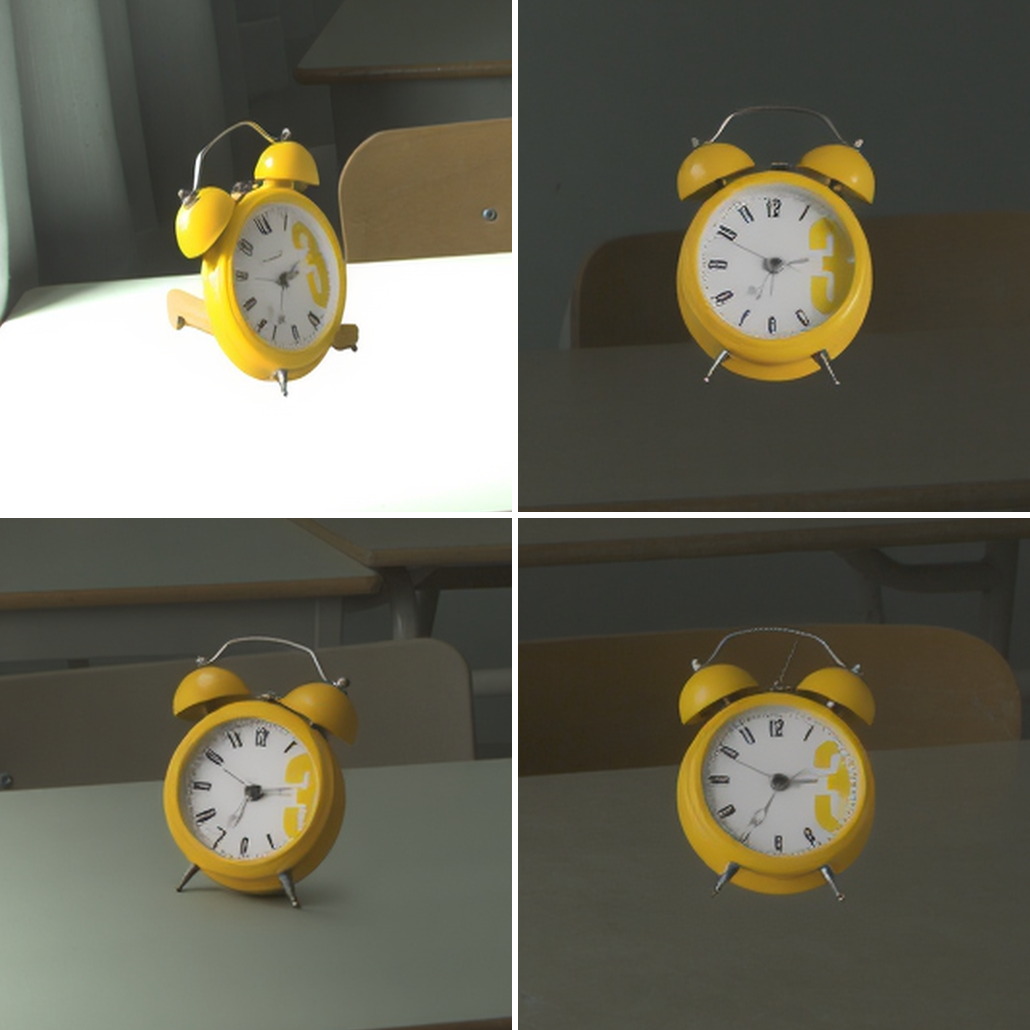}            &
		\includegraphics[width=.17\linewidth]{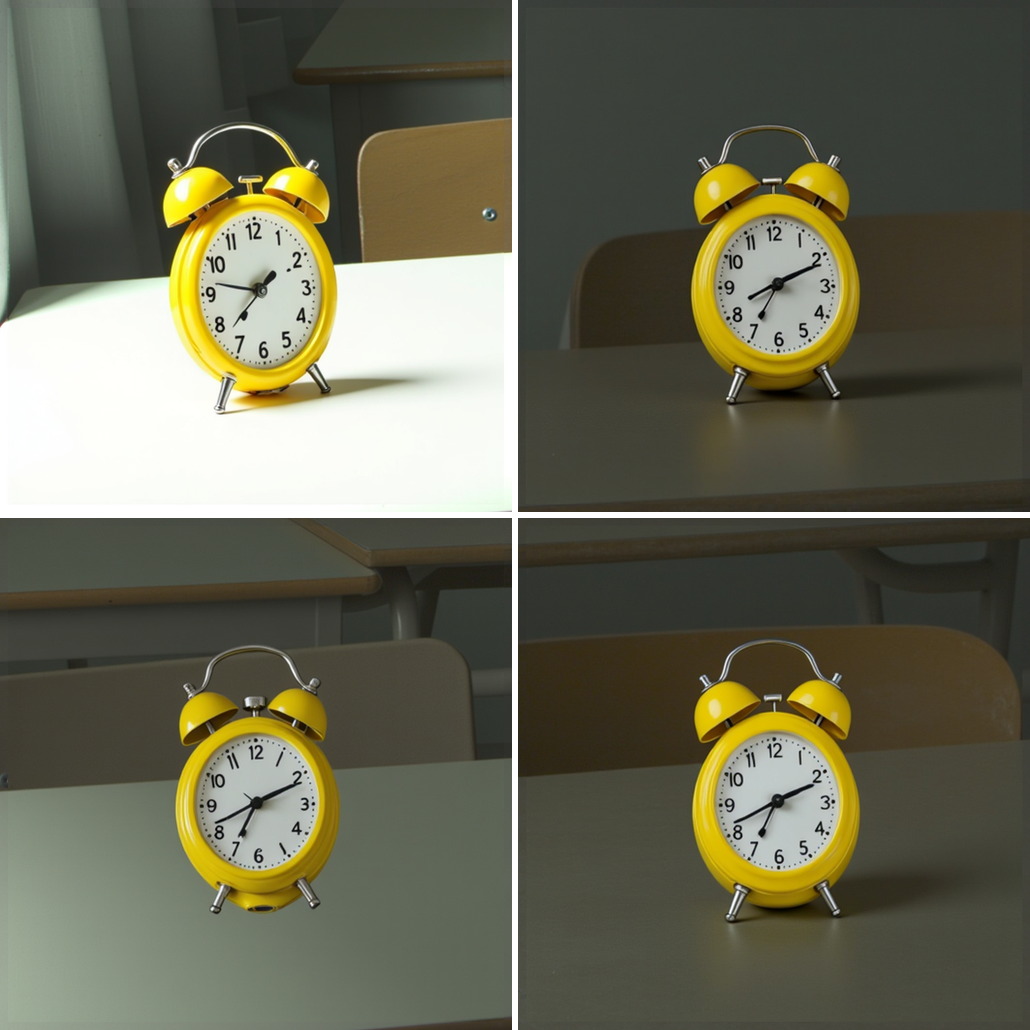}             &
		\includegraphics[width=.17\linewidth]{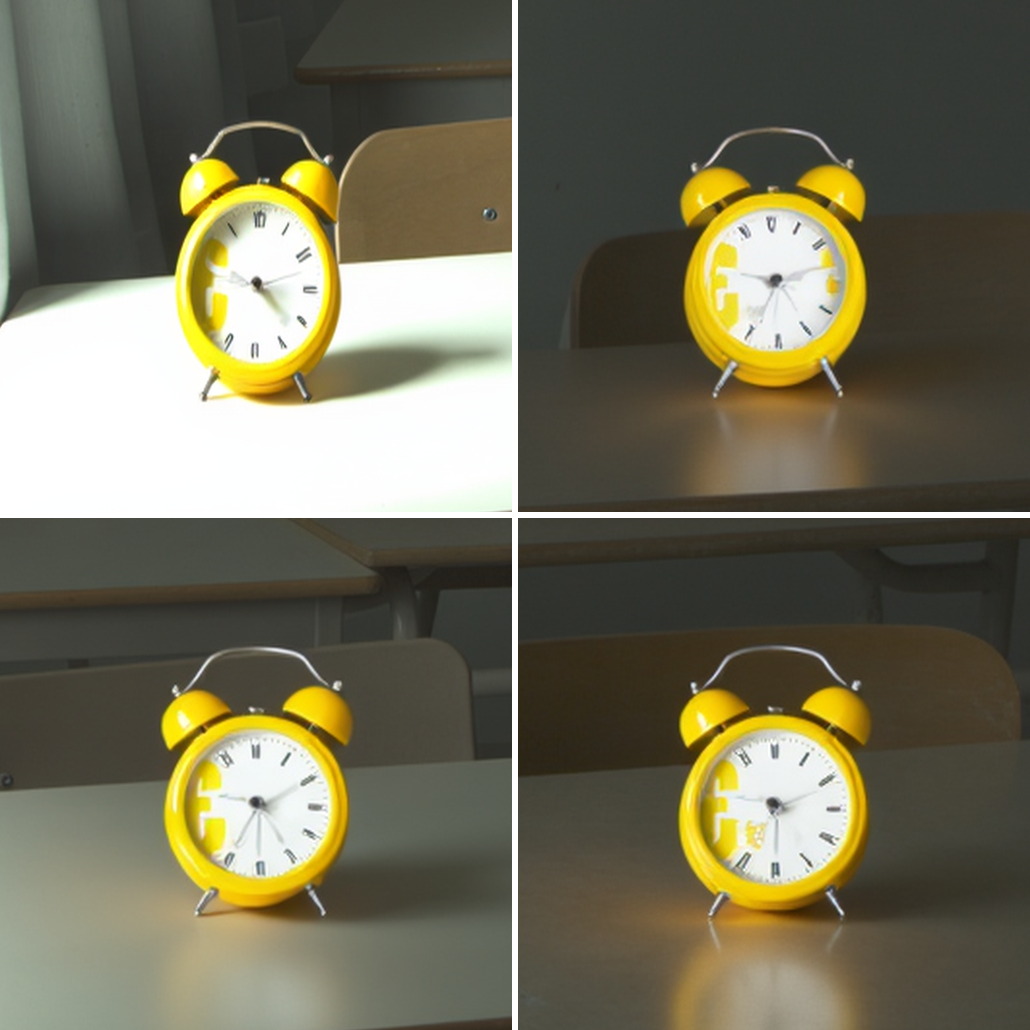}             &
		\includegraphics[width=.17\linewidth]{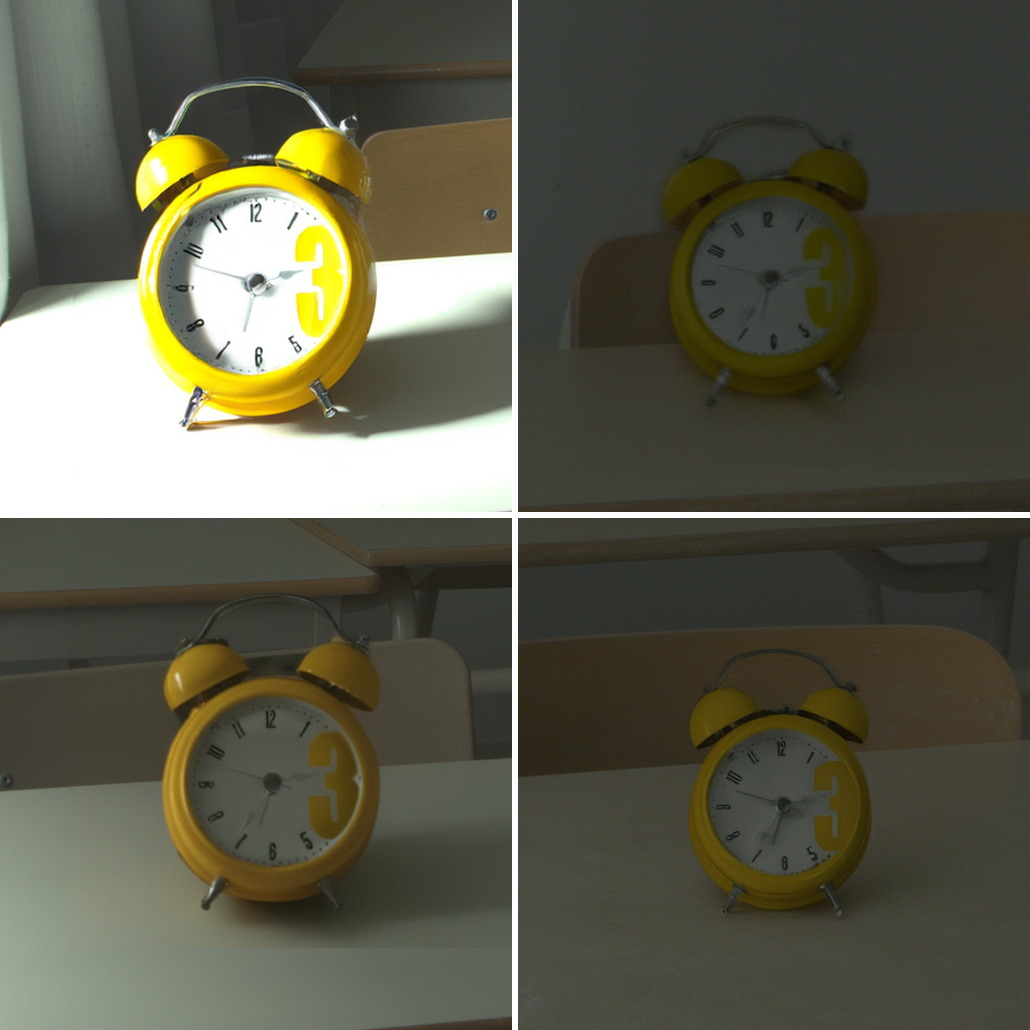}                  &
		\rotlabel{Composite}                                                                                                \\
		\includegraphics[width=.17\linewidth]{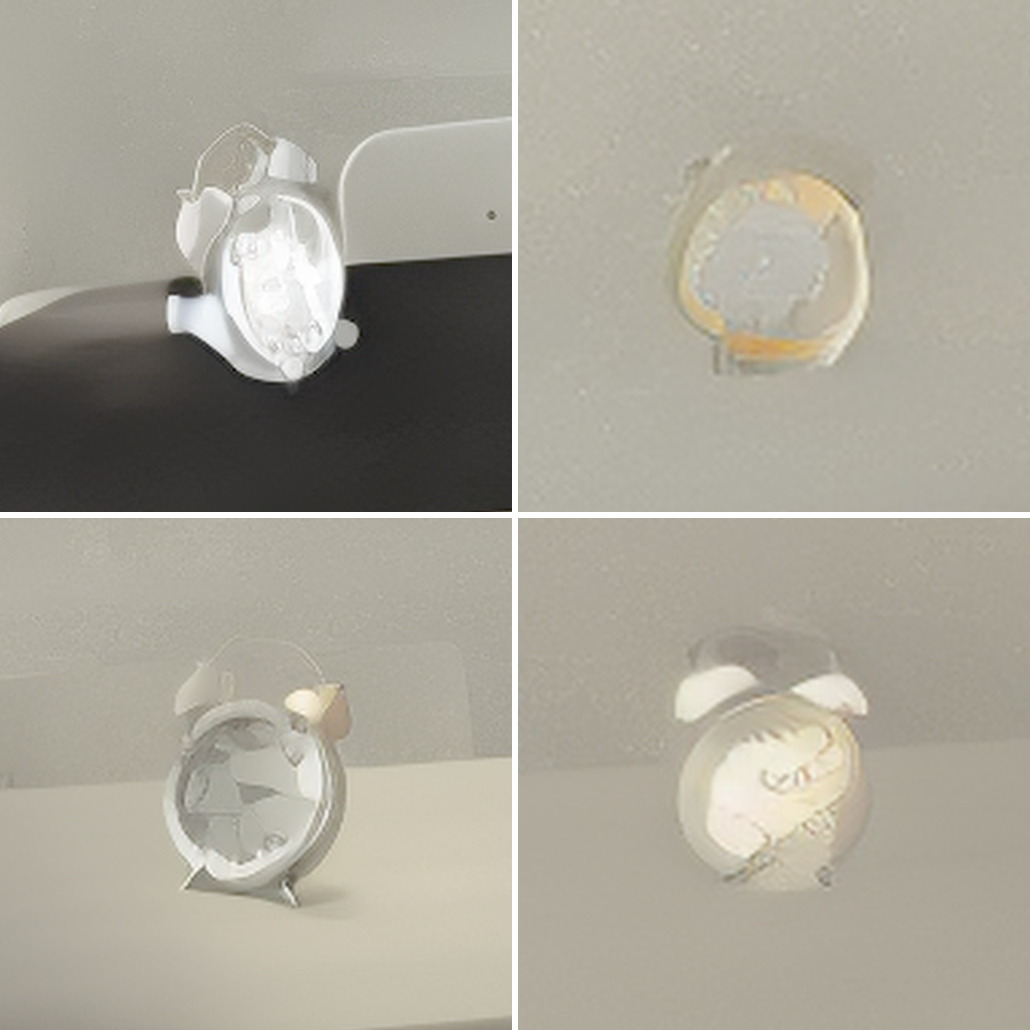} &
		\includegraphics[width=.17\linewidth]{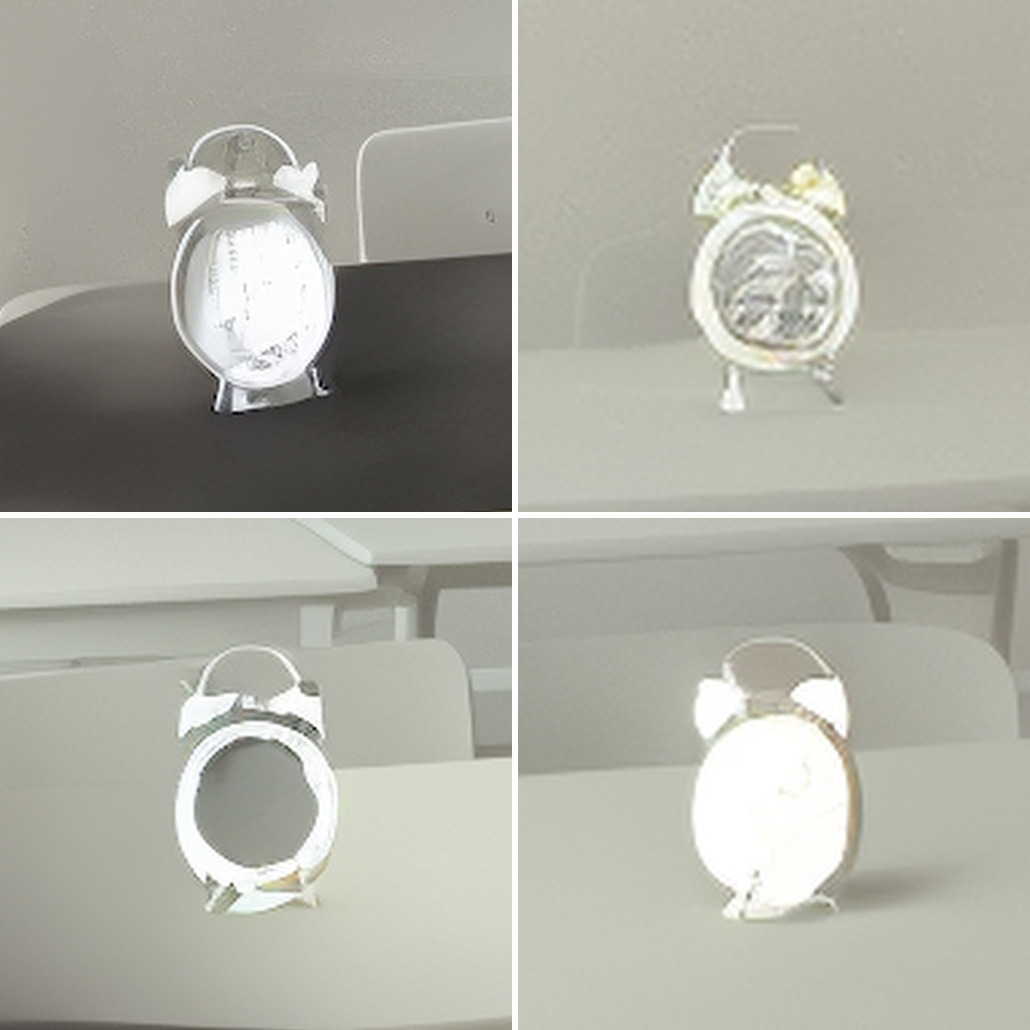}  &
		\includegraphics[width=.17\linewidth]{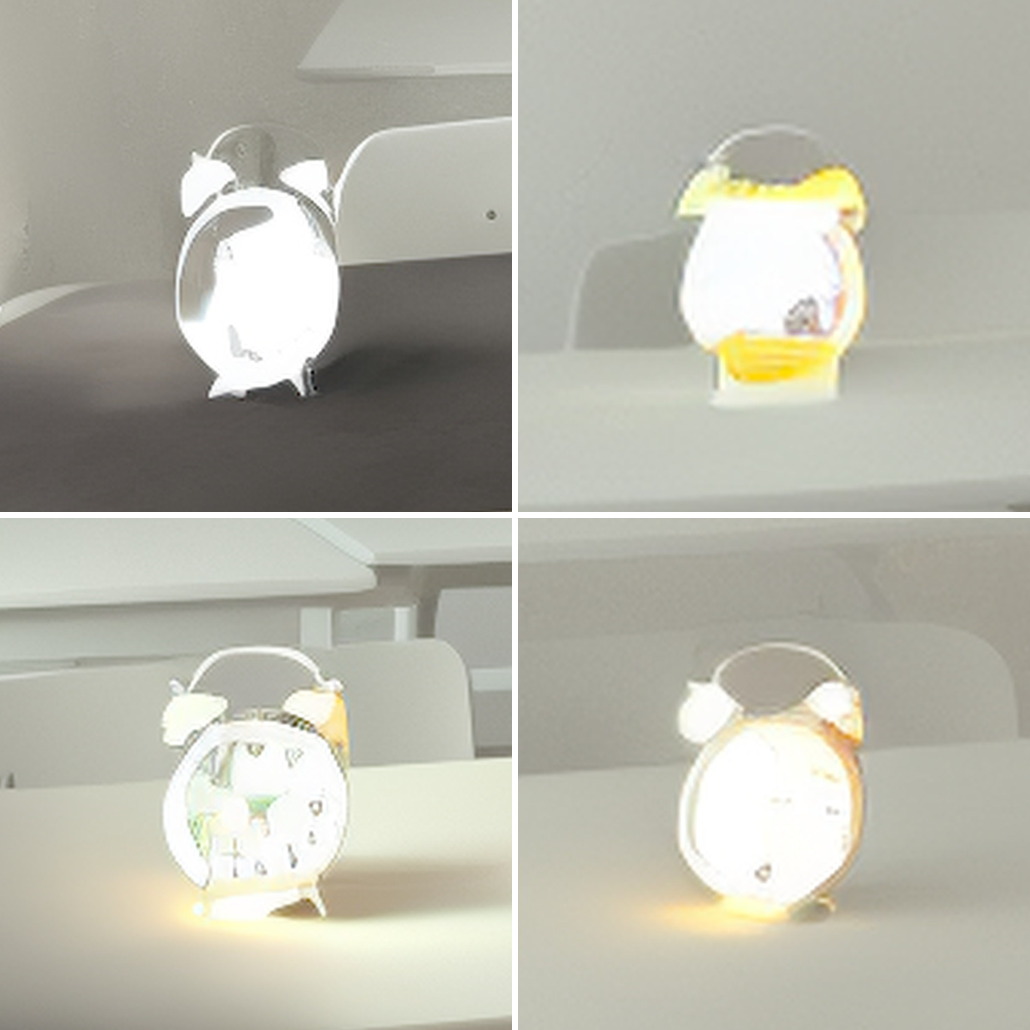}  &
		\includegraphics[width=.17\linewidth]{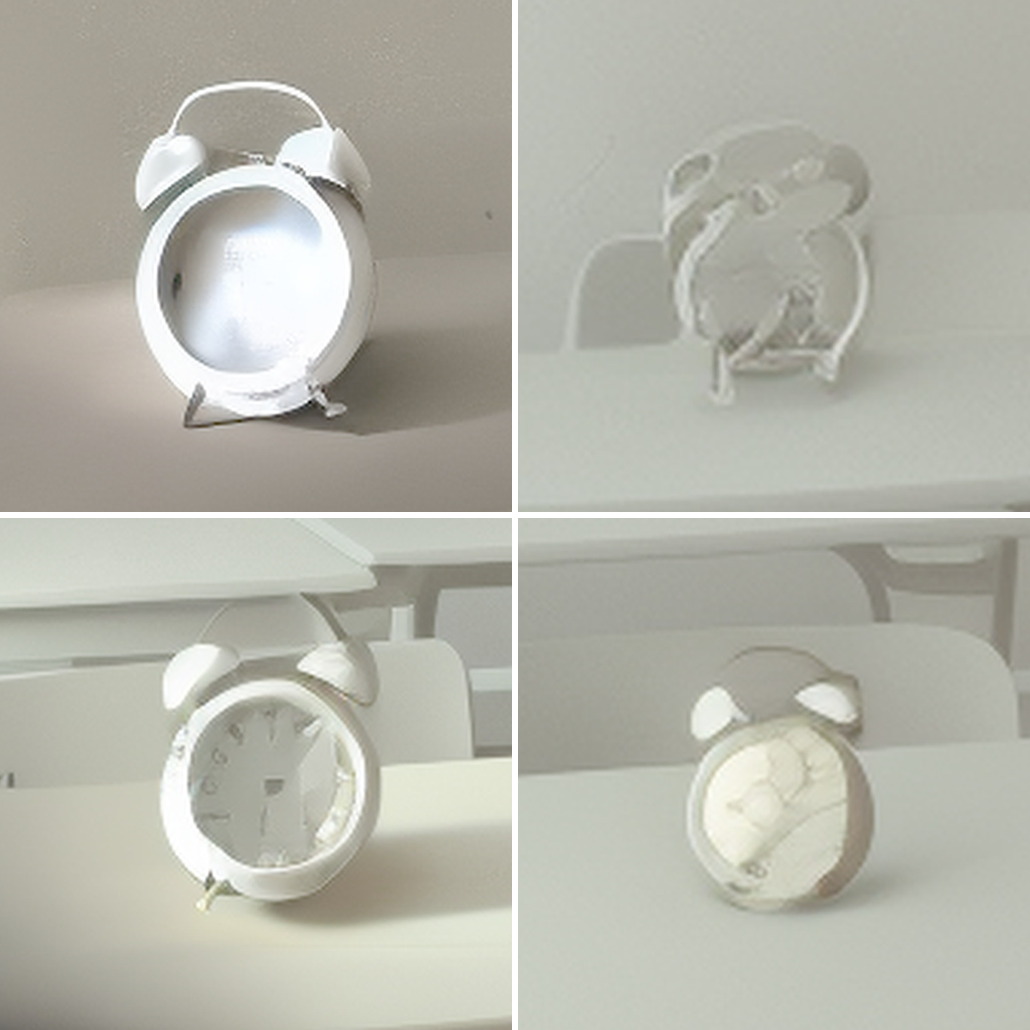}       &
		\rotlabel{Irradiance}                                                                                               \\
	\end{tabular}
	\caption{
		Estimated-irradiance comparison~\citep{zeng2024rgb} across generative compositing baselines.
		For baselines, the irradiance on the inserted object often differs from its surroundings, with the boundary discontinuity becoming more pronounced as the composite's shading deviates further from the local lighting.
		In contrast, \modelname{} yields irradiance that aligns with the surroundings and transitions smoothly across the boundary.
		For each input, the top row shows the composite output and the bottom row shows the estimated irradiance computed from that composite.
	}
	\label{fig:irradiance}
\end{figure}

\FloatBarrier
\vspace{4pt}
\noindent
\textbf{Comparison with ground truth on \benchname{}.}
Fig.~\ref{fig:bench_examples} compares \modelname{} with captured ground truth at two insertion locations per scene.
The examples highlight remaining differences in fine shading, partially observed regions, and transmissive materials.

\newcommand{\benchexample}[5]{%
	\begin{minipage}[c]{0.42\linewidth}
		\centering
		\begin{tikzpicture}[baseline=(img.south)]
			\node[inner sep=0] (img) {\includegraphics[width=\linewidth]{#1}};
			\useasboundingbox (img.south west) rectangle (img.north east);
			\node[anchor=south, font=\scriptsize, text=black, fill=white, inner xsep=3pt, inner ysep=1pt, yshift=2pt] at (img.north) {Input};
		\end{tikzpicture}
	\end{minipage}\hspace{2pt}%
	\begin{minipage}[c]{0.54\linewidth}
		\centering
		\begin{tabular}{@{}c@{\hspace{2pt}}c@{}}
			\makecell{\modelname}                                          &
			\makecell{GT}                                                    \\
			\includegraphics[width=.49\linewidth,height=.49\linewidth]{#2} &
			\includegraphics[width=.49\linewidth,height=.49\linewidth]{#3}   \\[2pt]
			\includegraphics[width=.49\linewidth,height=.49\linewidth]{#4} &
			\includegraphics[width=.49\linewidth,height=.49\linewidth]{#5}   \\
		\end{tabular}
	\end{minipage}%
}

\begin{figure}[htbp]
	\centering
	\scriptsize
	\begin{minipage}{0.49\linewidth}
		\centering
		\benchexample{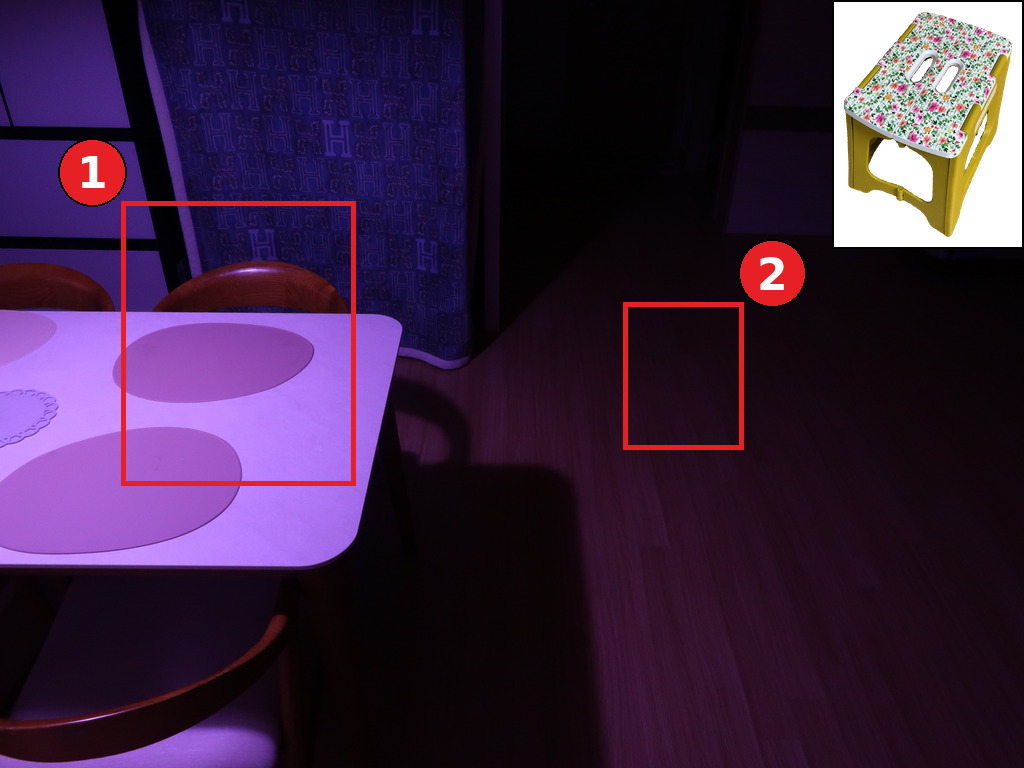}{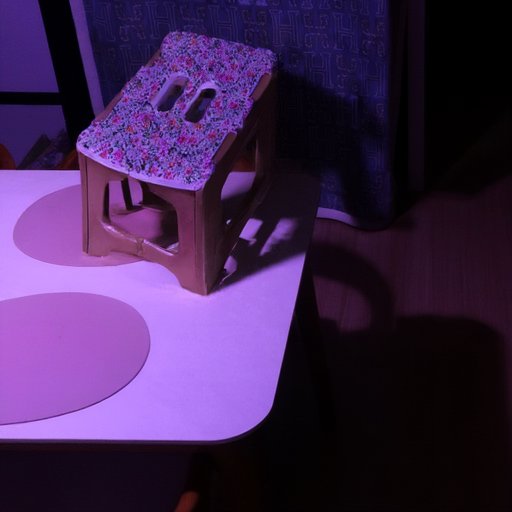}{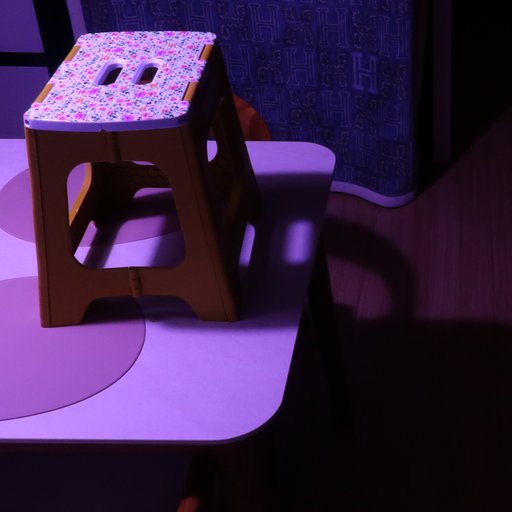}{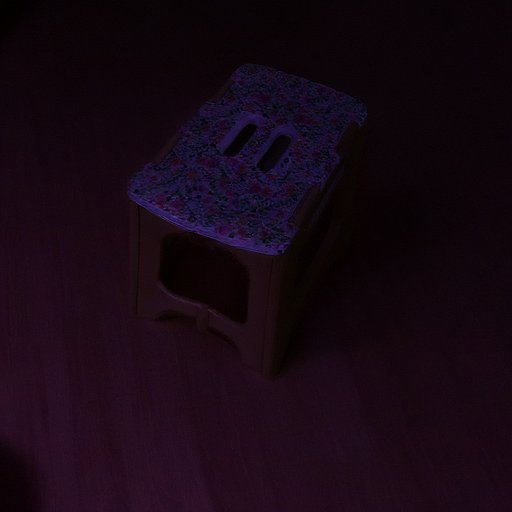}{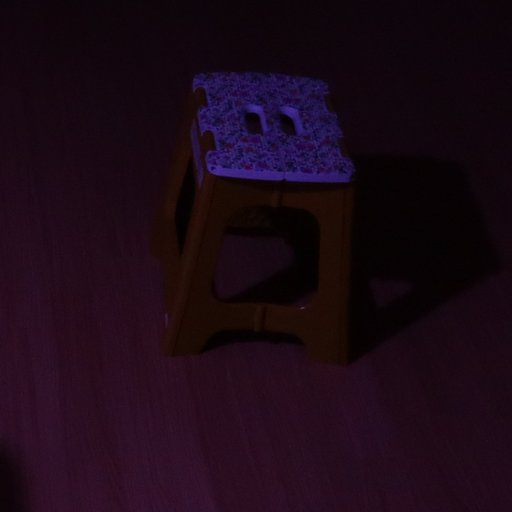}
	\end{minipage}\hfill
	\begin{minipage}{0.49\linewidth}
		\centering
		\benchexample{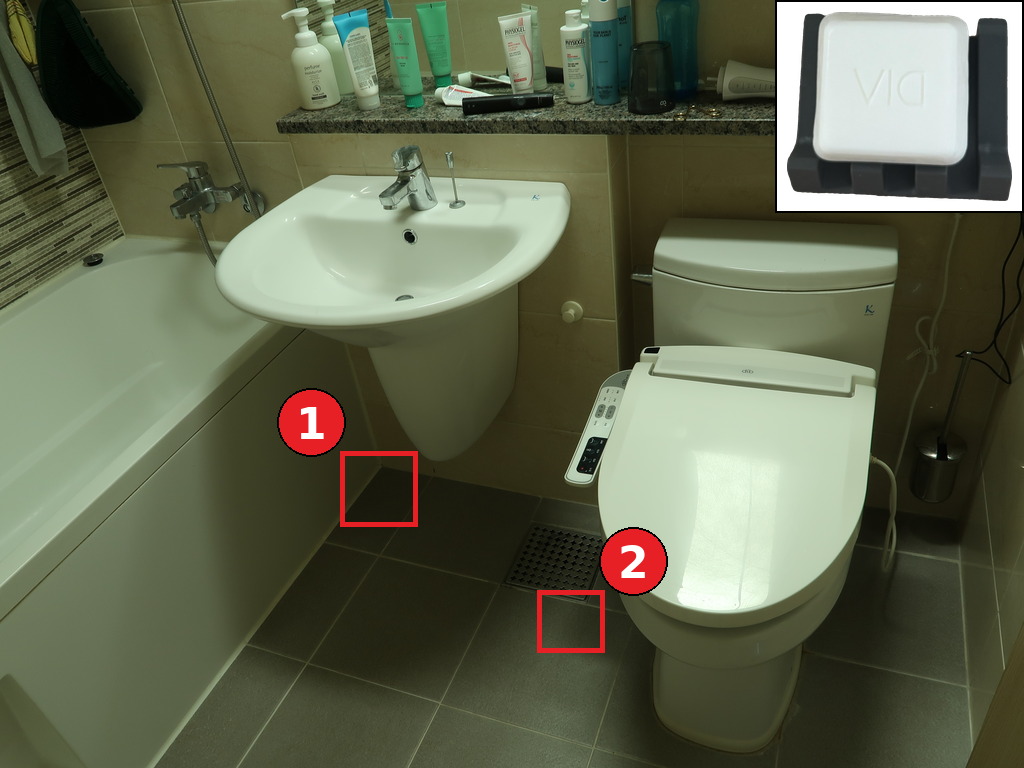}{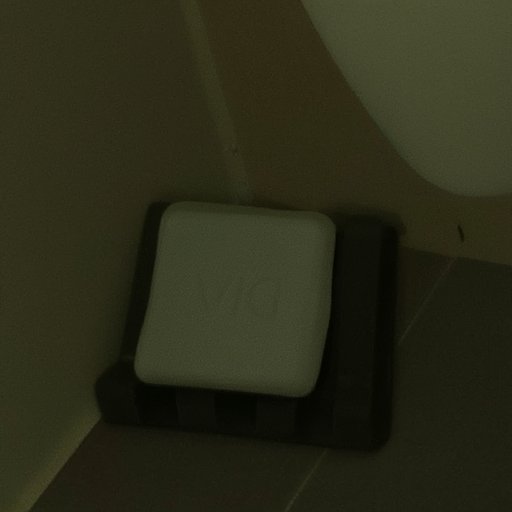}{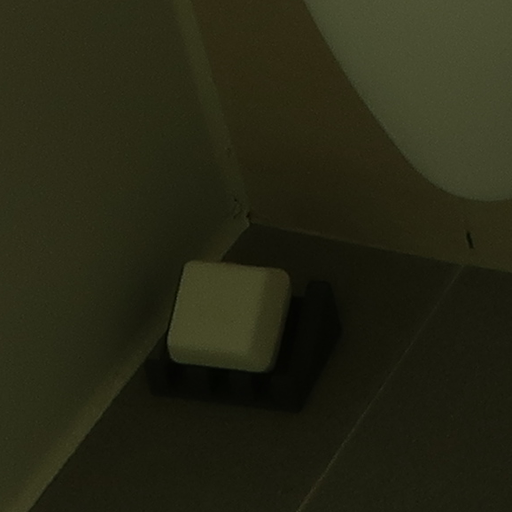}{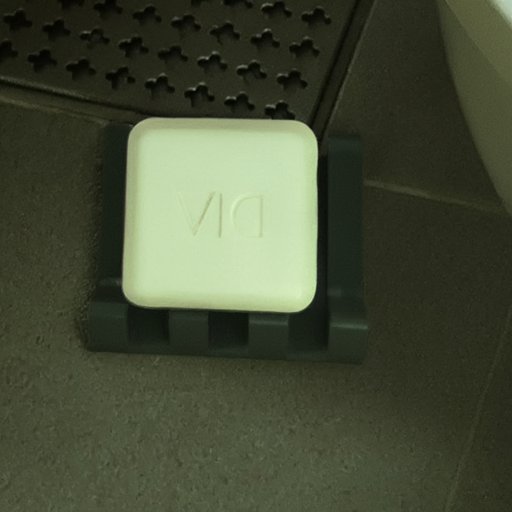}{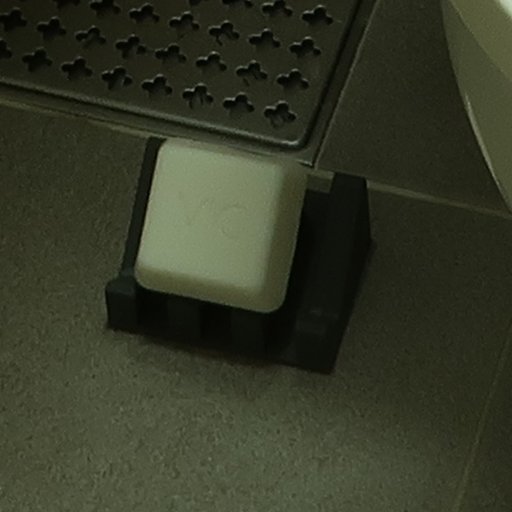}
	\end{minipage}\\[6pt]
	\begin{minipage}{0.49\linewidth}
		\centering
		\benchexample{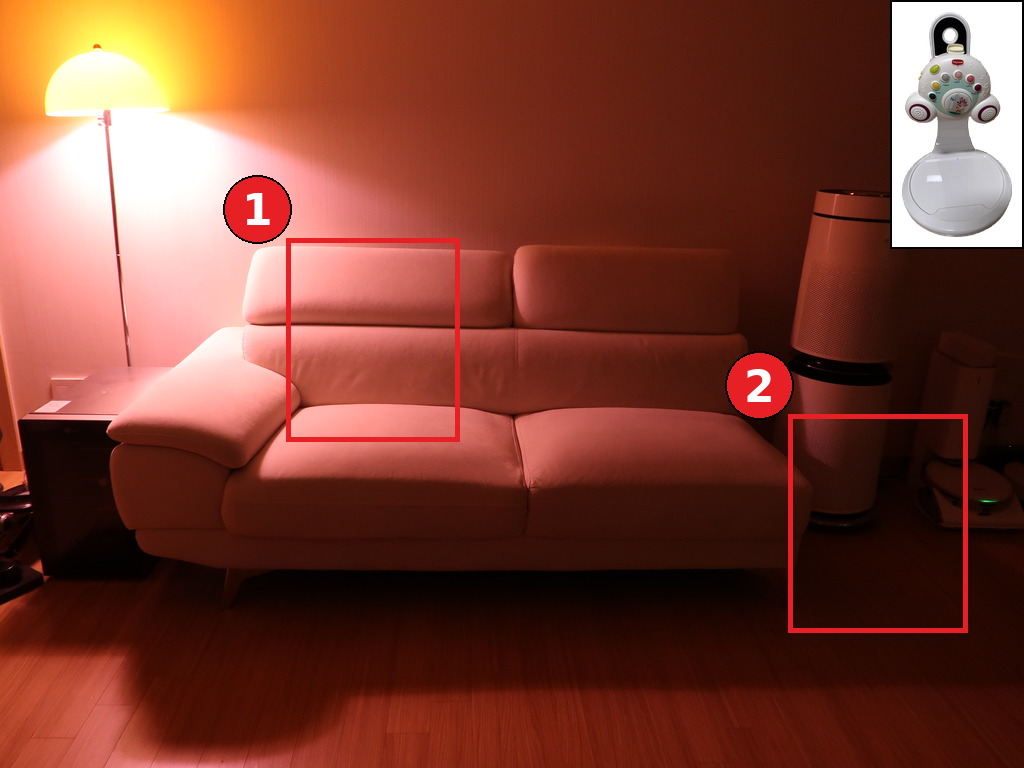}{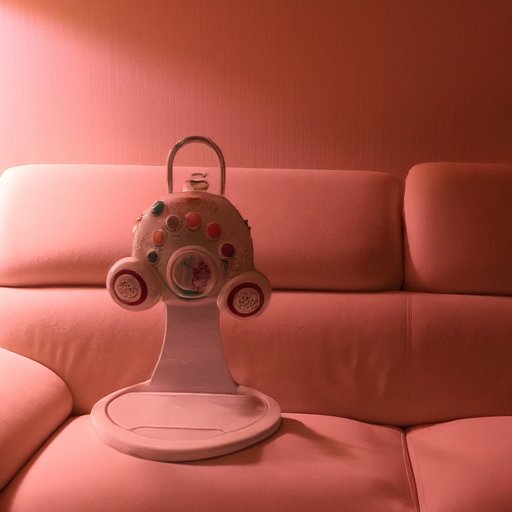}{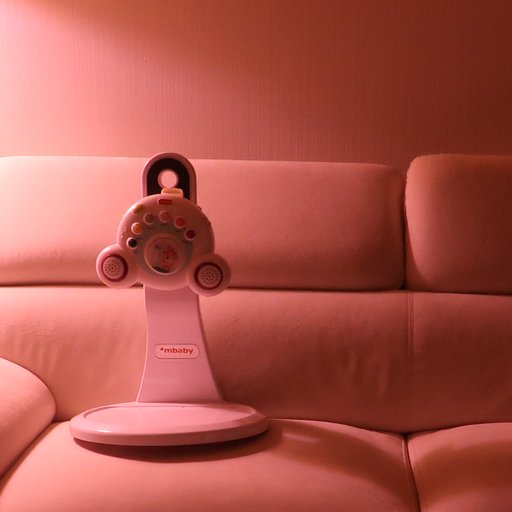}{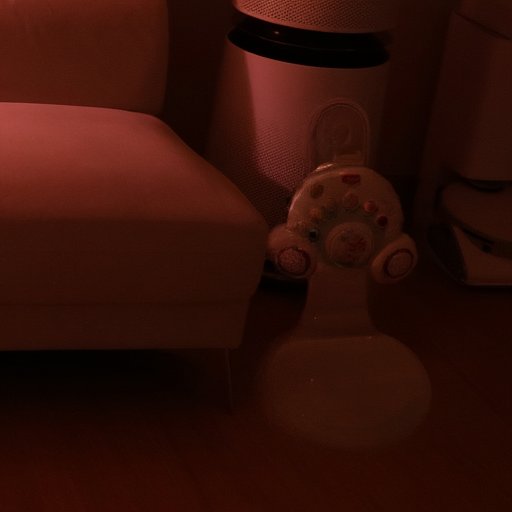}{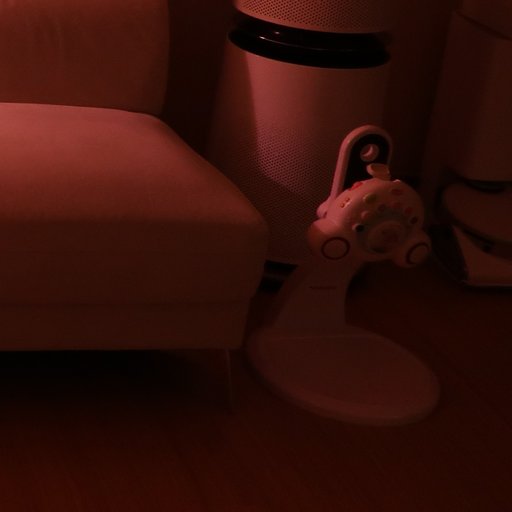}
	\end{minipage}\hfill
	\begin{minipage}{0.49\linewidth}
		\centering
		\benchexample{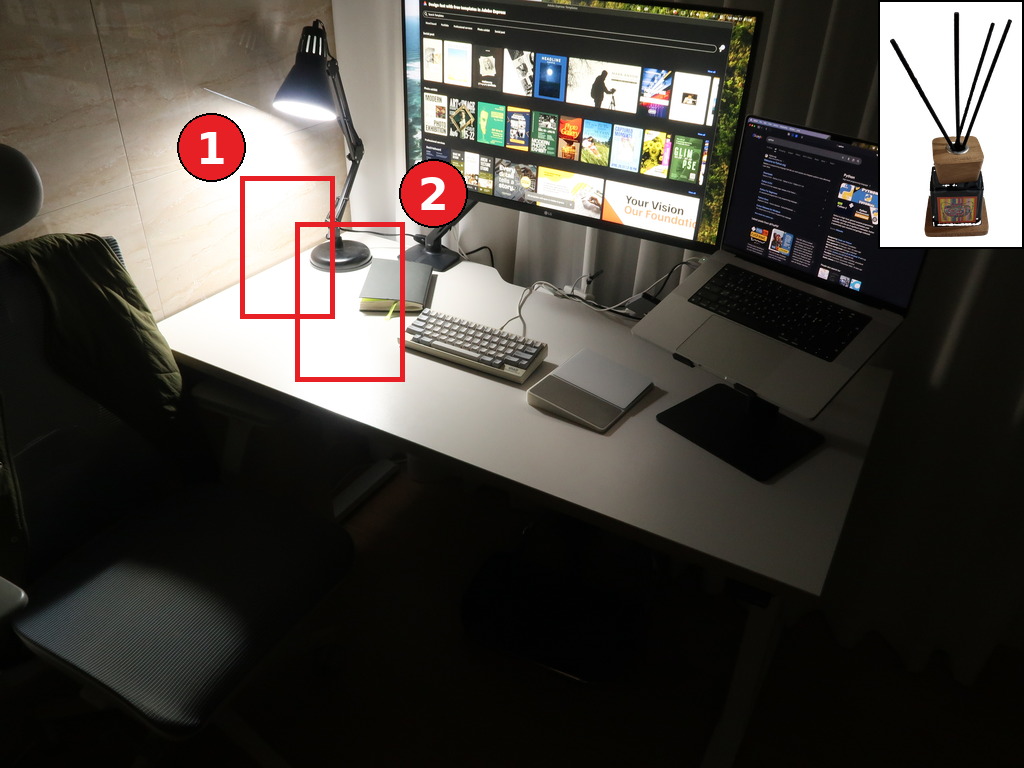}{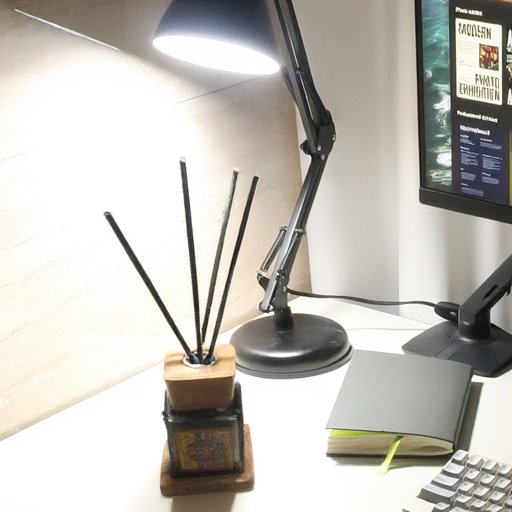}{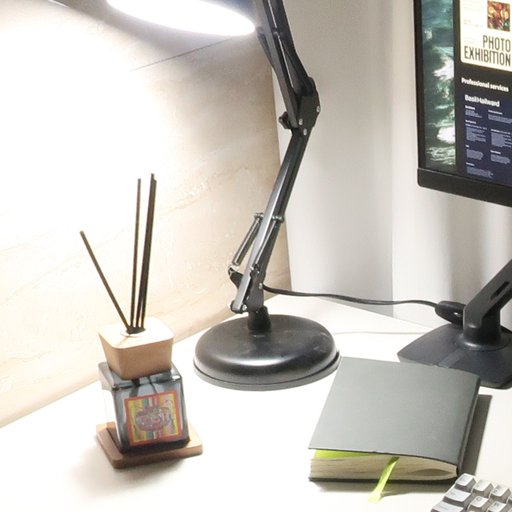}{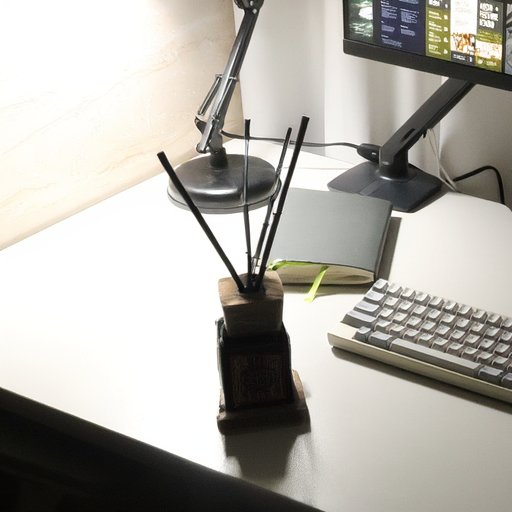}{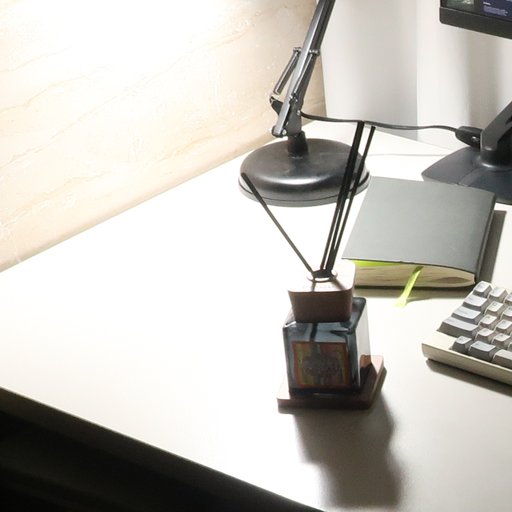}
	\end{minipage}
	\caption{
		Challenging samples from \benchname{} under spatially varying illumination.
		Each panel shows one object-scene example: the input composite on the left, and two (\modelname{}, ground truth) comparisons on the right, corresponding to two locations within the same scene.
		The examples include strong colored lighting, transmissive materials, and viewpoints where parts of the object are weakly observed in the input.
		While \modelname{} reproduces the dominant lighting effects, comparison with the ground truth reveals remaining differences in fine shading details, particularly on regions partially occluded in the input view, as well as transmissive effects.
	}
	\label{fig:bench_examples}
\end{figure}

\FloatBarrier
\subsection{Evaluation Protocol Details}
\label{app:eval_protocol}

This subsection documents four evaluation choices: how the user study was conducted, why we adopt DreamSim as the primary perceptual metric on \benchname, the rationale and computation of the IC metric, and the direct comparison with ground-truth composites that complements IC on \benchname.

\vspace{4pt}
\noindent
\textbf{User study.}
\label{app:user_study}
We run a pairwise user study comparing \modelname{} against each baseline.
Each trial presents \modelname{} and a randomly sampled baseline on the same (object, scene) pair, with randomized left/right ordering.
Above the candidates, we show the input object and the target background scene.
The candidates contain the object inserted at multiple locations; for \benchname{}, the ground-truth composite is additionally shown.

Participants then judge the two candidates on three criteria:
\begin{itemize}[leftmargin=1.2em,itemsep=2pt,topsep=2pt]
	\item \emph{Harmony}: whether the inserted object's shading matches the surrounding illumination
	\item \emph{Identity}: whether the object's shape, texture, and color are preserved
	\item \emph{Quality}: overall compositing quality, including illumination plausibility and visual coherence
\end{itemize}
For each criterion the participant chooses one of \emph{A} (left), \emph{B} (right), or \emph{Tie}.
Laval Indoor SV HDR is evaluated without a reference, while \benchname{} includes the ground-truth composite.

\textit{Protocol.}
The study is conducted through a custom web-based interface (Fig.~\ref{fig:user_study_ui}).
Each participant completes 10 trials per dataset.
We collect 38 responses on \benchname{} and 33 on Laval after excluding inattentive responses; we exclude a participant if they choose the same option (A, B, or Tie) across all three criteria in all 10 trials.

\begin{figure}[htbp]
	\centering
	\scriptsize
	\setlength{\tabcolsep}{2pt}
	\renewcommand{\arraystretch}{1.1}
	\begin{tabular}{@{}cc@{}}
		Laval Indoor SV HDR (no reference)                                                      & \benchname{} (ground-truth composite shown) \\
		\includegraphics[width=.49\linewidth]{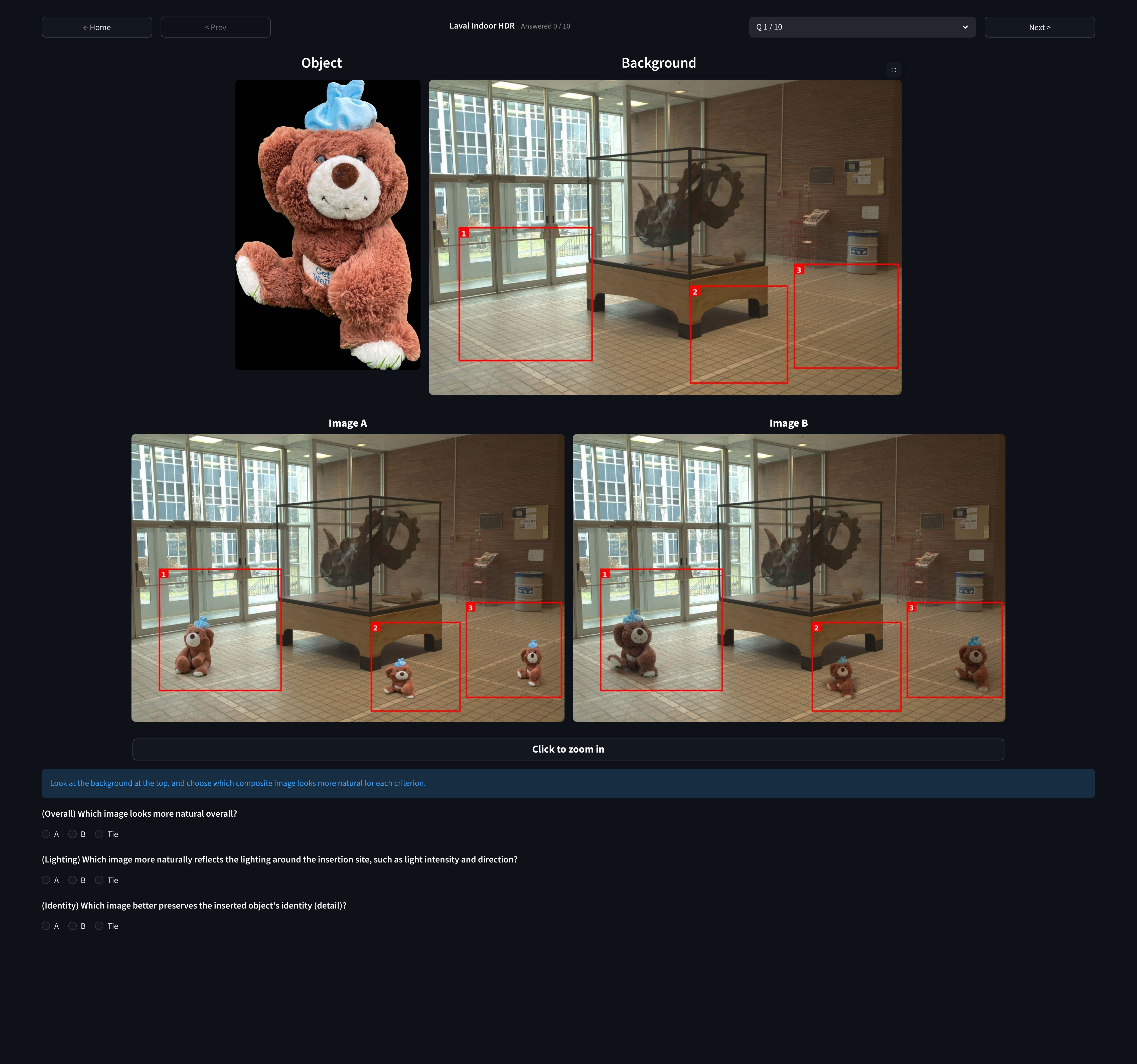} &
		\includegraphics[width=.49\linewidth]{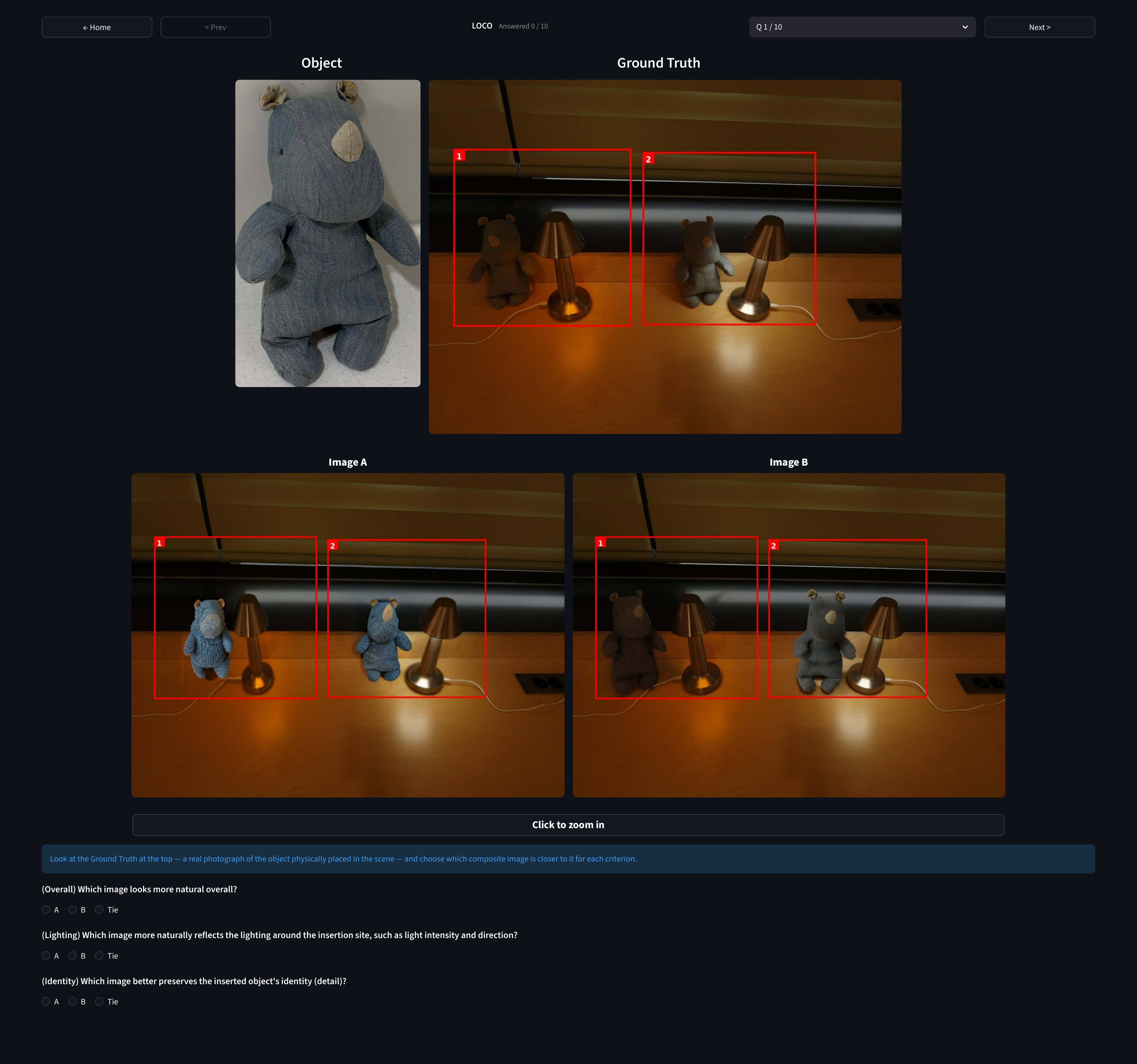}                                                \\
	\end{tabular}
	\caption{
		Web interface used for the user study.
		For each trial, two composites are shown side by side, with a reference (ground-truth composite) above on \benchname{} and no reference on Laval Indoor SV HDR.
		Participants choose A / B / Tie for each of the three criteria.
	}
	\label{fig:user_study_ui}
\end{figure}

\FloatBarrier
\vspace{4pt}
\noindent
\textbf{Choice of perceptual metric.}
\label{app:dreamsim}
On \benchname{}, the ground truth contains the object \emph{relit} to match the local illumination, so a suitable metric should reflect alignment in appearance rather than identity alone.
DINO produces features designed to be robust to style variations~\citep{simeoni2025dinov3}, placing greater emphasis on shape and texture similarity than on appearance differences such as shading.
As a result, a \emph{copy-and-paste} prediction (directly inserting the input object without relighting) can still achieve a high DINO score.

DreamSim~\citep{fu2023dreamsim} is a learned perceptual metric trained on human similarity judgments over mid-level attributes such as style, color, pose, and layout, making it more sensitive to appearance differences that include shading.
Predictions whose shading better matches the ground truth tend to receive lower DreamSim distance, while copy-and-paste composites are penalized.
Fig.~\ref{fig:dino_vs_dreamsim} illustrates this on \benchname{} samples: Insert Anything largely preserves the input object's appearance and scores higher on DINO than \modelname{}, whereas \modelname{}'s relit prediction is closer to the ground truth under DreamSim.
We therefore adopt DreamSim as the primary perceptual metric on \benchname{}.

\begin{figure}[htbp]
	\centering
	\setlength{\tabcolsep}{2pt}
	\renewcommand{\arraystretch}{1.05}
	\scriptsize
	\newcommand{\dvdpanelh}{.165\linewidth}
	\begin{tabular}{@{}cccc@{}}
		Input                                                                                                       & Insert Anything                 & \modelname{} (Ours) & GT \\
		\includegraphics[height=\dvdpanelh]{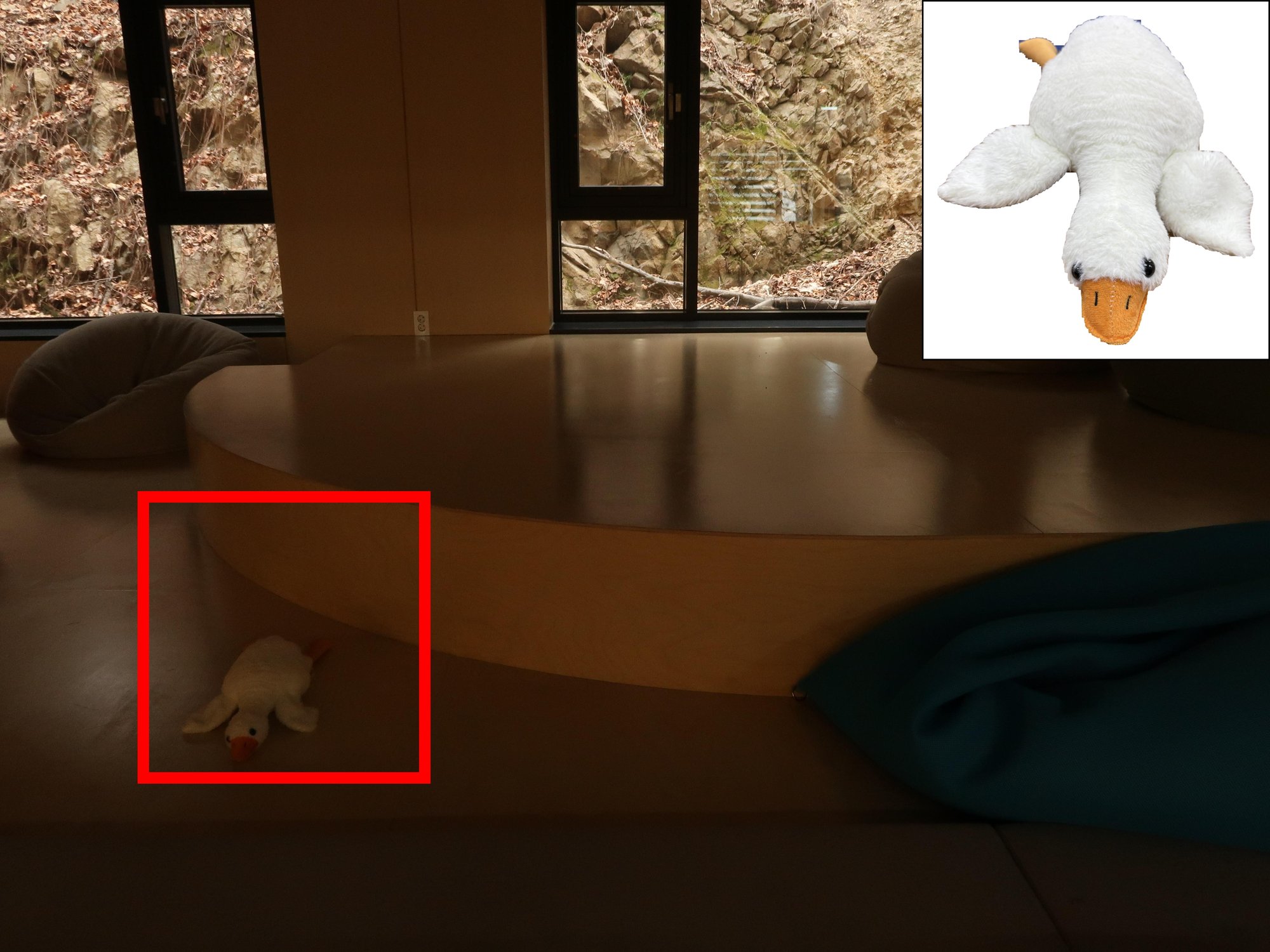}              &
		\includegraphics[height=\dvdpanelh]{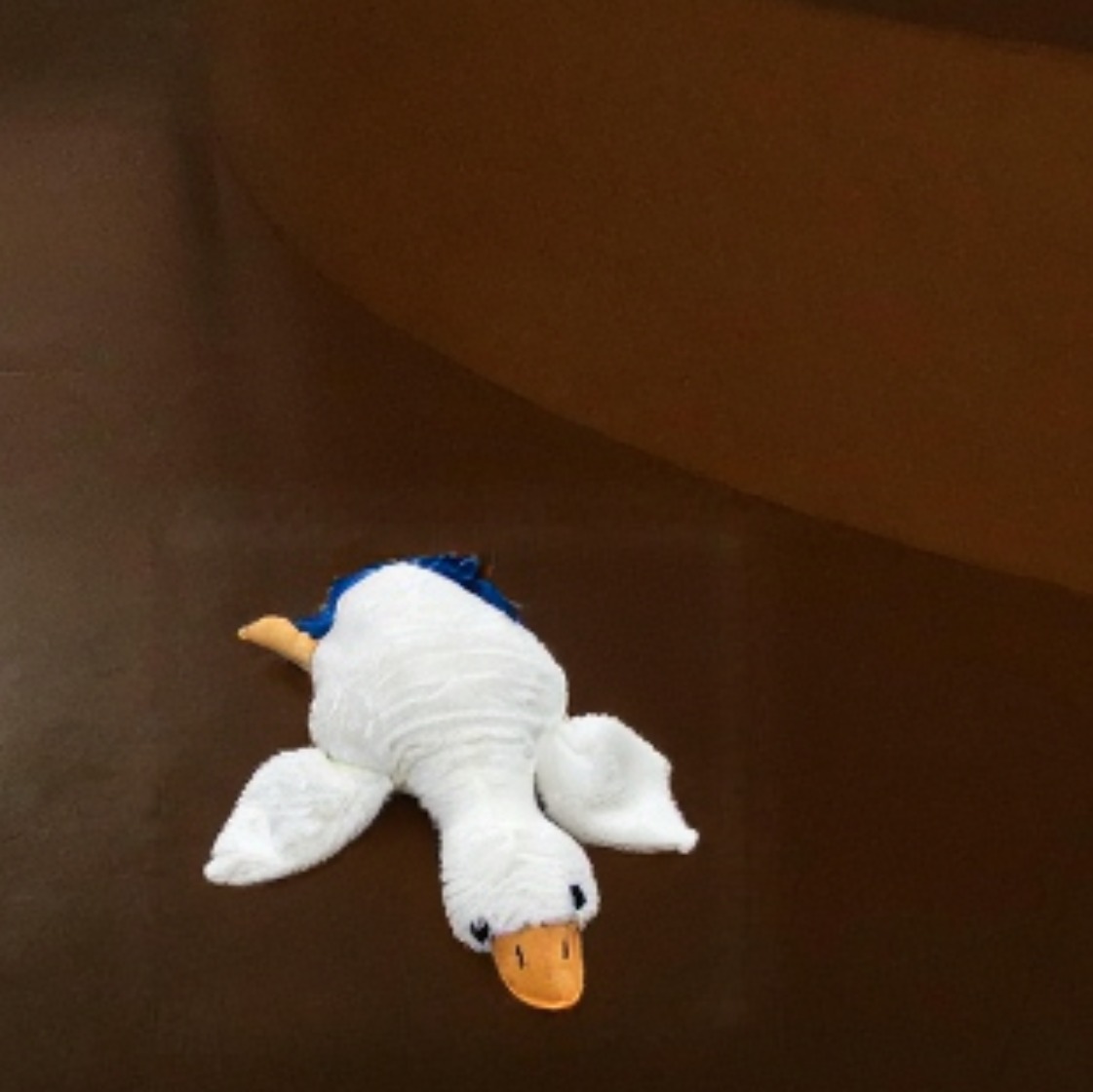} &
		\includegraphics[height=\dvdpanelh]{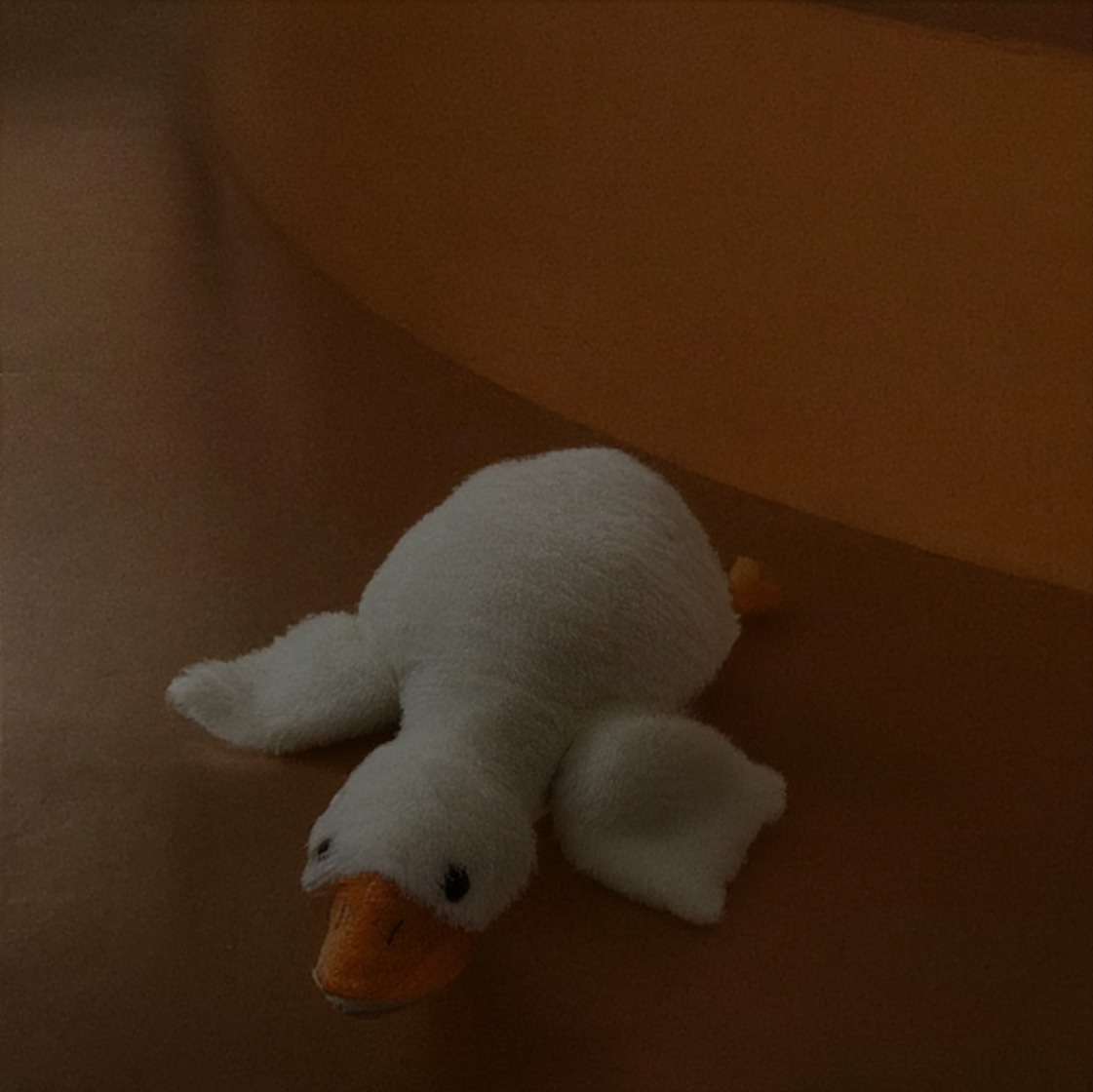}               &
		\includegraphics[height=\dvdpanelh]{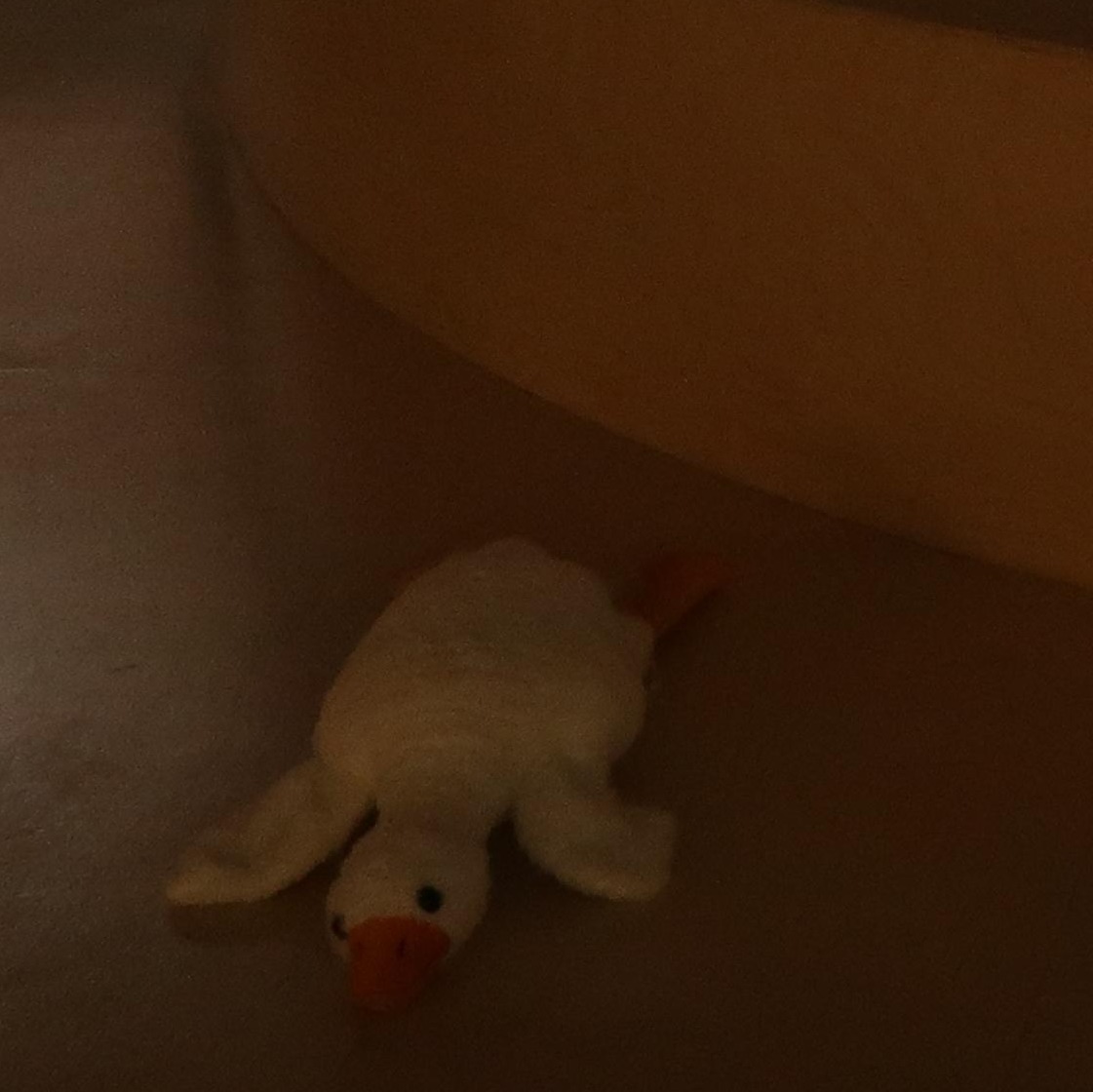}                                                                              \\
		                                                                                                            & \makecell{DINO\,$\uparrow$0.733                            \\DreamSim\,$\downarrow$0.444}
		                                                                                                            & \makecell{DINO\,$\uparrow$0.542                            \\DreamSim\,$\downarrow$0.268} & \\[4pt]
		\includegraphics[height=\dvdpanelh]{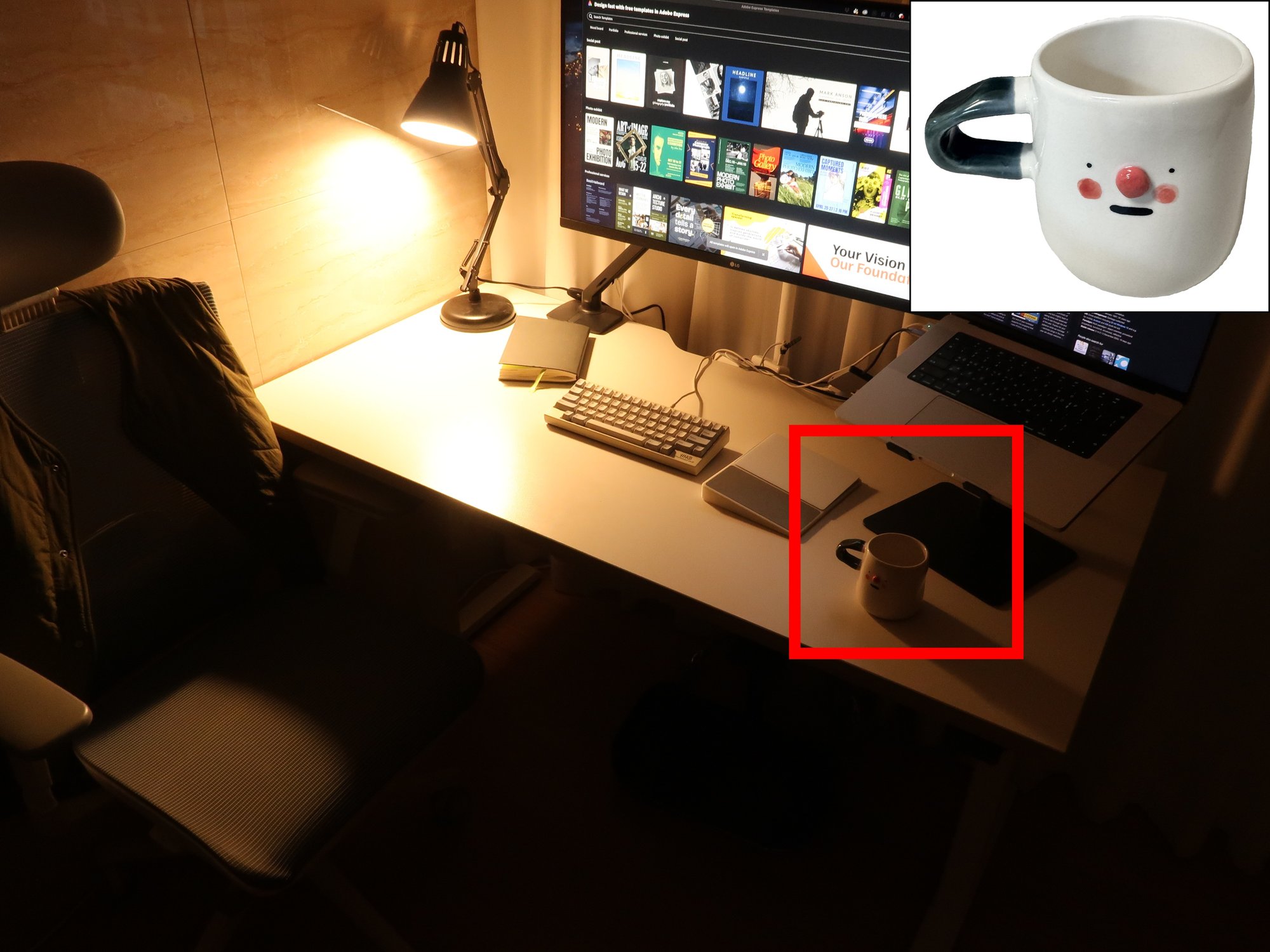}                &
		\includegraphics[height=\dvdpanelh]{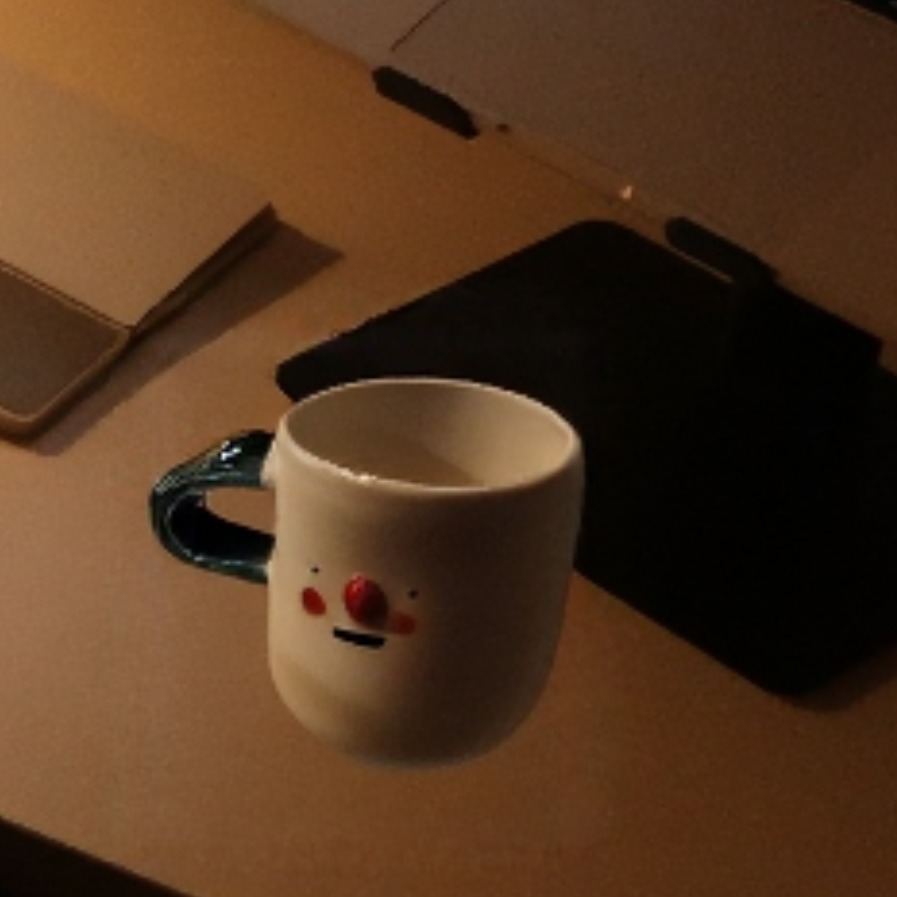}   &
		\includegraphics[height=\dvdpanelh]{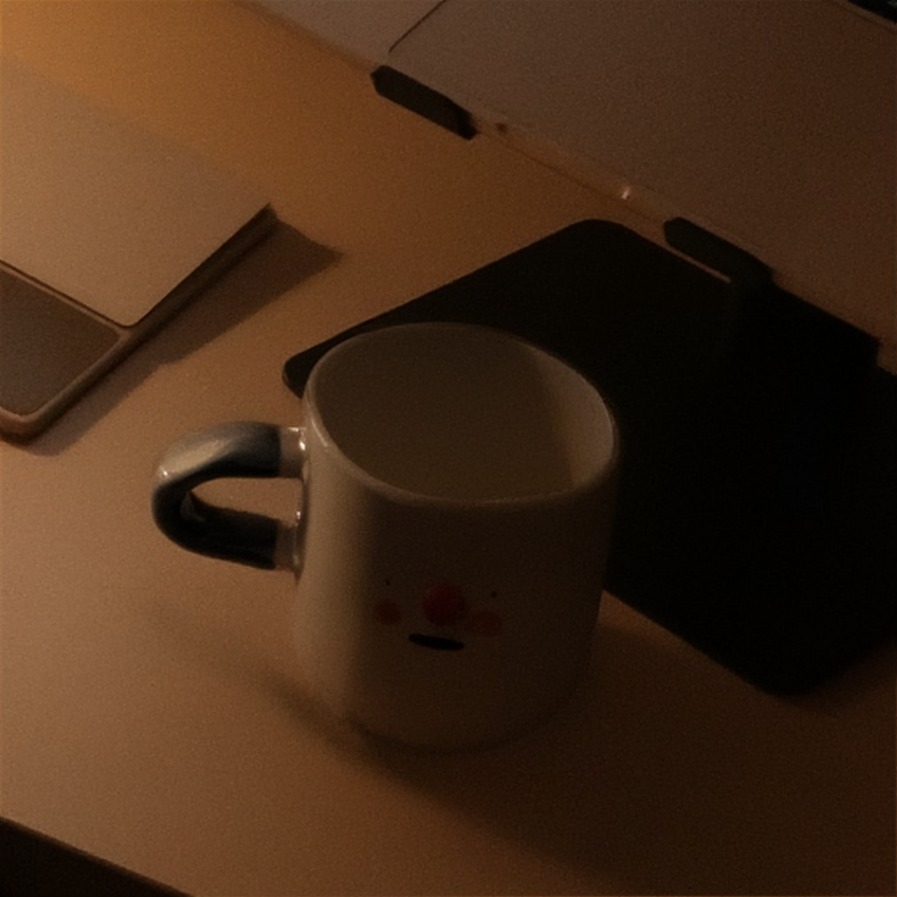}                 &
		\includegraphics[height=\dvdpanelh]{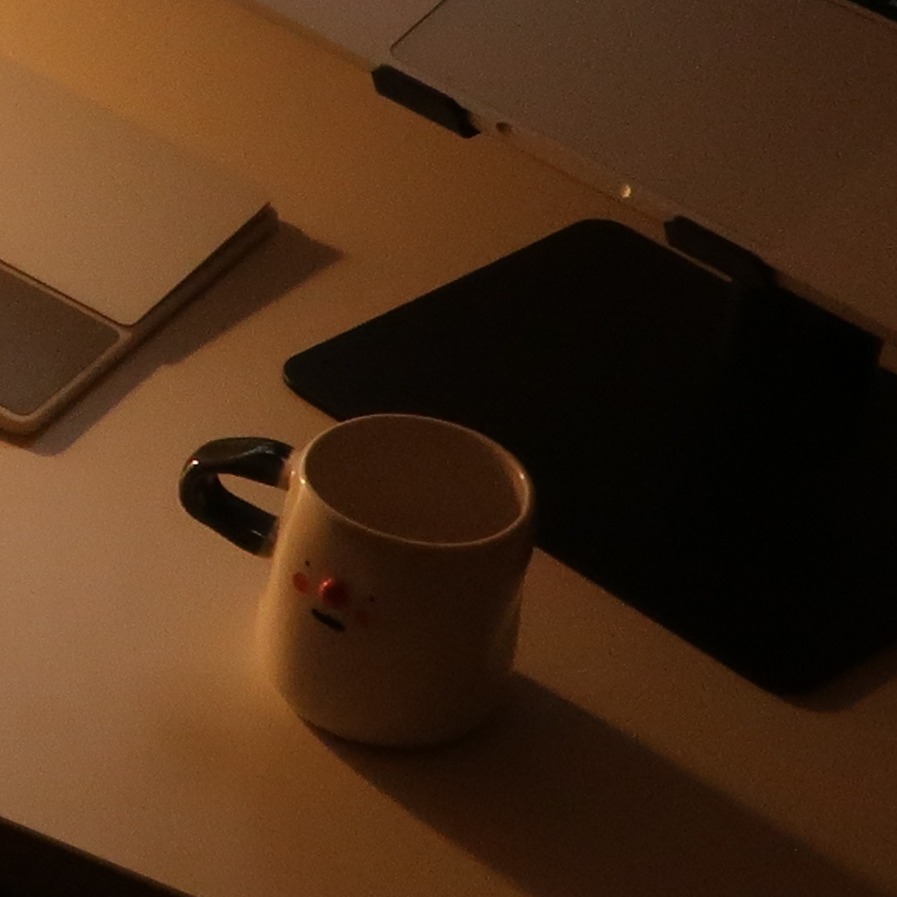}                                                                                \\
		                                                                                                            & \makecell{DINO\,$\uparrow$0.921                            \\DreamSim\,$\downarrow$0.229}
		                                                                                                            & \makecell{DINO\,$\uparrow$0.898                            \\DreamSim\,$\downarrow$0.127} & \\
	\end{tabular}
	\caption{
		Sample-level disagreement between DINO and DreamSim on \benchname{}.
		Insert Anything largely preserves the input object's appearance, so its prediction remains close in shape and texture and receives a higher DINO score than \modelname{}'s relit output.
		DreamSim, which reflects perceptual differences in attributes such as color and shading, assigns a lower (better) distance to \modelname{}, consistent with visual similarity to the ground truth.
		Each score is computed between the prediction and the ground-truth crop in the same row.
	}
	\label{fig:dino_vs_dreamsim}
\end{figure}

\FloatBarrier
\vspace{4pt}
\noindent
\textbf{IC metric details.}
\label{app:icmetric}
DreamSim requires a ground-truth composite as reference, which limits its applicability to in-the-wild scenes such as Laval Indoor SV HDR.
For such settings, we additionally use a reference-free illumination consistency (IC) measure, which operates on an estimated lighting field and provides a signal largely independent of object identity.

The intuition behind IC is that the scene's light sources remain unchanged by insertion.
If the inserted object's shading matches the local lighting, an illumination estimator's prediction at the insertion region should remain close to its pre-compositing prediction; mismatched shading may shift it.
We use Li~\etal~\citep{li2020inverse} for its per-pixel predictions.

\textit{Computation.}
For each pixel inside the insertion bounding box, the estimator outputs an $8\times16$ equirectangular environment map (decoded from 12 spherical Gaussians) oriented to the pixel's surface normal.
To preserve angular detail under normal-based rotations, we reproject each per-pixel envmap onto a shared higher-resolution grid ($64\times128$) in world coordinates and average over all pixels in the bounding box.
IC is computed as the PSNR between this bbox-level envmap before and after compositing.

\textit{Validation with ground truth.}
\benchname{} provides ground-truth composites, allowing a direct examination of the metric.
All evaluated methods score below the ground truth ($14.52$), with scores ranging from $7.88$ (AnyDoor) to $13.99$ (\modelname{}), and UniCombine as the strongest baseline at $11.80$.
\modelname{} is $2.19$ PSNR above the strongest baseline and $0.53$ PSNR below the ground truth.

\textit{Relation to human judgments.}
We further examine how IC relates to human preferences by comparing each baseline's IC with its harmony preference rate in the user study (\S\ref{app:user_study}).
The two tend to move in the same direction.
On Laval (no-reference), the two highest-IC baselines, DreamFuse ($13.27$) and UniCombine ($13.17$), correspond to lower \modelname{} harmony win rates ($82\%$ and $90\%$), while the four lowest-IC baselines are at $94\%$ or above (Spearman $\rho{=}0.87$).
On \benchname{} (reference-based), a similar but weaker trend is observed (Spearman $\rho{=}0.64$).
These observations suggest that IC captures aspects of local-lighting plausibility reflected in human judgments, although it does not replace human evaluation.

A practical caveat is the limited availability of publicly available per-pixel illumination estimators.
Li~\etal{}, the most recent we are aware of, outputs an $8\times16$ environment map per pixel.
The IC definition is independent of the estimator; substituting a stronger per-pixel model would likely improve measurement fidelity without altering the underlying metric.

\Needspace{8\baselineskip}
\vspace{4pt}
\noindent
\textbf{Illumination and appearance accuracy.}
\label{app:lighting_measurements}
IC compares the estimated lighting field before and after insertion, but it does not tell which aspect of the local illumination a method reproduces.
To complement it, we compare each prediction directly with the captured ground-truth composite on all 760 \benchname{} triplets, inside each insertion bounding box, on three quantities: the implied light direction, and the brightness and color of the inserted object.

\textit{Light direction.}
Extracting a single shadow axis is unreliable under overlapping indoor lights, and some methods change the object's orientation.
We therefore compare the light direction implied by each prediction with that of the ground truth.
From the per-pixel environment maps of the IC estimator, we compute the intensity-weighted mean direction over the bounding box and report (i) the L2 distance between the weighted mean directions and (ii) the angular error between their normalized directions.
The unnormalized L2 distance down-weights diffuse regions, where direction is poorly defined.

\textit{Local brightness.}
We convert the mean RGB inside the bounding box to HSV and report the absolute difference in value, $|\Delta V|$, against the ground truth.

\textit{Color shift.}
From the same HSV values, we report the cyclic hue difference, $\min(|\Delta h|,\,360^\circ - |\Delta h|)$, and the absolute saturation difference $|\Delta S|$.
We use hue and saturation rather than a scalar color temperature, which is undefined for arbitrary RGB illumination.

Tab.~\ref{tab:lighting_measurements_full} reports all five readouts for the generative compositing baselines of the main paper and for the harmonization and intrinsic-based methods of \S\ref{app:baseline_analysis}.
\modelname{} attains the best score on every measure.
Against the generative compositing baselines, it improves on the per-column best by 12\% in light direction (both readouts), 22\% in hue, 38\% in brightness, and 53\% in saturation.
The strongest baseline on the direction and brightness axes is PIH, a harmonization method, which \modelname{} leads by 12\% and 28\%.
Fig.~\ref{fig:irradiance} provides the corresponding spatial evidence: the irradiance estimated on \modelname{}'s composites follows the local brightness and shading variation of the scene.

\begin{table}[htbp]
	\centering
	\footnotesize
	\setlength{\tabcolsep}{5pt}
	\caption{
		Illumination and appearance accuracy on \benchname{} (760 triplets).
		Each quantity is measured inside the insertion bounding box against the captured ground-truth composite; lower is better.
		Light direction is the intensity-weighted mean direction of the per-pixel environment maps from the IC estimator, reported as an L2 distance and as an angular error.
		Brightness, hue, and saturation are differences of the bbox-mean color in HSV.
		The generative compositing rows are also shown in main paper Tab.~\ref{tab:lighting_measurements}.
	}
	\label{tab:lighting_measurements_full}
	\vspace{2pt}
	\begin{tabular}{l ccccc}
		\toprule
		                     & \multicolumn{2}{c}{Light direction} & \multicolumn{3}{c}{Color (HSV)}                                                                                \\
		\cmidrule(lr){2-3}\cmidrule(lr){4-6}
		Method               & L2$\downarrow$                      & Angle ($^\circ$)$\downarrow$    & Hue ($^\circ$)$\downarrow$ & Brightness$\downarrow$ & Saturation$\downarrow$ \\
		\midrule
		\multicolumn{6}{l}{\textit{Generative compositing (main paper)}}                                                                                                            \\
		AnyDoor              & 0.191                               & 34.29                           & 36.96                      & 0.177                  & 0.253                  \\
		Personalize Anything & 0.188                               & 32.82                           & 22.88                      & 0.172                  & 0.248                  \\
		Insert Anything      & 0.147                               & 26.36                           & 14.33                      & 0.107                  & 0.175                  \\
		UniCombine           & 0.150                               & 26.55                           & 13.06                      & 0.103                  & 0.178                  \\
		DreamFuse            & 0.149                               & 26.23                           & 10.17                      & 0.099                  & 0.148                  \\
		OmniPaint            & 0.171                               & 30.30                           & 14.48                      & 0.150                  & 0.194                  \\
		FreeCompose          & 0.202                               & 33.93                           & 22.58                      & 0.158                  & 0.214                  \\
		\midrule
		\multicolumn{6}{l}{\textit{Harmonization and intrinsic-based editing (\S\ref{app:baseline_analysis})}}                                                                      \\
		PIH                  & 0.146                               & 26.21                           & 18.34                      & 0.085                  & 0.145                  \\
		RGBX                 & 0.237                               & 40.05                           & 11.23                      & 0.268                  & 0.235                  \\
		ZeroComp             & 0.178                               & 31.52                           & 15.16                      & 0.090                  & 0.126                  \\
		DiffusionRenderer    & 0.149                               & 26.40                           & 20.24                      & 0.119                  & 0.114                  \\
		\midrule
		\modelname{}         & \textbf{0.129}                      & \textbf{23.00}                  & \textbf{7.96}              & \textbf{0.061}         & \textbf{0.070}         \\
		\bottomrule
	\end{tabular}
\end{table}

\clearpage
\subsection{\benchname{} Construction}
\label{app:dataset}

\textbf{Overview.}
\benchname{} consists of 760 (object, scene, ground-truth composite) triplets over 62 objects and 49 scenes.
For each scene, we first photograph the object under a lighting environment different from the target, then physically place it into the scene and capture the composite under the target lighting.
This deliberate illumination mismatch between object reference and target scene tests a method's ability to adapt object appearance to local lighting.

\vspace{4pt}
\noindent
\textbf{Capture protocol.}
We capture all scenes indoors with a tripod-mounted camera so that the empty background and the composite share viewpoints.
For each scene we first capture the empty background, then place the object at up to 8 distinct positions within the scene (median 3) and capture the composite at every position.
Object reference images are captured separately under a different lighting environment than the target scene.
Cameras include a Canon G7X, Galaxy smartphone, iPhone 15 Pro, and GoPro 10, with per-scene fixed white balance and exposure.

\vspace{4pt}
\noindent
\textbf{Scene diversity.}
The 49 scenes span living rooms, kitchens, bedrooms, offices, and corridors.
Lighting ranges from single overhead fixtures to multi-source setups with colored smart bulbs, table lamps, and window light.
We collect scenes with strong spatially varying illumination: visible cast shadows, color-temperature gradients, and localized highlights.

\vspace{4pt}
\noindent
\textbf{Object categories.}
The 62 objects span ten functional categories:
\emph{household goods} (step stool, soap, dispensers, containers, paper towels, cups; 11 objects),
\emph{electronics \& appliances} (speakers, thermometers, air purifier, razor, VR headset, scanner, laptop, toaster; 11),
\emph{bags \& wearables} (backpacks, caps, shoes, slippers, pouch, ear muffs, neck pillow; 11),
\emph{toys \& textiles} (stuffed animals, baby toy, blankets, cushion; 6),
\emph{furniture} (table, stools, chair, ottoman; 5),
\emph{d\'{e}cor \& lighting} (table lamp, diffuser, artificial plant, balloon, gift box; 5),
\emph{safety equipment} (fire extinguishers, traffic cone; 4),
\emph{kitchen \& tableware} (cup, bowl, tumblers; 4),
\emph{stationery} (stapler, notebook, notepad; 3),
and \emph{food items} (bananas, orange; 2).
Materials span plastic, ceramic, metal, wood, fabric, glass, paper, and fruit peel; shapes range from compact convex forms (\eg, orange, mug) to thin structures (\eg, lamp, traffic cone) and deformable items (\eg, blanket, stuffed animal).
This diversity exercises illumination adaptation across a wide range of shapes and reflectance properties.

\vspace{4pt}
\noindent
\textbf{Annotation.}
Each triplet ships with an object segmentation mask from SAM2~\citep{ravi2025sam2segmentimages} (manually corrected) and a tight insertion bounding box.

\vspace{4pt}
Fig.~\ref{fig:bench_samples} shows representative triplets across the lighting setups and object categories described above.

\newcommand{\benchcell}[1]{\includegraphics[width=.19\linewidth]{figures/data/bench_samples/#1}}
\begin{figure}[p]
	\centering
	\scriptsize
	\setlength{\tabcolsep}{2pt}
	\begin{tabular}{@{} c@{\hspace{6pt}} c@{\hspace{6pt}} *{3}{c@{\hspace{2pt}}} @{}}
		Object               & Background           & \multicolumn{3}{c}{G.T. Composites (varied insertion locations)}                                                             \\
		\benchcell{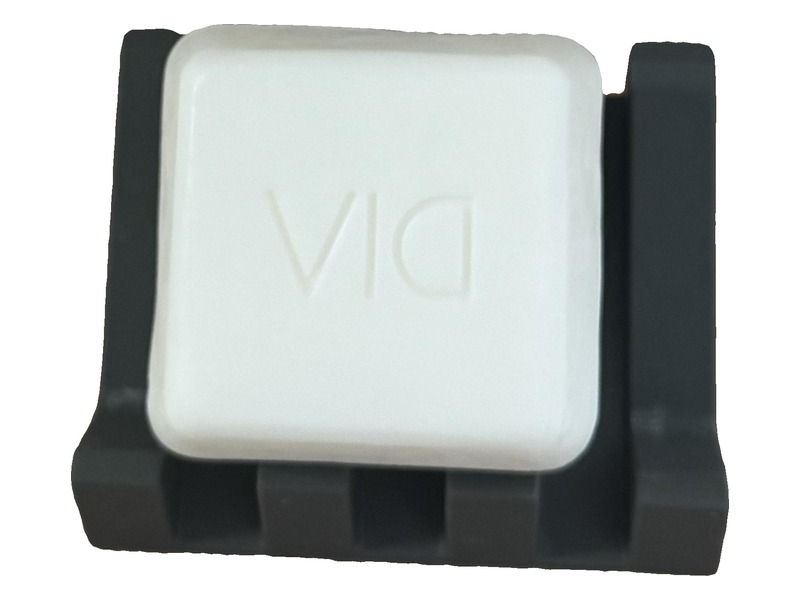} & \benchcell{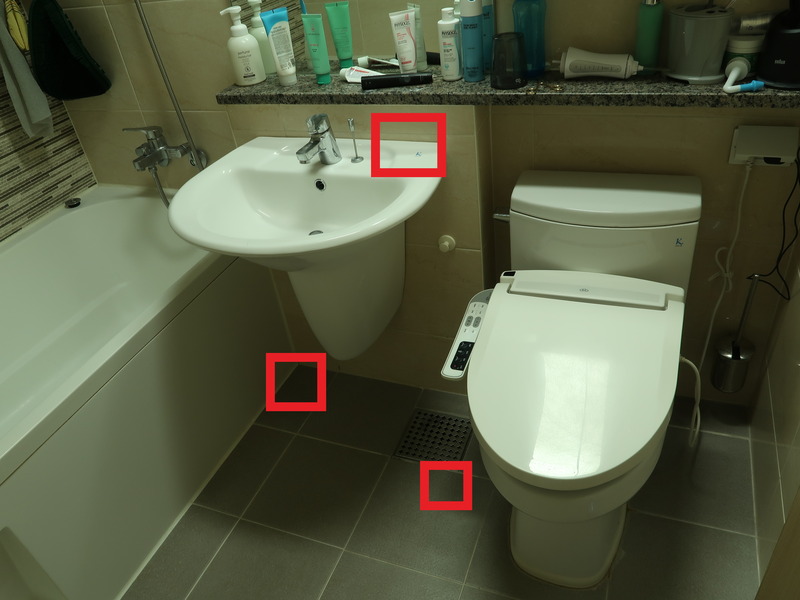} & \benchcell{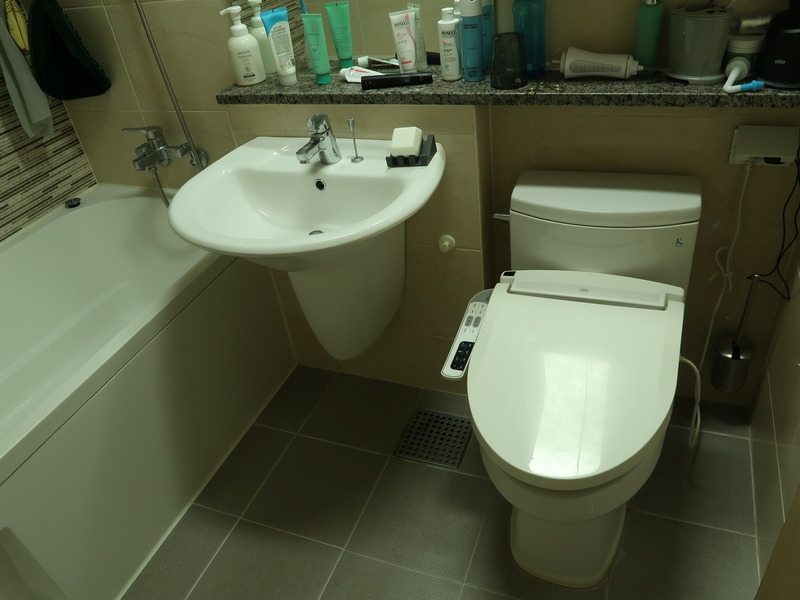}                                      & \benchcell{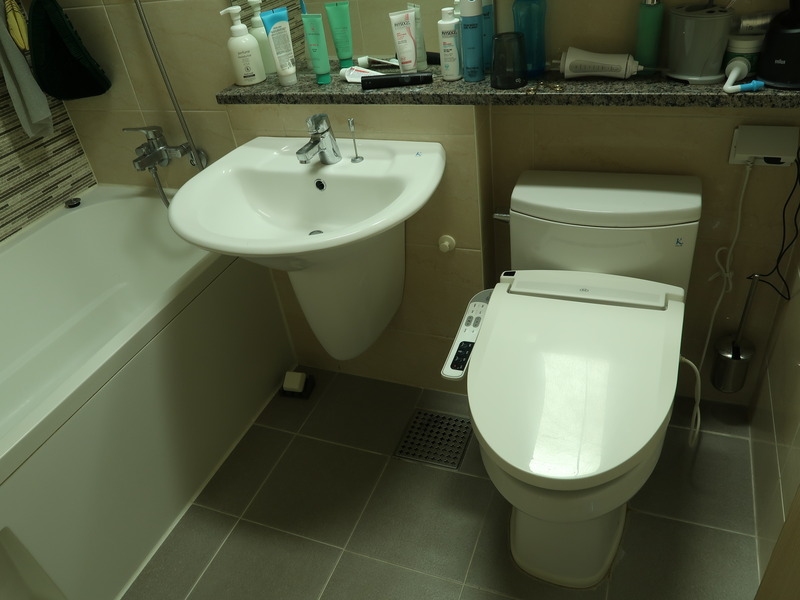} & \benchcell{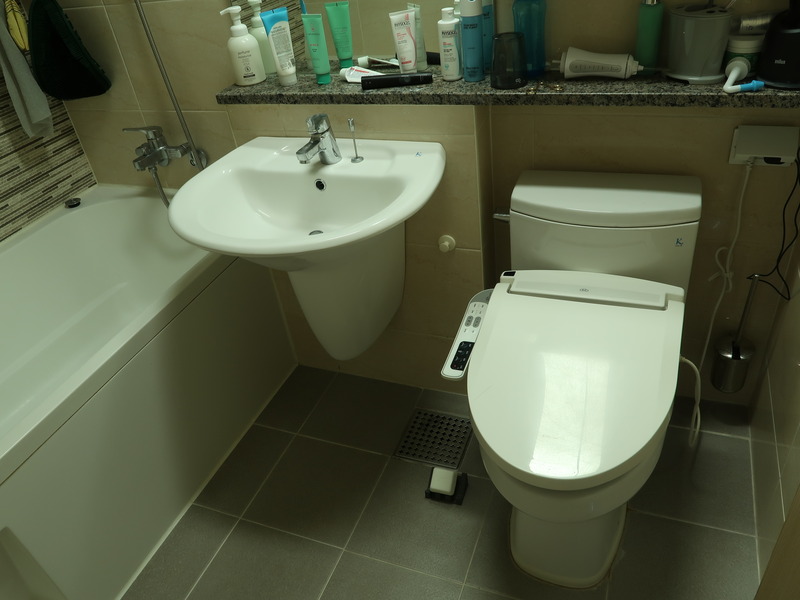} \\
		\benchcell{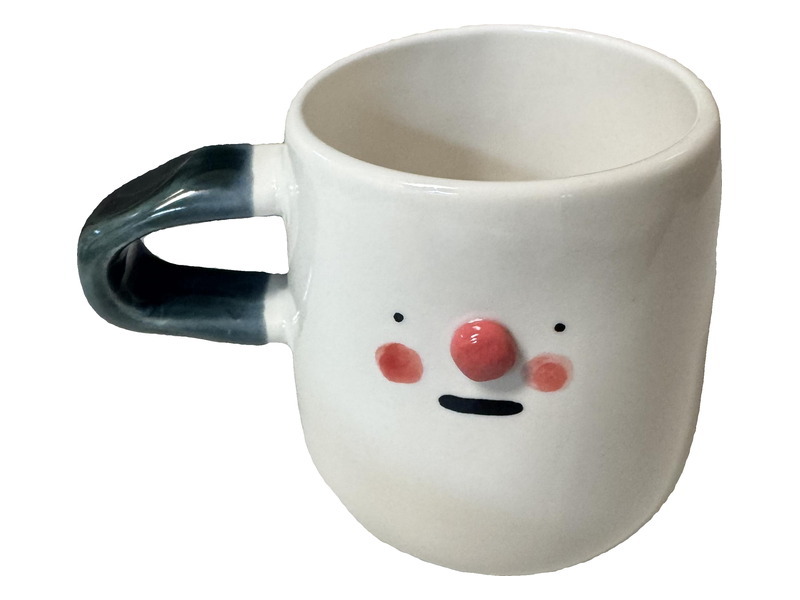}  & \benchcell{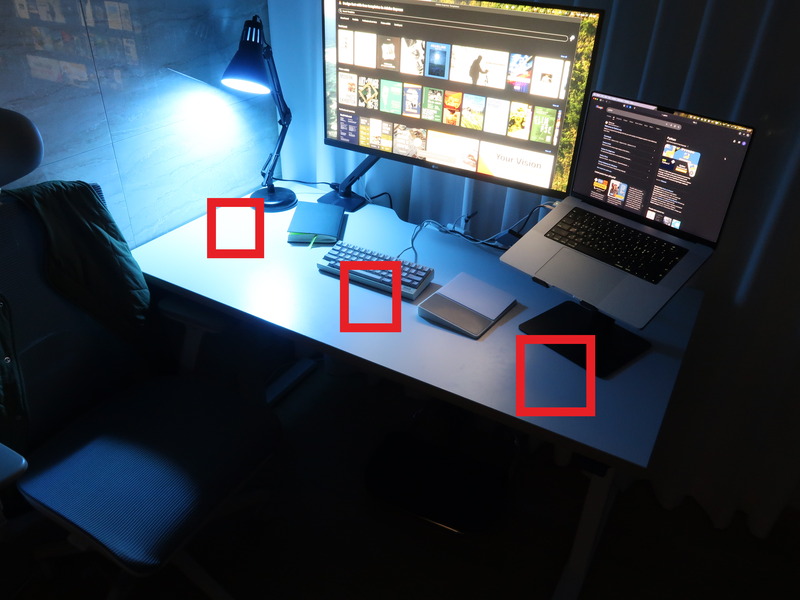}  & \benchcell{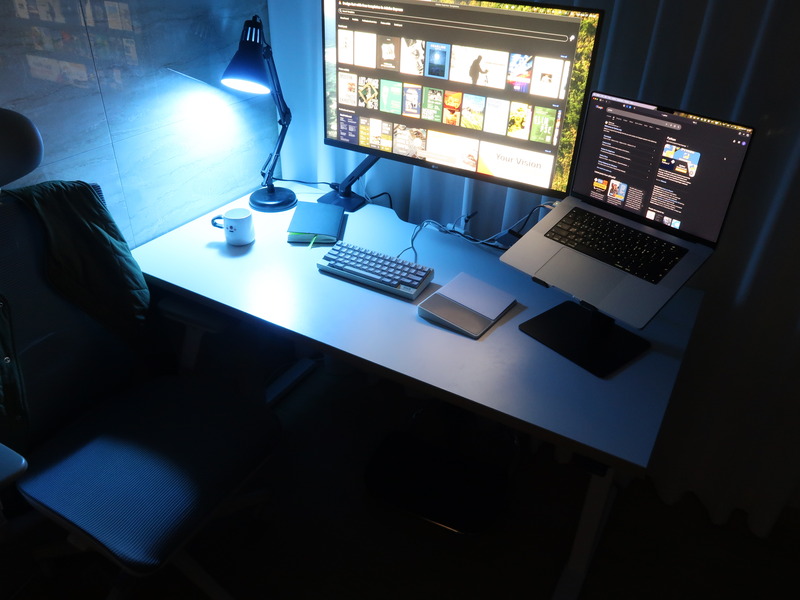}                                        & \benchcell{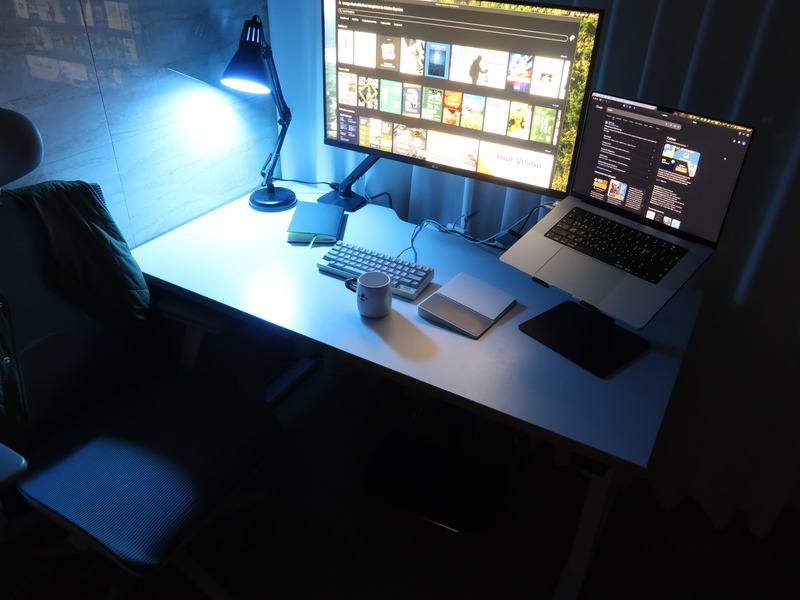}   & \benchcell{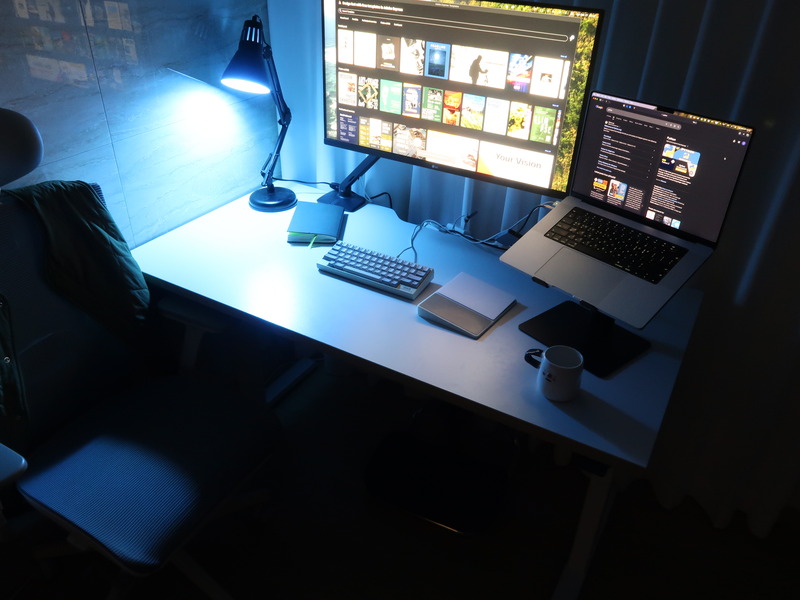}   \\
		\benchcell{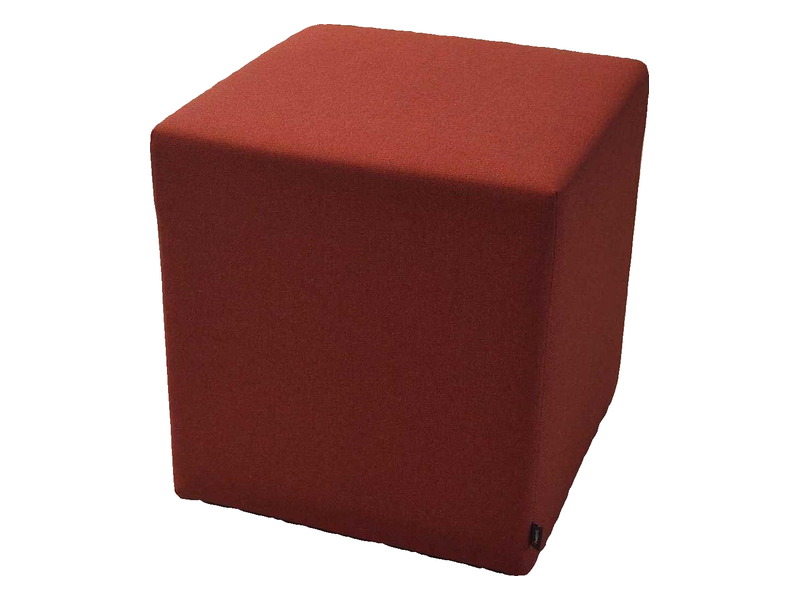} & \benchcell{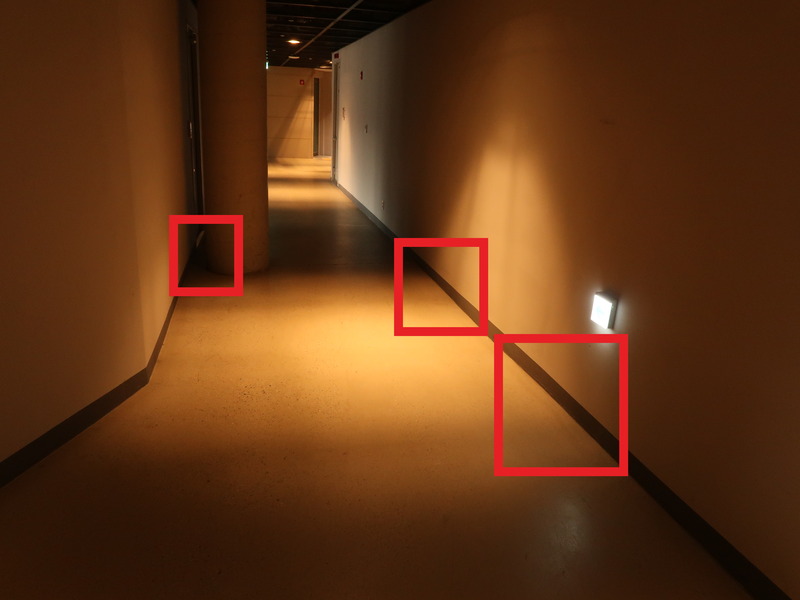} & \benchcell{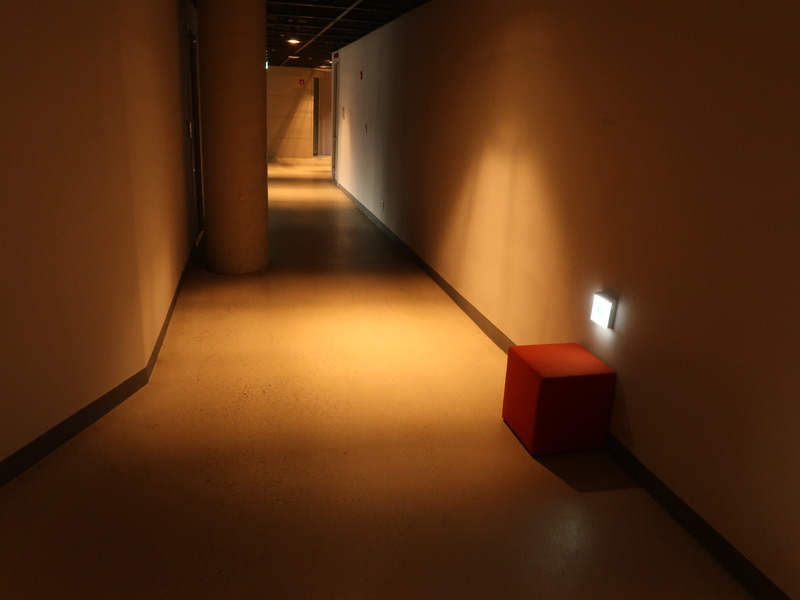}                                      & \benchcell{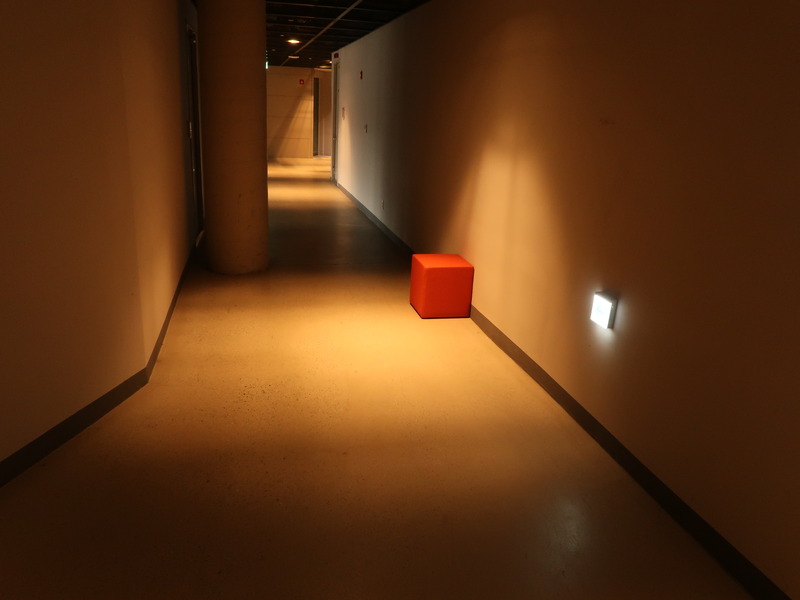} & \benchcell{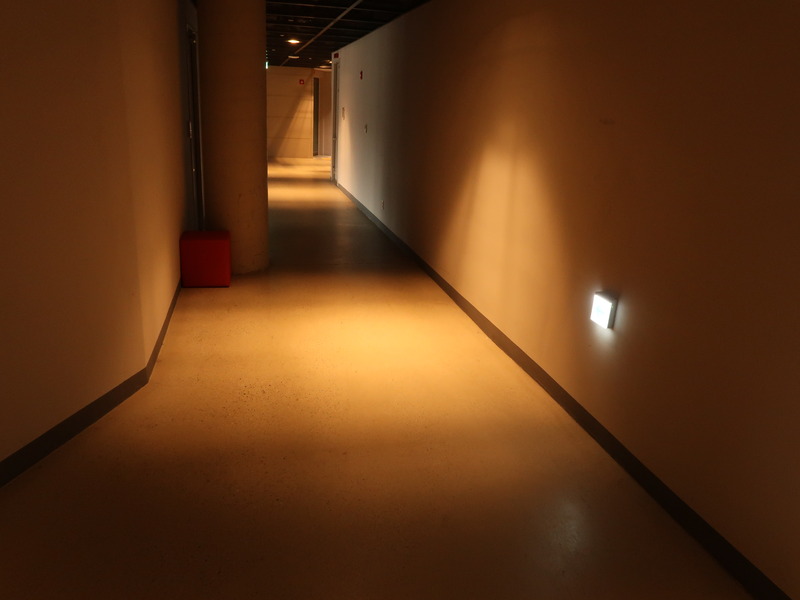} \\
		\benchcell{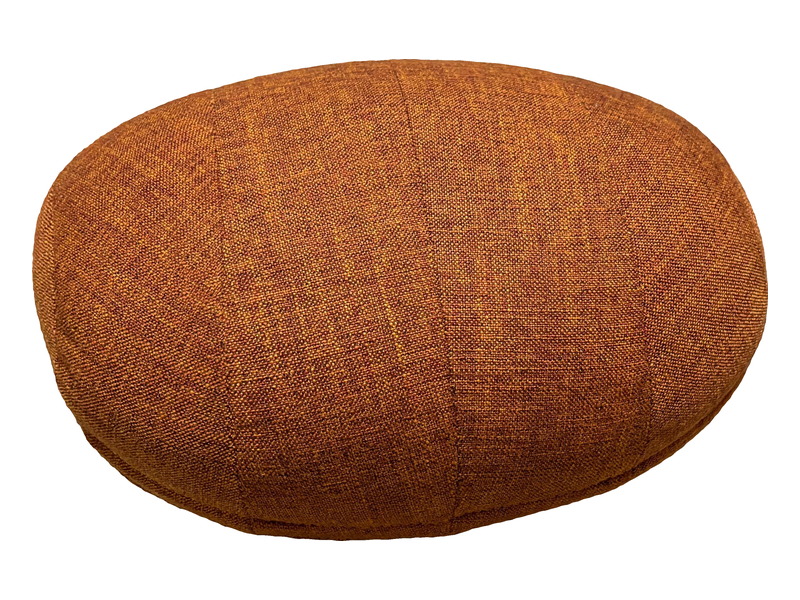} & \benchcell{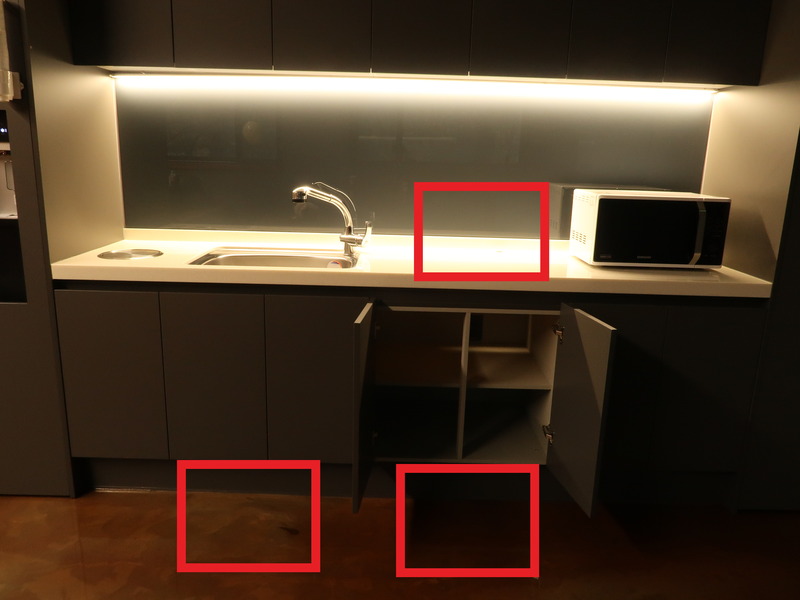} & \benchcell{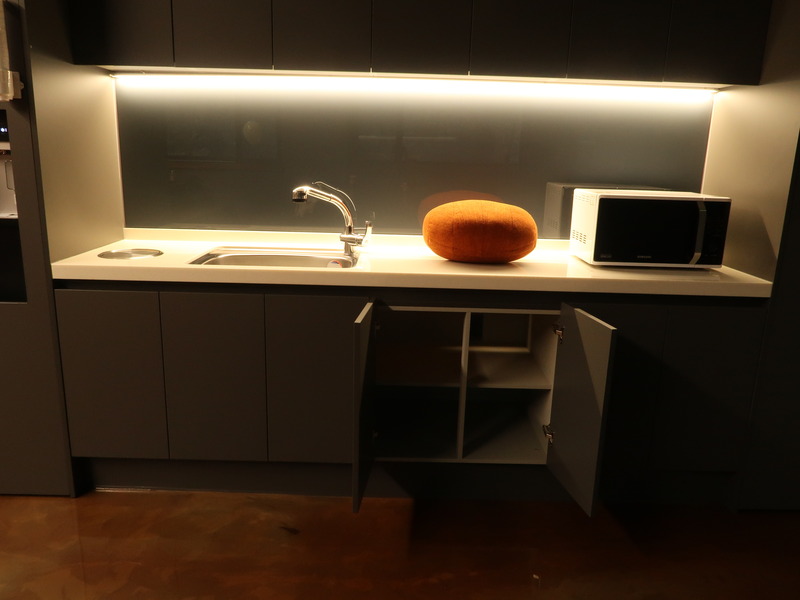}                                      & \benchcell{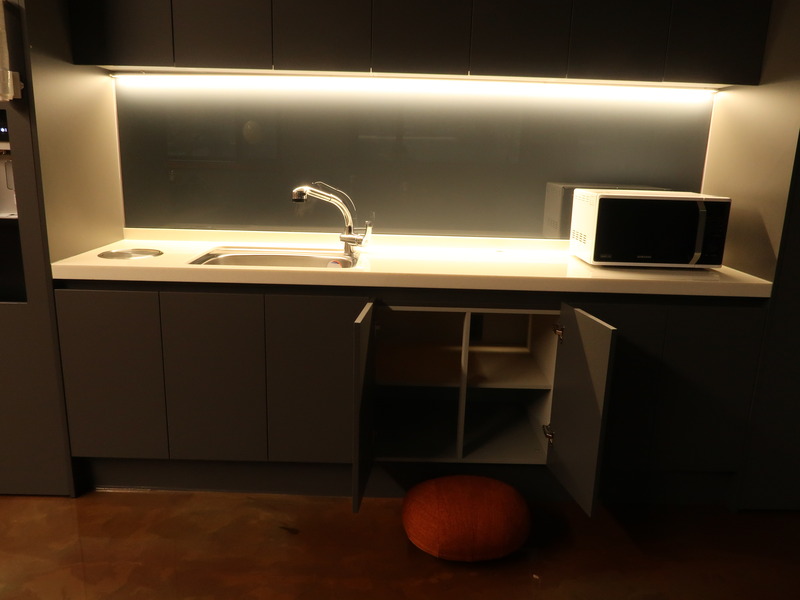} & \benchcell{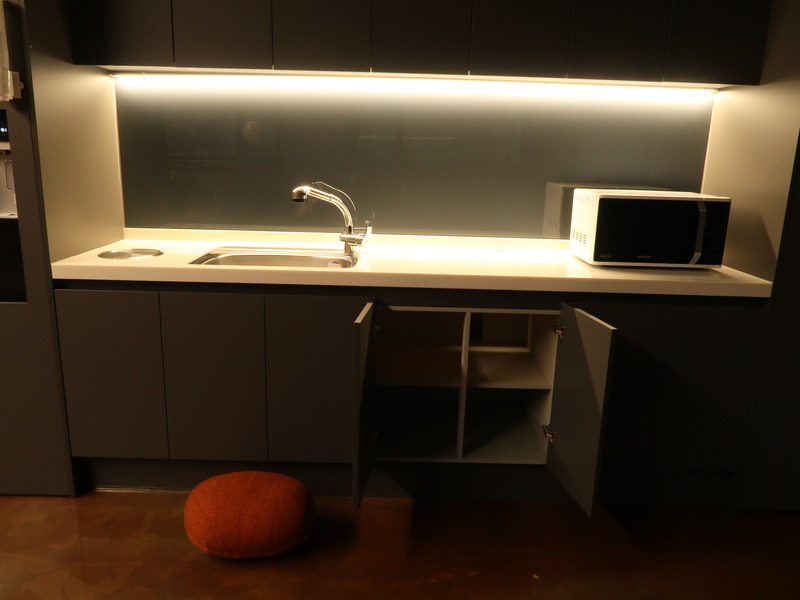} \\
		\benchcell{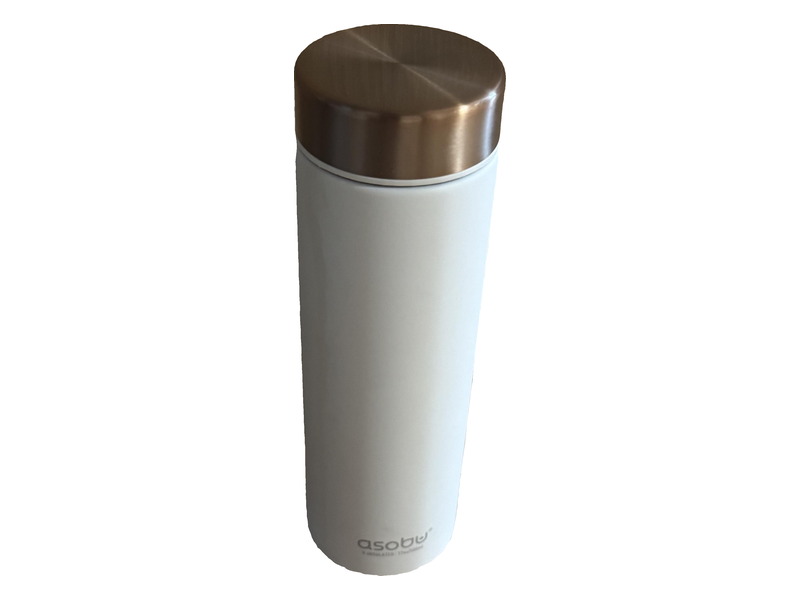} & \benchcell{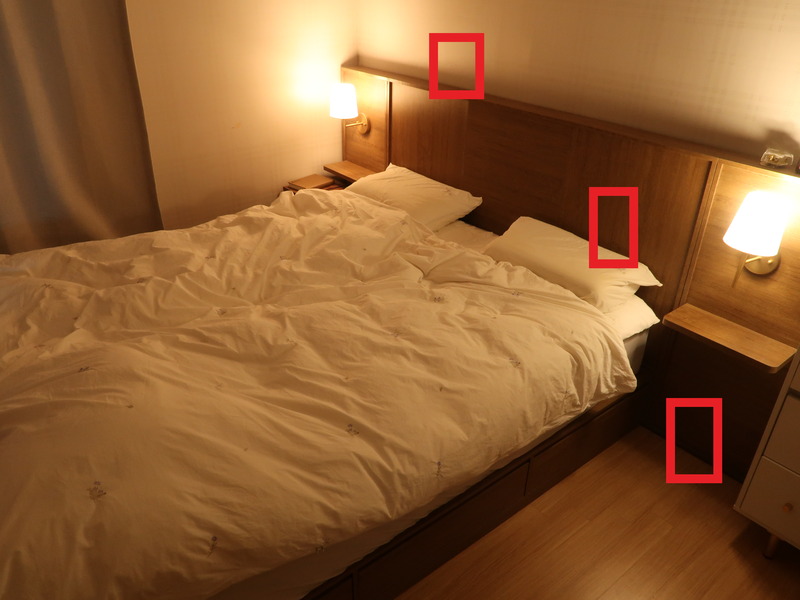} & \benchcell{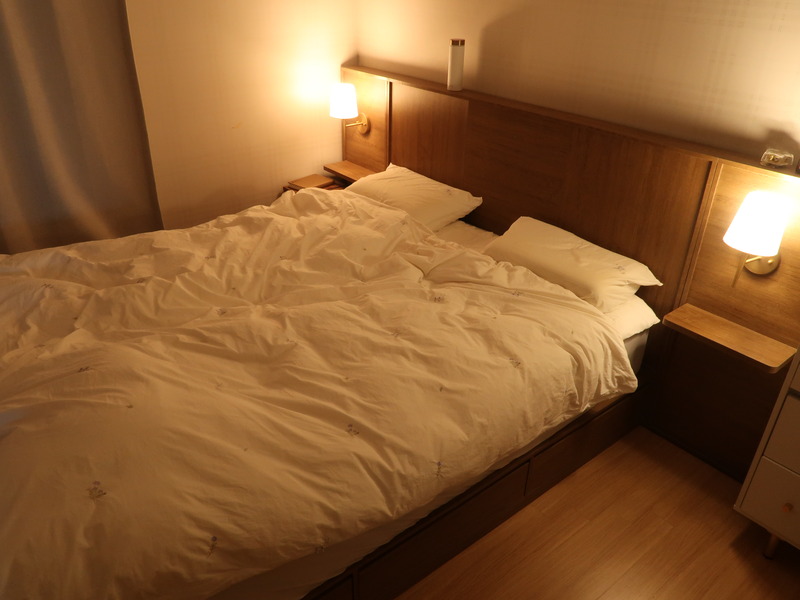}                                      & \benchcell{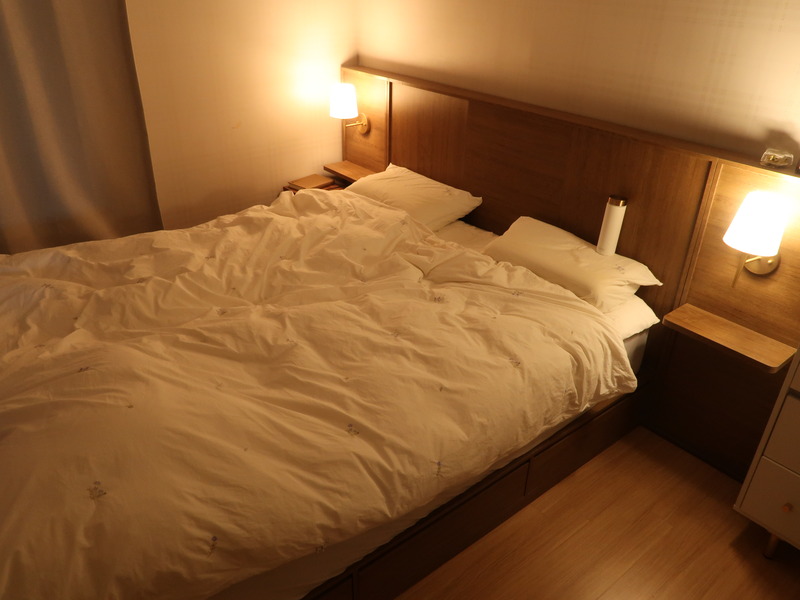} & \benchcell{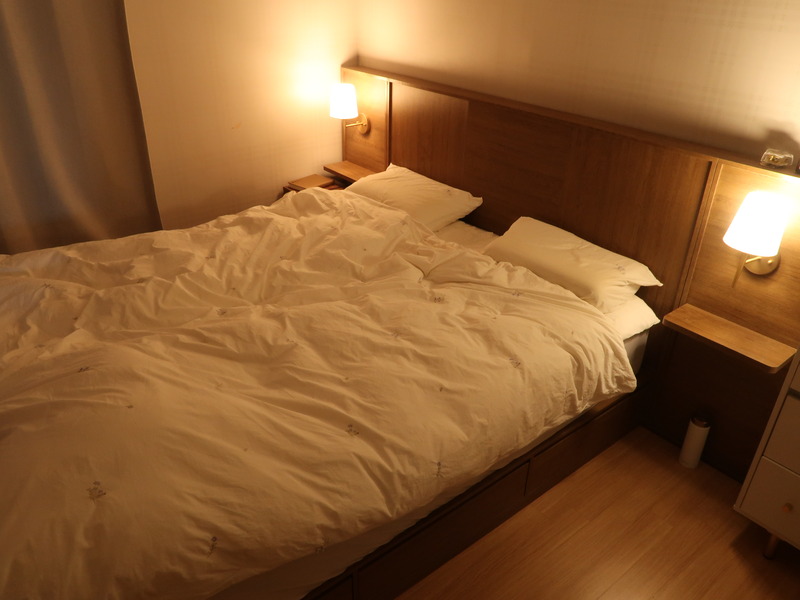} \\
		\benchcell{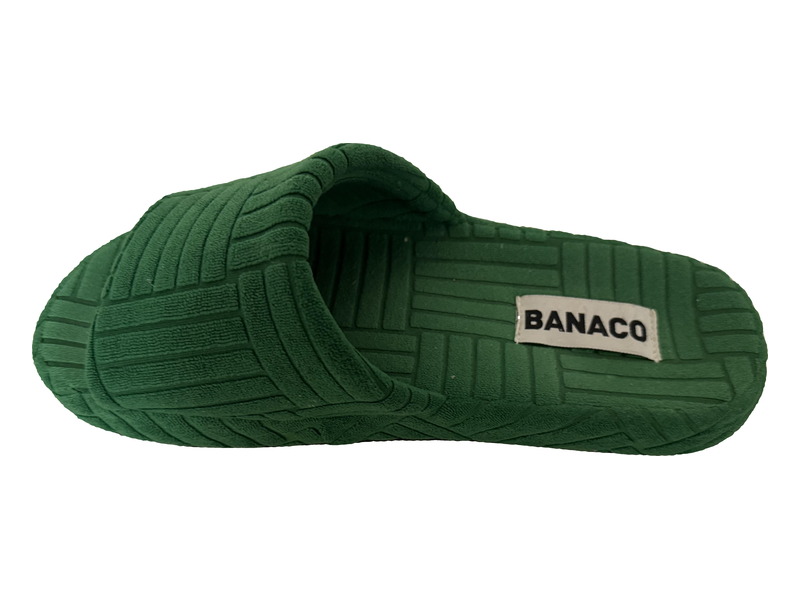} & \benchcell{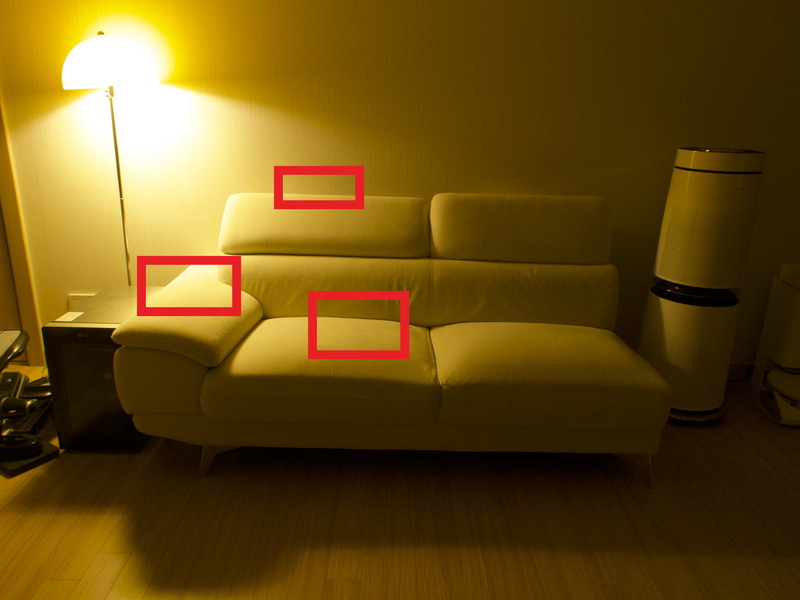} & \benchcell{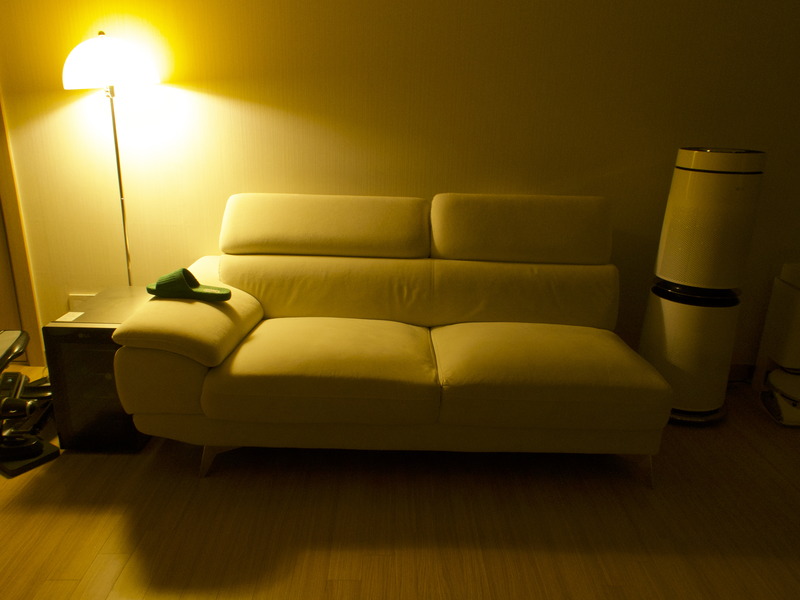}                                      & \benchcell{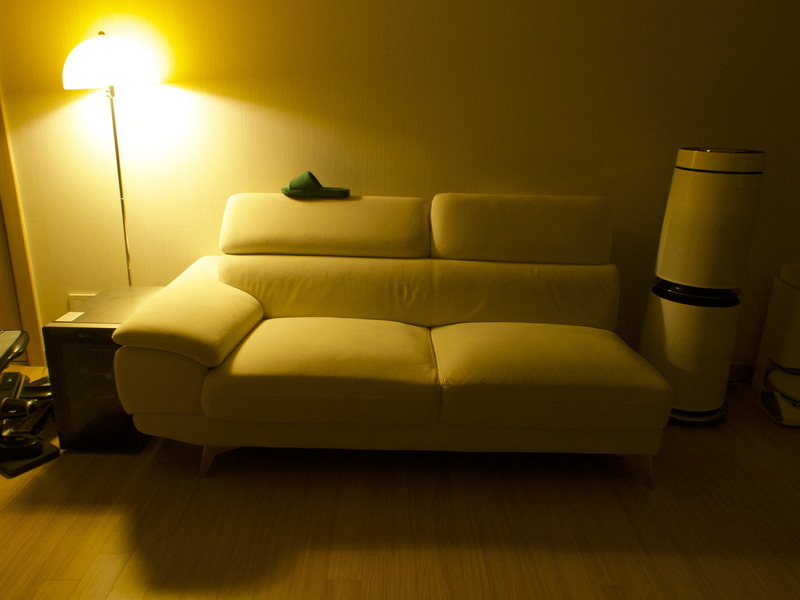} & \benchcell{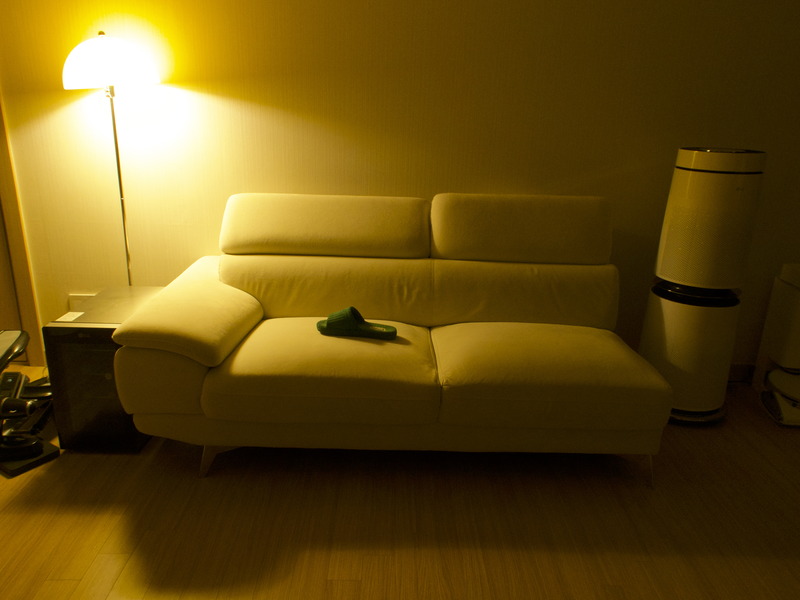} \\
		\benchcell{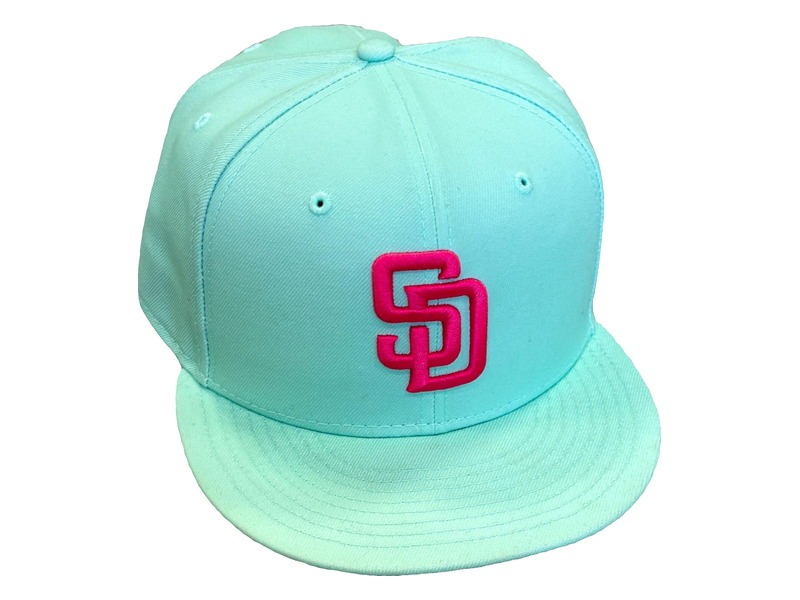} & \benchcell{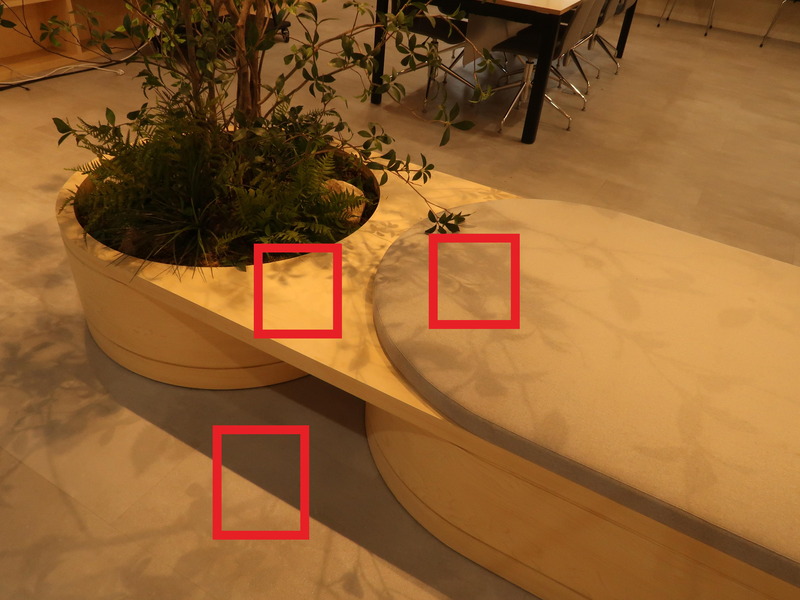} & \benchcell{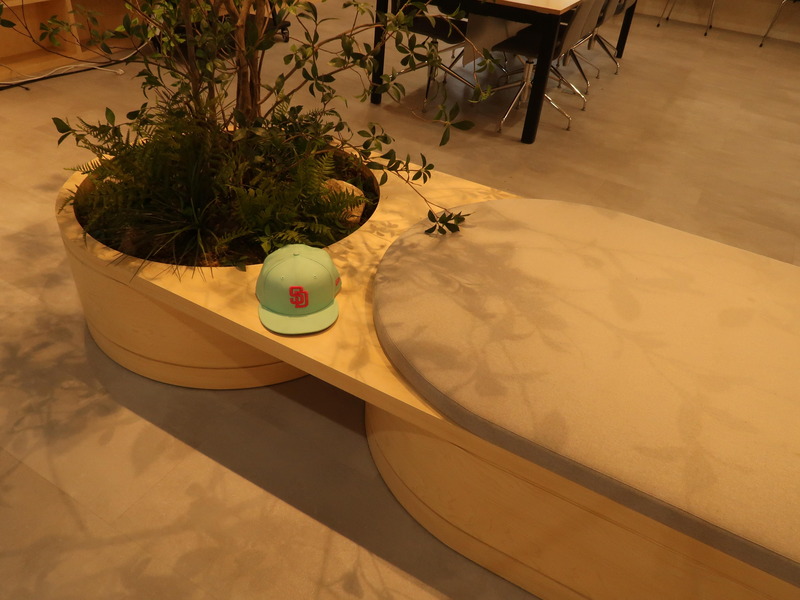}                                      & \benchcell{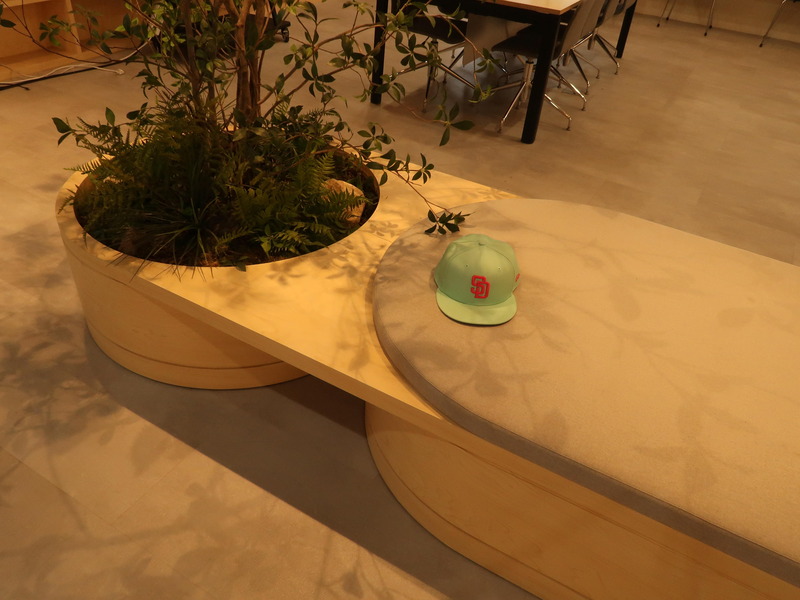} & \benchcell{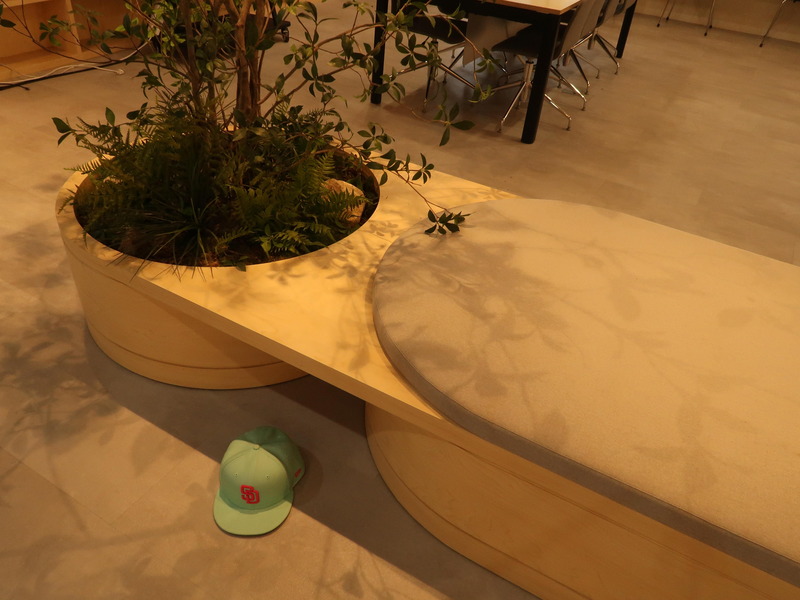} \\
	\end{tabular}
	\caption{
		\benchname{} samples.
		Each row shows one object and the matching empty background overlaid with three red bboxes marking the insertion locations, followed by the ground-truth composite at each bbox.
		Rows cover a tiled bathroom under overhead lighting, a cool-lit office desk, a warm-lit corridor, a kitchen counter under cove lighting, a bedroom with bedside lamps, a sofa under a floor lamp, and a planter bench in an indoor lobby.
	}
	\label{fig:bench_samples}
\end{figure}

\FloatBarrier
\subsection{Scene Preparation Details}
\label{app:scene_prep}

This subsection details how the scene parameters consumed by the on-the-fly relighting pipeline (Sec.~\ref{sec:augmentation}) are estimated and cached.
Every step runs once per scene before training; at training time the augmentation only reads from this cache.
Fig.~\ref{fig:scene_prep} summarizes the pipeline.

\begin{figure}[htbp]
	\centering
	\includegraphics[width=0.9\linewidth]{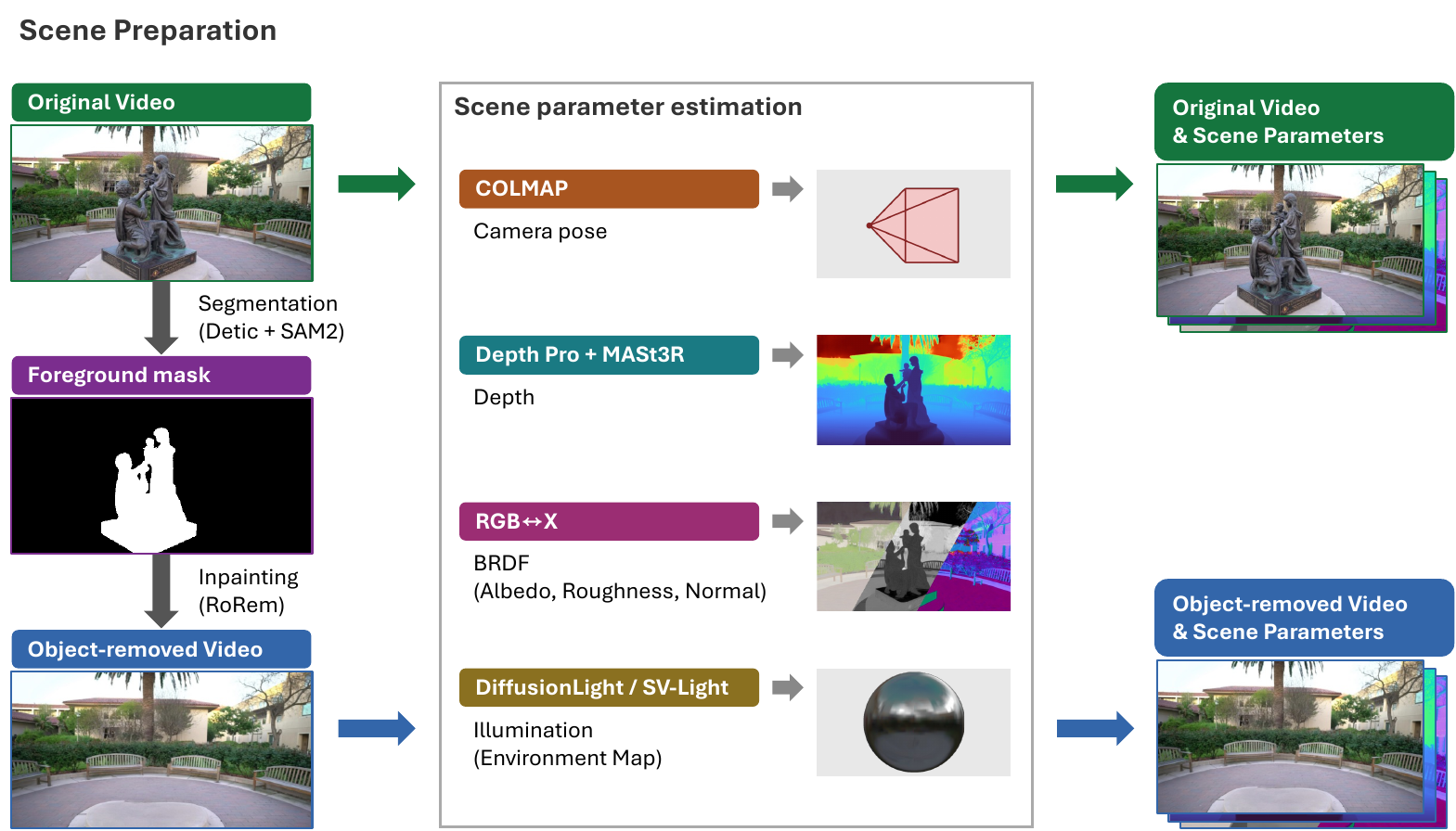}
	\caption{
		\textbf{Scene preparation pipeline.}
		Foreground objects are masked with Detic\,+\,SAM2 and inpainted out with RoRem to yield clean backgrounds.
		Camera poses are shared between the original and object-removed videos; depth and BRDF maps for the latter are completed in the removed regions, conditioned on the original estimates.
		All quantities are cached offline and reused throughout the on-the-fly relighting augmentation.
	}
	\label{fig:scene_prep}
\end{figure}

\noindent
\textbf{Foreground segmentation.}
We detect per-frame foreground objects with Detic~\citep{zhou2022detecting} using its open-vocabulary class set, then track each detected object across the whole video with SAM2~\citep{ravi2025sam2segmentimages} to obtain temporally consistent instance masks.
Spurious detections and unreliable masks are removed by the filters detailed in Appendix~\ref{app:data_filtering}.

\noindent
\textbf{Background inpainting.}
Each tracked mask is dilated by a small margin so that the inpainter does not reconstruct the object from leaked boundary cues, then the masked region is inpainted by RoRem~\citep{li2024RORem}, yielding an \emph{object-removed} variant of the video.
As a check for object re-hallucination, we measure the DINO similarity between the original object and its inpainted region. The similarity is 0.252, closer to unrelated background patches (0.203) than to the same object in another frame (0.691).

\noindent
\textbf{Geometry and reflectance estimation.}
Camera intrinsics and extrinsics are reconstructed once from the original video with COLMAP~\citep{colmap} and shared with the object-removed variant.
For depth, we first predict per-frame monocular depth with Depth Pro~\citep{bochkovskii2025depthprosharpmonocular}. For each frame, we obtain correspondences to other frames with MASt3R~\citep{mast3r_eccv24}, estimate a candidate scale for each pair using RANSAC with fixed COLMAP poses, and take the median of these estimates as the frame's scale factor.
For reflectance parameters, RGB$\leftrightarrow$X~\citep{zeng2024rgb} estimates albedo, roughness, and normal, which feed the simplified Disney BRDF detailed in Appendix~\ref{app:implementation}.
For the object-removed frames, we complete the removed regions in the depth and BRDF maps with DepthLab~\citep{liu2024depthlab} and RGB$\leftrightarrow$X~\citep{zeng2024rgb}, respectively, conditioned on the original-frame estimates. The completed maps are blended back using a dilated, blurred object mask, leaving the original estimates unchanged elsewhere.

\Needspace{12\baselineskip}
\noindent
\textbf{Original scene illumination.}
Indoor and outdoor scenes require different illumination models, so we classify each scene with a vision-language model~\citep{bai2025qwen25vl}.
We sample up to 10 frames from the video, query the model on each for an indoor/outdoor label with a confidence score, and take the majority vote.
We estimate a single HDR environment map with DiffusionLight~\citep{Phongthawee2024DiffusionLight} for outdoor scenes, and per-pixel environment maps~\citep{li2020inverse} for indoor scenes.
In both cases, we precompute the per-pixel shading induced by the estimated illumination ($L_{\text{orig}}$),
\begin{equation}
	S_{\text{orig}}(x) = \int_{\Omega} L_{\text{orig}}(x, \omega_i)\, R(x, \omega_o, \omega_i)\, d\omega_i,
\end{equation}
evaluated by discrete quadrature over the environment-map directions, using a single envmap outdoors and per-pixel envmaps indoors.
Rather than caching the illumination maps directly, we cache $S_{\text{orig}}$ and use it in the ratio-based illumination transfer of Eq.~3 (Sec.~\ref{sec:augmentation}).

\FloatBarrier
\subsection{Training Data Statistics}
\label{app:data_stats}
We draw monocular videos from five sources: Mip-NeRF 360~\citep{360mipnerf2022}, Tanks\&Temples~\citep{Knapitsch2017}, DL3DV~\citep{ling2024dl3dv}, uCO3D~\citep{liu24uco3d}, and our own casual captures.
Our casual captures are handheld phone-camera videos recorded in everyday scenes, typically 30 seconds to 2 minutes long, without staging scenes specifically for spatially varying illumination.
Fig.~\ref{fig:aug} of the main paper shows a raw triplet from these videos and its illumination-augmented counterpart.
Each video is uniformly sampled to up to 100 frames and passed through the filtering pipeline of \S\ref{app:data_filtering}.
Tab.~\ref{tab:data_stats} reports per-source counts after filtering.
Representative augmented triplets and their cross-permutations appear in Figs.~\ref{fig:training_samples} and~\ref{fig:permutation_16}.

\begin{table}[htbp]
	\centering
	\caption{Training data statistics per source. \emph{Frames} counts unaugmented training triplets after filtering.}
	\label{tab:data_stats}
	\small
	\begin{tabular}{lrrrr}
		\toprule
		Source                               & Scenes           & Objects           & Frames             & Avg.\ Views/Obj \\
		\midrule
		MipNeRF360~\citep{360mipnerf2022}    & 9                & 39                & 1{,}980            & 50.8            \\
		Tanks\&Temples~\citep{Knapitsch2017} & 19               & 62                & 1{,}809            & 29.2            \\
		DL3DV~\citep{ling2024dl3dv}          & 5{,}731          & 21{,}373          & 536{,}132          & 25.1            \\
		uCO3D~\citep{liu24uco3d}             & 2{,}239          & 2{,}460           & 118{,}545          & 48.2            \\
		Casual Captures                      & 51               & 247               & 12{,}433           & 50.3            \\
		\midrule
		\textbf{Total}                       & \textbf{8{,}049} & \textbf{24{,}181} & \textbf{670{,}899} & \textbf{27.7}   \\
		\bottomrule
	\end{tabular}
\end{table}

\begin{figure}[p]
	\centering
	\scriptsize
	\begin{tabular}{@{} c@{\hspace{2pt}}c@{\hspace{2pt}}c @{\hspace{10pt}} c@{\hspace{2pt}}c@{\hspace{2pt}}c @{}}
		\makecell{B.G. \& BBox}                                                         & Object                                                                           & G.T.                                                                              & \makecell{B.G. \& BBox}                                                         & Object                                                                           & G.T.                                                                              \\
		\includegraphics[width=.15\linewidth]{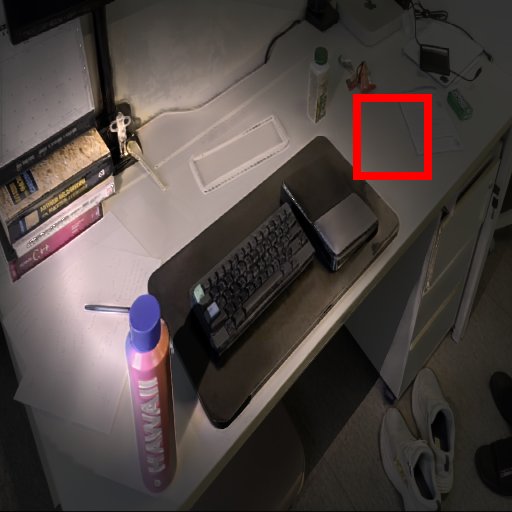}  & \includegraphics[width=.15\linewidth]{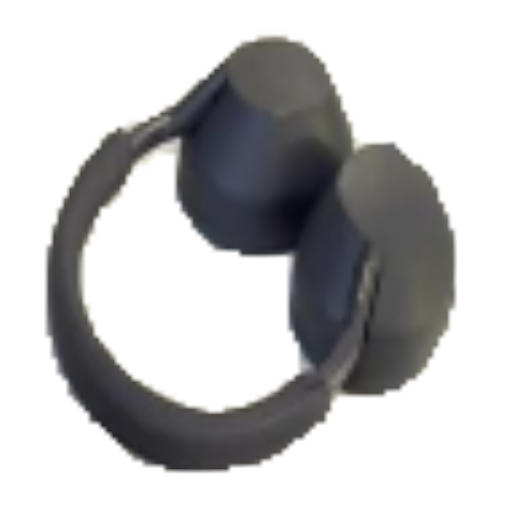}  & \includegraphics[width=.15\linewidth]{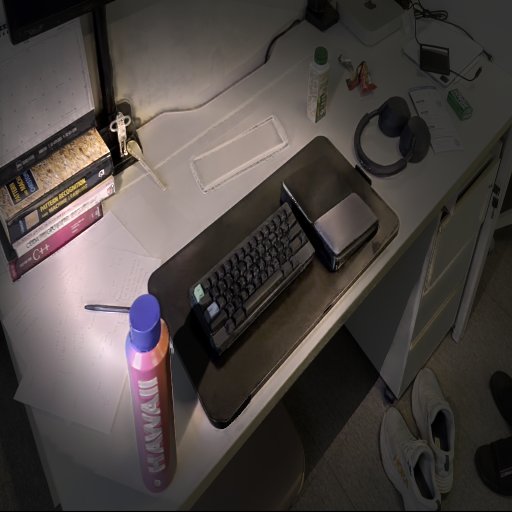}  & \includegraphics[width=.15\linewidth]{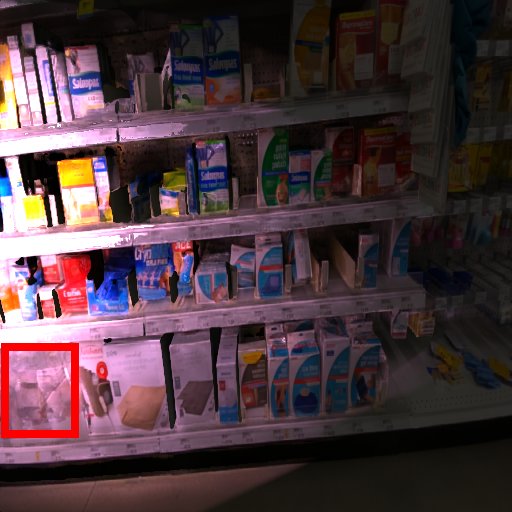}  & \includegraphics[width=.15\linewidth]{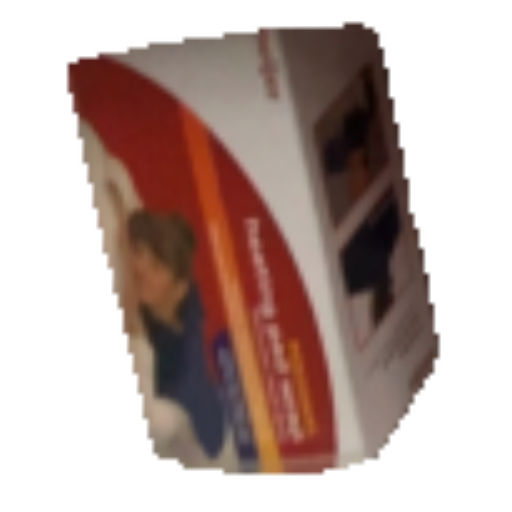}  & \includegraphics[width=.15\linewidth]{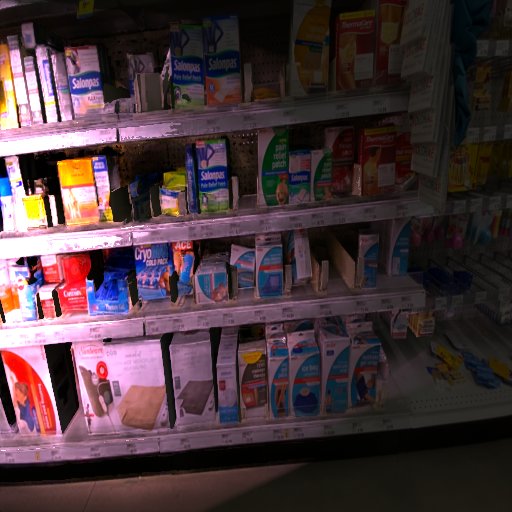}  \\
		\includegraphics[width=.15\linewidth]{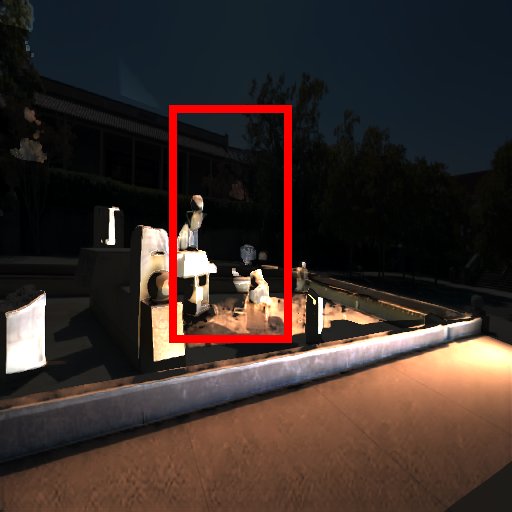}  & \includegraphics[width=.15\linewidth]{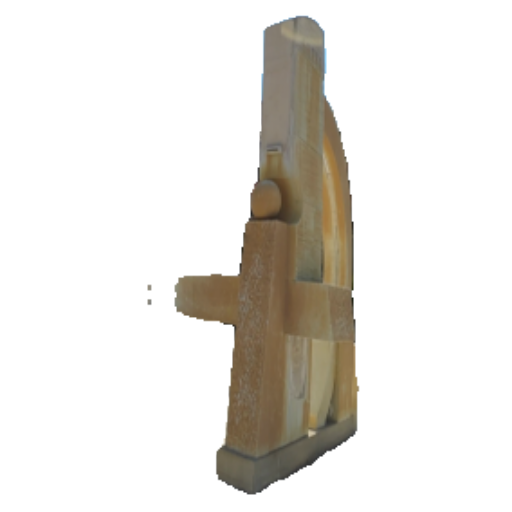}  & \includegraphics[width=.15\linewidth]{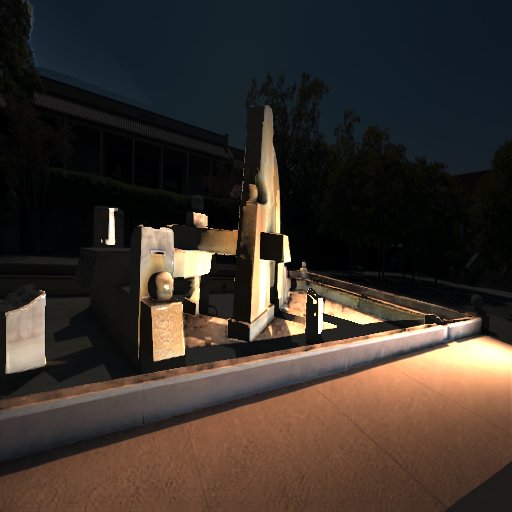}  & \includegraphics[width=.15\linewidth]{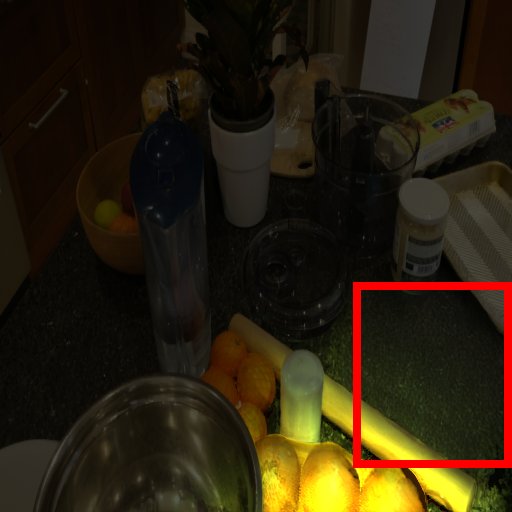}  & \includegraphics[width=.15\linewidth]{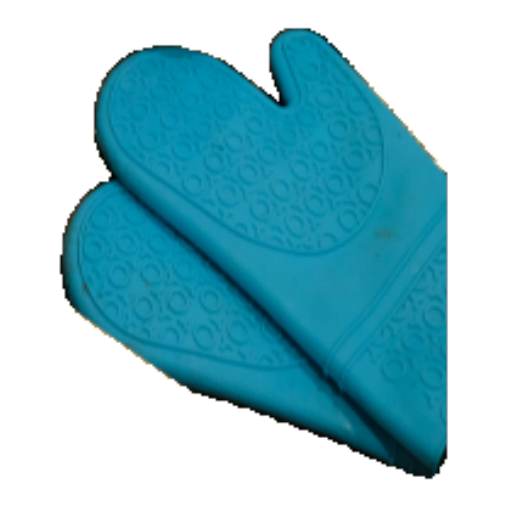}  & \includegraphics[width=.15\linewidth]{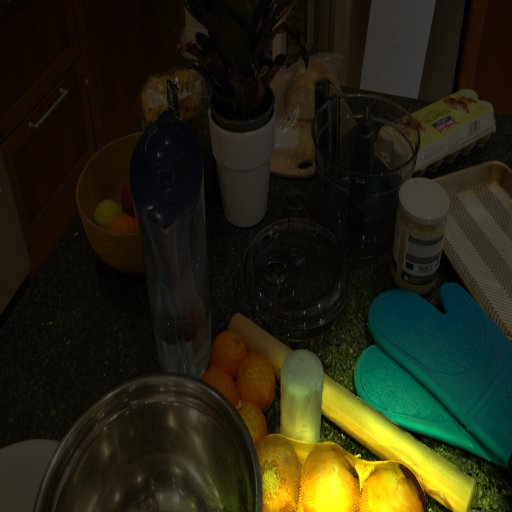}  \\
		\includegraphics[width=.15\linewidth]{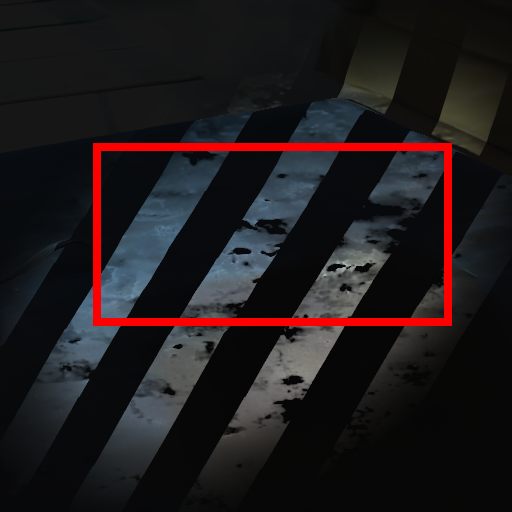}  & \includegraphics[width=.15\linewidth]{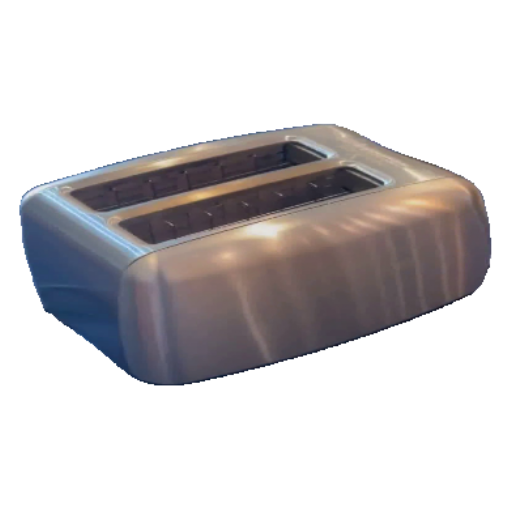}  & \includegraphics[width=.15\linewidth]{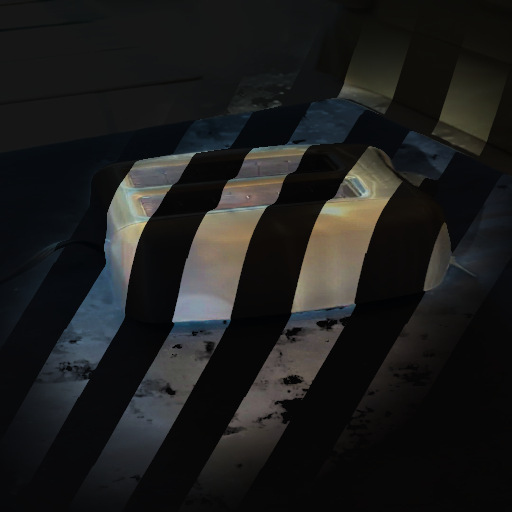}  & \includegraphics[width=.15\linewidth]{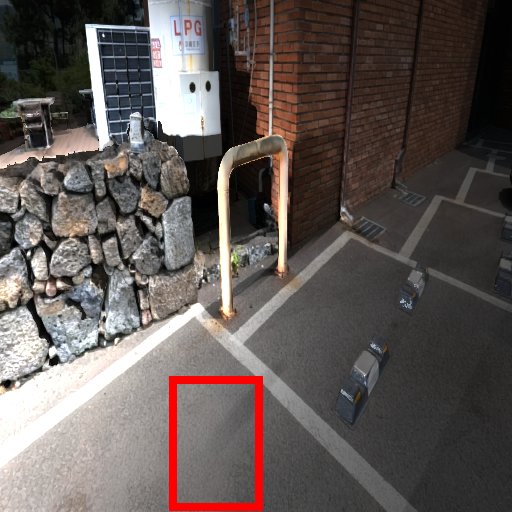}  & \includegraphics[width=.15\linewidth]{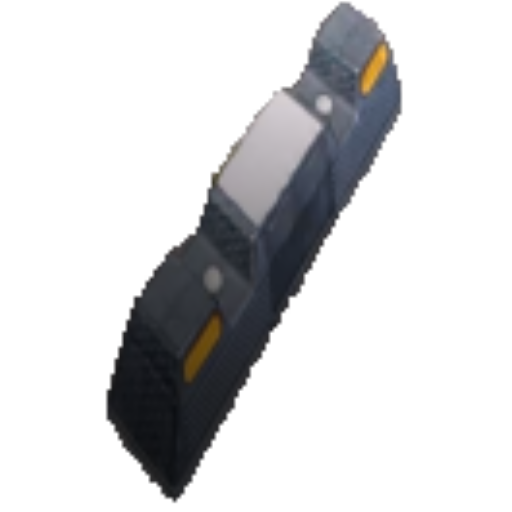}  & \includegraphics[width=.15\linewidth]{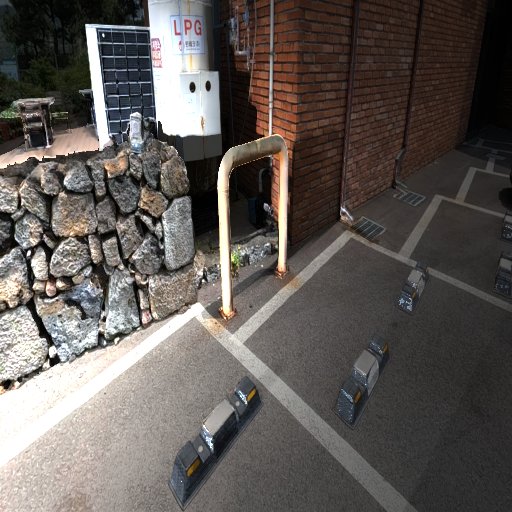}  \\
		\includegraphics[width=.15\linewidth]{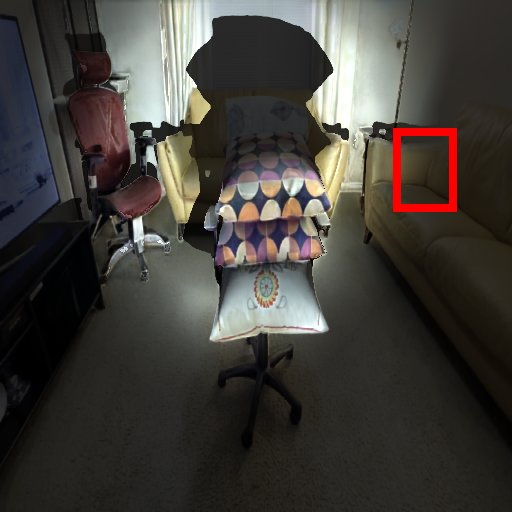}  & \includegraphics[width=.15\linewidth]{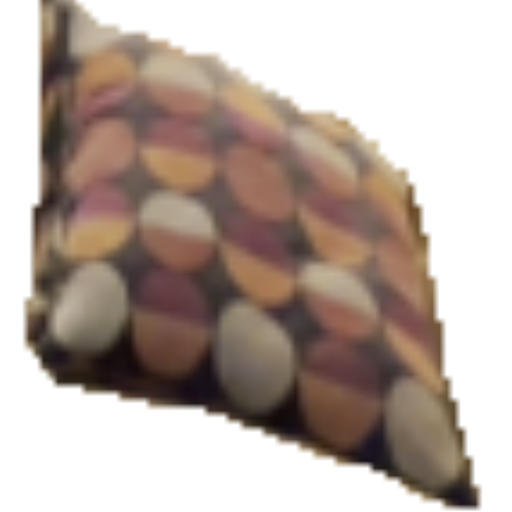}  & \includegraphics[width=.15\linewidth]{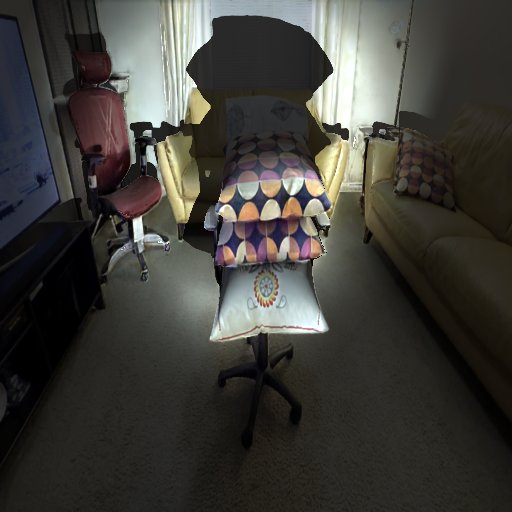}  & \includegraphics[width=.15\linewidth]{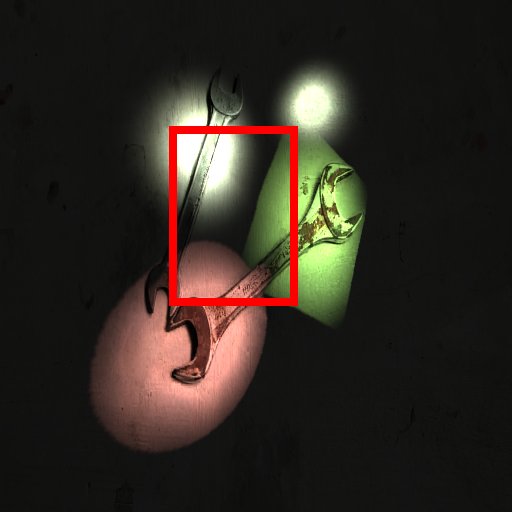}  & \includegraphics[width=.15\linewidth]{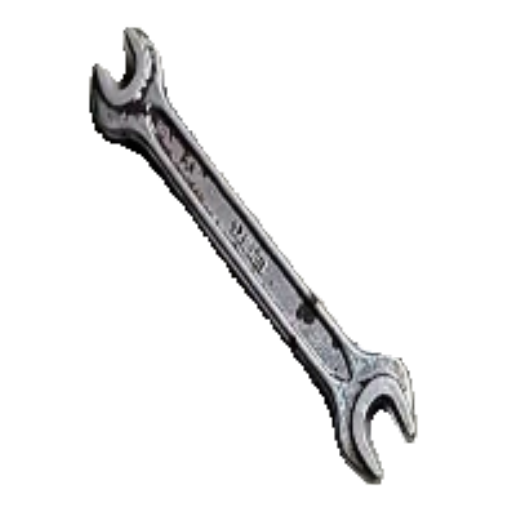}  & \includegraphics[width=.15\linewidth]{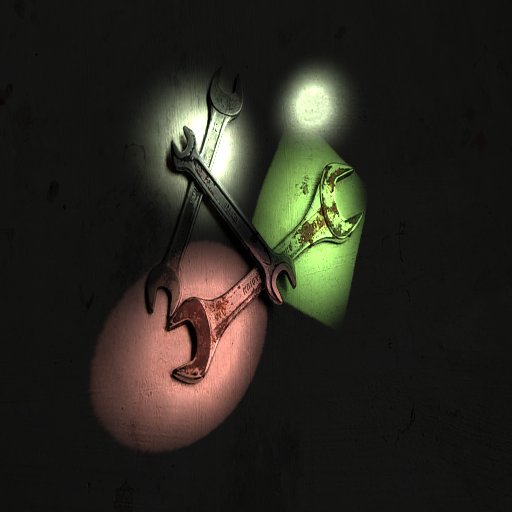}  \\
		\includegraphics[width=.15\linewidth]{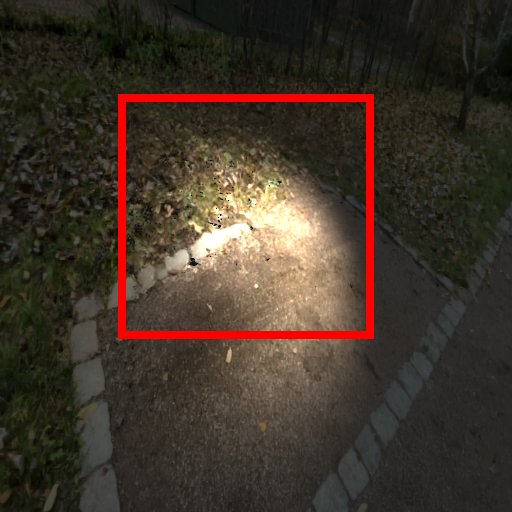}  & \includegraphics[width=.15\linewidth]{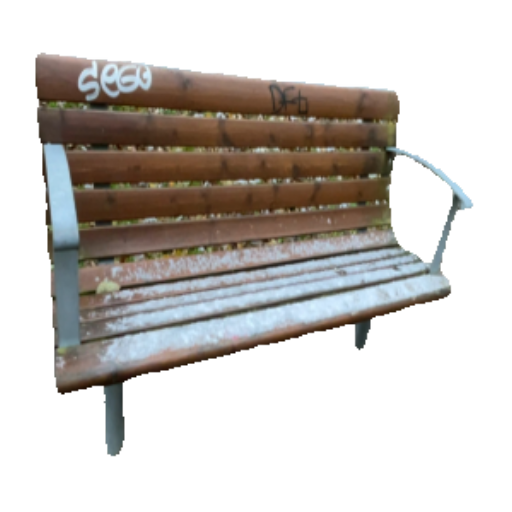}  & \includegraphics[width=.15\linewidth]{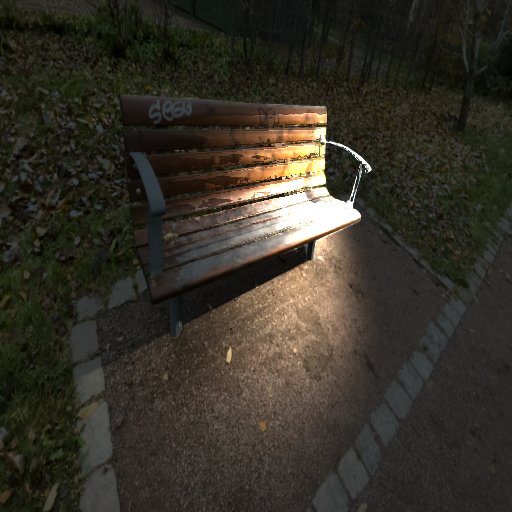}  & \includegraphics[width=.15\linewidth]{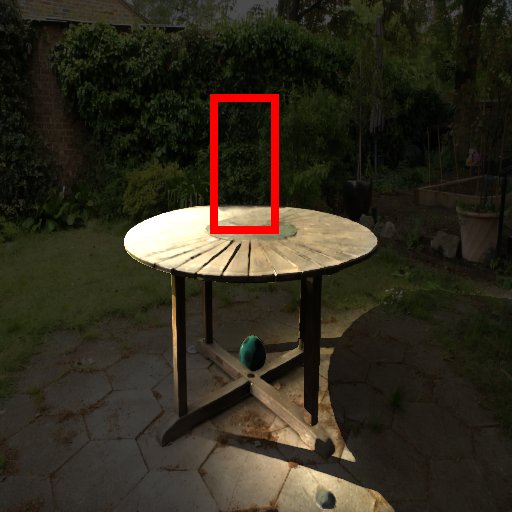} & \includegraphics[width=.15\linewidth]{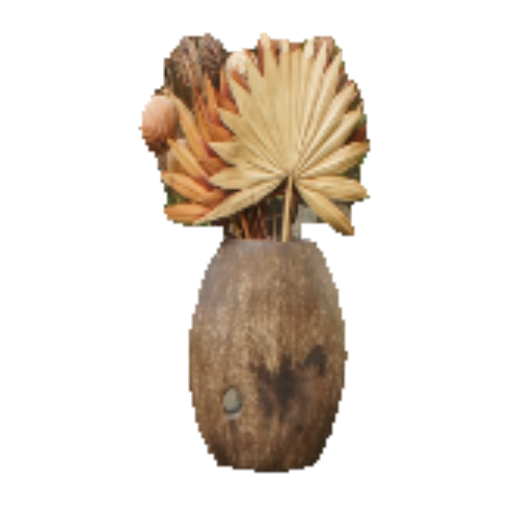} & \includegraphics[width=.15\linewidth]{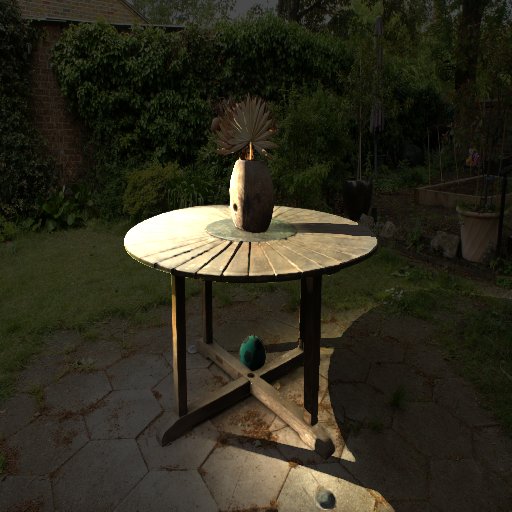} \\
		\includegraphics[width=.15\linewidth]{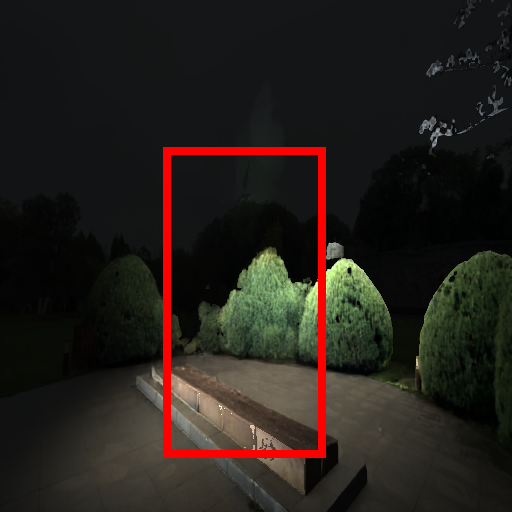} & \includegraphics[width=.15\linewidth]{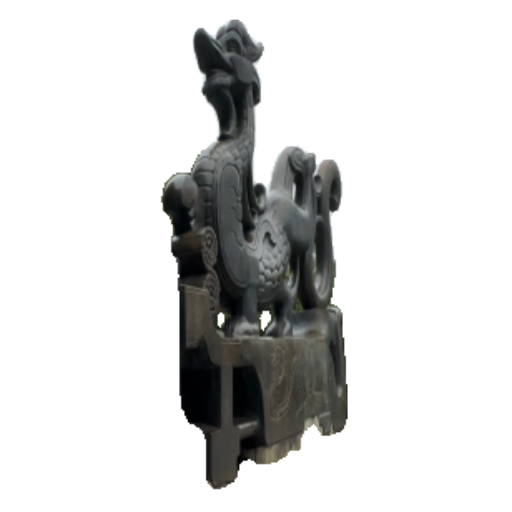} & \includegraphics[width=.15\linewidth]{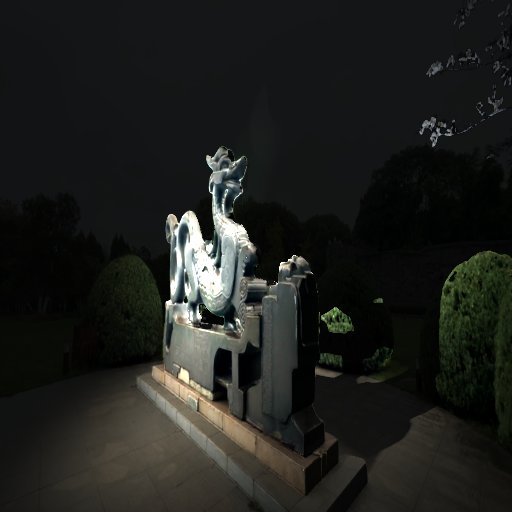} & \includegraphics[width=.15\linewidth]{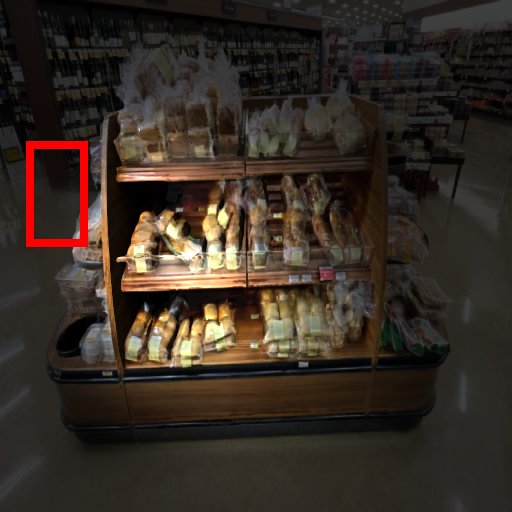} & \includegraphics[width=.15\linewidth]{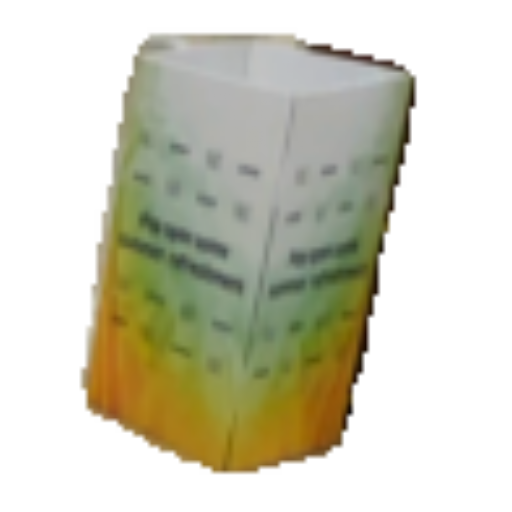} & \includegraphics[width=.15\linewidth]{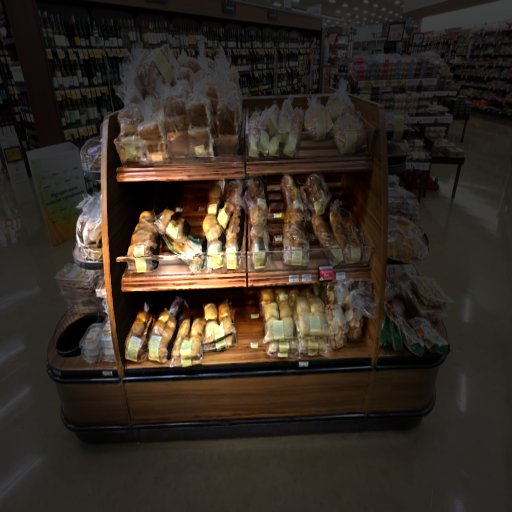} \\
		\includegraphics[width=.15\linewidth]{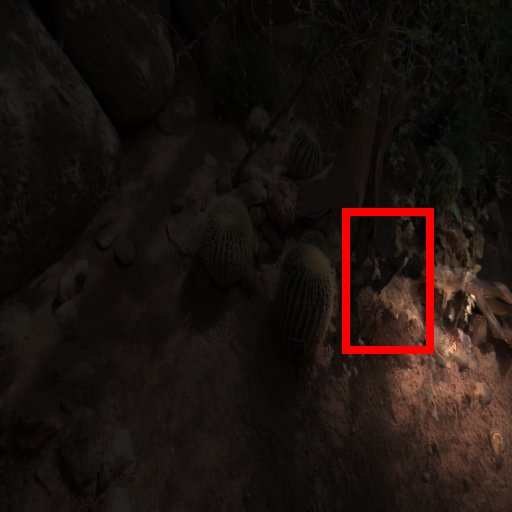} & \includegraphics[width=.15\linewidth]{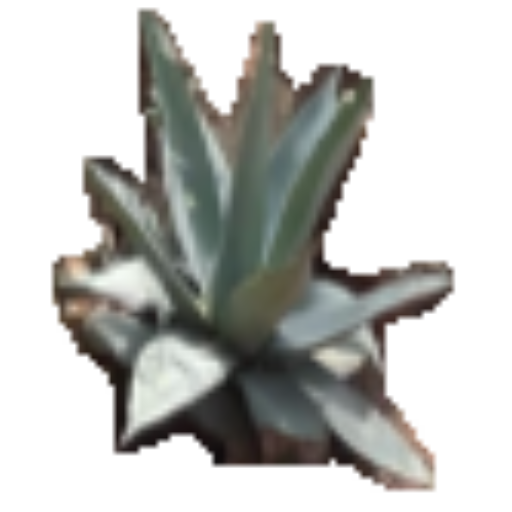} & \includegraphics[width=.15\linewidth]{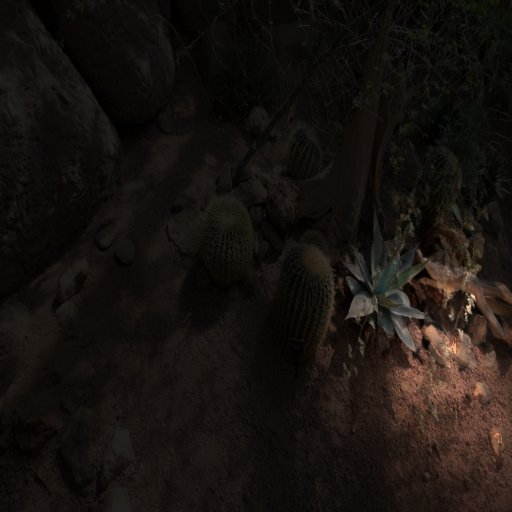} & \includegraphics[width=.15\linewidth]{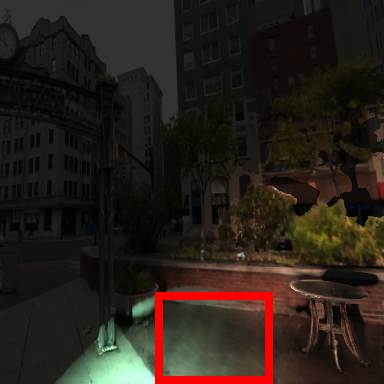} & \includegraphics[width=.15\linewidth]{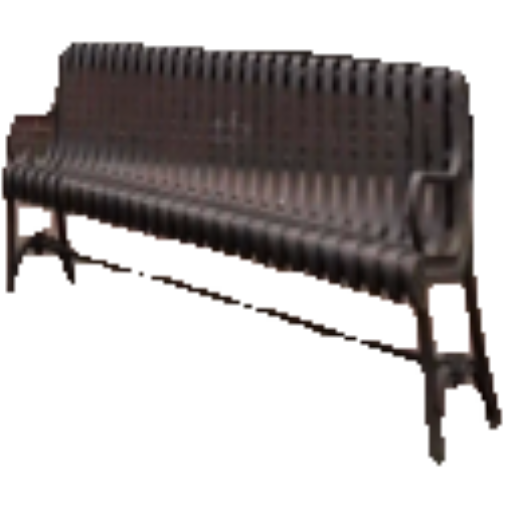} & \includegraphics[width=.15\linewidth]{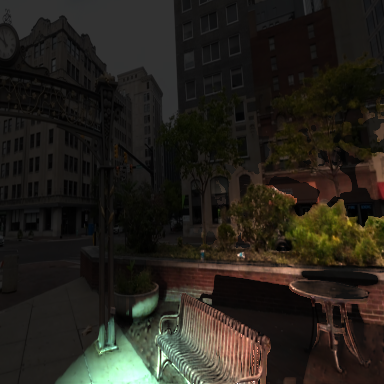} \\
	\end{tabular}
	\caption{
		Examples of illumination-augmented training triplets $(I_{\text{bg}}, I_{\text{obj}}, I_{\text{comp}})$ produced by the pipeline in Sec.~\ref{sec:augmentation}.
	}
	\label{fig:training_samples}
\end{figure}

\begin{figure}[p]
	\centering
	\scriptsize
	\begin{tabular}{@{}c@{\hspace{2pt}}c@{\hspace{2pt}}c@{\hspace{10pt}}c@{\hspace{2pt}}c@{\hspace{2pt}}c@{}}
		B.G. \& BBox                                                                  & Object                                                                         & G.T.                                                                            & B.G. \& BBox                                                                  & Object                                                                         & G.T.                                                                            \\
		\includegraphics[width=.15\linewidth]{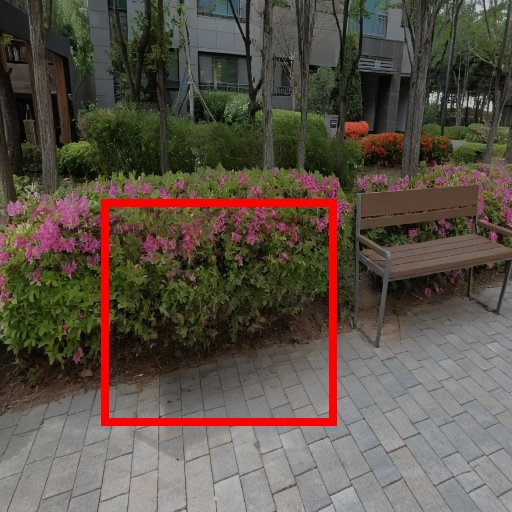}  & \includegraphics[width=.15\linewidth]{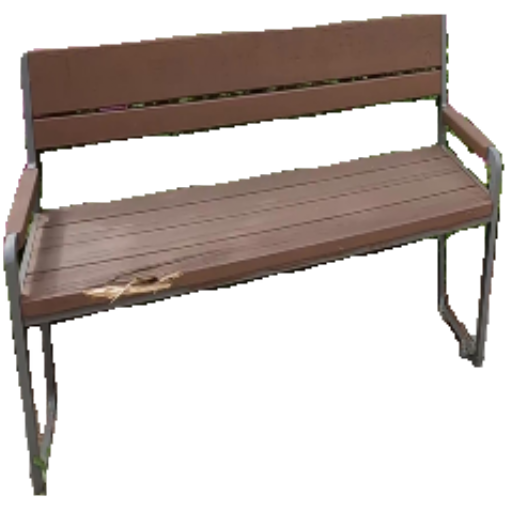}  & \includegraphics[width=.15\linewidth]{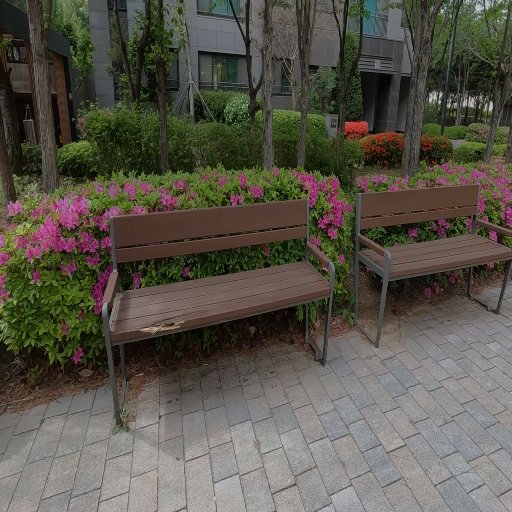}  & \includegraphics[width=.15\linewidth]{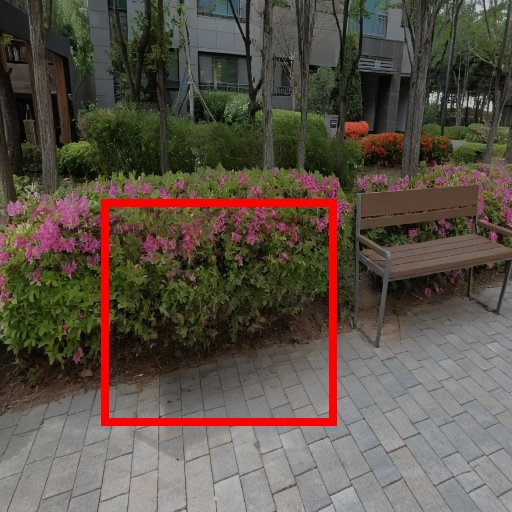}  & \includegraphics[width=.15\linewidth]{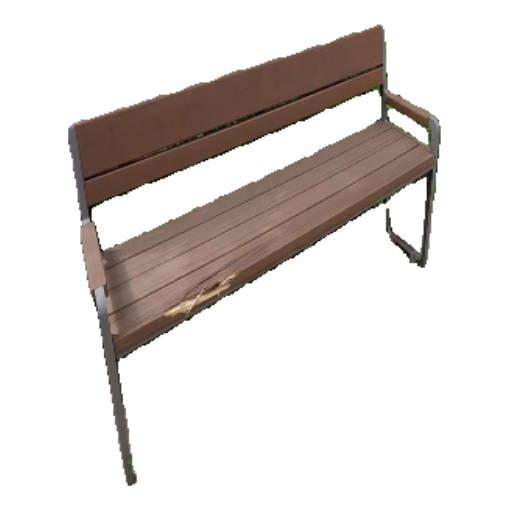}  & \includegraphics[width=.15\linewidth]{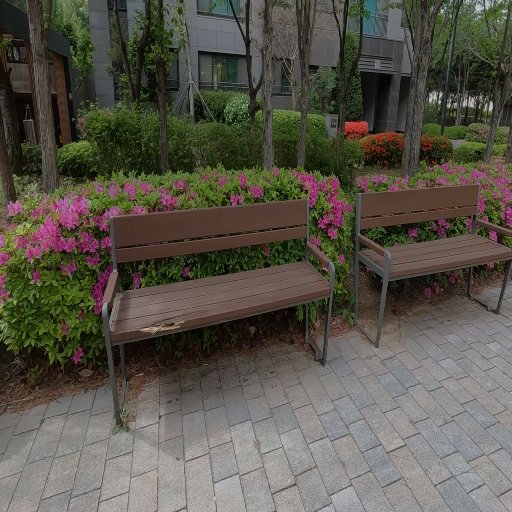}  \\
		\includegraphics[width=.15\linewidth]{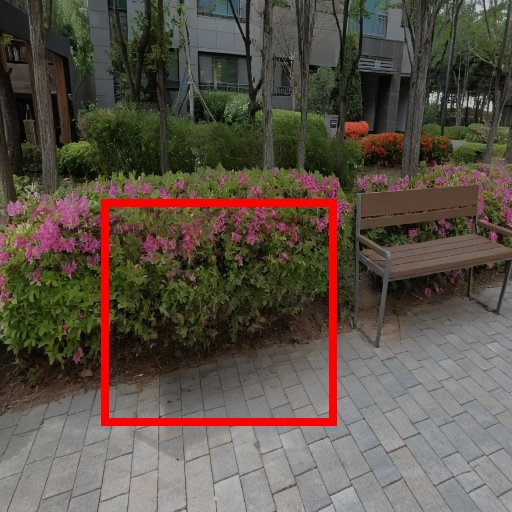}  & \includegraphics[width=.15\linewidth]{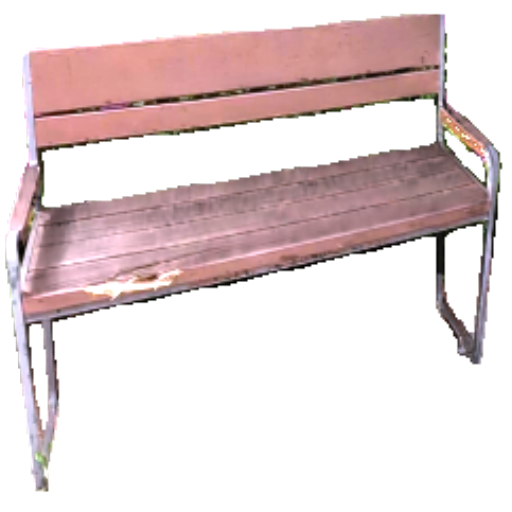}  & \includegraphics[width=.15\linewidth]{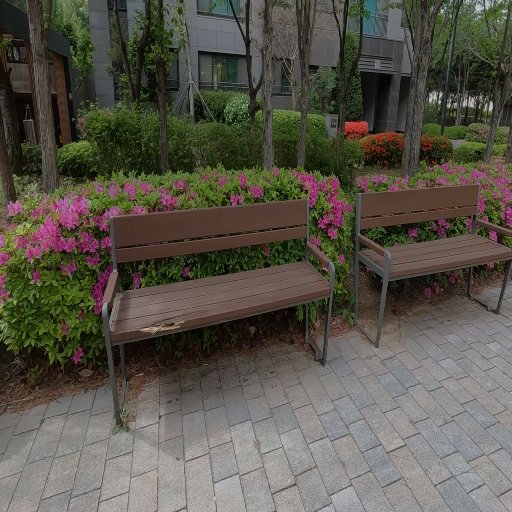}  & \includegraphics[width=.15\linewidth]{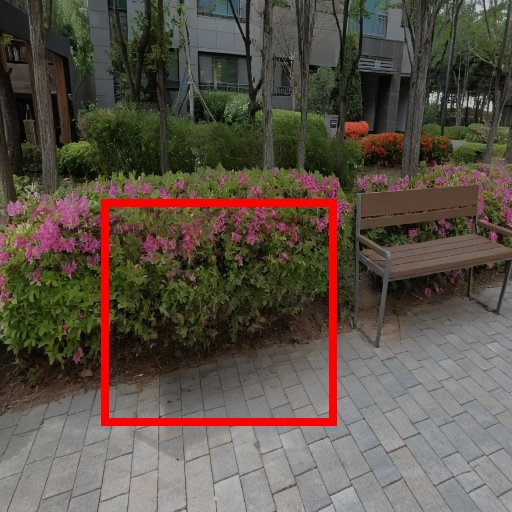}  & \includegraphics[width=.15\linewidth]{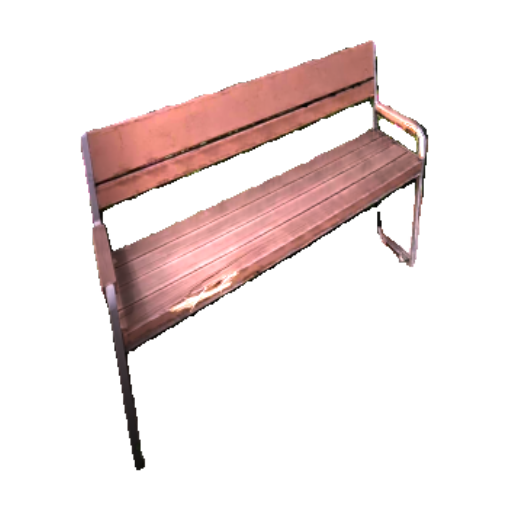}  & \includegraphics[width=.15\linewidth]{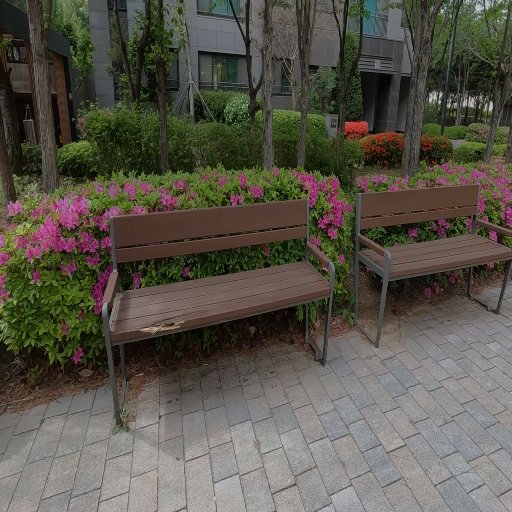}  \\
		\includegraphics[width=.15\linewidth]{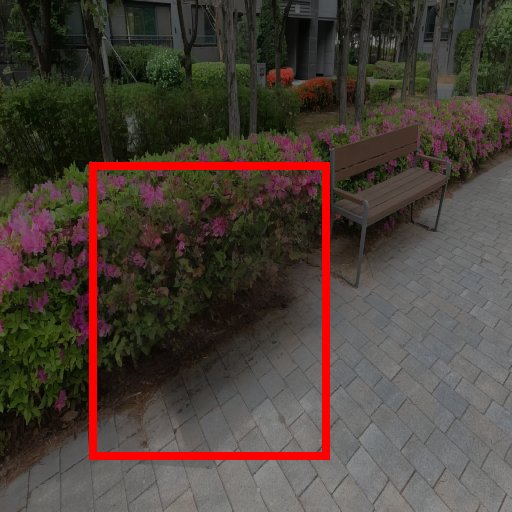}  & \includegraphics[width=.15\linewidth]{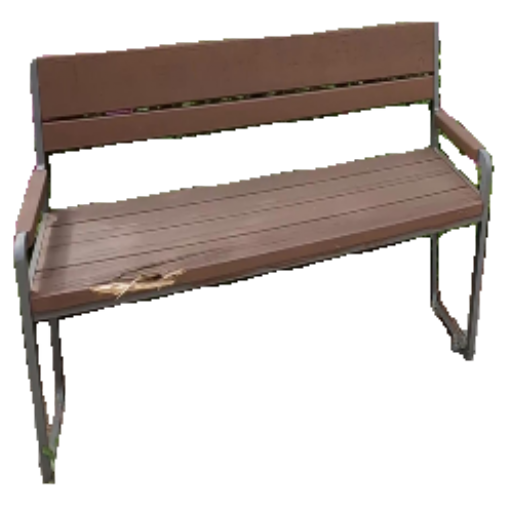}  & \includegraphics[width=.15\linewidth]{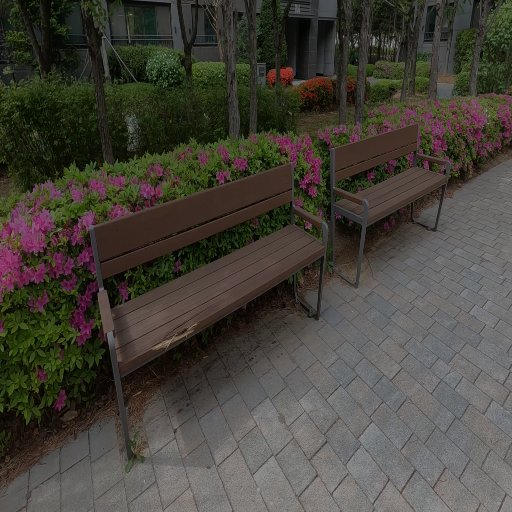}  & \includegraphics[width=.15\linewidth]{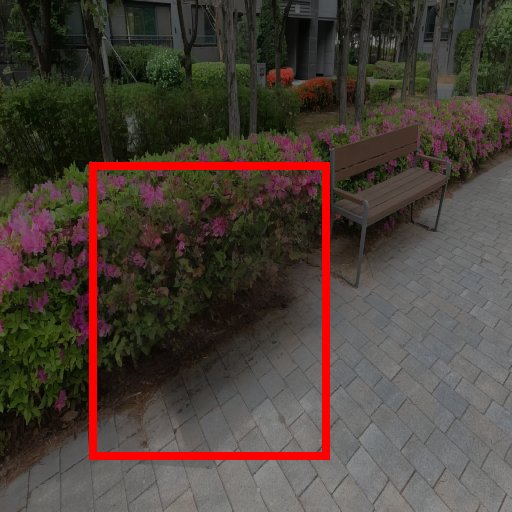}  & \includegraphics[width=.15\linewidth]{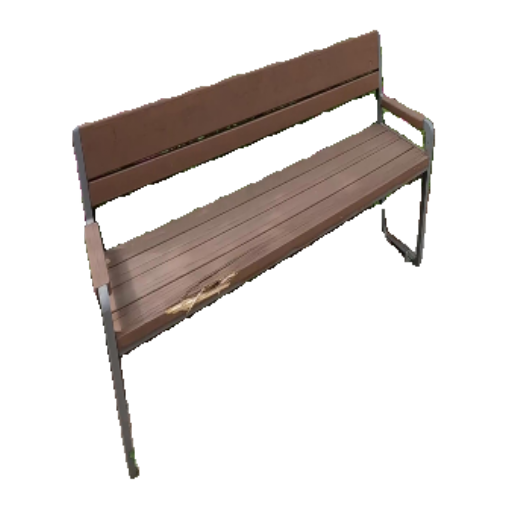}  & \includegraphics[width=.15\linewidth]{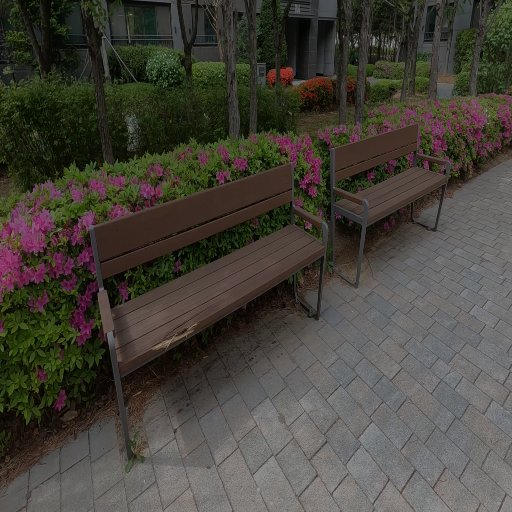}  \\
		\includegraphics[width=.15\linewidth]{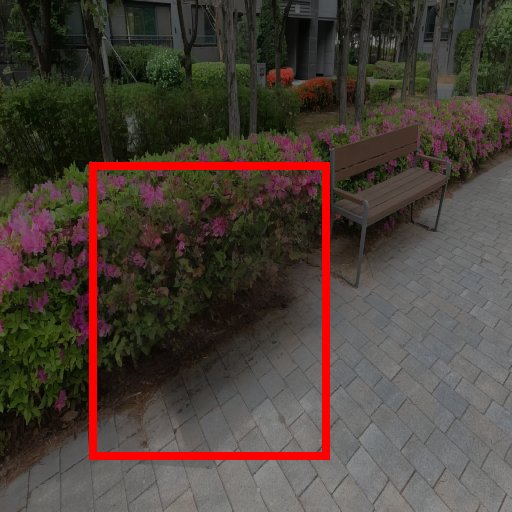}  & \includegraphics[width=.15\linewidth]{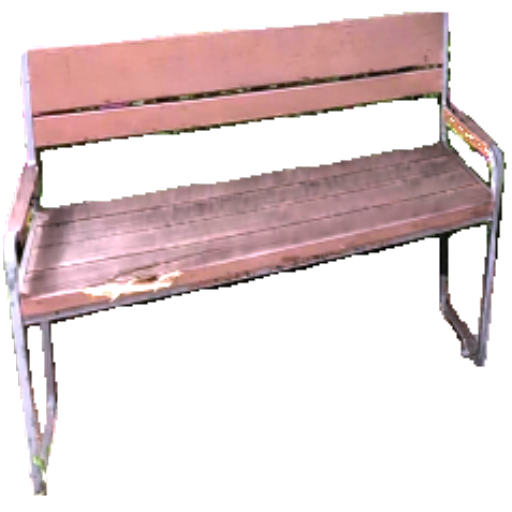}  & \includegraphics[width=.15\linewidth]{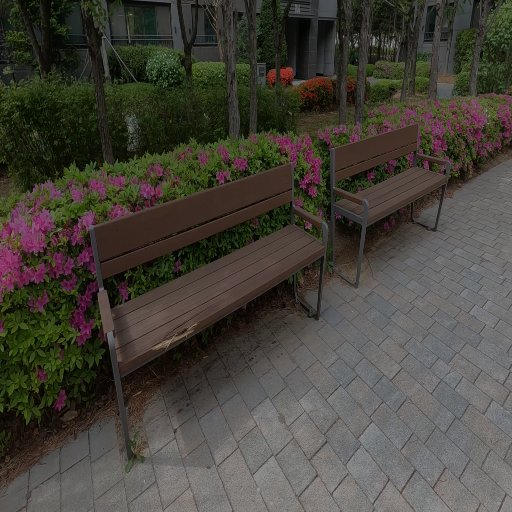}  & \includegraphics[width=.15\linewidth]{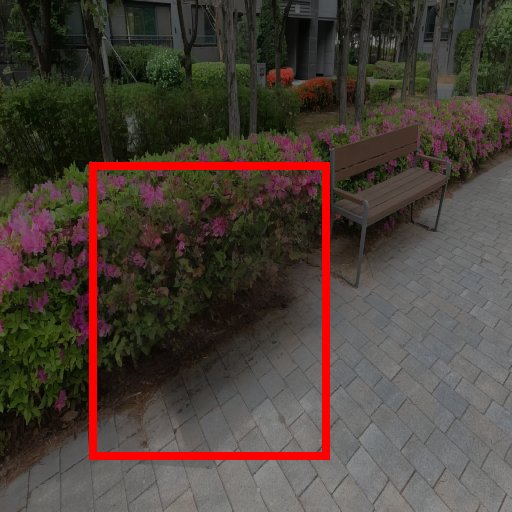}  & \includegraphics[width=.15\linewidth]{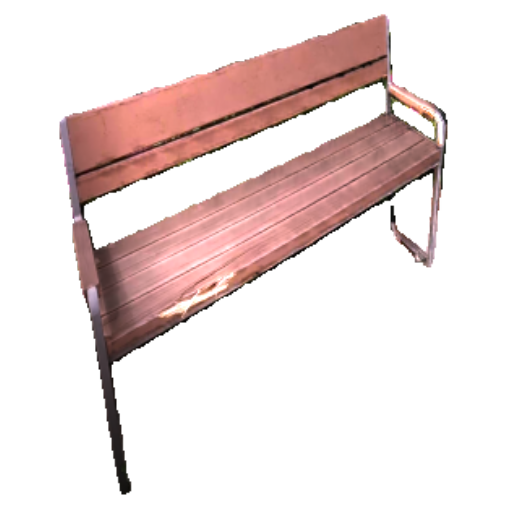}  & \includegraphics[width=.15\linewidth]{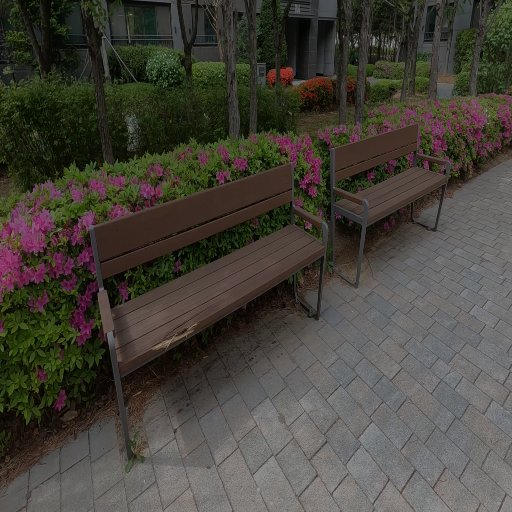}  \\
		\includegraphics[width=.15\linewidth]{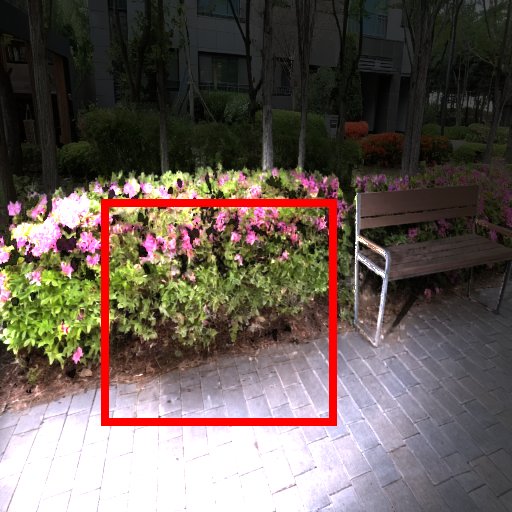}  & \includegraphics[width=.15\linewidth]{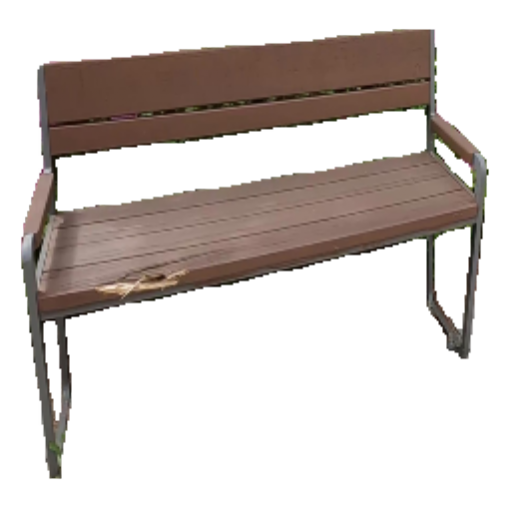}  & \includegraphics[width=.15\linewidth]{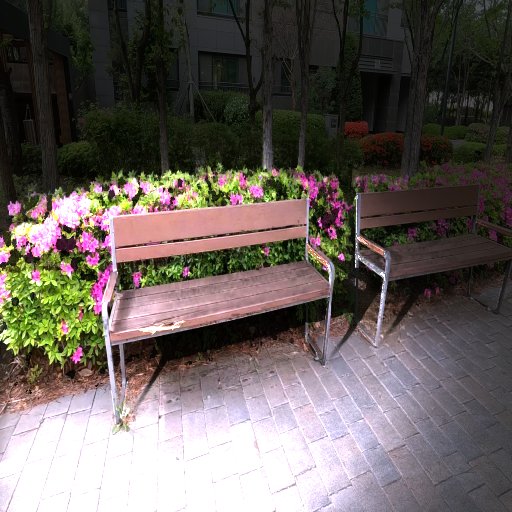}  & \includegraphics[width=.15\linewidth]{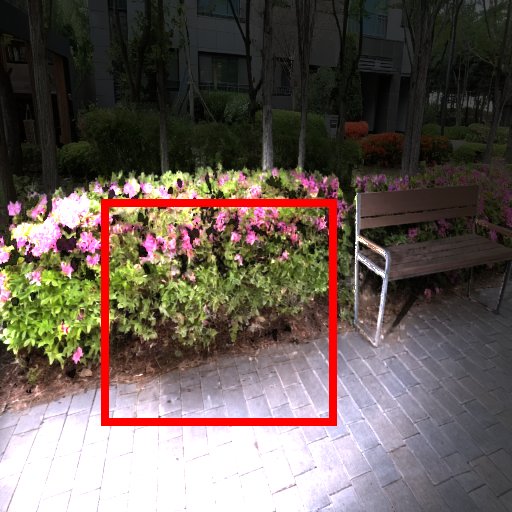}  & \includegraphics[width=.15\linewidth]{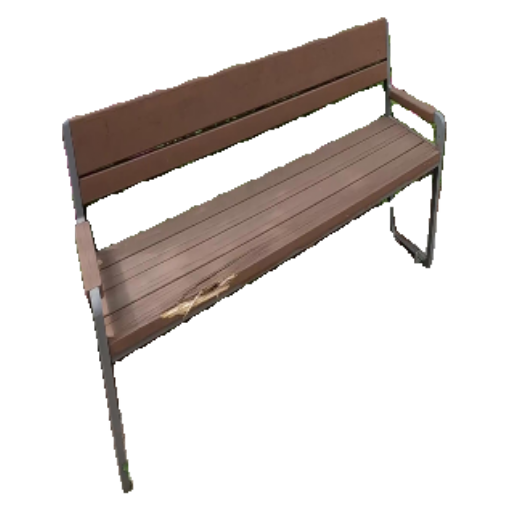}  & \includegraphics[width=.15\linewidth]{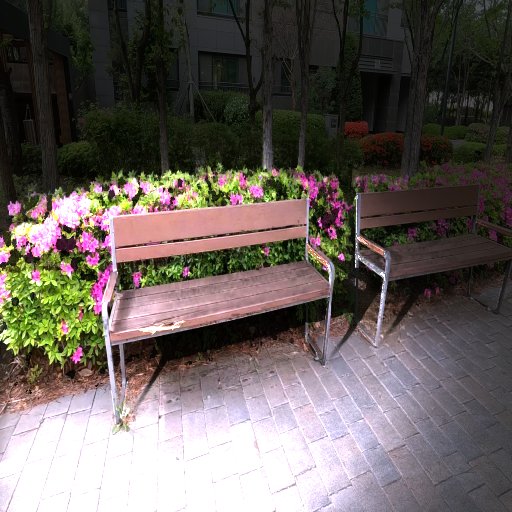}  \\
		\includegraphics[width=.15\linewidth]{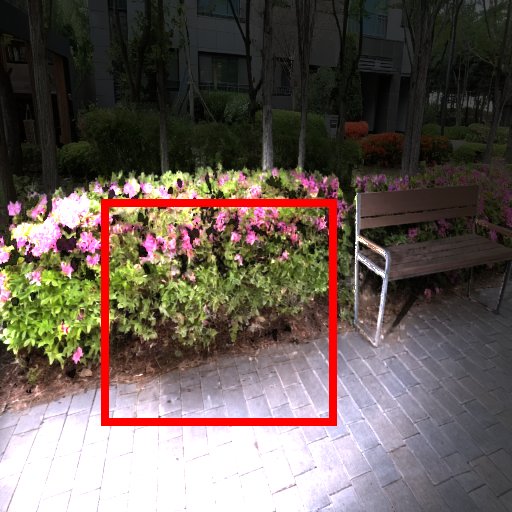} & \includegraphics[width=.15\linewidth]{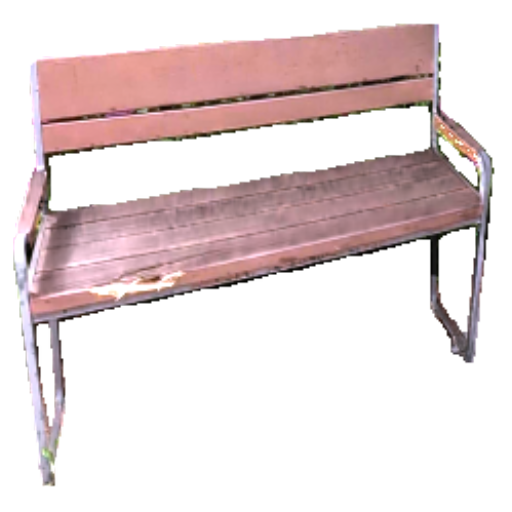} & \includegraphics[width=.15\linewidth]{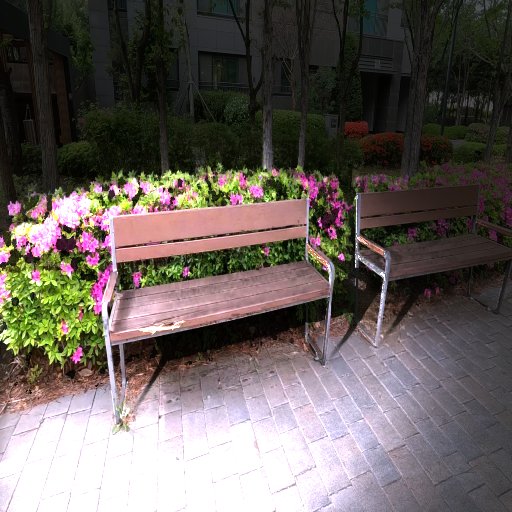} & \includegraphics[width=.15\linewidth]{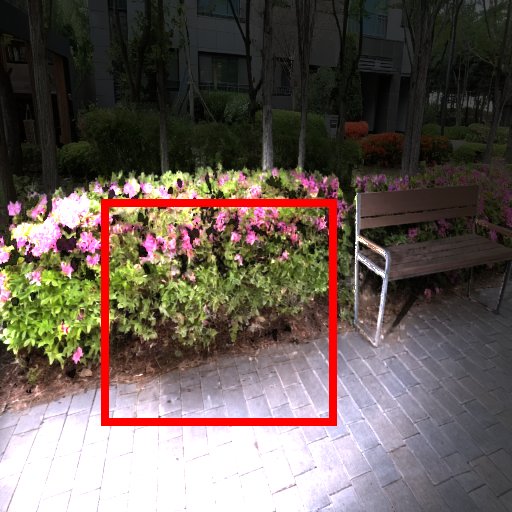} & \includegraphics[width=.15\linewidth]{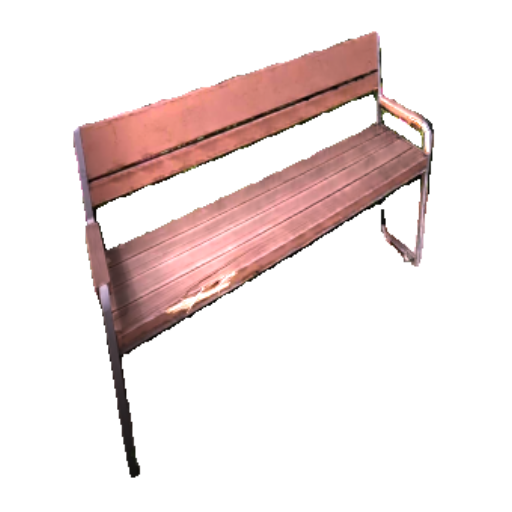} & \includegraphics[width=.15\linewidth]{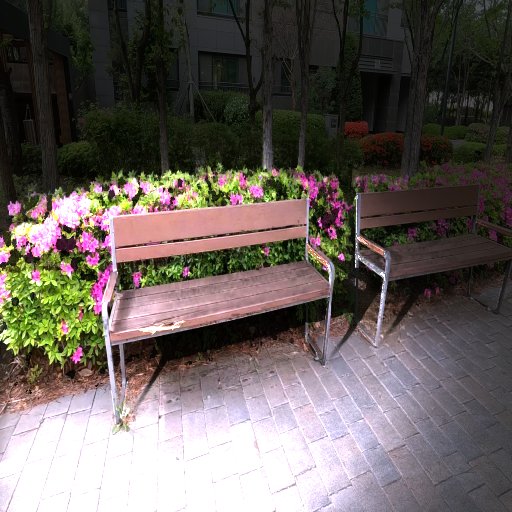} \\
		\includegraphics[width=.15\linewidth]{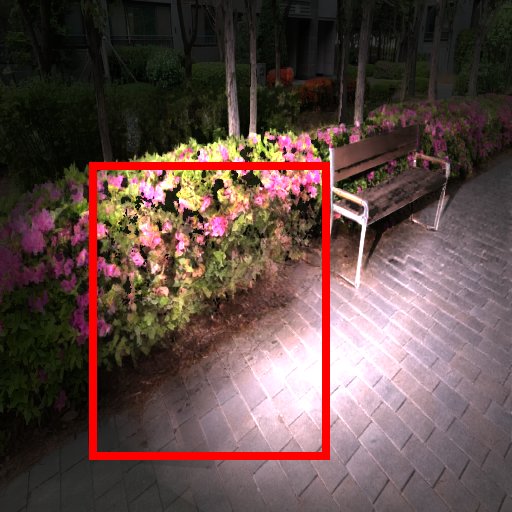} & \includegraphics[width=.15\linewidth]{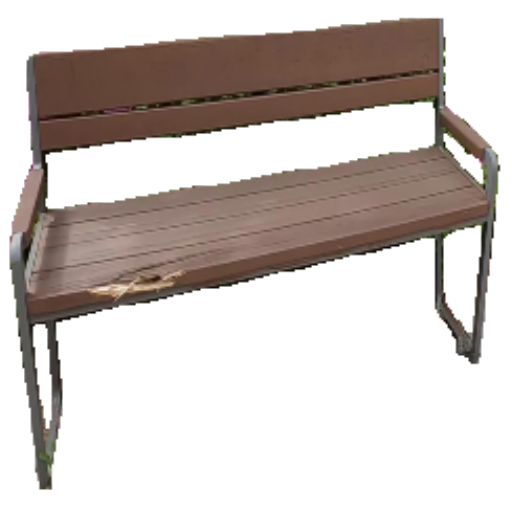} & \includegraphics[width=.15\linewidth]{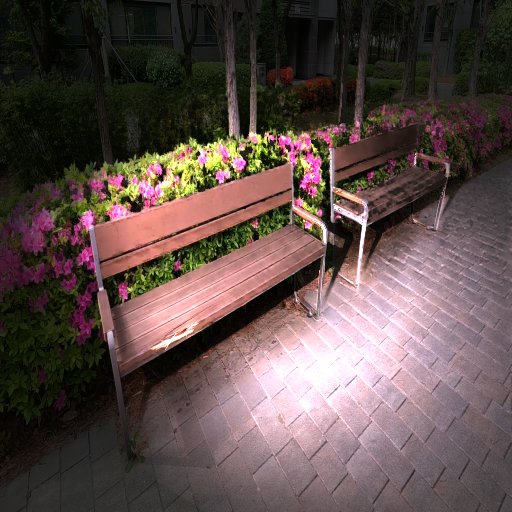} & \includegraphics[width=.15\linewidth]{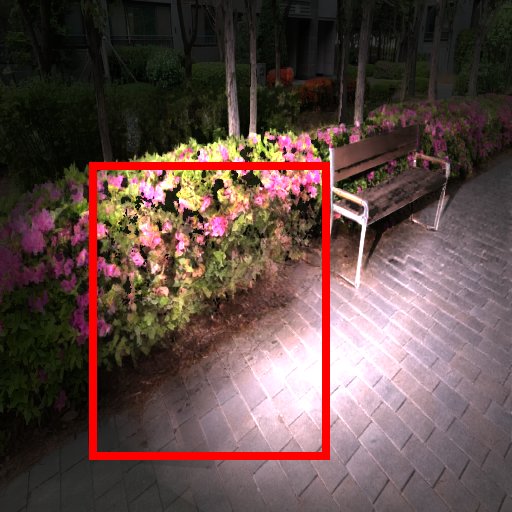} & \includegraphics[width=.15\linewidth]{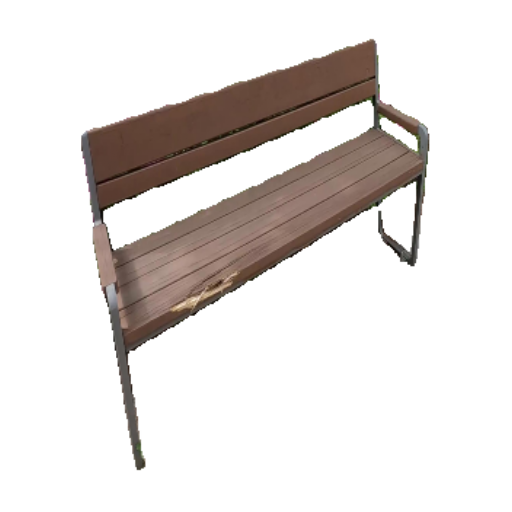} & \includegraphics[width=.15\linewidth]{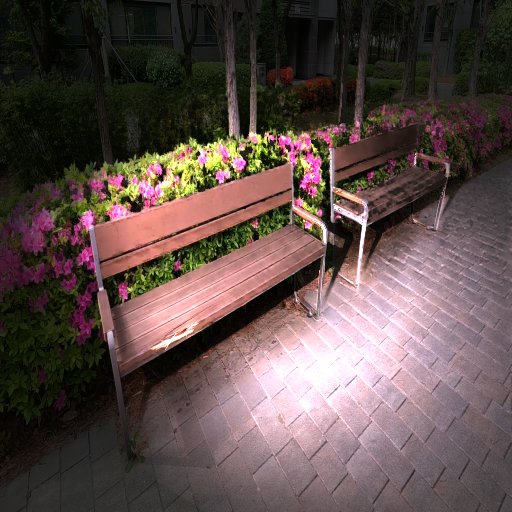} \\
		\includegraphics[width=.15\linewidth]{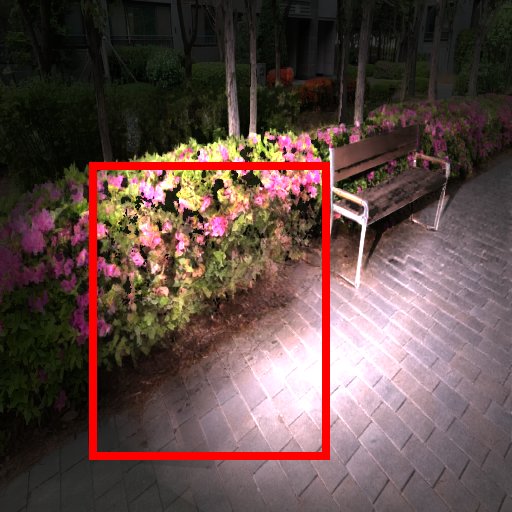} & \includegraphics[width=.15\linewidth]{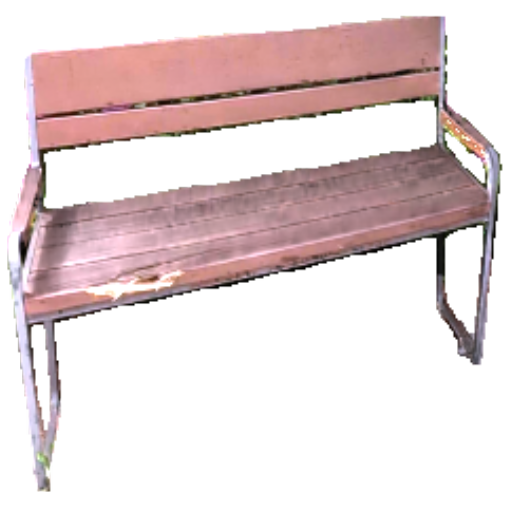} & \includegraphics[width=.15\linewidth]{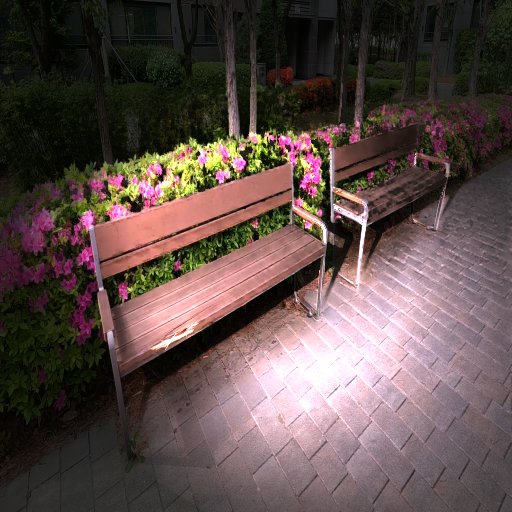} & \includegraphics[width=.15\linewidth]{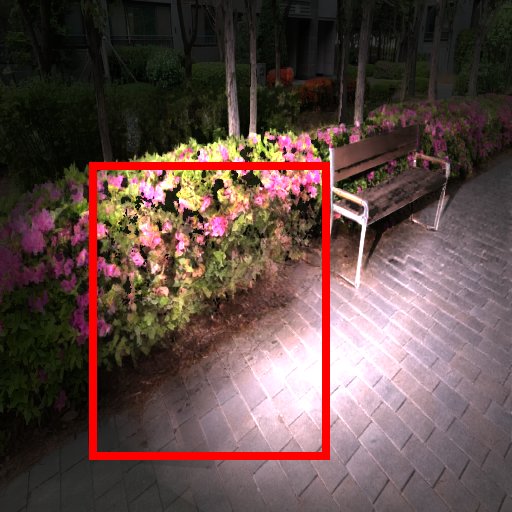} & \includegraphics[width=.15\linewidth]{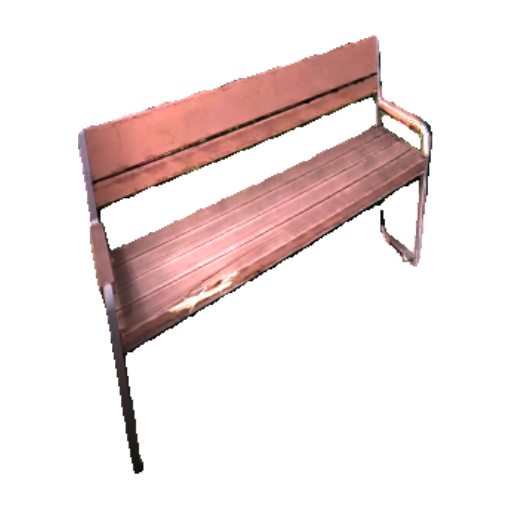} & \includegraphics[width=.15\linewidth]{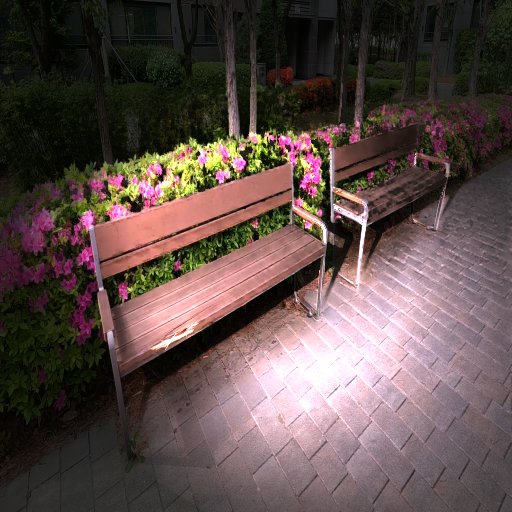} \\
	\end{tabular}
	\caption{
	All 16 training triplets from the batch-permutation scheme with $V{=}2$ reference frames (Sec.~\ref{sec:augmentation}).
	Each group of three columns shows one triplet: background with insertion bbox, input object, and target composite.
	Each of the $2V$ backgrounds is paired with an $I_{\text{obj}}$ from any of the $2V$ composites, yielding $(2V)^2{=}16$ triplets.
	}
	\label{fig:permutation_16}
\end{figure}

\FloatBarrier
\subsection{Training Data Filtering}
\label{app:data_filtering}

The automatically extracted training objects and scene estimates contain various artifacts, including inaccurate masks, tracking drift, degenerate geometry, and unstable normal predictions.
We apply three stages of input filtering before training and additionally reject unreliable relighting outputs.

\vspace{4pt}
\noindent
\textbf{Stage 1: detection filtering.}
From Detic~\citep{zhou2022detecting} detections, we discard boxes whose area (normalized by image area) lies outside $[\tau_{\min},\,\tau_{\max}]$, eliminating spurious tiny detections and scene-spanning false positives.
Among the remaining detections, we keep the top $N$ by confidence.

\vspace{4pt}
\noindent
\textbf{Stage 2: mask-shape filtering.}
We track each detected object across the video with SAM2~\citep{ravi2025sam2segmentimages} to obtain per-frame masks, then apply four complementary filters across all frames in which the object appears:
\begin{itemize}[leftmargin=1.2em,itemsep=2pt,topsep=2pt]
	\item \emph{Small-object filter.} Discard frames where the mask covers less than $\tau_a$ of the image (object barely visible), and drop objects retained in fewer than $N_f$ such frames.
	\item \emph{Thin-object filter.} We remove elongated or sparse shapes (\eg, cables, table legs, picture-frame edges). For each frame we measure how densely the mask fills its bounding box, $e = A_{\text{mask}} / A_{\text{bbox}}$, and how stretched the box is, $r = \max(h,w)/\min(h,w)$, flagging the frame as thin if $e < \tau_e$ or $r > \tau_r$. We drop the object when more than half of its surviving frames are flagged.
	\item \emph{Nested-object filter.} An object whose mask sits mostly inside another's is typically a sub-part of that object (\eg, a label on a bottle, a cushion on a sofa). For each ordered pair $(i,j)$ we compute $|M_i \cap M_j| / |M_i|$ per frame and drop $i$ if the ratio exceeds $\tau_{\text{nest}}$ in any frame.
	\item \emph{Under-object filter.} When two objects share part of the image, the one positioned lower is often partially occluded by the upper one. For each ordered pair $(i,j)$ whose bounding boxes overlap and where $i$ is not nested in $j$, we drop $i$ if its bounding-box center lies below $j$'s, $y_c(B_i) > y_c(B_j)$ (image $y$ grows downward), in any frame.
\end{itemize}

\vspace{4pt}
\noindent
\textbf{Stage 3: surface and feature filtering.}
Once scene decomposition and property estimation are complete, four further filters exploit estimated normals and learned features unavailable earlier:
\begin{itemize}[leftmargin=1.2em,itemsep=2pt,topsep=2pt]
	\item \emph{Flat-surface filter.} We remove planar surfaces such as floors and walls, keeping only foreground objects. For each frame we compute the variance of estimated normals inside the object mask and flag the frame if the variance falls below $\tau_{\text{flat}}$. We drop the object when more than half of its frames are flagged.
	\item \emph{Noisy-normal filter.} We remove objects whose normal predictions are unstable. Using the same normal variance as above, we drop object frames whose value exceeds $\tau_{\text{noisy}}$.
	\item \emph{DINO consistency filter.} Mask tracking can drift to a different instance or background, especially when the object becomes partially occluded or small. To catch such failures, we take the frame with the largest mask (where the object is most clearly visible) as reference and compare every other frame's crop against it using DINOv2~\citep{simeoni2025dinov3} cosine similarity. We drop frames whose similarity falls below $\tau_{\text{dino}}$ and discard objects with fewer than 10 surviving frames.
	\item \emph{Normal--ray consistency filter.} A surface visible to the camera should have its normal facing the camera; frames where many pixels violate this signal unreliable normal predictions. We compute the fraction of bounding-box pixels with $\mathbf{n}(x) \cdot (-\boldsymbol{\omega}_o(x)) < 0$ (normal pointing away from the camera) and drop frames whose fraction exceeds $\tau_{\text{nr}}$.
\end{itemize}
Finally, we cap each scene at 8 objects, keeping those with the largest total mask area summed across surviving frames.

\vspace{4pt}
\noindent
\textbf{Relighting quality filtering.}
We discard degenerate relighting outputs, including holes or speckles in illuminated regions, light leaks, flat-shading artifacts, and cases where the foreground object is brightened but the surrounding background remains barely illuminated, leaving insufficient contextual lighting cues.

On DL3DV, detection filtering removes 91.1\% of raw boxes, mask-shape filtering removes 53.1\% of tracked objects, and surface filtering removes 8.0\% of the remaining objects.
The DINO, normal--ray, and relighting quality filters remove 38.2\%, 4.0\%, and 33.7\% of frames, respectively.

\newcommand{\fcell}{\framebox[.22\linewidth][c]{\rule[-1.6cm]{0pt}{3.2cm}\textit{placeholder}}}
\newcommand{\fimg}[1]{\includegraphics[width=.22\linewidth]{#1}}

\FloatBarrier
\subsection{Limitations}
\label{app:limitations}

\modelname{} adapts the inserted object's shading to local lighting at the insertion location, but several limitations remain.
We illustrate failure cases in Fig.~\ref{fig:failure_cases}.

\begin{figure}[htbp]
	\centering
	\scriptsize
	\setlength{\tabcolsep}{2pt}
	\begin{tabular}{@{} *{3}{c@{\hspace{2pt}}} @{}}
		Background                               & Object                                    & \modelname{}                               \\
		\fimg{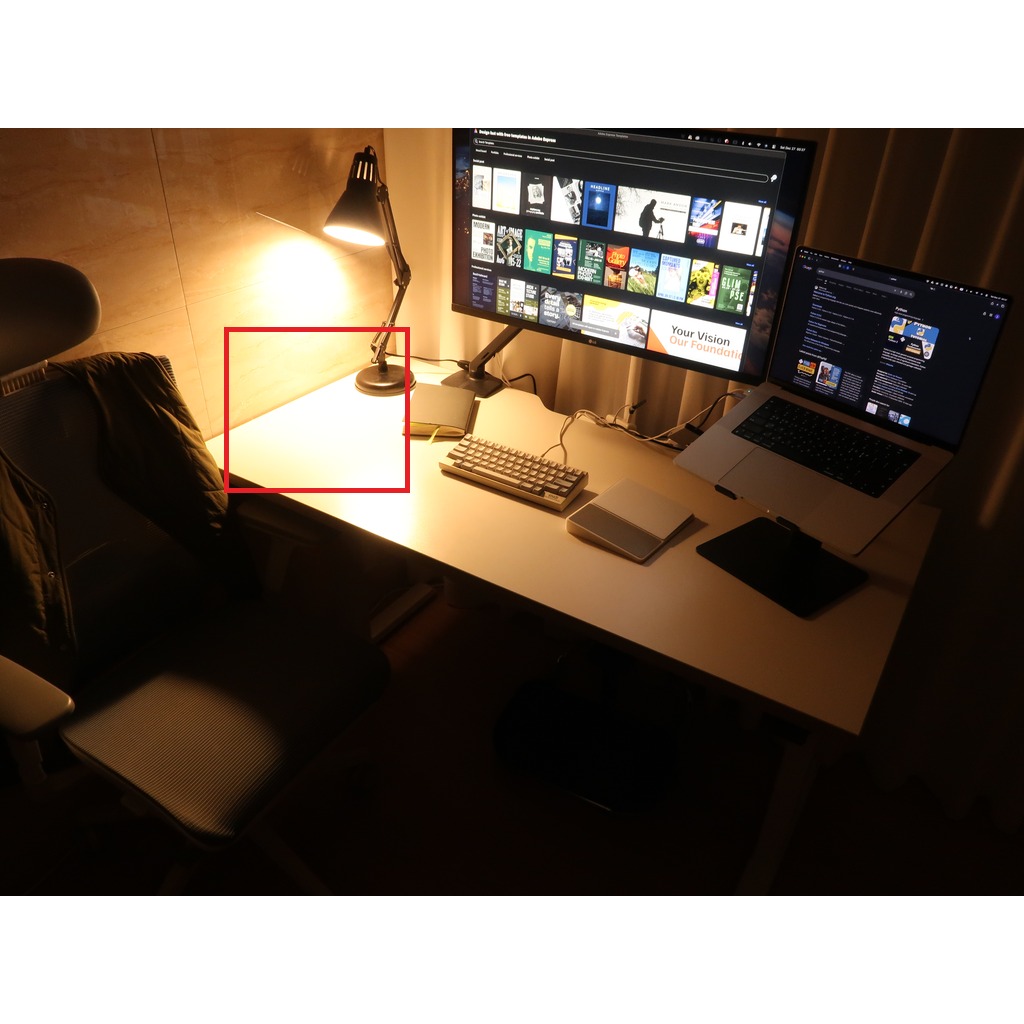} & \fimg{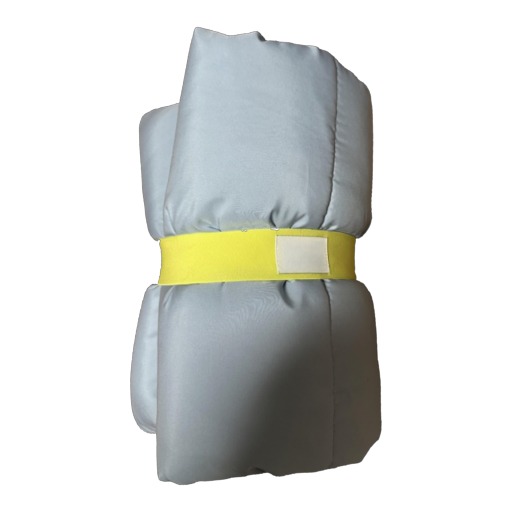} & \fimg{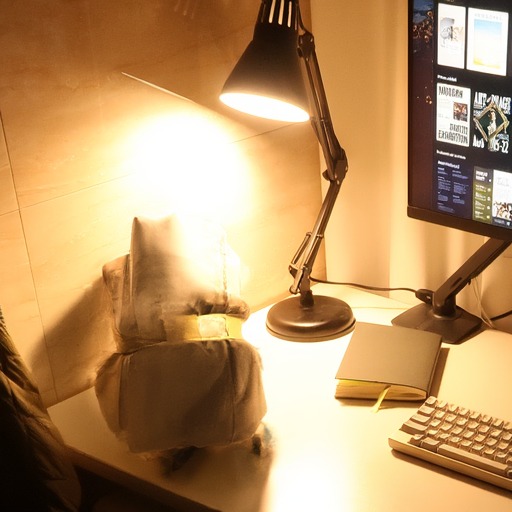} \\
		\fimg{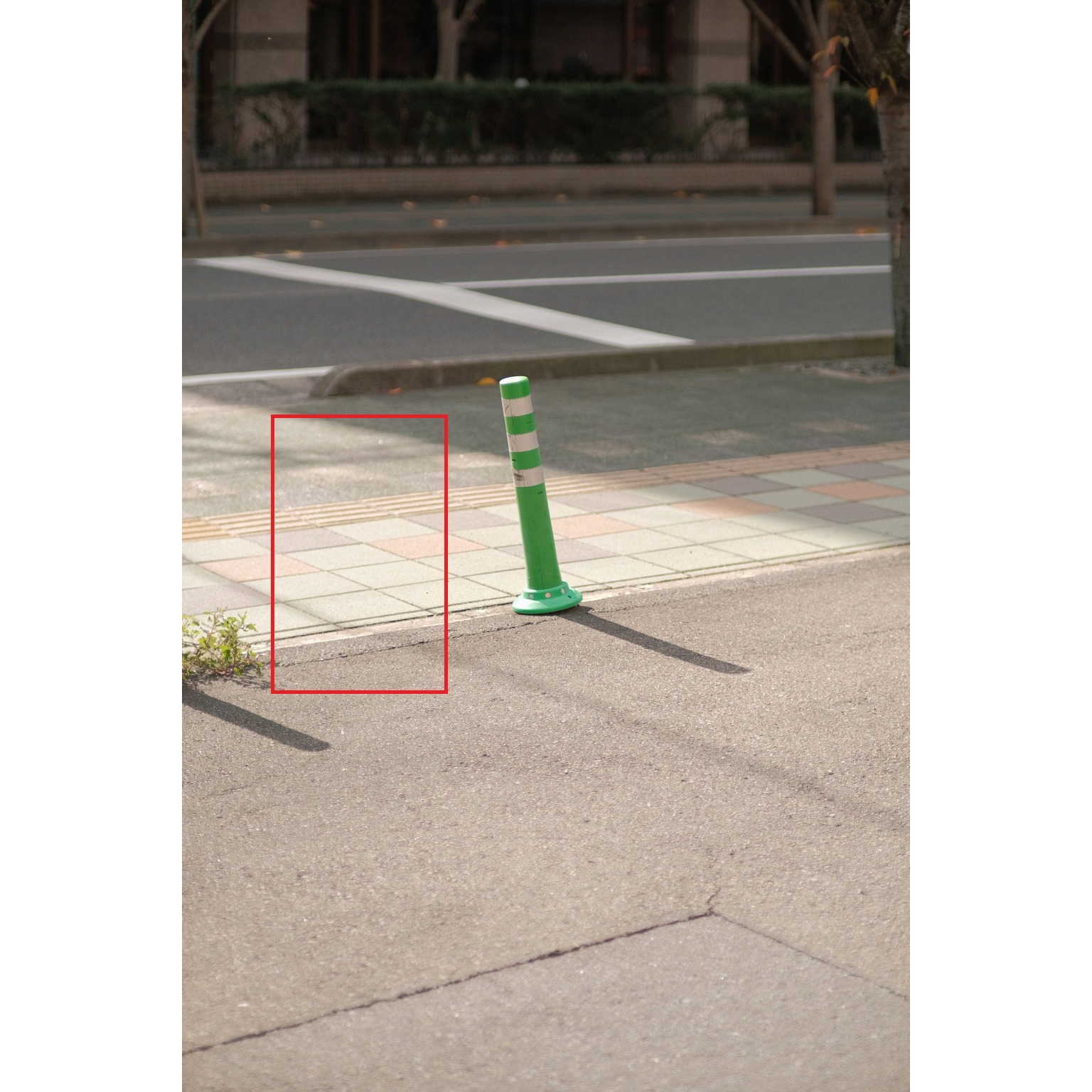} & \fimg{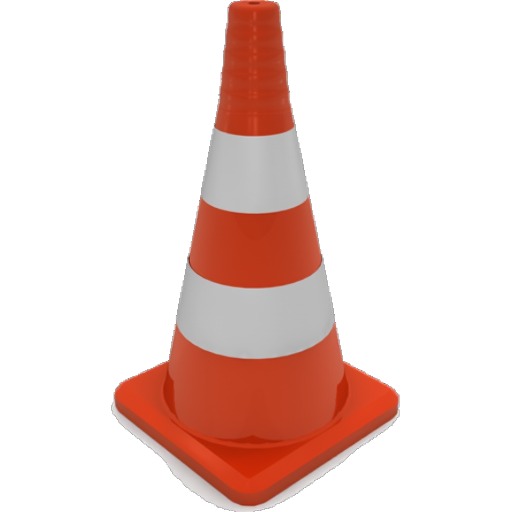} & \fimg{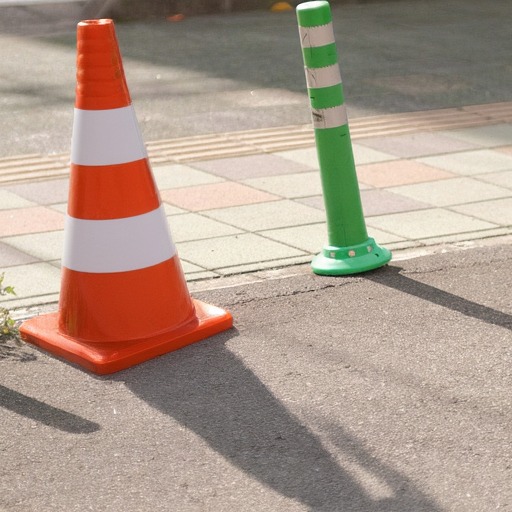} \\
	\end{tabular}
	\caption{
		Failure cases of \modelname.
		\textit{Top:} object affordance is ignored. A soft cushion-like object is placed upright on a desk, an unstable pose; the natural placement is to lay it flat so that it does not topple.
		\textit{Bottom:} Outdoor scenes follow a different lighting regime. A distant sun illuminates the scene from a single direction regardless of the location, resulting in parallel cast shadows. Our virtual-spotlight augmentation models point-source illumination and does not reproduce this parallel-shadow structure.
	}
	\label{fig:failure_cases}
\end{figure}

\noindent
\textbf{Resolution.}
Our $512\times512$ generation resolution limits fine-scale object appearance and lighting detail.
A possible extension is to first generate the illumination-adapted crop at $512\times512$ and then refine the inserted region at higher resolution, conditioned on the original object and background.
We have not evaluated this extension and leave it to future work.

\noindent
\textbf{Articulated and deformable objects.}
\modelname{} treats the inserted object as rigid: its appearance adapts to the location, but joints do not bend and shape does not deform.
For example, a cup adapts to a tilted surface, whereas a standing figure does not re-pose to sit on a chair.
Affordance-driven re-posing is a separate problem from illumination adaptation.

\noindent
\textbf{Complex geometry and materials.}
The augmentation relies on monocular depth and approximate BRDF estimates, which are unreliable for thin structures, highly specular surfaces, and transparent materials, leading to relighting artifacts.
Glossy objects are partially supported through the microfacet specular term of the BRDF, but small high-frequency reflections are unreliable.
Thin or complex objects are more susceptible to segmentation and depth errors; the filters of \S\ref{app:data_filtering} protect training quality but leave such objects under-represented in the training data.

\noindent
\textbf{Outdoor scenes.}
We focus on indoor scenes, where multiple localized emitters such as lamps and windows produce spatially varying illumination over short distances.
Outdoor scenes exhibit a different lighting structure: a distant sun acts as the dominant source, so light rays arrive at nearly the same angle throughout the scene and cast shadows are parallel~\citep{holdgeoffroy-cvpr-17,zhu2021spatiallyvaryingoutdoorlightingestimation}.
Our virtual spotlights model point-source illumination and may therefore not fully capture this parallel-shadow structure.
Covering both indoor and outdoor lighting within a unified framework remains an open problem~\citep{dastjerdi2023everlight}; extending our augmentation to outdoor scenes is left to future work.

\FloatBarrier
\Needspace{8\baselineskip}
\subsection{Broader Impact}
\label{app:broader_impact}

\modelname{} extends object compositing to scenes with spatially varying lighting, with applications in content creation, interior-design visualization, and augmented reality.
The same capability can be misused for deceptive imagery and fraud; \modelname{} does not introduce a new manipulation capability, but improves the local-lighting plausibility of an existing task already served by diffusion-based compositing.
As generative quality rises, the community should continue to invest in detection and provenance tools.

\FloatBarrier
\Needspace{9\baselineskip}
\subsection{Image Attribution}
\label{app:image_attribution}

The outdoor failure case in Fig.~\ref{fig:failure_cases} (bottom row) uses third-party assets:
\begin{itemize}[leftmargin=1.5em,topsep=2pt,itemsep=2pt]
	\item \textbf{Background:} ``A green and white tube on a sidewalk'' by Mak (\url{https://unsplash.com/photos/a-green-and-white-tube-on-a-sidewalk-Js1NAy3LQUQ}), Unsplash License (\url{https://unsplash.com/license}).
	\item \textbf{Object (traffic cone):} ``Cones PNG 13323'' from \url{https://pngimg.com/image/13323}, pngimg License (\url{https://pngimg.com/license}).
\end{itemize}

\clearpage
\endgroup

\end{document}